\documentclass[10pt,twocolumn,letterpaper]{article}

\usepackage{cvpr}
\usepackage{times}
\usepackage{epsfig}
\usepackage{graphicx}
\usepackage[table]{xcolor}%
\usepackage{amsmath}
\usepackage{amssymb}
\usepackage{subcaption}
\usepackage{adjustbox}

\usepackage[pagebackref=true,breaklinks=true,letterpaper=true,colorlinks,bookmarks=false]{hyperref}

\newcommand{\cS}{\mathcal{S}}

\newcommand{\bbR}{\mathbb{R}}
\newcommand{\bx}{\mathbf{x}}
\newcommand{\by}{\mathbf{y}}

\newcommand\shenlong[1]{\textcolor{black}{#1}}

\newcommand\simon[1]{\textcolor{black}{#1}}

\makeatletter
\usepackage{xspace}
\def\@onedot{\ifx\@let@token.\else.\null\fi\xspace}
\DeclareRobustCommand\onedot{\futurelet\@let@token\@onedot}

\newcommand{\figref}[1]{Fig\onedot~\ref{#1}}

\newcommand{\tabref}[1]{Tab\onedot~\ref{#1}}

\def\eg{\emph{e.g}\onedot} 
\def\ie{\emph{i.e}\onedot} 
 
\def\etc{\emph{etc}\onedot}

\cvprfinalcopy %

\def\cvprPaperID{2565} %

\ifcvprfinal\fi
\begin{document}

\title{Deep Parametric Continuous Convolutional Neural Networks}

\author{ Shenlong Wang$^{1, 3,^\ast}$  \ \ Simon Suo$^{2,3,^\ast}$  \ \ Wei-Chiu Ma$^{3}$  \ \ Andrei Pokrovsky$^{3}$ \ \ Raquel Urtasun$^{1, 3}$ \\
$^1$University of Toronto, $^2$University of Waterloo, $^3$Uber Advanced Technologies Group\\
{\tt\small \{slwang, suo, weichiu, andrei, urtasun\}@uber.com}
}

\maketitle

\begin{abstract}

Standard convolutional neural networks assume a \simon{grid structured input} is available and exploit discrete convolutions as their fundamental building blocks. This limits their applicability to many real-world applications. 
\simon{In this paper we propose Parametric Continuous Convolution, a new learnable operator that operates over non-grid structured data.} The key idea is to exploit parameterized kernel functions that span the full continuous vector space. 
\simon{This generalization allows us to learn over} 
arbitrary data structures as long as their support relationship is computable.
Our experiments show significant improvement over the state-of-the-art in point cloud segmentation of indoor and outdoor scenes, and lidar motion estimation of driving scenes.

\end{abstract}

\section{Introduction}

Discrete convolutions are the most fundamental building block of modern deep learning architectures. 
Its efficiency and effectiveness \simon{relies on} 
the fact that the data appears naturally in a \simon{dense} grid structure (e.g., 2D grid for images,  3D grid for videos). 
However, many real world applications such as visual perception from 3D point clouds,  mesh registration and non-rigid shape correspondences rely on making statistical predictions from non-grid structured data. 
Unfortunately, standard convolutional operators cannot be directly applied \simon{in these cases}. 

Multiple approaches have been proposed to handle non-grid structured data. The simplest approach is to voxelize the space to form  a grid where standard 
\simon{discrete} convolutions can be performed \cite{modelnet40, octnet}. However, most of the volume is typically empty, and thus this results in both  memory inefficiency and wasted computation.  %
Geometric deep learning \cite{geometric-deep-learning, gcn} and graph neural network approaches \cite{gnn, ggnn} exploit the graph structure of the data and model the relationship between nodes. 
Information is then propagated through the graph edges. However, they either have difficulties generalizing well or require strong feature representations as input to perform competitively. End-to-end learning is typically performed via back-propagation through time, but \simon{it} is difficult to learn very deep networks due to the memory limitations of modern GPUs. %

In contrast to the aforementioned approaches, in this paper we propose a new learnable operator, which we call {\it parametric continuous convolution}. %
The key idea is a parameterized kernel function that spans the full continuous vector space. 
In this way, it can handle arbitrary data structures as long as its support relationship is computable. %
This is a natural extension since objects in the real-world such as point clouds 
captured from 3D sensors are distributed unevenly in continuous domain. Based upon this we build a new family of deep neural networks that can be applied on generic non-grid structured data. %
The proposed networks are both expressive and memory efficient. %

We demonstrate the effectiveness of our approach in both semantic labeling and motion estimation of point clouds. \simon{Most} importantly,  we show that very deep networks can be learned over raw point clouds in an end-to-end manner. Our experiments show that the proposed approach outperforms the state-of-the-art by a large margin in both outdoor and indoor 3D point cloud segmentation tasks, as well as lidar motion estimation in driving scenes. 
Importantly, our  outdoor semantic labeling and lidar flow experiments are conducted on a very large scale dataset, containing 223 billion points captured by a  3D sensor mounted on the roof of  a self-driving car.
To our knowledge, this is 2 orders of magnitude larger than any existing benchmark.

\section{Related Work}

\paragraph{Deep Learning for 3D Geometry:} Deep learning approaches that exploit 3D geometric data have recently become populer in the computer vision  community.  
Early approaches convert the 3D data into a  two-dimensional  RGB + depth image \cite{fcn, rgbd-detection-seg} and exploit  conventional convolutional neural networks (CNNs). Unfortunately, this \simon{representation} does not capture the true  geometric relationships between  3D points (i.e. neighboring pixels could be potentially far away geometrically). %
Another popular approach is to  conduct 3D convolutions over volumetric representations \cite{modelnet40, subvolume, octnet, sparseconvnet, voxnet}.  Voxelization is \simon{employed to}
convert point clouds 
into a 3D grid that encodes the geometric information. 
These approaches have been popular in medical imaging and indoor scene understanding, where the volume is relatively small. However, typical voxelization approaches sacrifice  precision and the 3D volumetric representation is not memory efficient. Sparse convolutions \cite{sparseconvnet} and advanced data structures such as oct-trees \cite{octnet} have been used to overcome these difficulties.  Learning directly over point clouds has only been studied very recently. The pioneer work of PointNet \cite{pointnet}, learns an MLP over individual points and aggregates global information using pooling. PointNet++ \cite{pointnet2}, the follow-up, improves the ability to capture local structures through a multi-scale grouping strategy. 

\paragraph{Graph Neural Networks:}  Graph neural networks (GNNs) \cite{gnn} are generalizations of neural networks to graph structured data. Early approaches apply neural networks either over the hidden representation of each node or the messages passed between adjacent nodes in the graph, and use back-propagation through time to conduct learning. Gated graph neural networks (GGNNs) \cite{ggnn} exploit  gated recurrent units along with modern optimization techniques, resulting in improved performance.
In \cite{3dgnn}, GGNNs are applied to point cloud segmentation, \simon{achieving} significant improvements over the state-of-the-art.  %
One of the major difficulties of graph neural networks is that  propagation is conducted in a synchronous manner and thus it is hard to scale up to graphs with millions of nodes.   
Inference in graphical models as well as recurrent neural networks can be seen as special cases of graph neural networks.

\paragraph{Graph Convolution Networks:} An alternative \simon{formulation is} to learn convolution operations over graphs.  %
These methods  can be categorized into spectral  and spatial approaches depending on which domain the convolutions are applied to.  For spectral methods, convolutions are converted to multiplication %
by computing the  graph Laplacian in Fourier domain \cite{spectralnn, anisotropic-cnn, syncspeccnn}. Parameterized spectral filters can be incorporated to reduce overfitting \cite{spectralnn}. These methods are not feasible for large scale data due to the expensive computation, since there is no FFT-like trick over generic graph.  Spatial approaches directly propagate information along the node neighborhoods in the graph. This can be implemented either through low-order approximation of spectral filtering\cite{chebnet, gcn, cnn-graph-hash}, or diffusion in a support domain \cite{monet, anisotropic-cnn, ecc, syncspeccnn, schnet}. Our approach  generalizes spatial approaches  in two ways: first, we use more expressive convolutional kernel functions; second, the output of the convolution could be any point in the whole continuous domain.

\begin{figure}
  \includegraphics[width=1.02\linewidth]{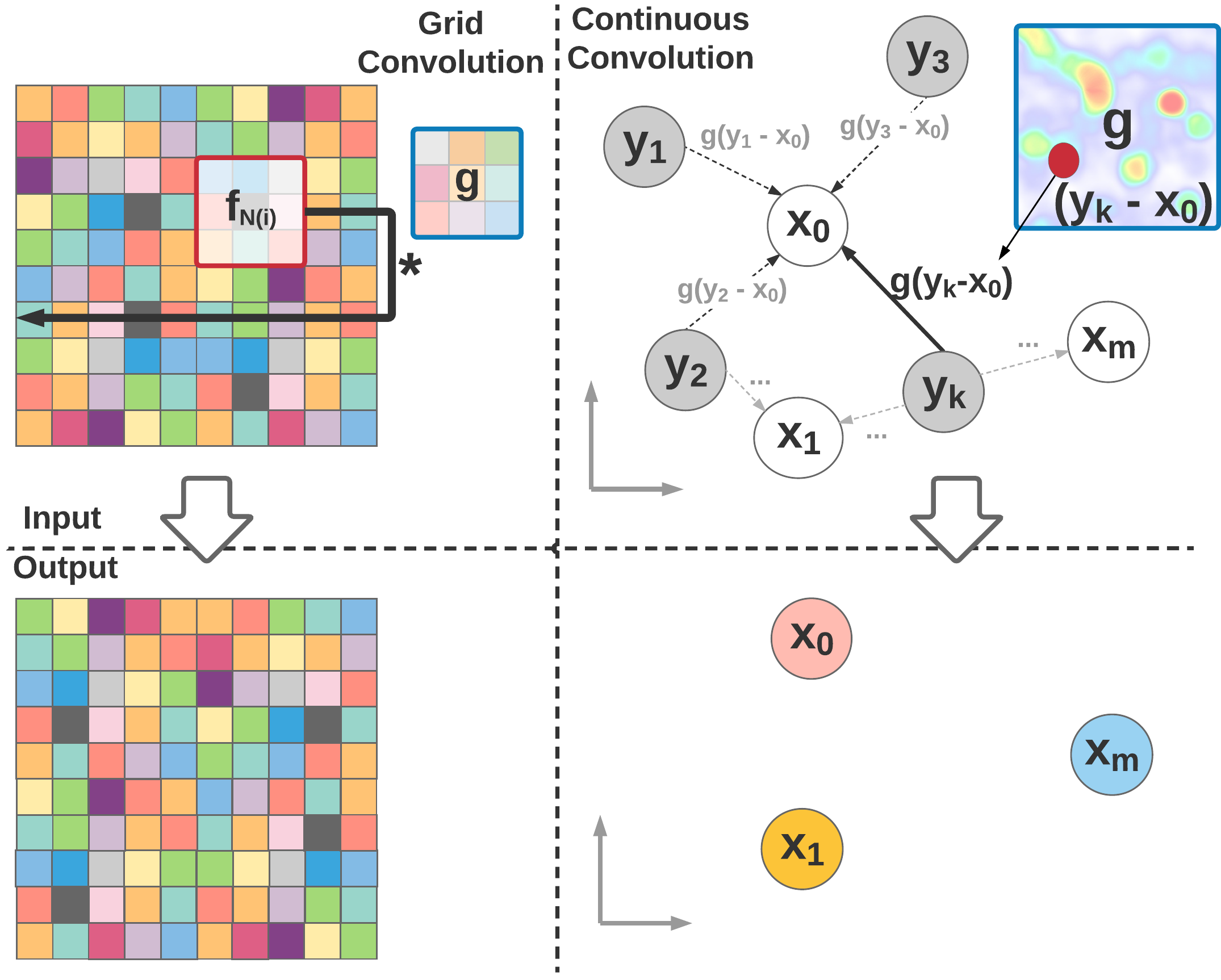}
  \vspace{-3mm}
  \caption{\simon{Unlike grid convolution, parametric continuous convolution uses kernel functions that are defined for arbitrary points in the continuous support domain. As a result, it is possible to output features at points not seen in the input.}}

  \label{fig:idea}
\end{figure}

\paragraph{Other Approaches:} Edge-conditioned filter networks \cite{ecc} use a weighting network to communicate between adjacent nodes on the graph \cite{dfn} conditioned on edge labels, which is primarily formulated as relative point locations. In contrast, our approach 
\simon{is not constrained to} a fixed graph structure, and 
\simon{has the flexibility} to output features at  arbitrary points over the continuous domain.  In a concurrent work,  \cite{schnet} uses similar parametric function form $f(\bx_i - \bx_j)$ to aggregate information \shenlong{between points}. However,  they only use shallow isotropic gaussian kernels to represent the weights, while we use expressive deep networks to parameterize the continuous filters.

\begin{figure*}
  \centering
  \includegraphics[width=0.95\linewidth]{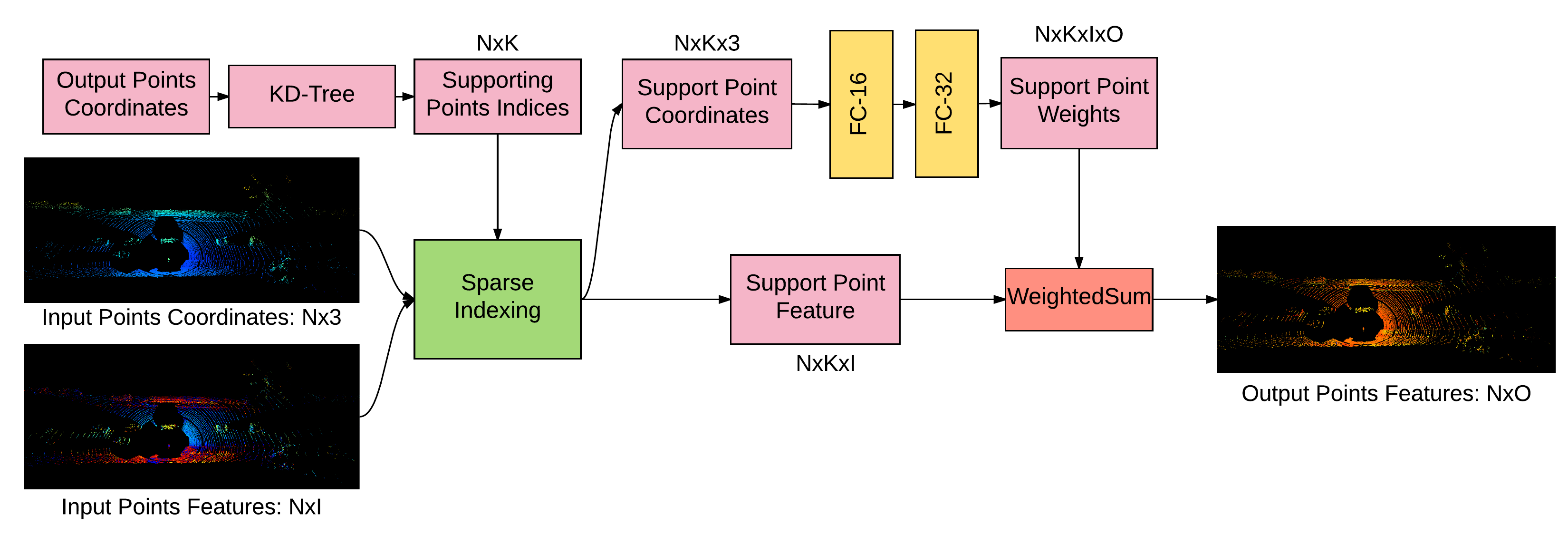}
  \vspace{-3mm}
  \caption{Detailed Computation Block for the Parametric Continuous Convolution Layer.} 
  \label{fig:layer}
\end{figure*}

\section{Deep Parametric Continuous CNNs}

\subsection{Parametric Continuous Convolutions}

Standard CNNs  use discrete convolutions (i.e., convolutions defined over discrete domain) as basic operations. 
\[
h[n] = (f \ast g)[n] = \sum_{m = -M}^M f[n - m] g[m] 
\] 
where $f: \mathcal{G} \rightarrow \bbR $ and $g: \mathcal{S} \rightarrow \bbR$ are functions defined over the support domain of finite integer set: $\mathcal{G} = \mathcal{Z}^D$ and $ \mathcal{S} = \{ -M, -M + 1, ..., M - 1, M\}^D$ respectively. 

In contrast, continuous convolutions can be defined as 
\begin{equation}
\label{conti-conv}
h(\bx) = (f \ast g) (\bx) = \int_{-\infty}^{\infty} f(\by) g(\bx -\by) d\by
\end{equation}
where both the kernel $g: \mathcal{S} \rightarrow \bbR$ and the feature $f: \mathcal{G} \rightarrow \bbR $ are defined as continuous functions over the support domain $\mathcal{G}=\bbR^D$ and $\mathcal{S}=\bbR^D$ respectively. %

Continuous convolutions  require the integration in Eq. \eqref{conti-conv}  to be analytically tractable.  %
Unfortunately, this is not  possible for real-world applications, where the input features are complicated and non-parametric, and the observations are sparse points sampled over the continuous domain. 

Motivated by monte-carlo integration \cite{mc} we derive our continuous convolution operator.  
 In particular, given continuous functions $f$ and $g$ with a finite number of input points ${\by_i}$ sampled from the domain, the convolution at an arbitrary point $\bx$ can be approximated as: 
\[
\label{sampled-conti-conv}
h(\bx) = \int_{-\infty}^{\infty} f(\by) g(\bx-\by) d\by \approx \sum_i^N \frac{1}{N} f(\by_i) g(\bx - \by_i)
\]

The next challenge we need to solve is %
\simon{constructing} the continuous convolutional kernel function $g$. 
Conventional 2D and 3D discrete convolution kernels are parameterized in a way that each point in the support domain is assigned a value (\ie the kernel weight). Such a parameterization is 
\simon{infeasible} for continuous convolutions, \simon{since} the kernel function $g$ is defined over an infinite number of points (i.e., has infinite support). %
Instead, in this paper we propose to use parametric continuous functions to model $g$. 
We \simon{name} our approach  {\it Parametric Continuous Convolutions}.
In particular, we use a multi-layer perceptron (MLP) \simon{as the approximator}. With reference to the universal approximation theorem of \cite{mlp-theory}, MLPs are expressive and capable of approximating continuous functions over $\bbR^n$. %
Thus we define:
\[
g(\mathbf{z}; \theta) = MLP(\mathbf{z}; \theta)
\]
The kernel function $g(\mathbf{z}; \theta): \mathbb{R}^D \rightarrow \mathbb{R}$ spans the full continuous support domain while remaining parametrizable  by a finite number of parameters. 
Note that other choices such as polynomials are possible, however low-order polynomials are not expressive, whereas  learning  high-order polynomials can be numerically 
\simon{unstable} for back-propagation. 

\begin{figure*}
  \centering
  \includegraphics[width=0.95\linewidth]{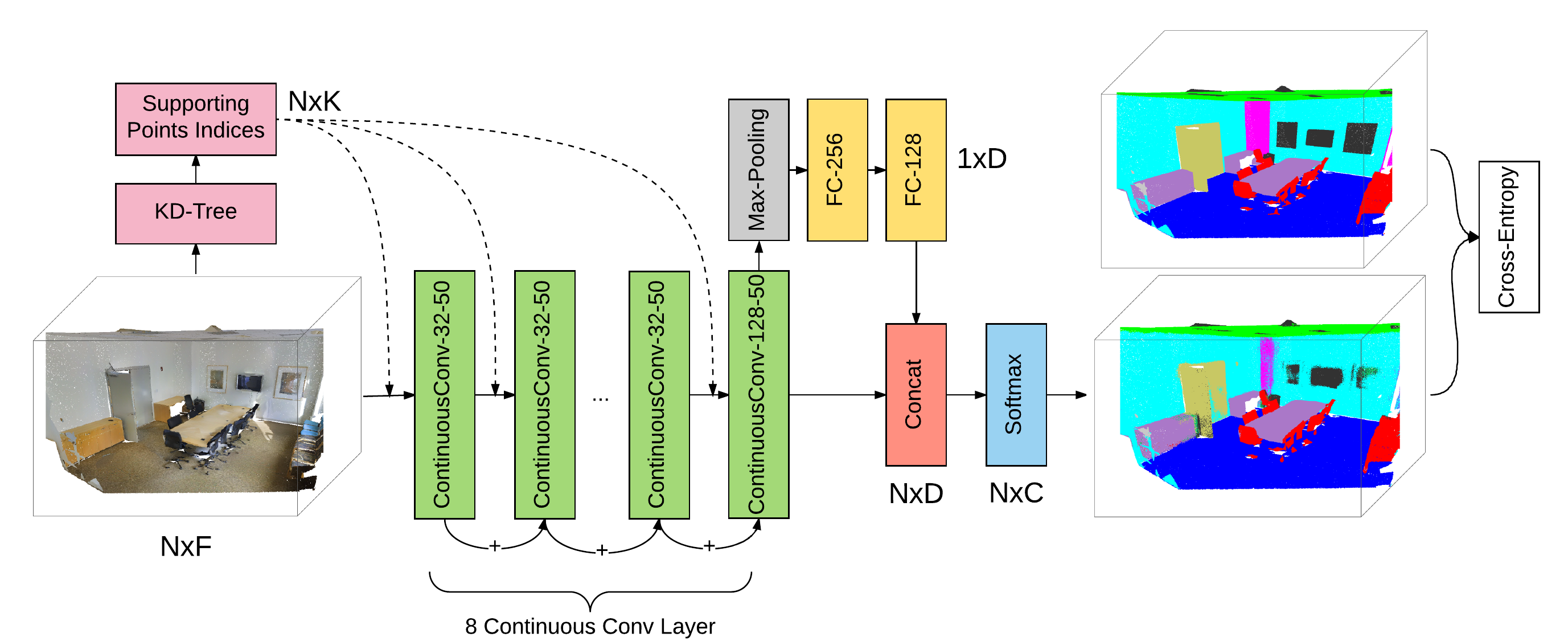}
  \vspace{-3mm}
  \caption{Architecture of \shenlong{the Deep Parametric Continuous CNNs for Semantic Labeling Task.}} 
  \label{fig:network}
\end{figure*}

\subsection{From Convolutions to Deep Networks}

In this section, we first design a new convolution layer based on the  parametric continuous convolutions derived in the previous subsection. We then propose a deep learning architecture using this new convolution layer.

\paragraph{Parametric Continuous Convolution Layer: }
Note that, unlike standard discrete convolutions which  are conducted over the same point set, the input and output points of our parametric continuous  convolution layer can be different. This is important for many practical applications, \simon{where we want to make dense predictions based on partial observations.} 
Furthermore, this allow us to abstract information from redundant input points (i.e., pooling). 
As a consequence, the input of each convolution  layer contains three parts: the input feature vector $\mathcal{F} = \{ \mathbf{f}_{\mathrm{in}, j} \in \mathbb{R}^F\}$,  the associated locations  
in the support domain $\mathcal{S} = \{ \by_j \}$, as well as the output domain locations %
$\mathcal{O} = \{ \bx_i \}$.   %
For each layer, we first  evaluate the kernel function 
$g_{d, k}(\by_i - \bx_j; \theta)$  for all $\bx_j \in \cS$ and   all $ \by_i \in \mathcal{O}$, given the  parameters $\theta$. Each element of the output feature vector is then computed as:
\[
{h}_{k, i} = \sum_d^{F} \sum_j^{N} g_{d, k}(\by_i - \bx_j) f_{d, j}
\]
Let $N$ be the number of input points,  $M$ be the number of output points, and $D$  the dimensionality of the support domain. Let 
$F$ and $O$ be predefined input  and output feature dimensions respectively. Note that \simon{these} are hyperparameters of the continuous convolution layer analogous to \shenlong{input and output feature dimensions in standard grid convolution layers.} 
 \figref{fig:idea} 
 \simon{depicts} our parametric continuous convolutions in comparison with conventional grid convolution. \shenlong{Two major differences are highlighted: 1) the kernel function is continuous given the relative location in support domain; 2) the input/ouput points could be any points in the continuous domain as well and can be different. } %

\begin{figure*}
\setlength\tabcolsep{0.5pt} %
\renewcommand{\arraystretch}{0.8}
\begin{tabular}{cccccc}
  \includegraphics[width=0.16\linewidth]{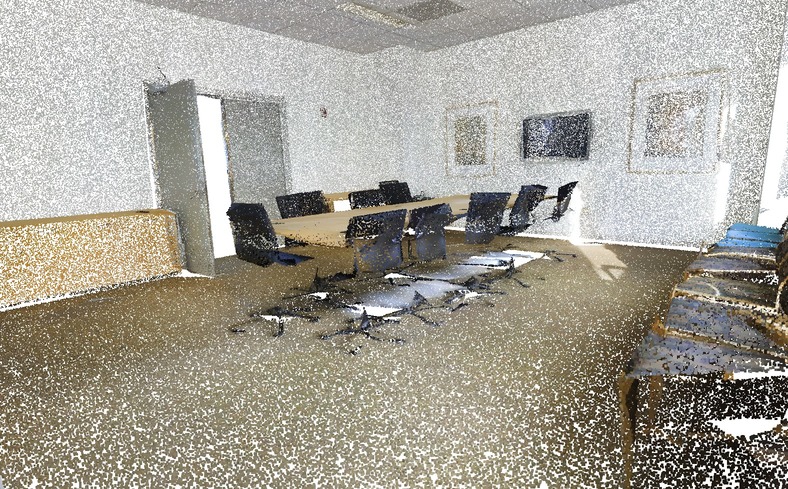} &   \includegraphics[width=0.16\linewidth]{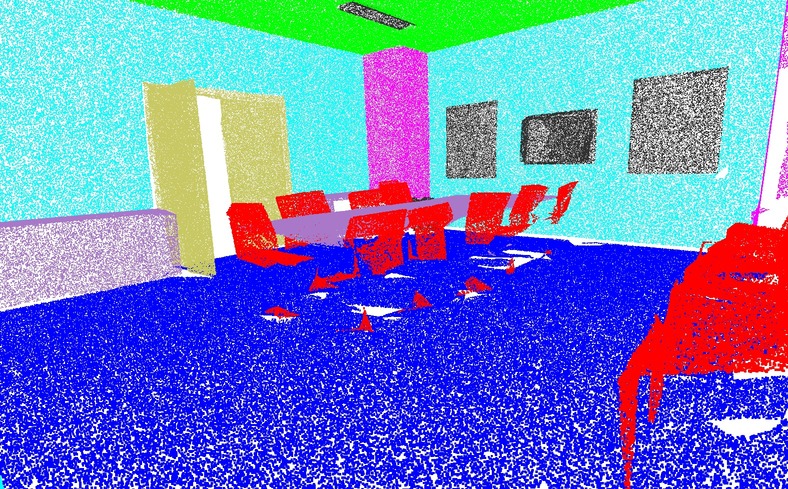} & \includegraphics[width=0.16\linewidth] {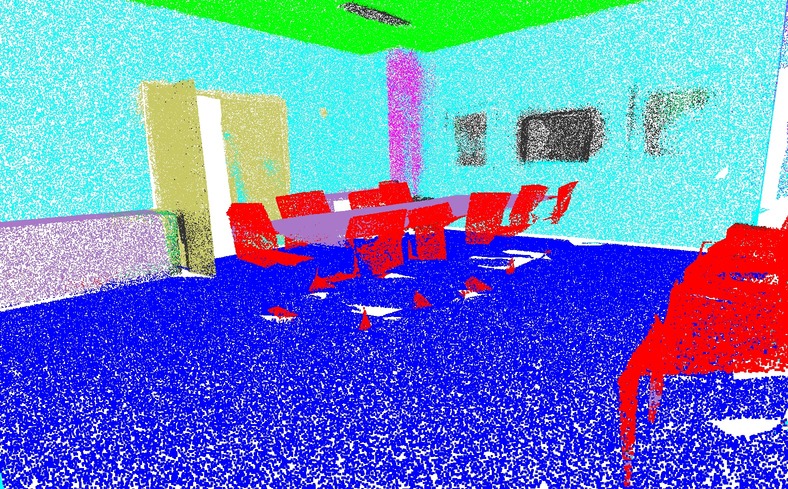} &  
   \includegraphics[width=0.16\linewidth]{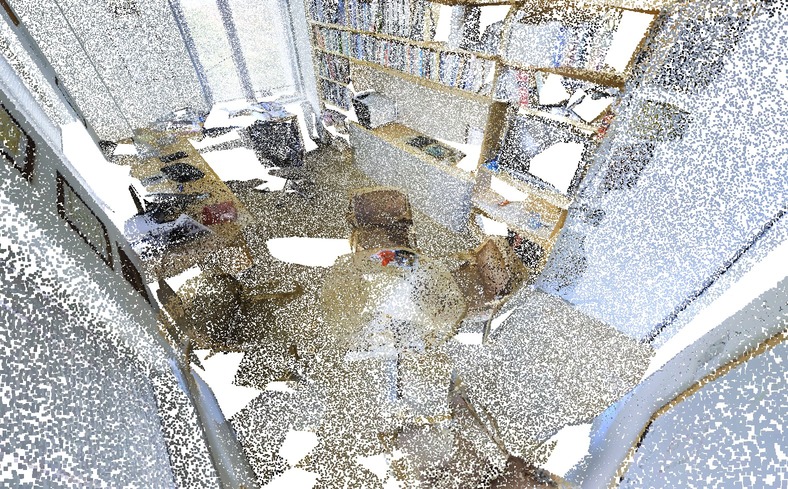} &   \includegraphics[width=0.16\linewidth]{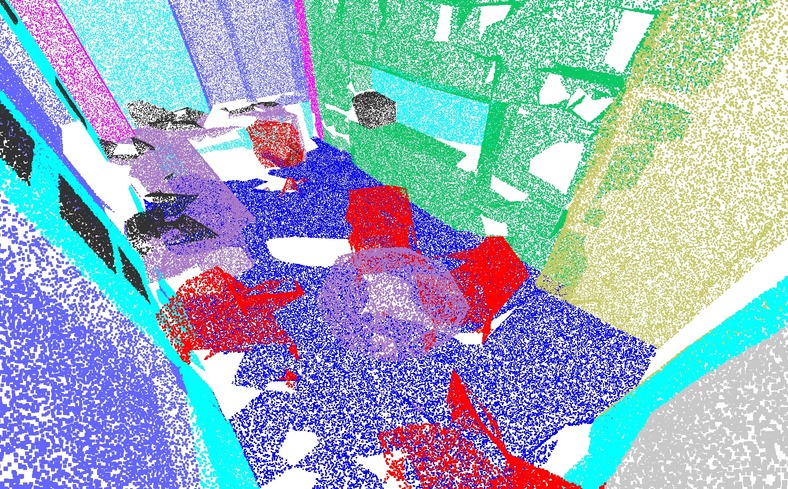} & \includegraphics[width=0.16\linewidth] {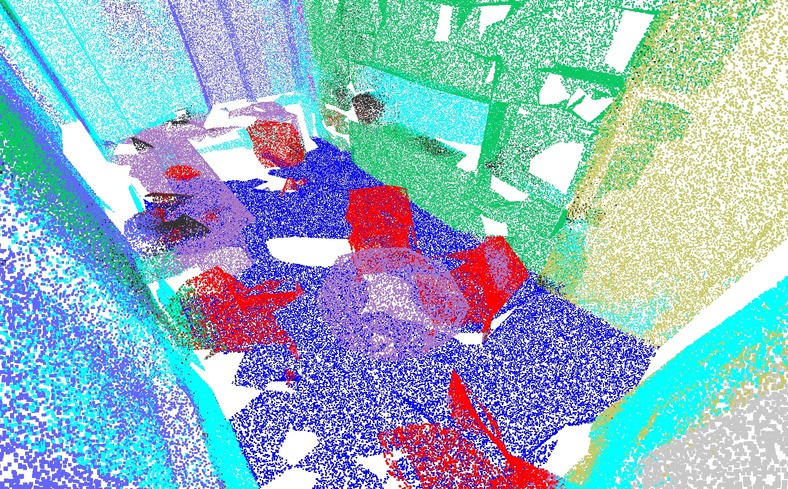} \\  
      \includegraphics[width=0.16\linewidth]{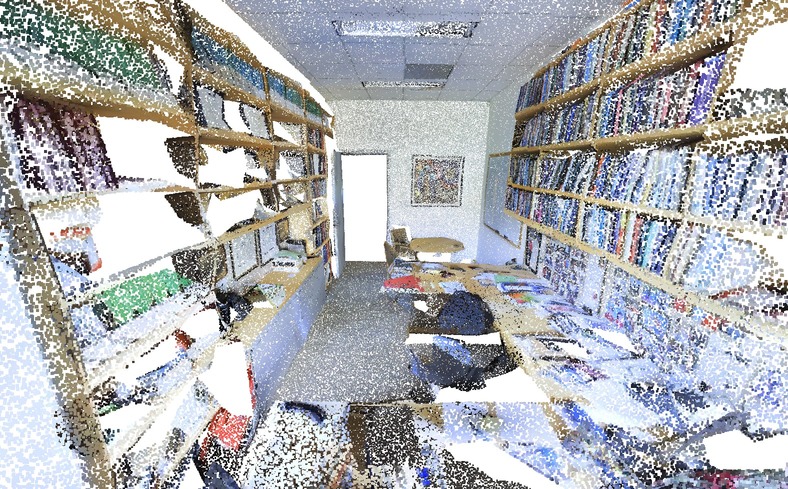} &   \includegraphics[width=0.16\linewidth]{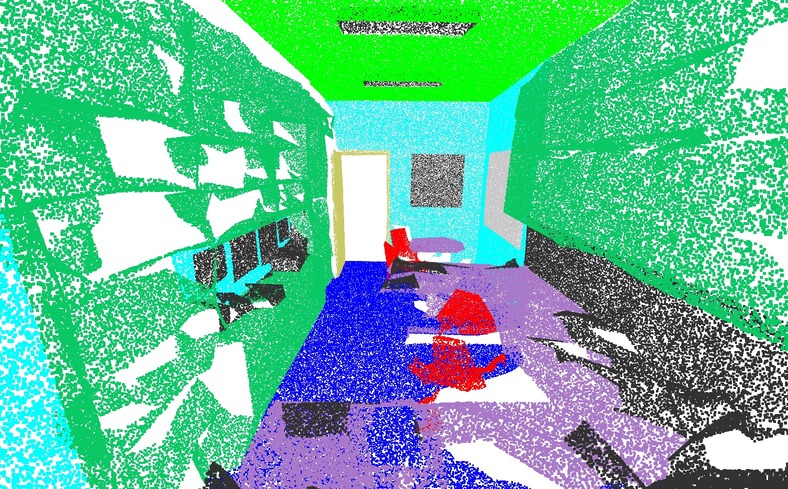} & \includegraphics[width=0.16\linewidth] {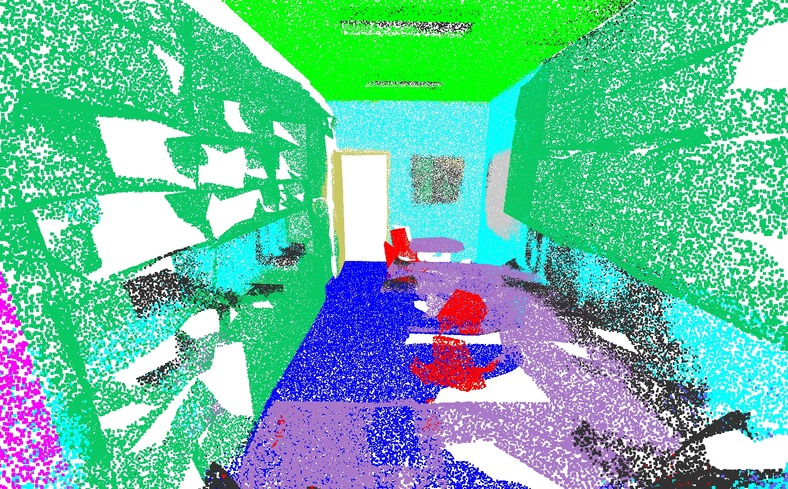} &  
    \includegraphics[width=0.16\linewidth]{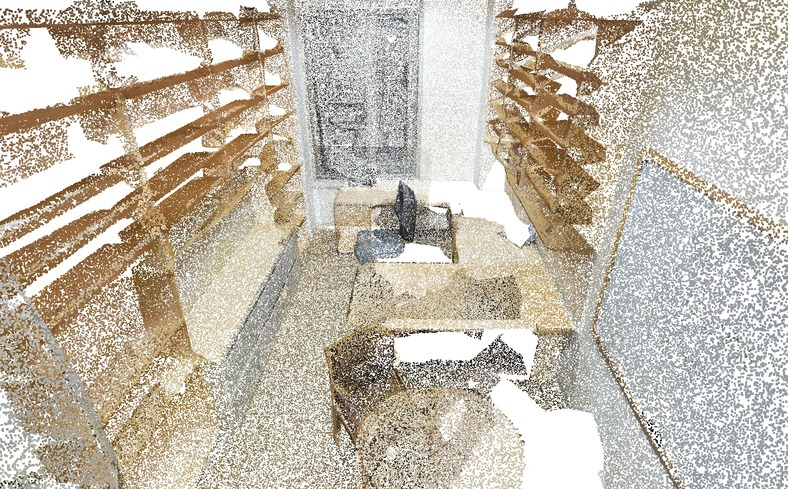} &   \includegraphics[width=0.16\linewidth]{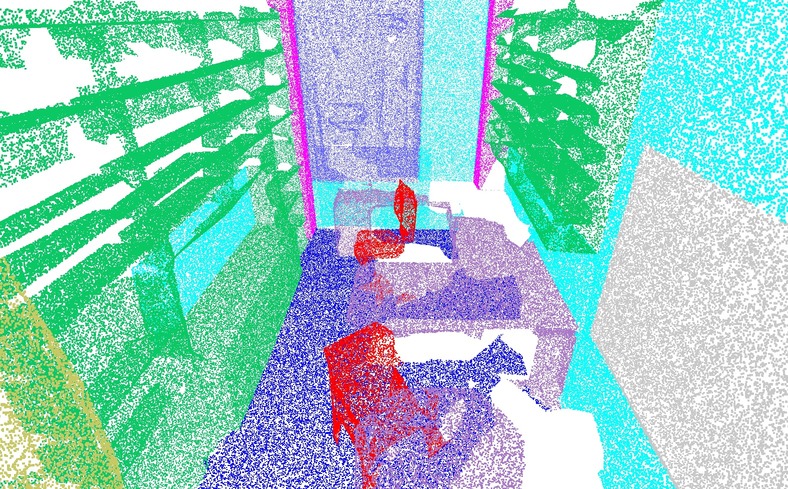} & \includegraphics[width=0.16\linewidth] {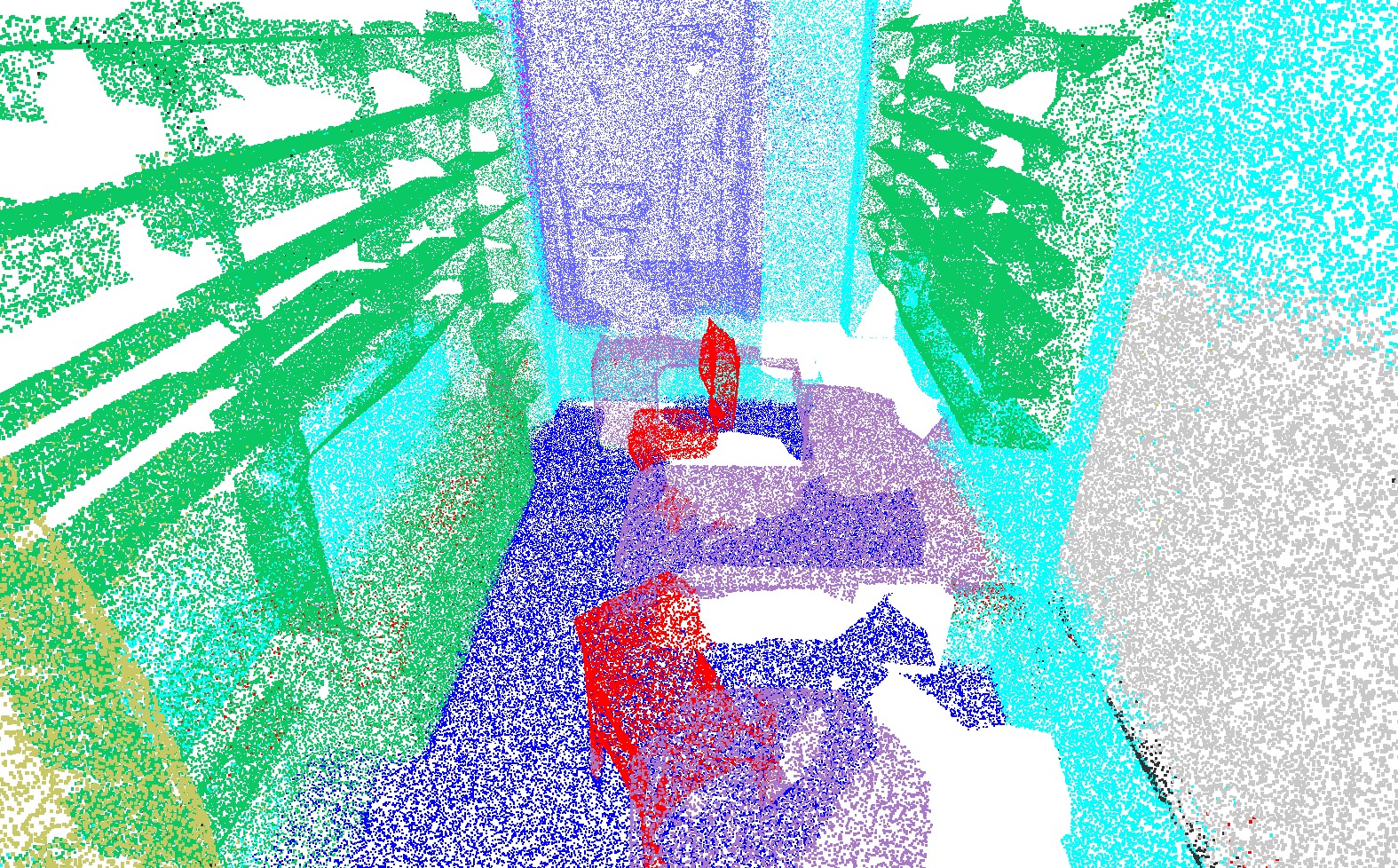} \\
          \includegraphics[width=0.16\linewidth]{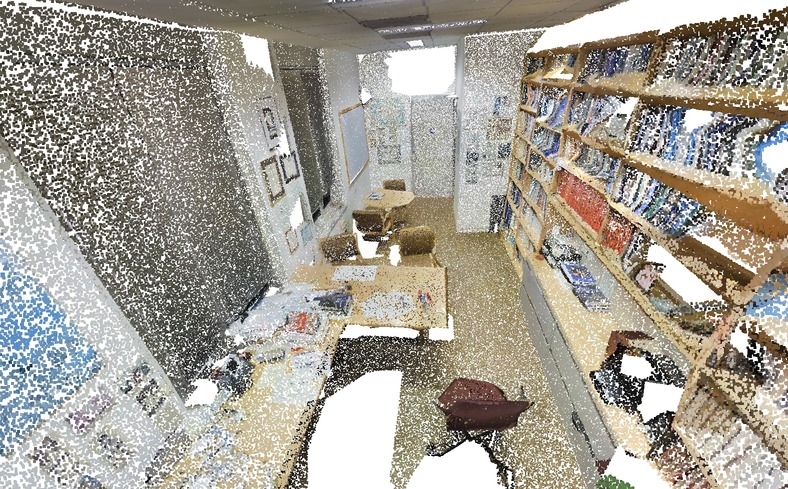} &   \includegraphics[width=0.16\linewidth]{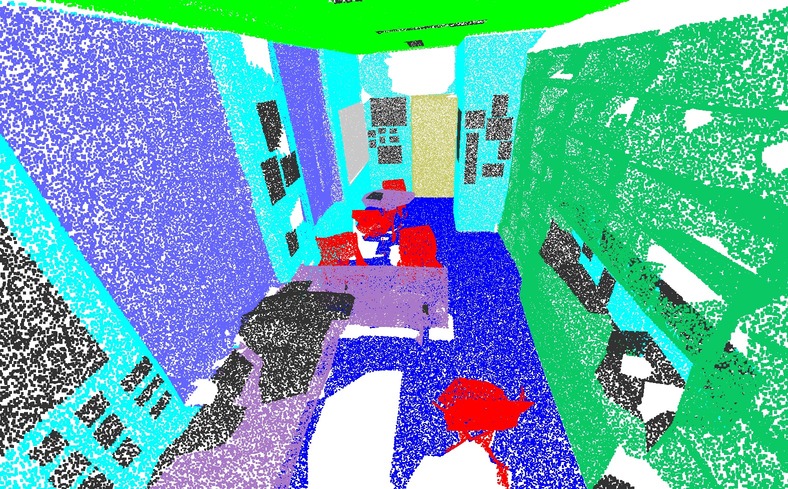} & \includegraphics[width=0.16\linewidth] {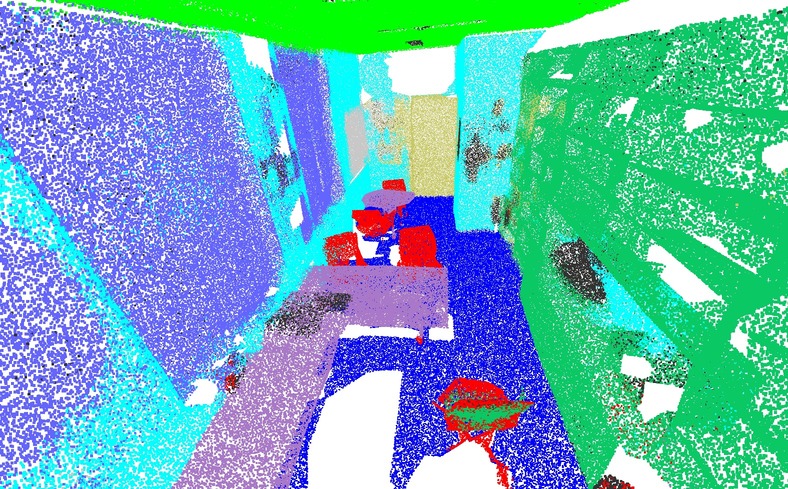} &  
    \includegraphics[width=0.16\linewidth]{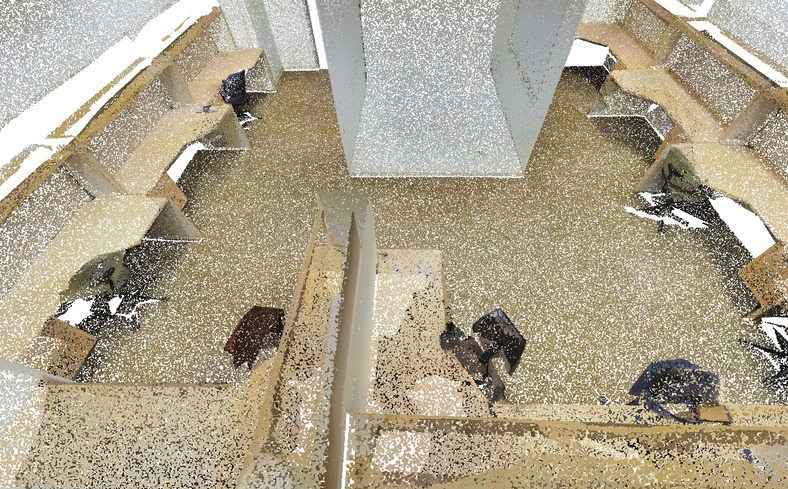} &   \includegraphics[width=0.16\linewidth]{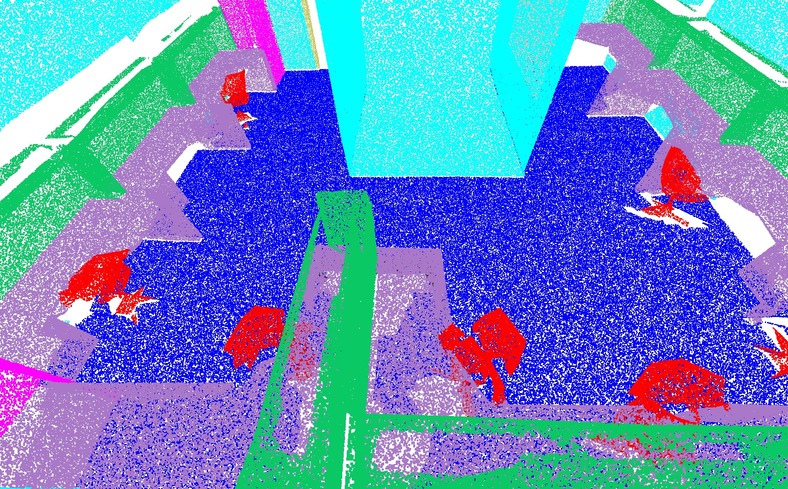} & \includegraphics[width=0.16\linewidth] {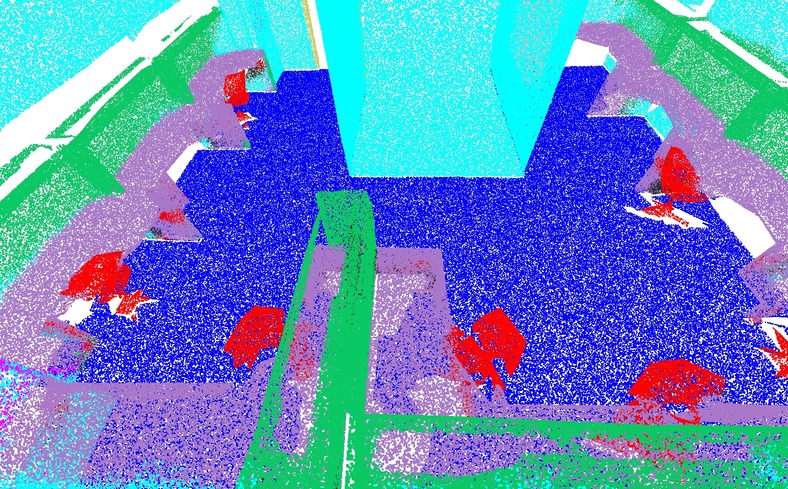} \\
          \includegraphics[width=0.16\linewidth]{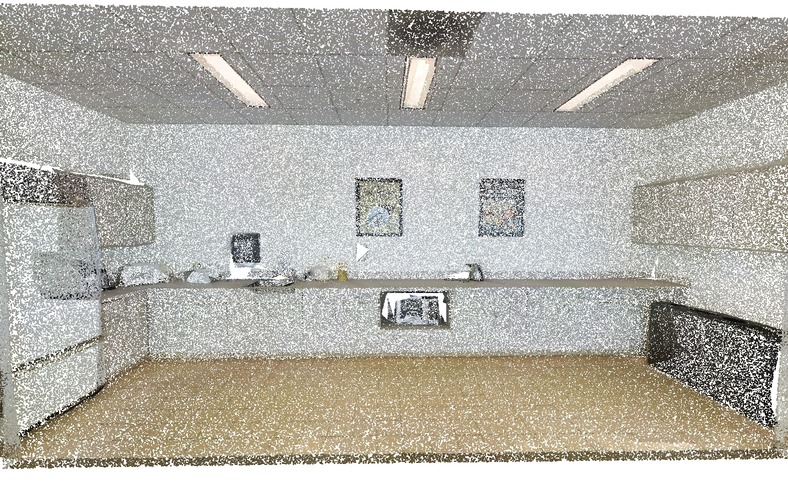} &   \includegraphics[width=0.16\linewidth]{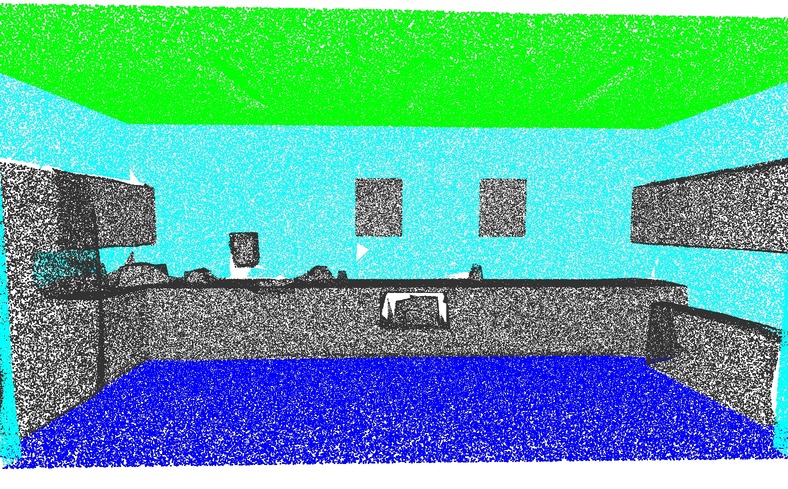} & \includegraphics[width=0.16\linewidth] {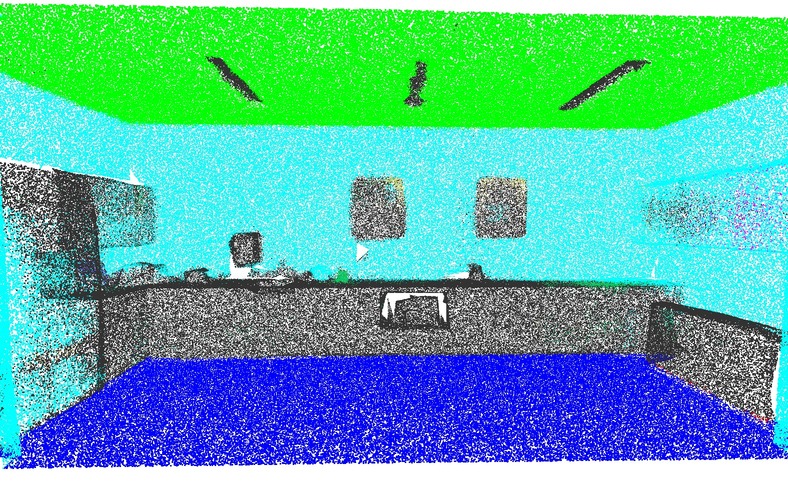} &  
          \includegraphics[width=0.16\linewidth]{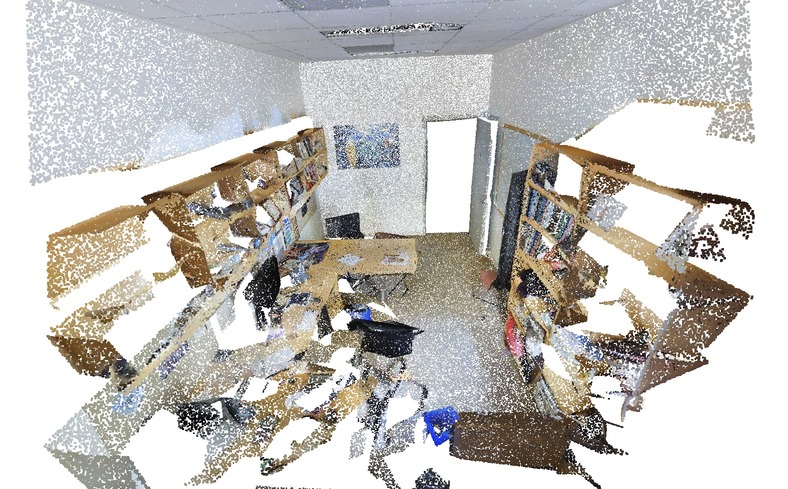} &   \includegraphics[width=0.16\linewidth]{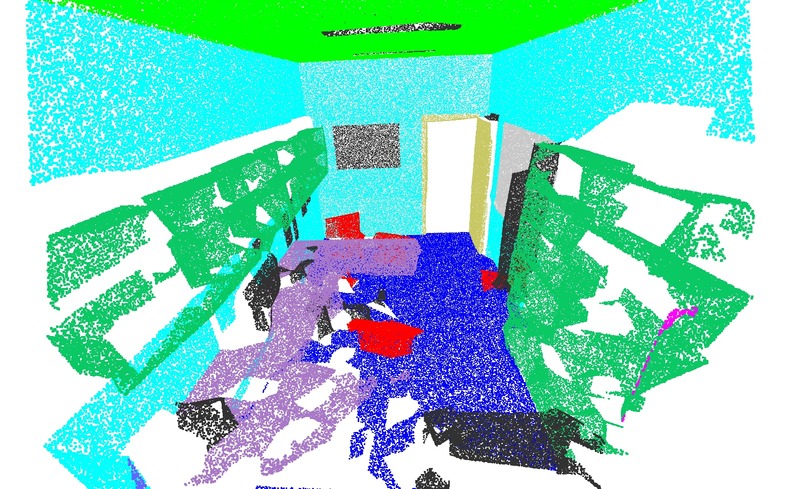} & \includegraphics[width=0.16\linewidth] {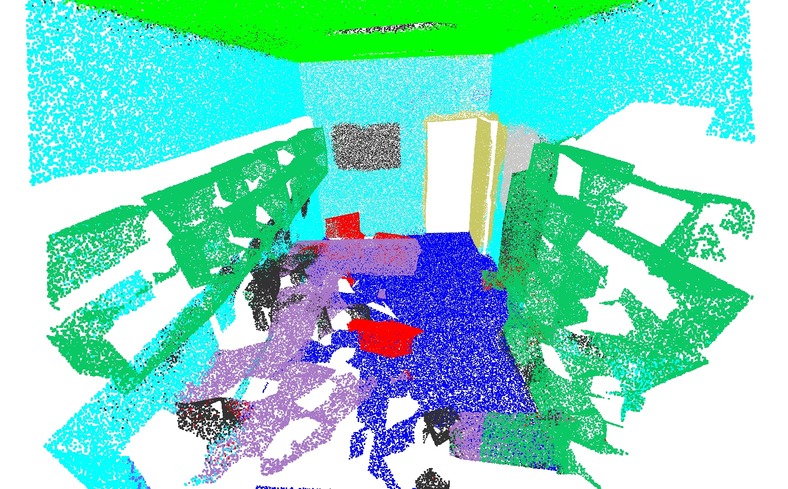} \\
              \includegraphics[width=0.16\linewidth]{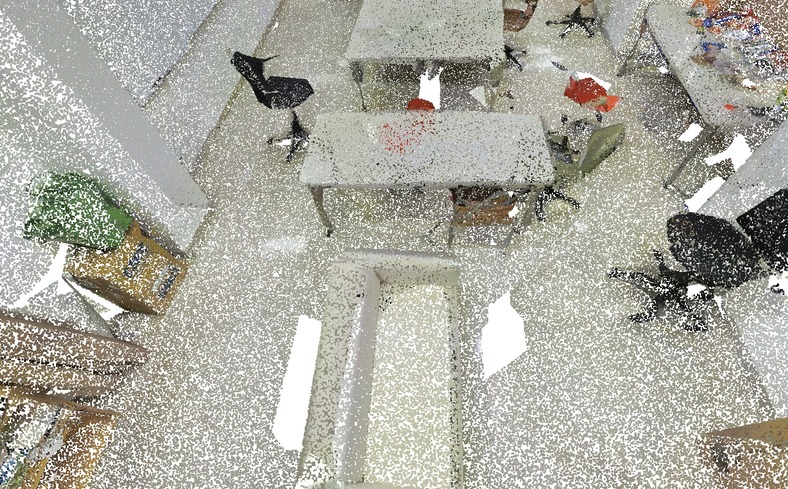} &   \includegraphics[width=0.16\linewidth]{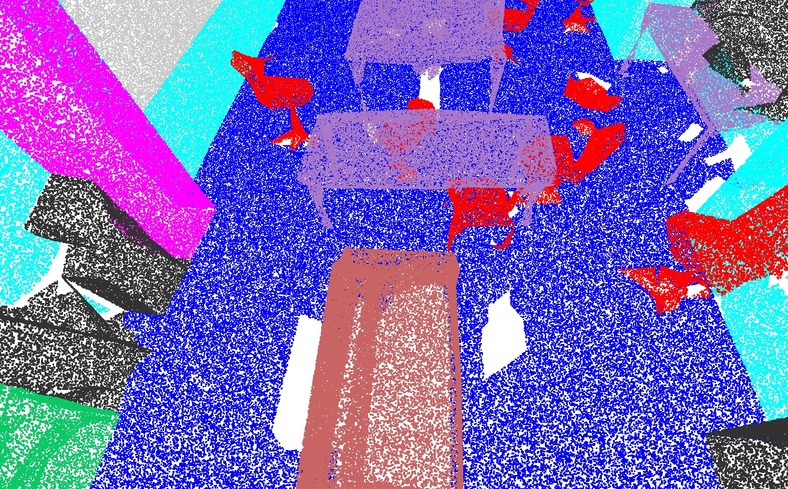} & \includegraphics[width=0.16\linewidth] {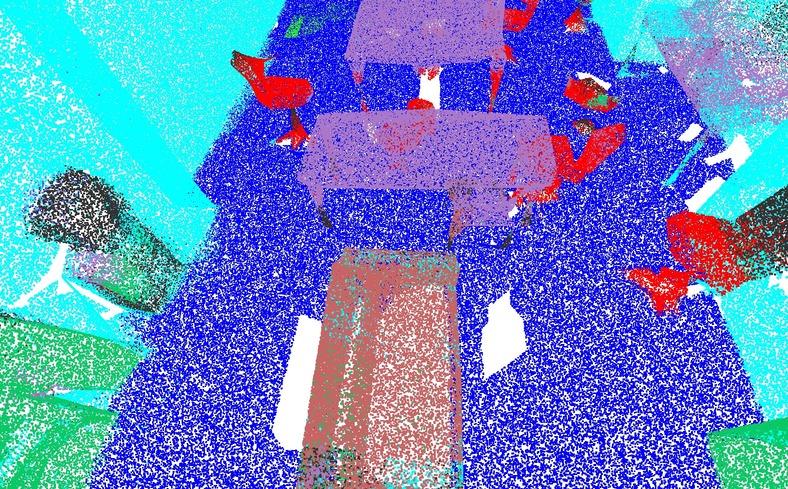} &  
    \includegraphics[width=0.16\linewidth]{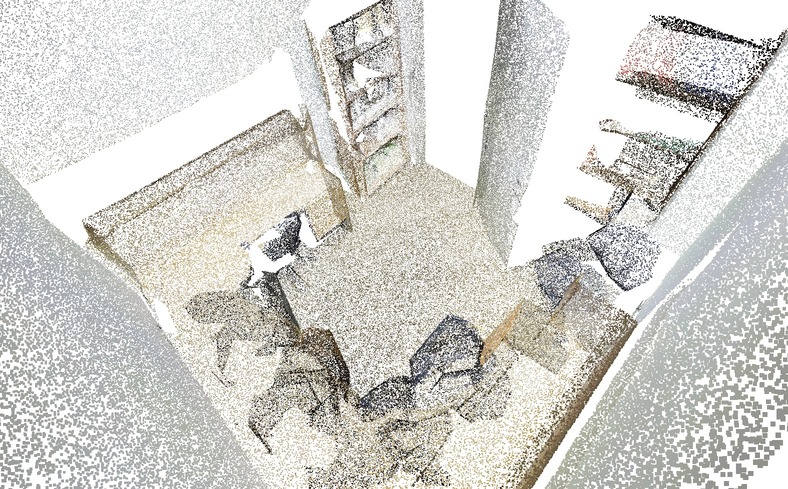} &   \includegraphics[width=0.16\linewidth]{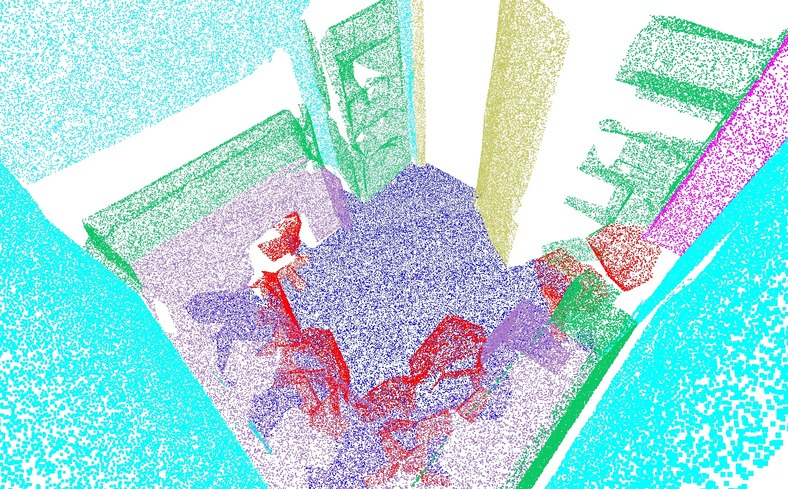} & \includegraphics[width=0.16\linewidth] {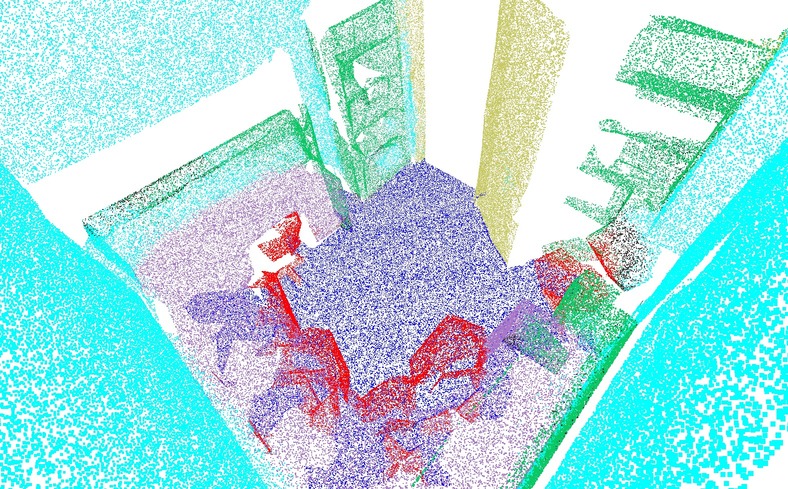} \\
              \includegraphics[width=0.16\linewidth]{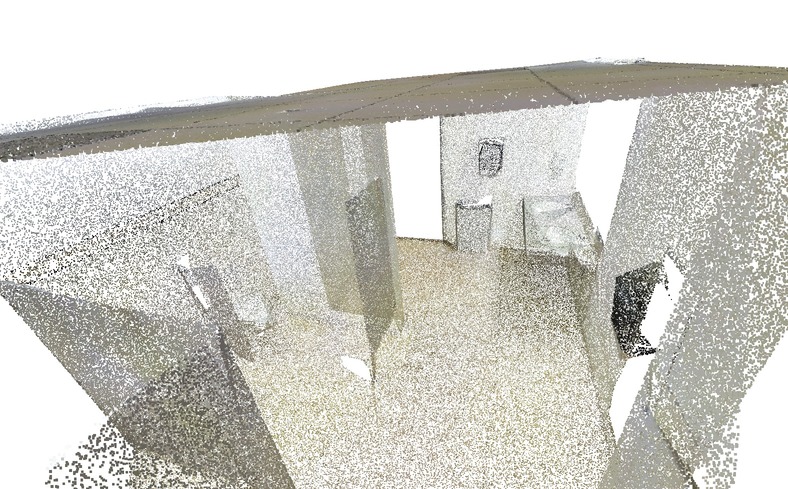} &   \includegraphics[width=0.16\linewidth]{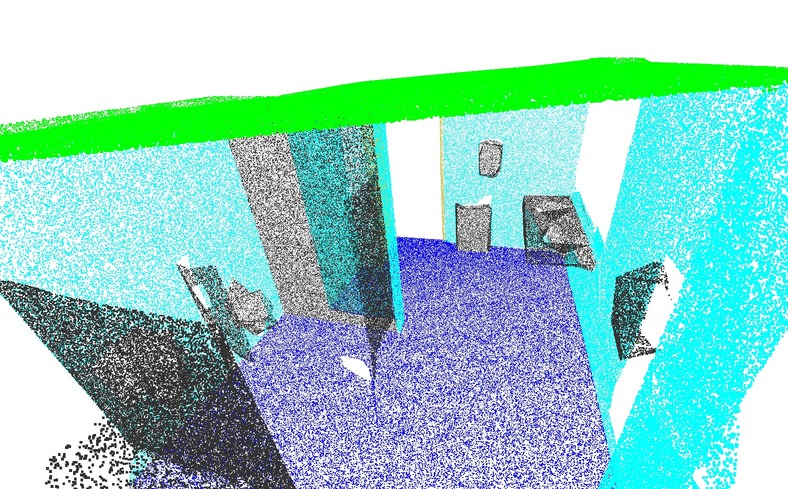} & \includegraphics[width=0.16\linewidth] {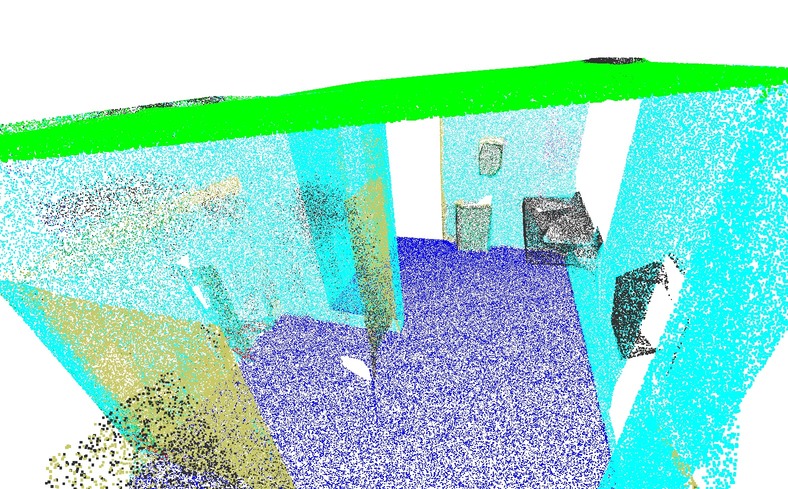} &
        \includegraphics[width=0.16\linewidth]{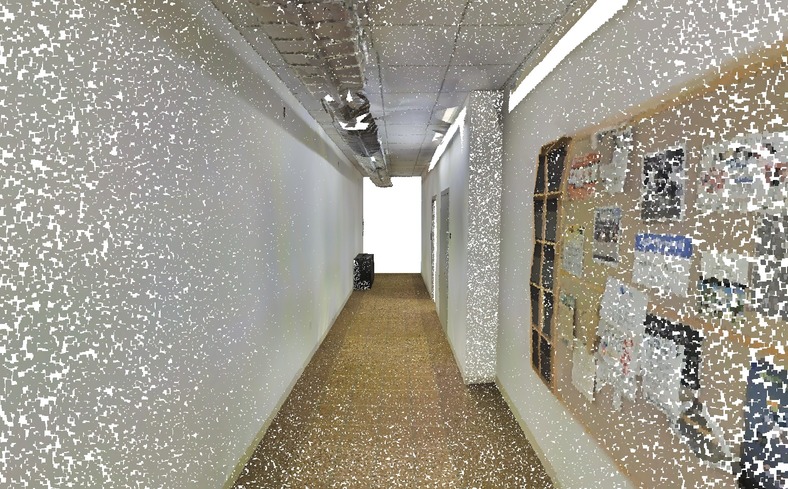} &   \includegraphics[width=0.16\linewidth]{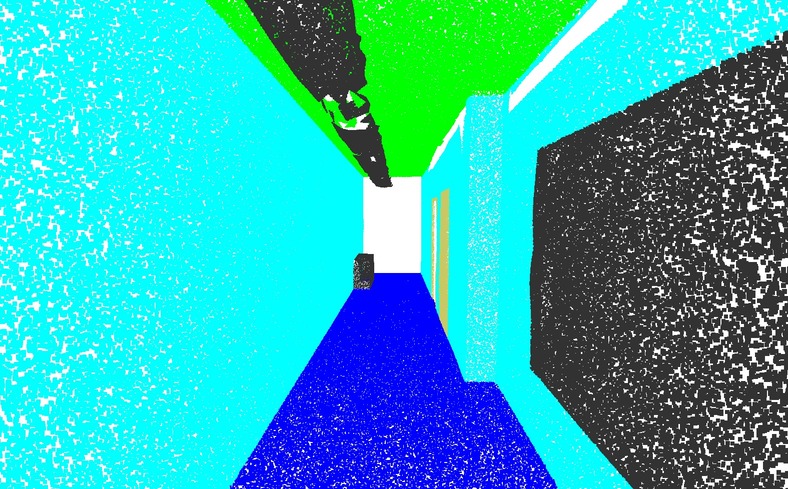} & \includegraphics[width=0.16\linewidth] {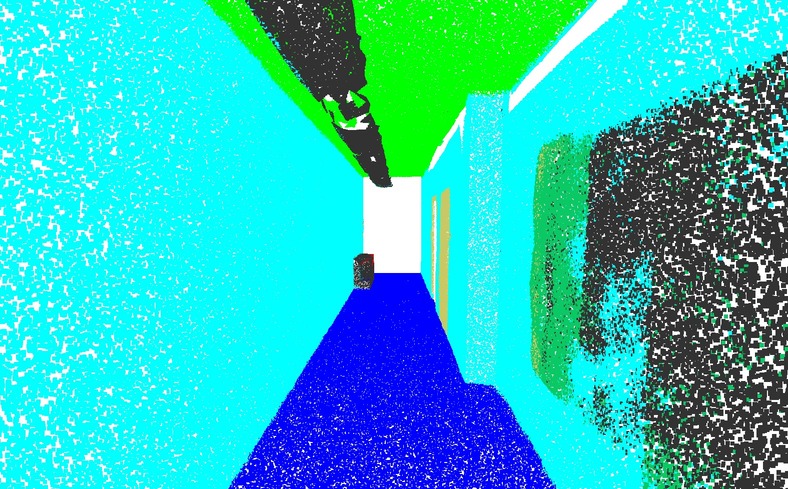} \\
Input & Ground Truth & Ours PCCN & Input & Ground Truth & Ours PCCN
\end{tabular}
\vspace{-3mm}
\caption{Semenatic Segmentation Results on Stanford Indoor3D Dataset}
\label{fig-indoor3d}
\end{figure*}

\paragraph{Deep Parametric Continuous CNNs:} Using the parametric continuous convolution layers as building blocks, we can construct a new family of deep networks which operates on  unstructured data  defined in a topological group under addition. In the following {discussions}, we will focus on  multi-diumensional euclidean space, and note that this is  a special case. 
The network takes the input features and \shenlong{their} associated positions in the support domain as input. %
Then the hidden representations are generated %
from successive parametric continuous  convolution layers. 
Following standard CNN \simon{architectures}, we can add batch normalization, non-linearities  and residual connections between layers. 
Pooling can also be employed over the support domain to aggregate  information. \simon{In practice, we find adding residual connection between parametric continuous convolution layers is critical to help convergence.} 
Please refer to  \figref{fig:layer} for an example of the computation graph of a single layer, and to \figref{fig:network} for an example of the network architecture employed for  our \shenlong{indoor} %
semantic segmentation task. 

\paragraph{Learning:} All of our building blocks are differentiable, thus our networks can be learned through back-prop:
\[
\frac{\partial h}{\partial \theta} = \frac{\partial h}{\partial g} \cdot \frac{\partial g}{\partial \theta}  = \sum_d^F \sum_j^N f_{d, j} \cdot \frac{\partial g}{\partial \theta}
\]

\begin{figure*}
	\setlength\tabcolsep{0.5pt} %
	\renewcommand{\arraystretch}{0.8}
	\begin{tabular}{cccc}
		\adjincludegraphics[width=0.25\linewidth, trim={{.2\width} {.2\height} {.2\width} {.2\height}}, clip]{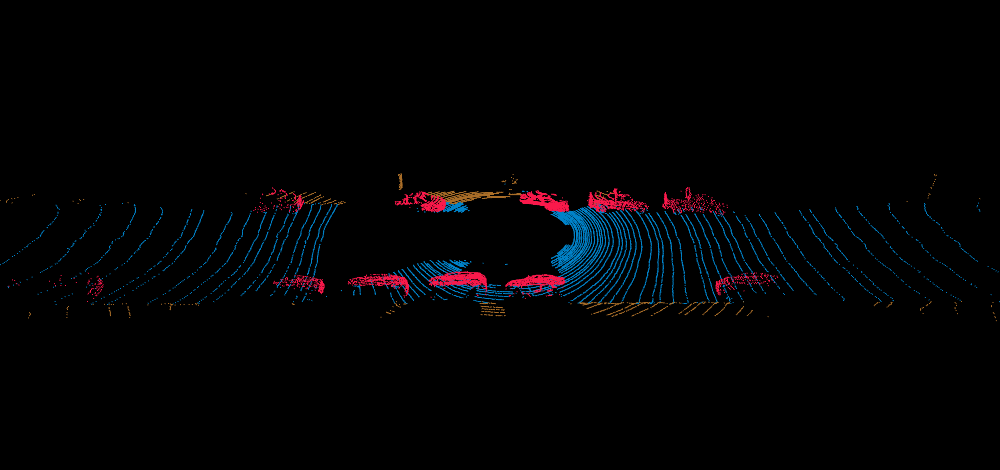} &  
		\adjincludegraphics[width=0.25\linewidth, trim={{.2\width} {.2\height} {.2\width} {.2\height}}, clip]{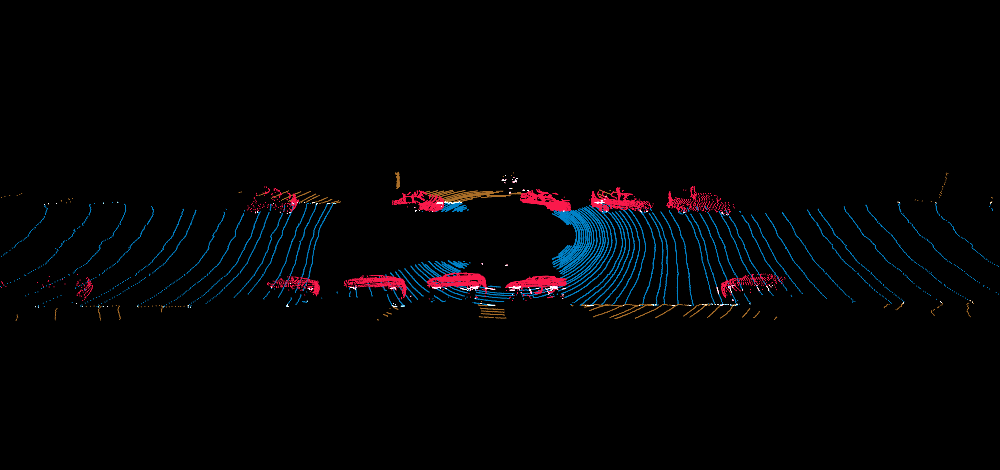} &
		\adjincludegraphics[width=0.25\linewidth, trim={{.2\width} {.2\height} {.2\width} {.2\height}}, clip]{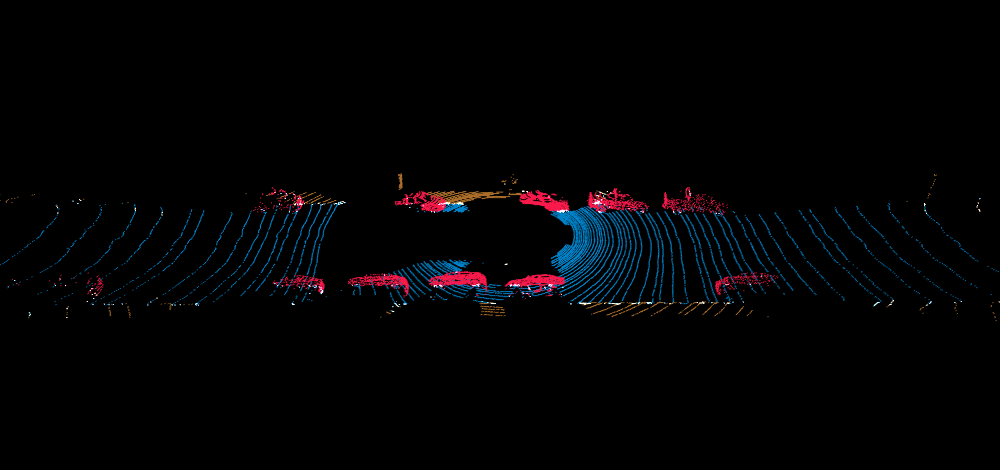} &
		\adjincludegraphics[width=0.25\linewidth, trim={{.2\width} {.2\height} {.2\width} {.2\height}}, clip]{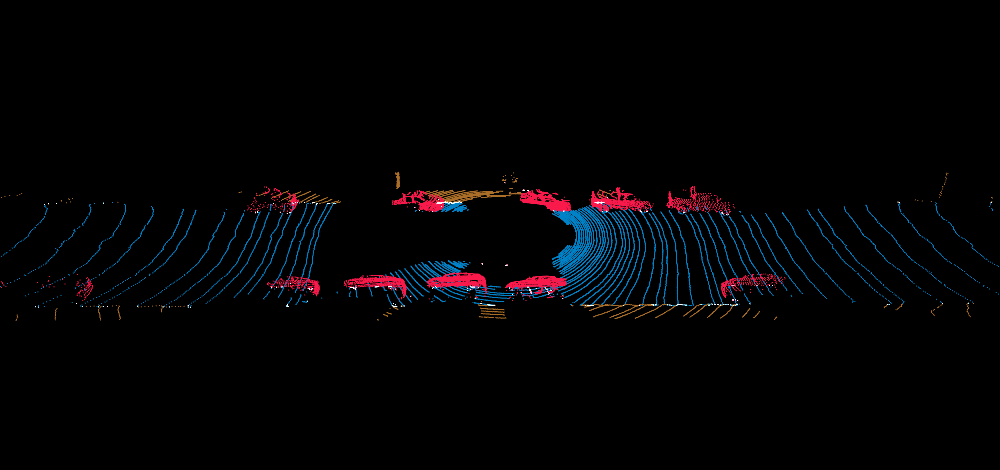} \\

		\adjincludegraphics[width=0.25\linewidth, trim={{.2\width} {.2\height} {.2\width} {.2\height}}, clip]{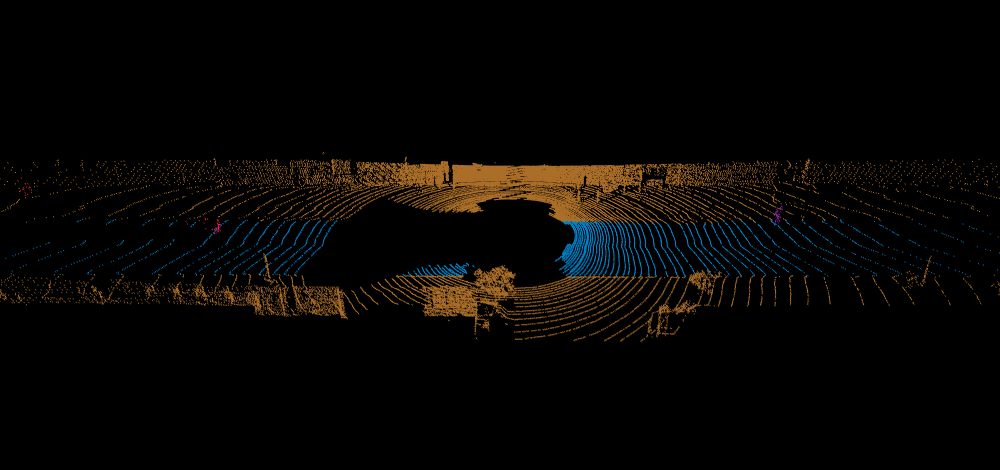} &  
		\adjincludegraphics[width=0.25\linewidth, trim={{.2\width} {.2\height} {.2\width} {.2\height}}, clip]{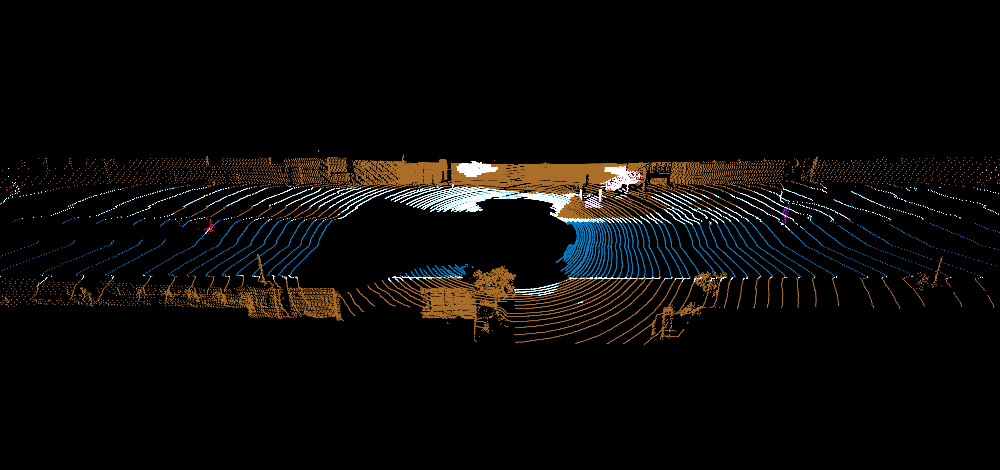} &
		\adjincludegraphics[width=0.25\linewidth, trim={{.2\width} {.2\height} {.2\width} {.2\height}}, clip]{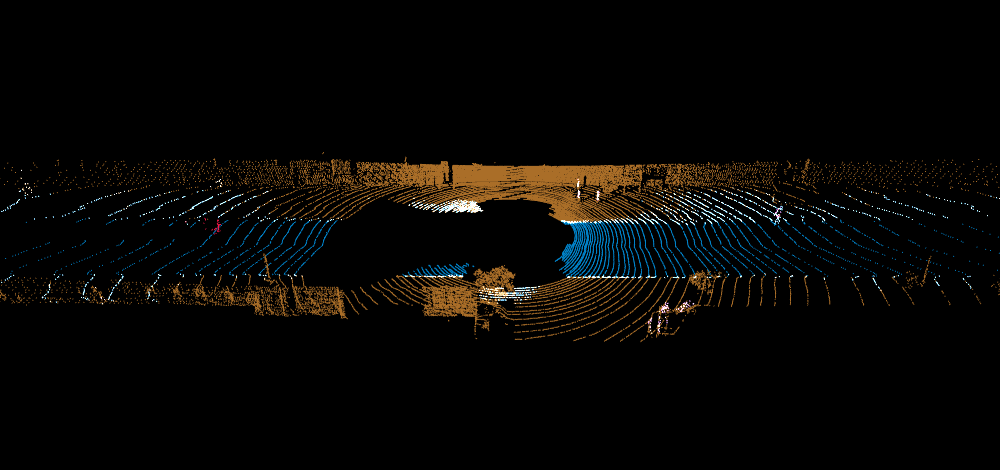} &
		\adjincludegraphics[width=0.25\linewidth, trim={{.2\width} {.2\height} {.2\width} {.2\height}}, clip]{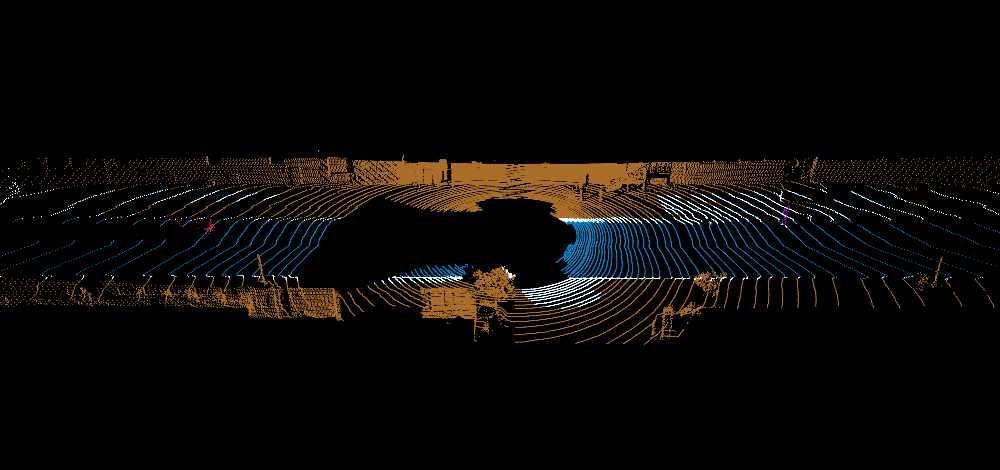} \\
		\adjincludegraphics[width=0.25\linewidth, trim={{.2\width} {.2\height} {.2\width} {.2\height}}, clip]{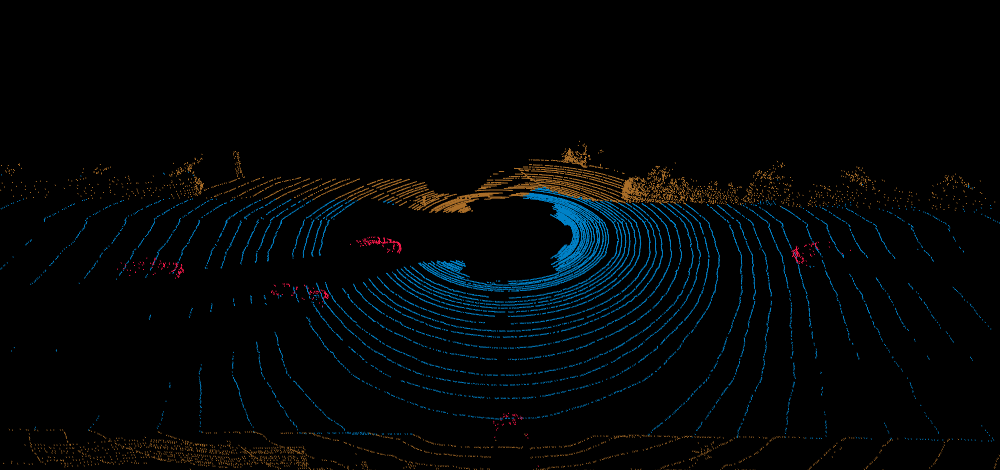} &  
		\adjincludegraphics[width=0.25\linewidth, trim={{.2\width} {.2\height} {.2\width} {.2\height}}, clip]{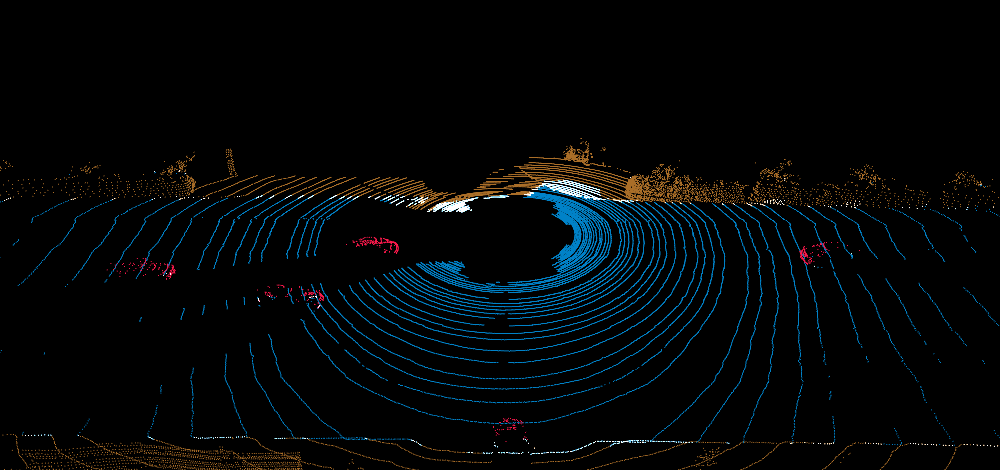} &
		\adjincludegraphics[width=0.25\linewidth, trim={{.2\width} {.2\height} {.2\width} {.2\height}}, clip]{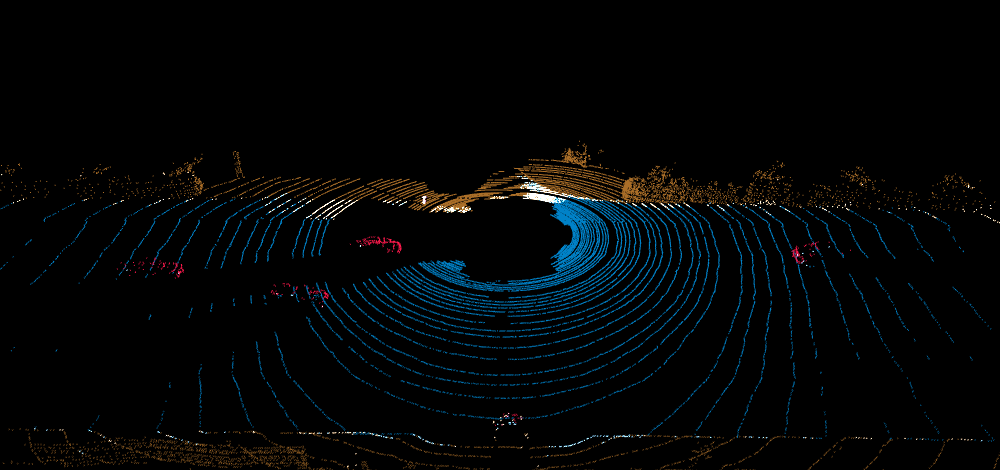} &
		\adjincludegraphics[width=0.25\linewidth, trim={{.2\width} {.2\height} {.2\width} {.2\height}}, clip]{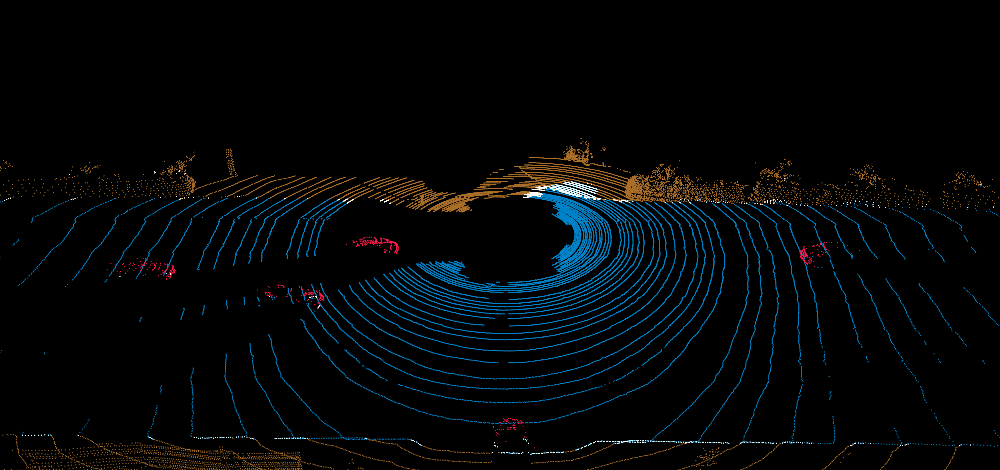} \\
		
		\adjincludegraphics[width=0.25\linewidth, trim={{.2\width} {.2\height} {.2\width} {.2\height}}, clip]{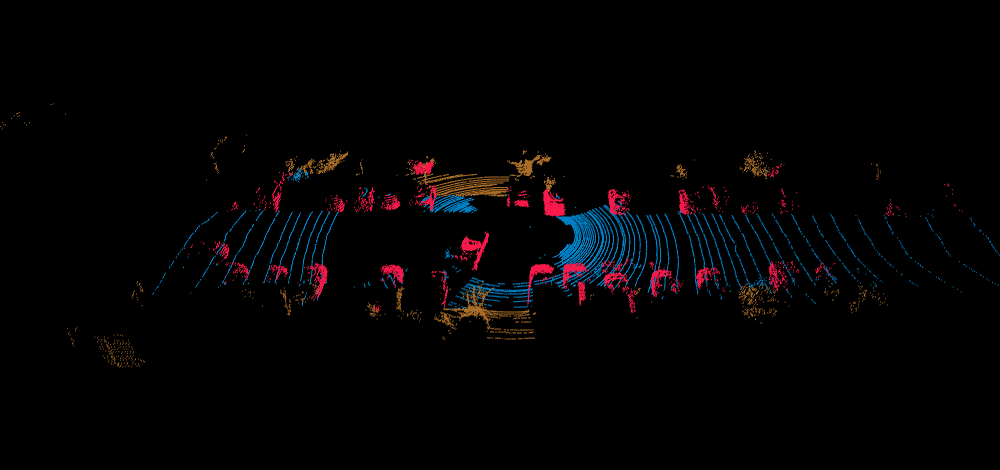} &   
		\adjincludegraphics[width=0.25\linewidth, trim={{.2\width} {.2\height} {.2\width} {.2\height}}, clip]{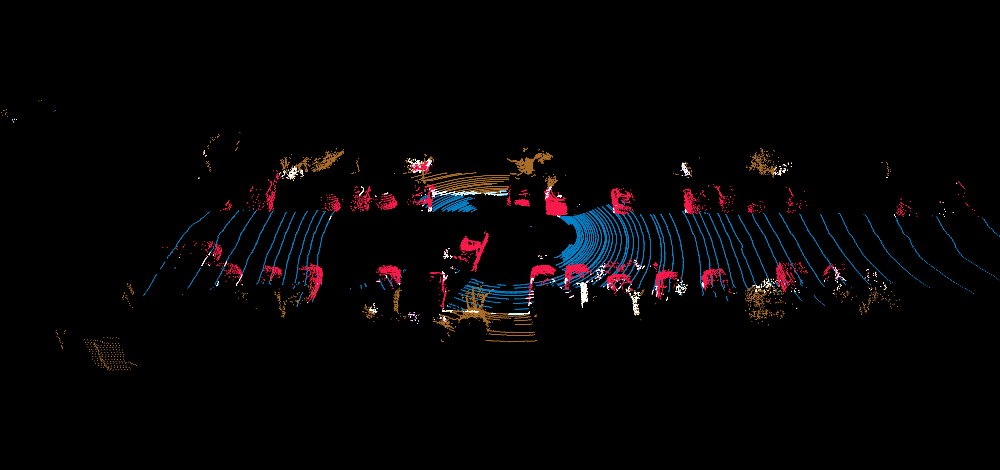} &
		\adjincludegraphics[width=0.25\linewidth, trim={{.2\width} {.2\height} {.2\width} {.2\height}}, clip]{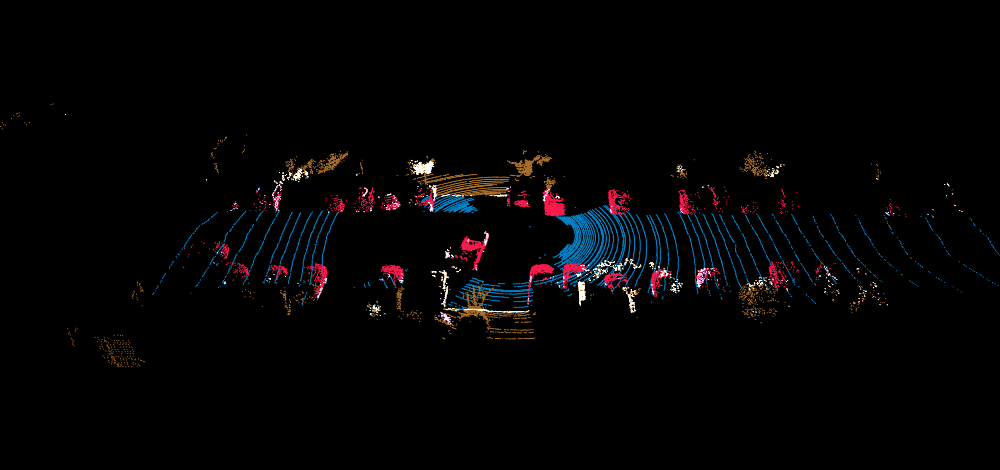} &
		\adjincludegraphics[width=0.25\linewidth, trim={{.2\width} {.2\height} {.2\width} {.2\height}}, clip]{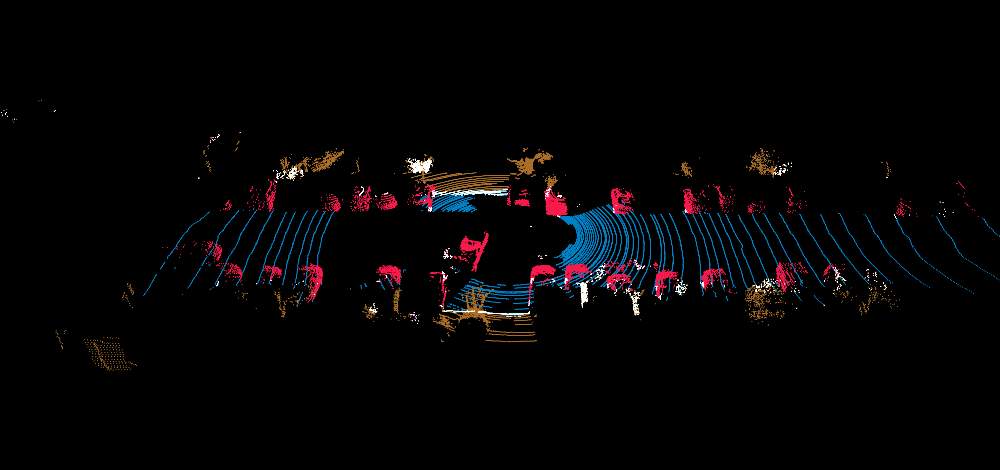} \\

		\adjincludegraphics[width=0.25\linewidth, trim={{.2\width} {.2\height} {.2\width} {.2\height}}, clip]{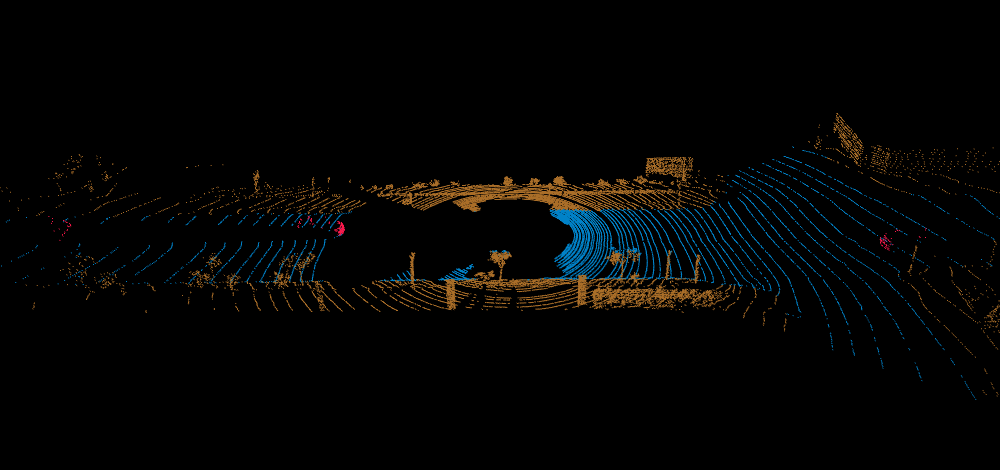} &  
		\adjincludegraphics[width=0.25\linewidth, trim={{.2\width} {.2\height} {.2\width} {.2\height}}, clip]{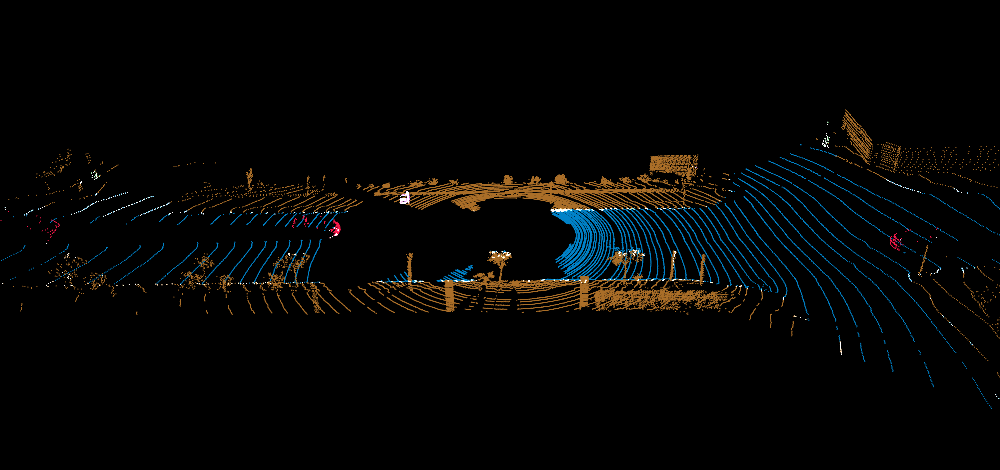} &
		\adjincludegraphics[width=0.25\linewidth, trim={{.2\width} {.2\height} {.2\width} {.2\height}}, clip]{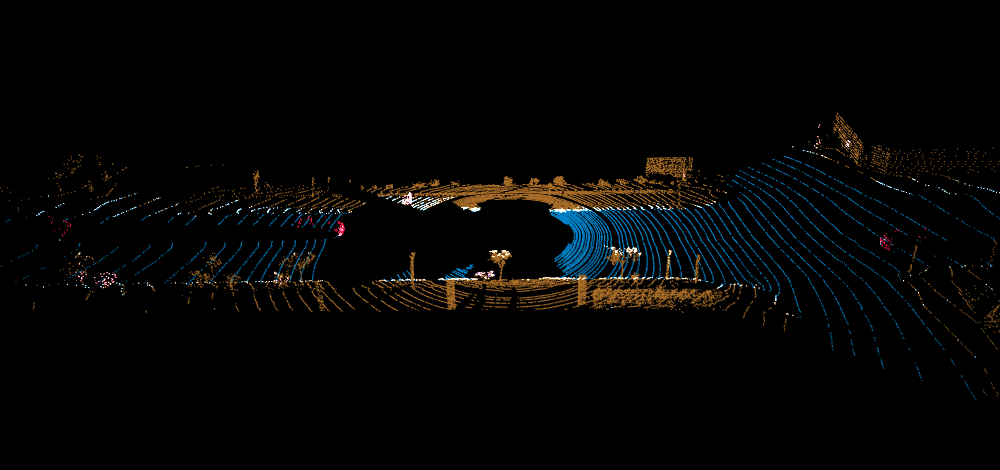} &
		\adjincludegraphics[width=0.25\linewidth, trim={{.2\width} {.2\height} {.2\width} {.2\height}}, clip]{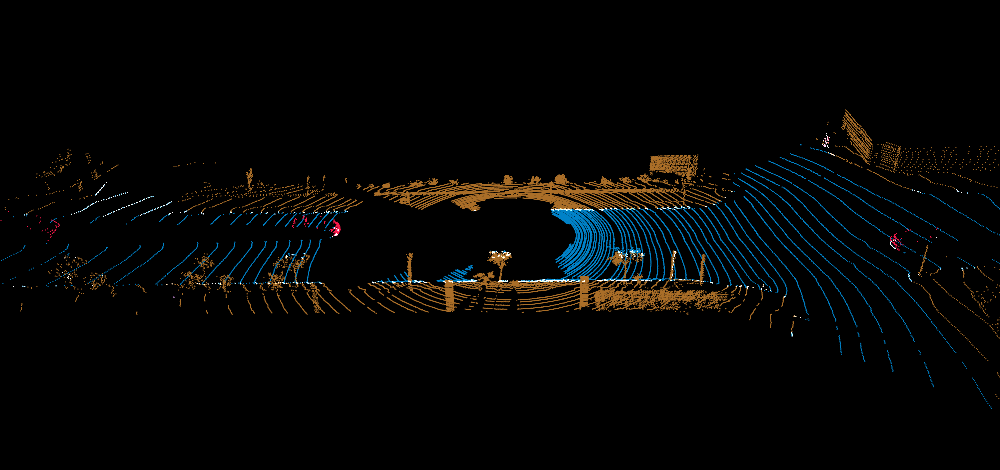} \\
				
		\adjincludegraphics[width=0.25\linewidth, trim={{.2\width} {.2\height} {.2\width} {.2\height}}, clip]{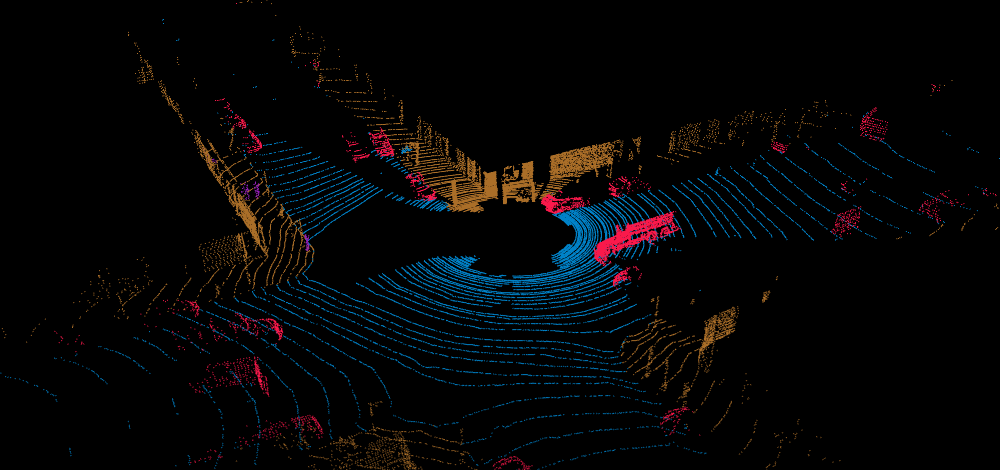} &   
		\adjincludegraphics[width=0.25\linewidth, trim={{.2\width} {.2\height} {.2\width} {.2\height}}, clip]{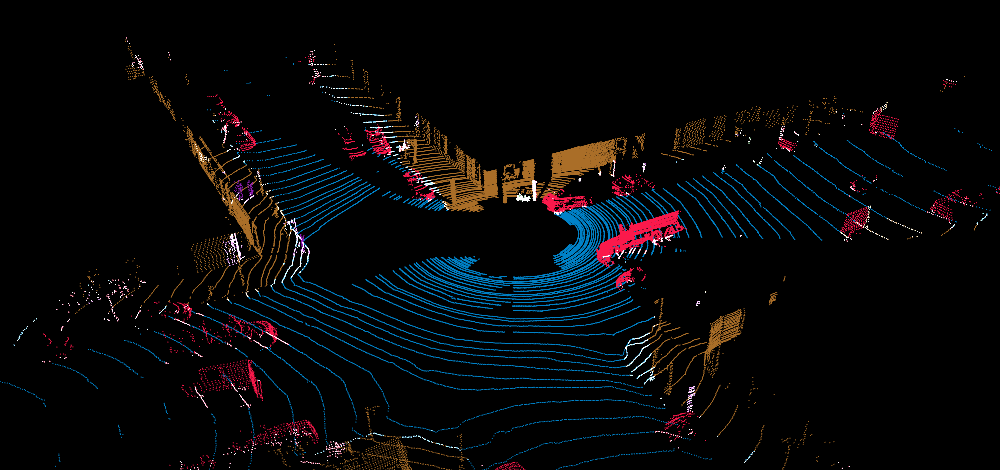} &
		\adjincludegraphics[width=0.25\linewidth, trim={{.2\width} {.2\height} {.2\width} {.2\height}}, clip]{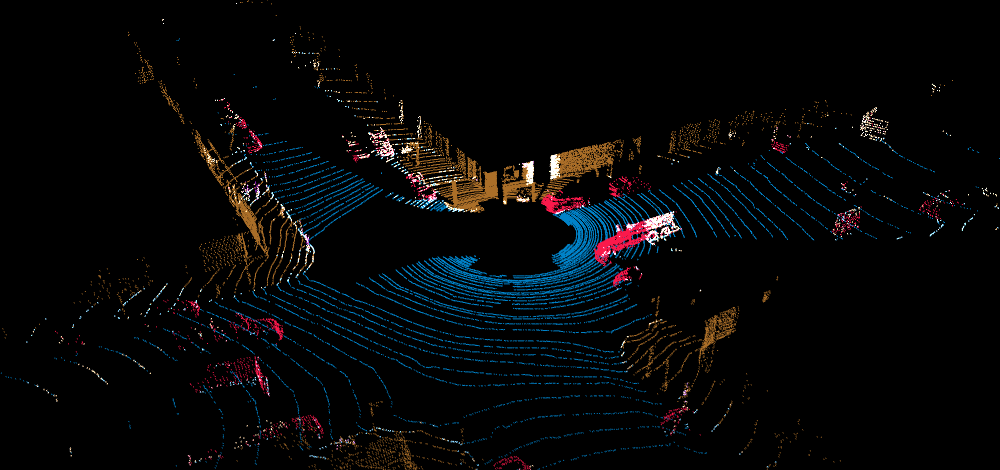} &
		\adjincludegraphics[width=0.25\linewidth, trim={{.2\width} {.2\height} {.2\width} {.2\height}}, clip]{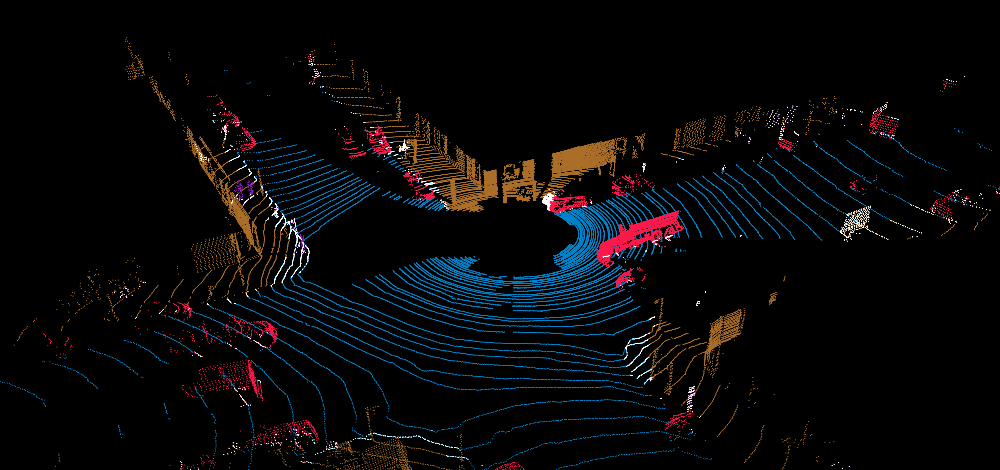} \\

		Ground Truth & 3D-FCN & Ours PCCN & Ours 3D-FCN+PCCN
	\end{tabular}
	\vspace{-3mm}
	\caption{Semenatic Segmentation Results on Driving Scene Dataset; Colored: correct prediciton; white: wrong prediciton.} 
	\label{fig-odtac}
\end{figure*}

\subsection{Discussions}

\paragraph{Locality \simon{Enforcing} Continuous Convolution:} Standard grid convolution are computed over a limited kernel size $M$ to keep locality. Similarly,  locality can be enforced in our parametric continuous convolutions by constraining the influence of the function $g$ to points close to $\bx$, \ie., %
\[
g(\mathbf{z}) = MLP(\mathbf{z}) w(\mathbf{z})
\]
where $w(\cdot)$ is a modulating window function. 
This can be achieved in differently.
First, we can constrain the cardinality of its local support domain and only keep non-zero kernel values for its K-nearest neighbors: $w(\mathbf{z}) = \mathbf{1}_{\mathbf{z} \in \mathrm{KNN}(\mathcal{S}, \mathbf{x})}$.
Alternatively we can keep non-zero kernel values for points within a fixed radius $r$: $w(\mathbf{z}) = \mathbf{1}_{||\mathbf{z}||_2 < r}$. %

\paragraph{Efficient Continuous Convolution:} For each continuous convolution layer, the kernel function is evaluated $N\times |\cS| \times F \times O$ times, where $|\cS|$ is the cardinality of the support domain, and the intermediate weight tensor is stored
\simon{for backpropagation}. \simon{This is expensive in practice, especially when both the number of points and the feature dimension are large.}  \simon{With the locality enforcing formulation}, we can constrain the cardinality of $\cS$. Furthermore, motivated by the idea of separable filters, we use the fact that this computation can be factorized if the kernel function value across different output dimensionality is shared. That is to say, we can decompose the weight tensor $W \in \bbR^{N\times |\cS| \times F \times O}$ into 
two tensors $W_1 = \bbR^{F\times O}$ and $W_2 = \bbR^{N\times |\cS| \times O}$ 
, where $W_1$ is a linear weight matrix and $W_2$ is evaluated through the MLP. 
 \simon{With this optimization,} only $N\times |\cS| \times O$  \simon{kernel} evaluations need to be computed and stored. Lastly, in inference stage, through merging the operations of batchnorm and fc layer in MLP, 3x speed boosting can be achieved. %

\paragraph{Special Cases:} Many previous convolutional layers are special cases of our approach. For instance, if the points are sampled over the finite 2D grid we recover conventional 2D convolutions. If the support domain is defined as concatenation of the spatial vector and feature vector with a gaussian kernel $g(\cdot)$, we recover the bilateral filter. If the support domain is defined as the neighboring vertices of a node we recover the first-order spatial graph convolution \cite{gcn}.  %

\begin{figure*}
\centering
	\setlength\tabcolsep{0.5pt} %
	\renewcommand{\arraystretch}{0.8}
	\begin{tabular}{ccc}
  		\adjincludegraphics[width=.32\linewidth, trim={{.15\width} {.15\height} {.15\width} {.15\height}}, clip]{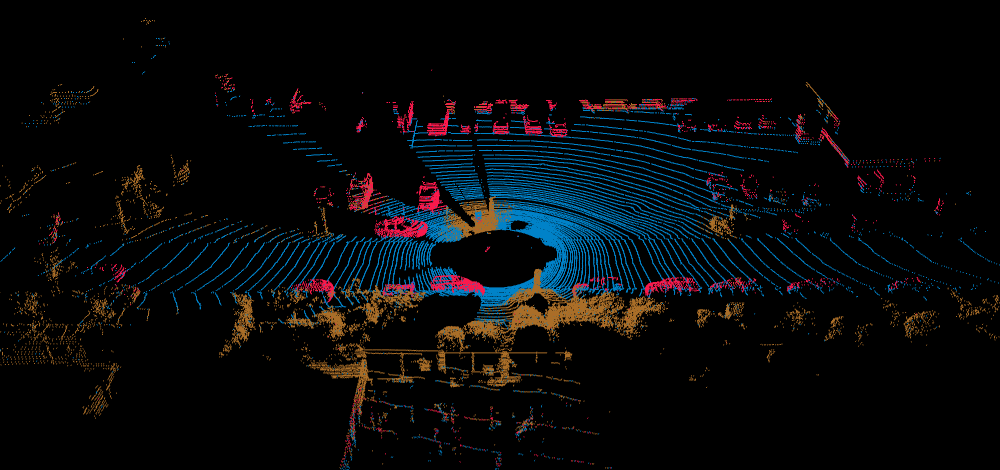} &   
  		\adjincludegraphics[width=.32\linewidth, trim={{.15\width} {.15\height} {.15\width} {.15\height}}, clip]{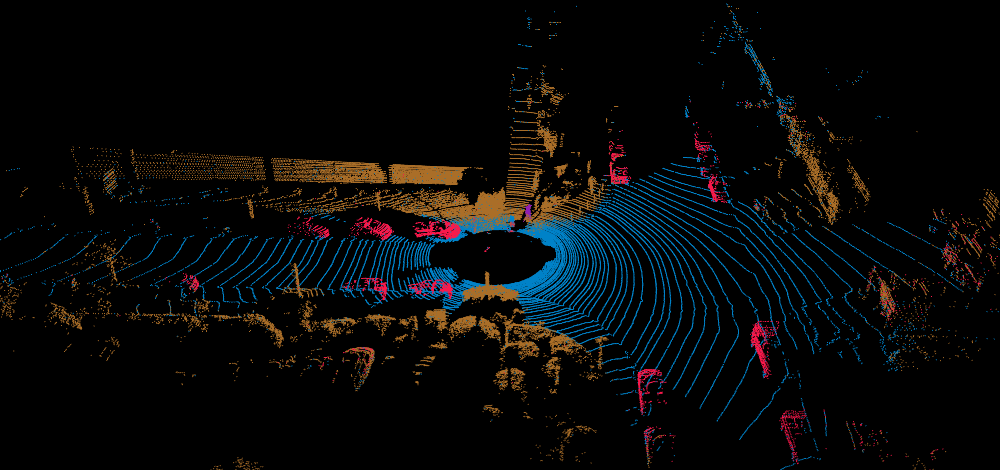}  & 					
  		\adjincludegraphics[width=.32\linewidth, trim={{.15\width} {.15\height} {.15\width} {.15\height}}, clip]{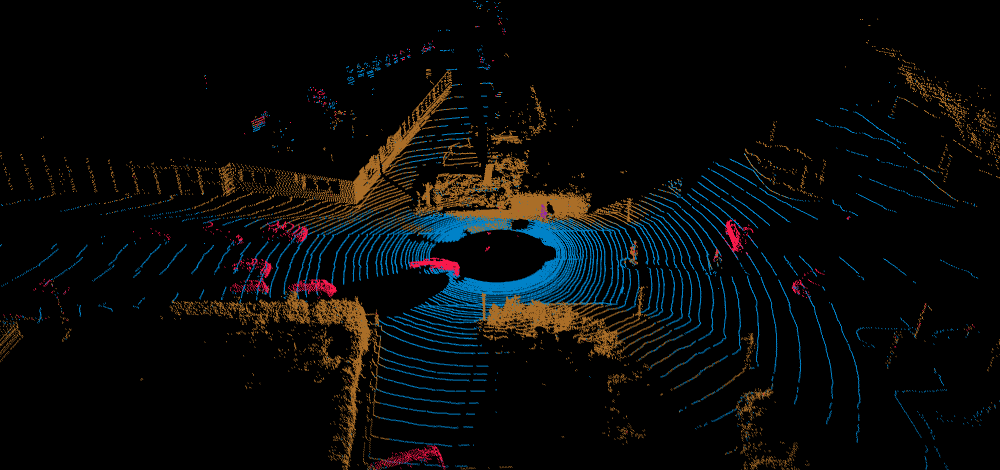} 
	\end{tabular}
		\vspace{-3mm}
	\caption{Semantic Labeling on KITTI Dataset without Retraining} %
	\label{fig:kitti}
\end{figure*}

\section{Experimental Evaluation}

We demonstrate the effectiveness of our approach in the tasks of semantic labeling and motion estimation of 3D point clouds, and show state-of-the-art performance. 
We conduct point-wise semantic labeling experiments over two datasets: a very large-scale outdoor lidar semantic segmentation dataset that we collected and labeled in house and \simon{a} large indoor semantic labeling dataset. To our knowledge, these are the largest real-world outdoor and indoor datasets that \simon{are available} for this task.
The datasets are fully labeled and contain 137 billion and 629 million points respectively. \shenlong{The lidar flow experiment is also conducted on this dataset with ground-truth 3D motion label for each point. }

\begin{table*}[]
\centering
\footnotesize
\setlength\tabcolsep{4pt} %
\begin{tabular}{c|cc|ccccccccccccc}
Method           				  &  \cellcolor{blue!25} \textbf{mIOU} & \cellcolor{blue!25}  \textbf{mAcc}     & ceiling        & floor          & wall           & beam          & column        & window         & door           & chair          & table          & bookcase       & sofa           & board          & clutter        \\ \hline
PointNet  \cite{pointnet}         &  \cellcolor{blue!25}41.09          &  \cellcolor{blue!25}48.98          	& 88.80          & \textbf{97.33} & 69.80          & 0.05          & 3.92           & 46.26          & 10.76          & 52.61          & 58.93          & 40.28          & 5.85           & 26.38          & 33.22          \\
3D-FCN-TI  \cite{segcloud}        &  \cellcolor{blue!25}47.46          & \cellcolor{blue!25} 54.91          	& {90.17} 		 & 96.48          & 70.16          & 0.00          & 11.40          & 33.36          & 21.12          & \textbf{76.12} & 70.07          & 57.89          & 37.46          & 11.16          & 41.61          \\
SEGCloud   \cite{segcloud}        & \cellcolor{blue!25} 48.92          &  \cellcolor{blue!25}57.35          	& 90.06          & 96.05          & 69.86          & 0.00          & \textbf{18.37} & 38.35          & 23.12          & 75.89          & \textbf{70.40} & \textbf{58.42} & 40.88          & 12.96          & {41.60} \\ \hline
Ours PCCN         				  & \cellcolor{blue!25} \textbf{58.27} & \cellcolor{blue!25} \textbf{67.01} 	& \textbf{92.26} & {96.20} & \textbf{75.89} & \textbf{0.27} & {5.98} & \textbf{69.49} & \textbf{63.45} & {66.87} & 65.63 & 47.28 & \textbf{68.91} & \textbf{59.10} & \textbf{46.22} \\
\end{tabular}
\vspace{-3mm}
\caption{Semantic Segmentation Results on Stanford Large-Scale 3D Indoor Scene Dataset }
\label{tab-indoor3d}
\end{table*}
\subsection{Semantic Segmentation of Indoor  Scenes} 

\paragraph{Dataset:} We use the  Stanford large-scale 3D indoor scene dataset \cite{indoor3d} and  follow the training and testing procedure used in \cite{segcloud}.
 We report the same metrics, i.e., mean-IOU, mean class accuracy (TP / (TP + FN)) and class-wise IOU. The input is six dimensional and is composed of   the xyz coordinates and RGB color intensity. Each point is labeled with one of 13 classes shown in \tabref{tab-indoor3d}. 

\paragraph{Competing Algorithms:} We compare our approach to  PointNet \cite{pointnet} and SegCloud \cite{segcloud}.
\shenlong{We evaluate the proposed end-to-end 
continuous convnet} with eight continuous convolution layers (\textit{Ours PCCN}). 
\simon{The kernels are defined over the continuous support domain of 3D Euclidean space.}
Each intermediate layer except the last has 32 dimensional hidden features followed by batchnorm and ReLU nonlinearity. The dimension of the last layer is 128. 
\simon{We observe that the distribution of semantic labels within a room is highly correlated with the room type (\eg office, hallway, conference room, \etc).}
Motivated by this, we 
apply max pooling over all the points in the last layer to obtain a global feature, which is then concatenated to the output feature of each points in the last layer, resulting in a 256 dimensional feature. \shenlong{A fully connected layer with softmax activation is used to produce the final logits. Our network is trained end-to-end with cross entropy loss, using Adam optimizer. }

\paragraph{Results:} As shown in  Tab.~\ref{tab-indoor3d} our approach outperforms the state-of-the-art by 9.3\% mIOU and 9.6\% mACC. \figref{fig-indoor3d}  shows  qualitative results. 
Despite the diversity of geometric structures, our approach works very well. 
Confusion mainly occurs between columns vs walls and  window vs bookcase. It is also worth noting that  our approach captures visual information encoded in RGB channels. The last row shows two failure cases. In the first one, the door in the washroom is labeled as clutter whearas our algorithm thinks is door. In the second one, the board on the right has a window-like texture, which makes the algorithm predict the wrong label. %

\begin{figure}
	\footnotesize
	\setlength\tabcolsep{0.5pt} %
	\renewcommand{\arraystretch}{0.8}
	\begin{tabular}{cc}
  		\adjincludegraphics[width=.5\linewidth, trim={{.2\width} {.2\height} {.2\width} {.25\height}}, clip]{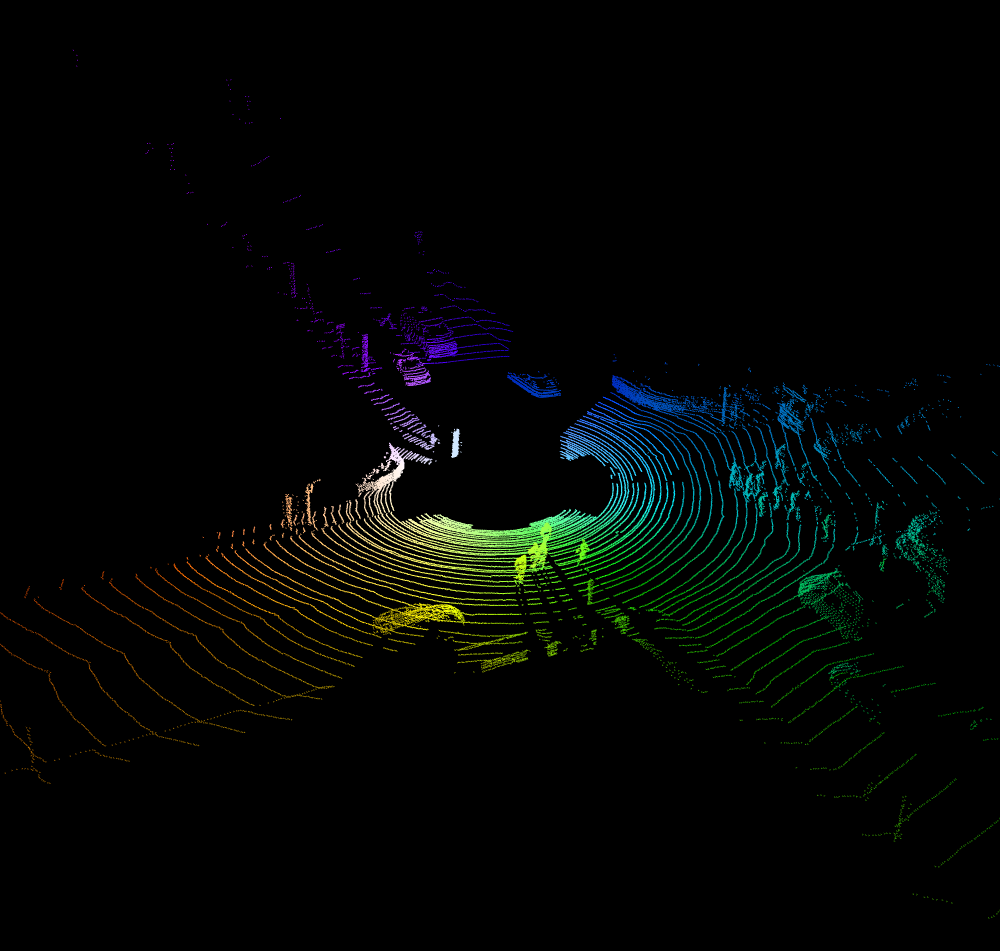} & 
		\adjincludegraphics[width=.5\linewidth, trim={{.2\width} {.2\height} {.2\width} {.25\height}}, clip]{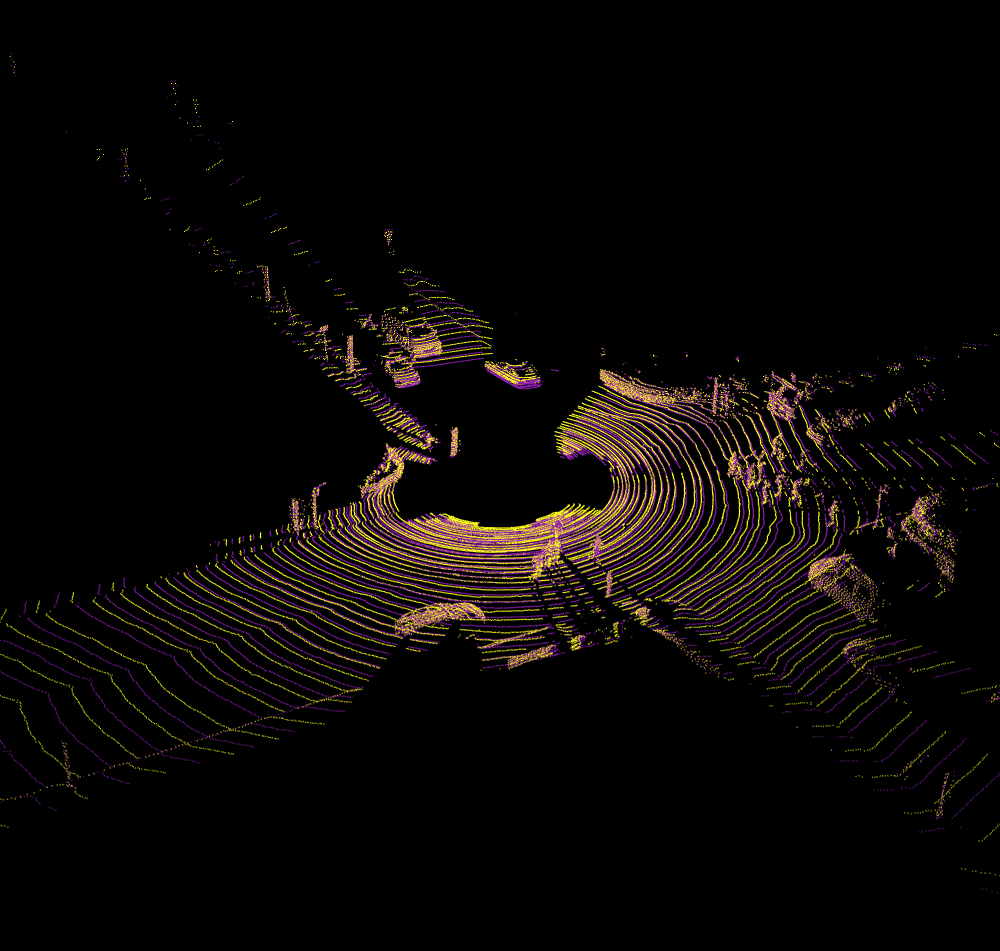} \\
		Flow Field & Overlay of Target and Warped Source\\
	\end{tabular}
	\vspace{-3mm}
	\caption{Right: \textcolor{purple}{purple} shows target frame, \textcolor{yellow}{yellow} shows source frame warped to target frame using ground truth flow}
	\label{fig:flow}
\end{figure}
\subsection{Semantic Segmentation of Driving Scenes} 
\label{sec:odtac}
\paragraph{Dataset:} We first conduct experiments on the task of point cloud segmentation in the context of autonomous driving. Each point cloud is produced by a full sweep of a roof-mounted Velodyne-64 lidar sensor driving in several cities in North America. 
The dataset is composed of snippets each having 300 consecutive frames. %
The training and validation set contains 11,337 snippets in total while the test set contains 1,644 snippets.  %
We  report metrics on a subset of the test set which is generated by sampling 10 frames from each snippet to avoid bias brought due to scenes where the ego-car is  static (e.g., when waiting at a traffic light). 
Each point is labeled with one of seven classes defined in Tab.~\ref{tab-odtac}.
We adopt mean intersection-over-union (meanIOU) and point-wise accuracy  (pointAcc) as our evaluation metrics. 
\begin{table*}[]
	\centering
	\footnotesize
	\setlength\tabcolsep{4pt} %
	\begin{tabular}{c|cc|ccccccc|c}
        Method	      							&  \cellcolor{blue!25}  \textbf{pACC}	&   \cellcolor{blue!25} \textbf{mIOU}   & vehicle         & bicyclist        & pedestrian       & motorcycle       & animal          & background       & road       & params size      \\ \hline
		PointNet \cite{pointnet}              	&  \cellcolor{blue!25}  91.96        	&  \cellcolor{blue!25}  38.05          & 76.73         & 2.85          & 6.62        & 8.02        & 0.0          & 89.83    & 91.96   & 20.34MB \\ 
		3D-FCN \cite{resnet}            	 	&   \cellcolor{blue!25} 94.31        	&  \cellcolor{blue!25}  49.28          & 86.74          & 22.30         	& 38.26          	& 17.22          & 0.98 	& 86.91          	& 92.56 &   74.66MB   \\ \hline
		Ours PCCN   							&   \cellcolor{blue!25} 94.56 &   \cellcolor{blue!25} 46.35 & 86.62 & 8.31 & 41.84 & 7.24 & 0.00 & 87.27 & \textbf{93.20} & \textbf{9.34MB} \\
		Ours 3D-FCN+PCCN 						&   \cellcolor{blue!25} \textbf{95.45}  &  \cellcolor{blue!25} 	\textbf{58.06}  &	\textbf{91.83}  & \textbf{40.23}  &	\textbf{47.74}  &	\textbf{42.91}  &	\textbf{1.25}  &	\textbf{89.27}  &	93.18  &  74.67MB \\ 
	\end{tabular}
	\vspace{-3mm}
	\caption{Semenatic Segmentation Results on Driving Scenes Dataset}
	\label{tab-odtac}
	\end{table*}

\paragraph{Baselines:} We compare our approach to  the point cloud segmentation network (\textit{PointNet}) \cite{pointnet} and  a 3D fully convolutional network (\textit{3D-FCN}) conducted over a 3D  occupancy grid.
We use a resolution of 0.2m for each voxel over a  160mx80mx6.4m range. This results in  an occupancy grid encoded as a  tensor of size  800x400x32. We define a voxel to be occupied if it contains at least one point. We use ResNet-50 as the backbone and replace the last average pooling and fully connected layer with two fully convolutional layers and a trilinear upsampling layer to obtain dense voxel predictions. The model is trained from scratch with the Adam optimizer\cite{adam} \shenlong{to minimize the class-reweighted cross-entropy loss.} %
Finally, the voxel-wise predictions are mapped back to the original points and metrics are computed over points.
\shenlong{We adapted the open-sourced PointNet model onto our dataset and trained from scratch. The architecture and loss function remain the same with the original paper, except that we removed the point rotation layer since it negatively \simon{impacts} validation performance on this dataset. } %

\begin{figure*}
	\footnotesize
	\setlength\tabcolsep{0.5pt} %
	\renewcommand{\arraystretch}{0.8}
	\begin{tabular}{cccc}
  		\adjincludegraphics[width=.24\linewidth, trim={{.15\width} {.25\height} {.15\width} {.25\height}}, clip]{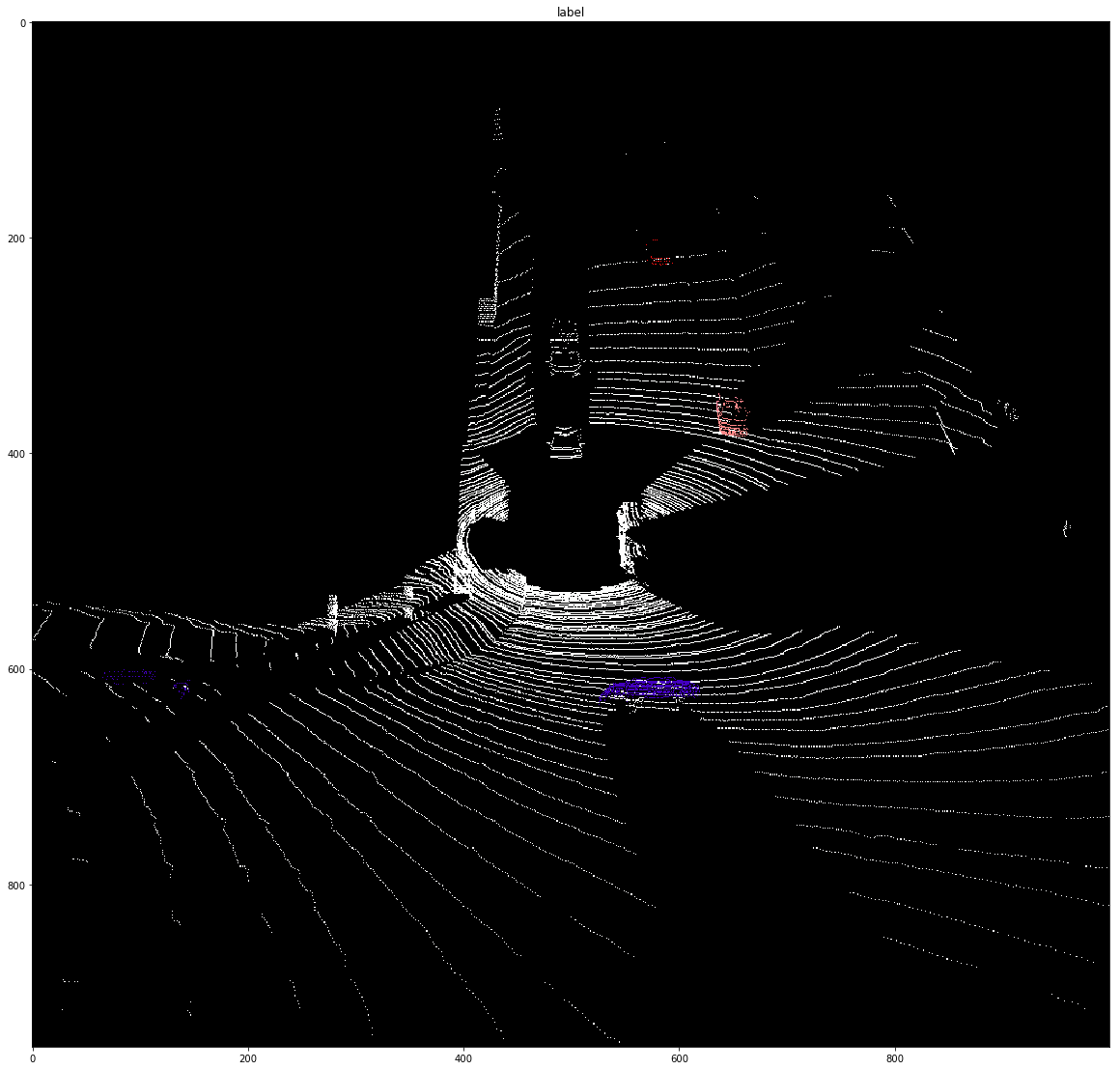} & 
		\adjincludegraphics[width=.24\linewidth, trim={{.15\width} {.25\height} {.15\width} {.25\height}}, clip]{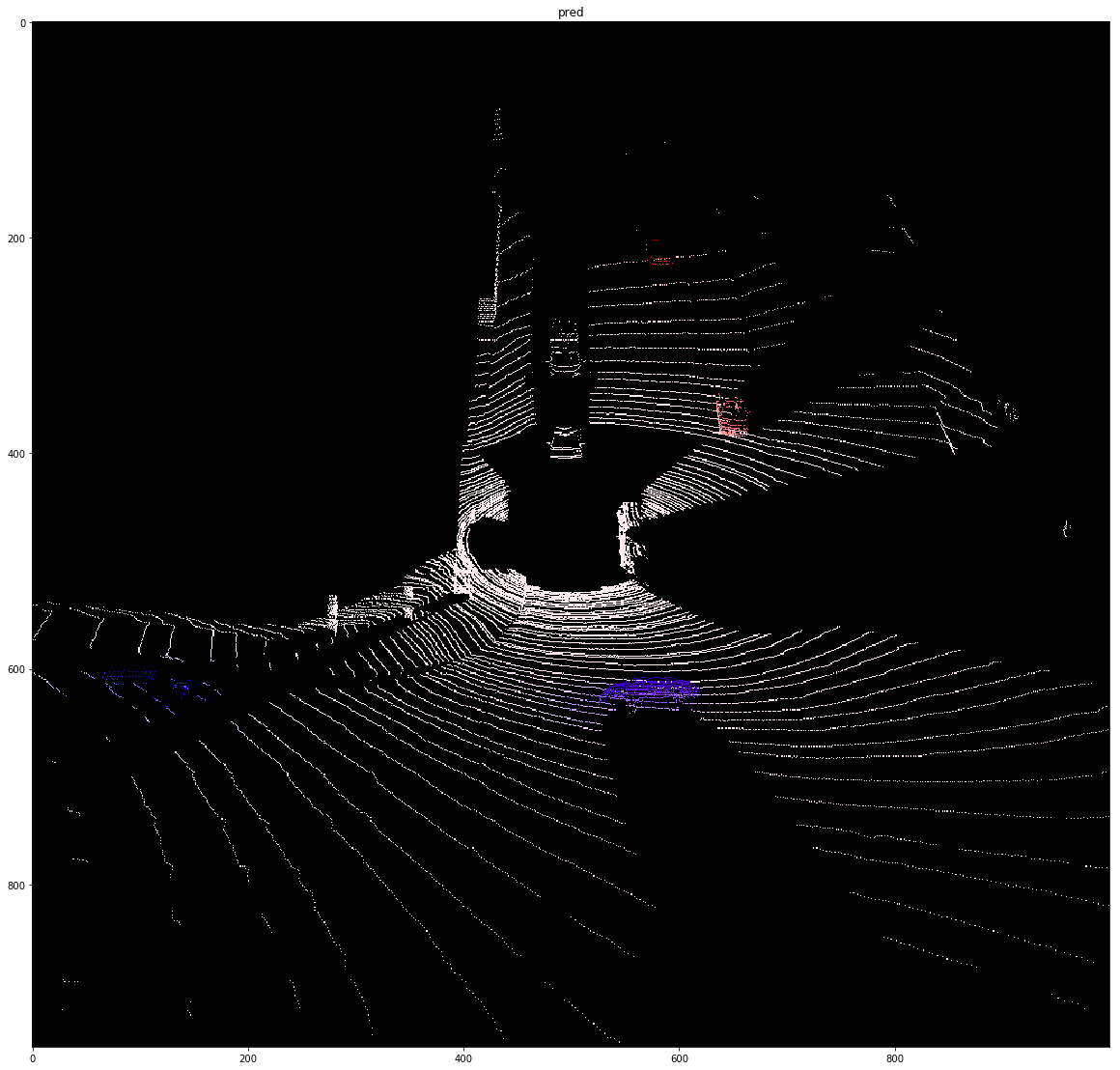} &
  		\adjincludegraphics[width=.24\linewidth, trim={{.15\width} {.25\height} {.15\width} {.25\height}}, clip]{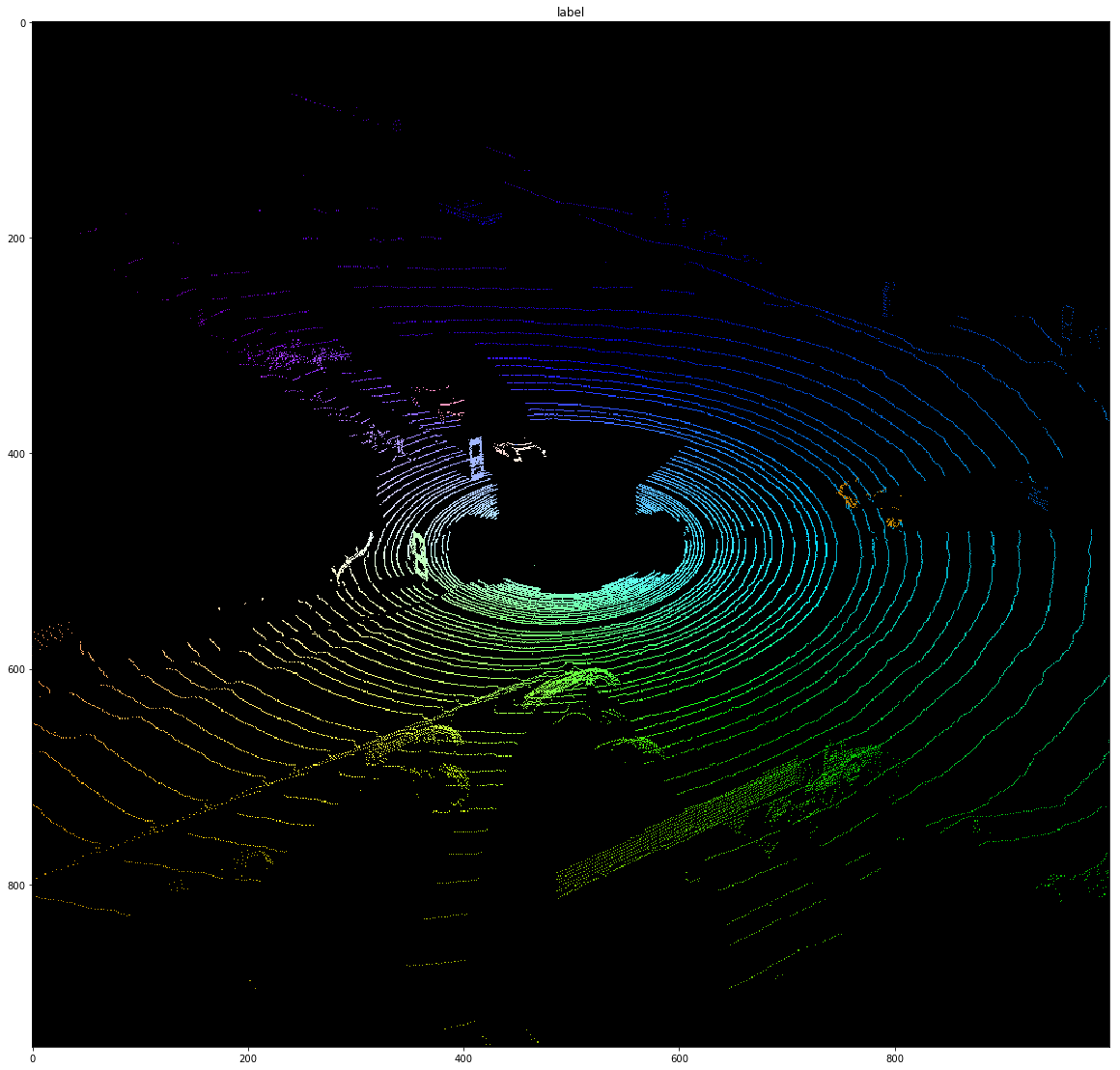} & 
		\adjincludegraphics[width=.24\linewidth, trim={{.15\width} {.25\height} {.15\width} {.25\height}}, clip]{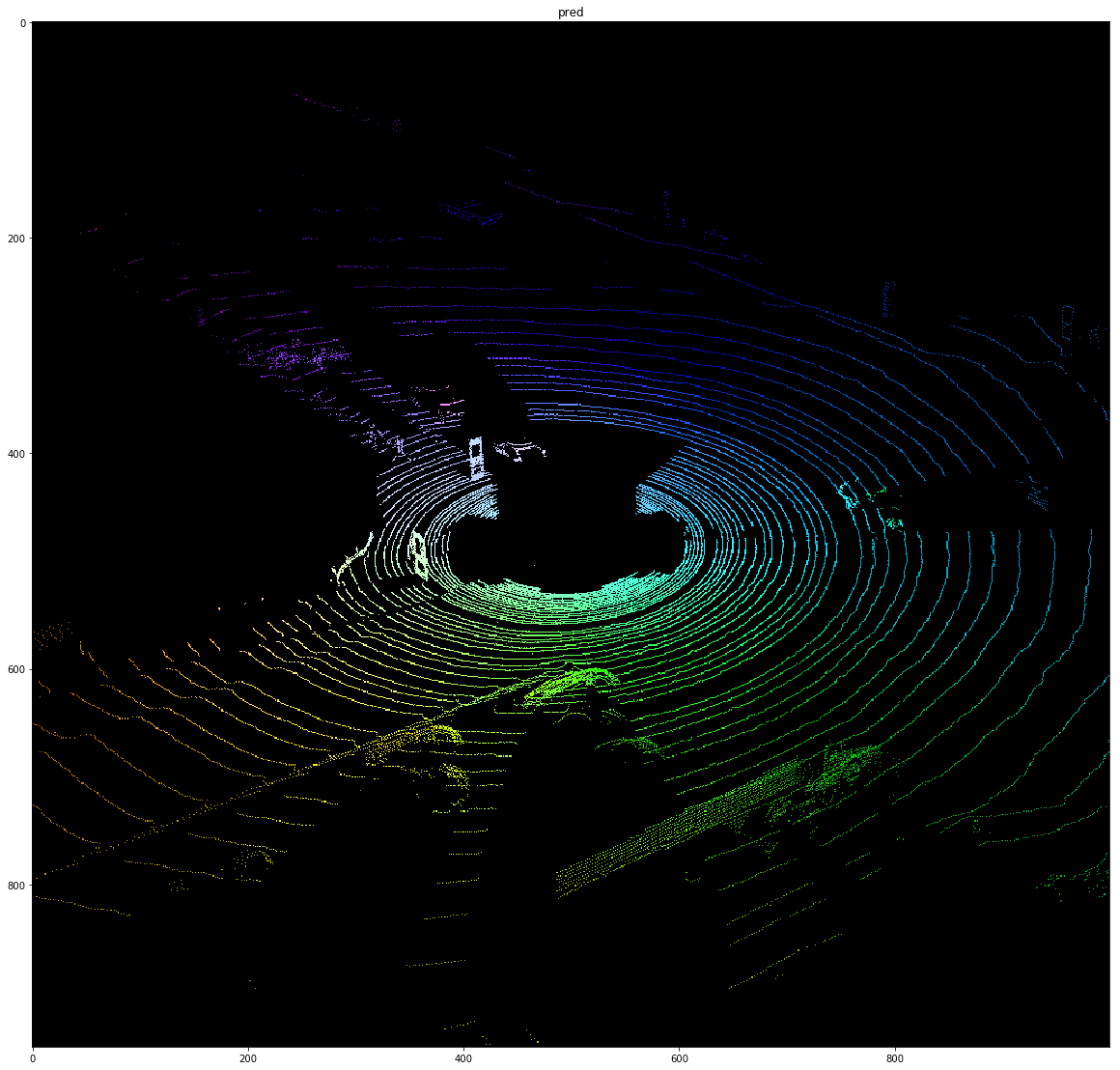} \\
		\adjincludegraphics[width=.24\linewidth, trim={{.15\width} {.25\height} {.15\width} {.25\height}}, clip]{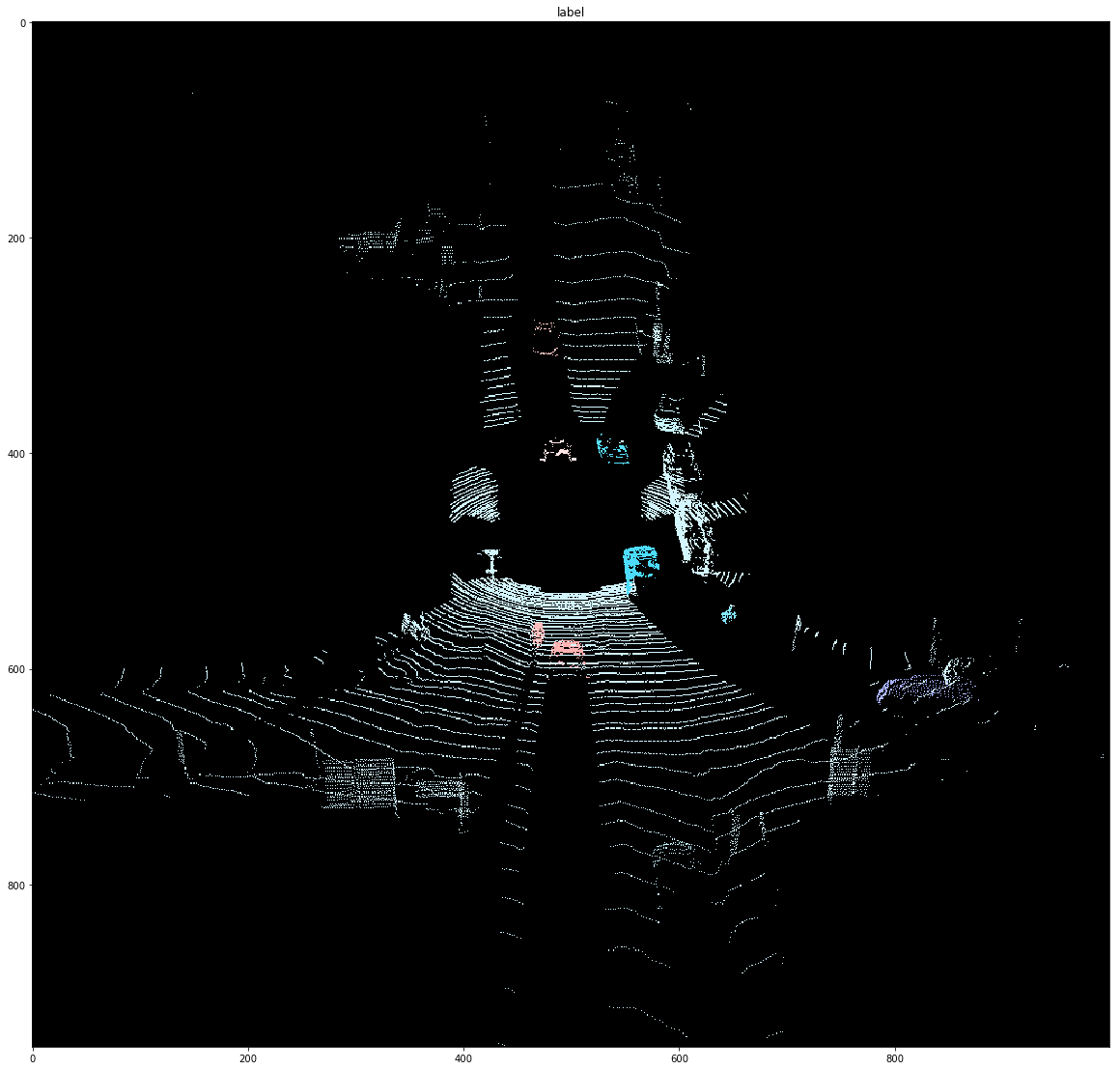} & 
		\adjincludegraphics[width=.24\linewidth, trim={{.15\width} {.25\height} {.15\width} {.25\height}}, clip]{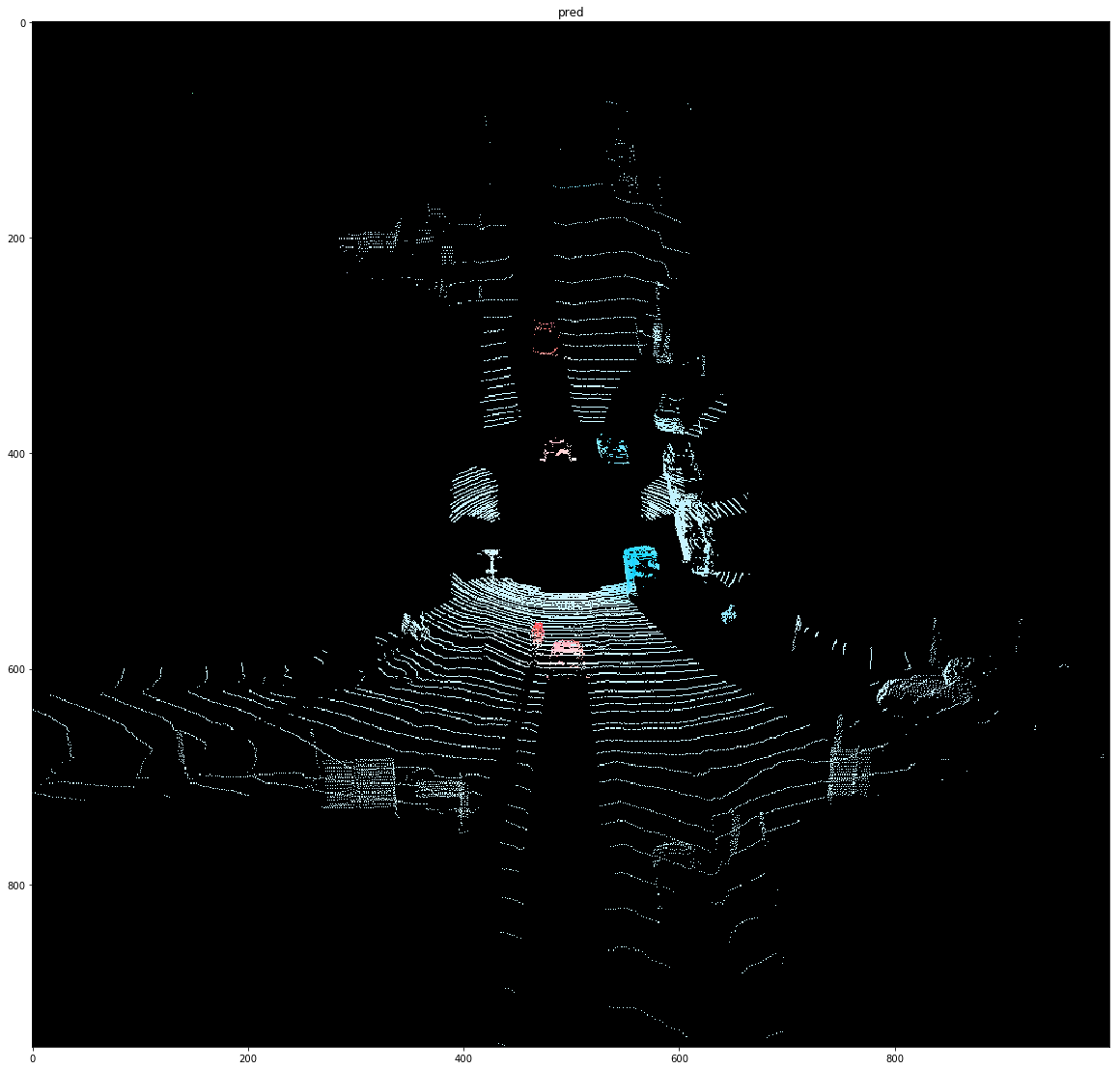} &
  		\adjincludegraphics[width=.24\linewidth, trim={{.15\width} {.25\height} {.15\width} {.25\height}}, clip]{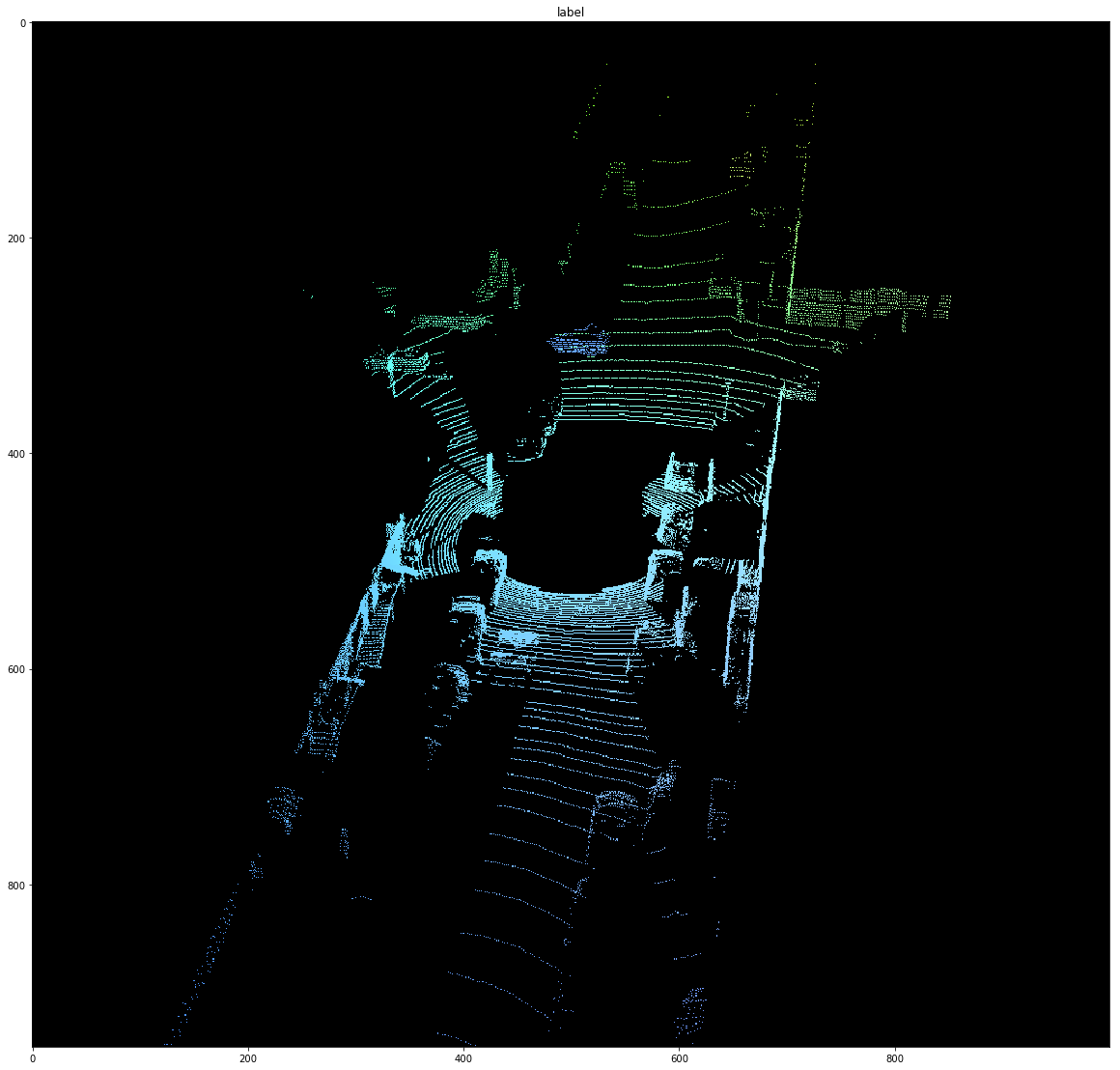} & 
		\adjincludegraphics[width=.24\linewidth, trim={{.15\width} {.25\height} {.15\width} {.25\height}}, clip]{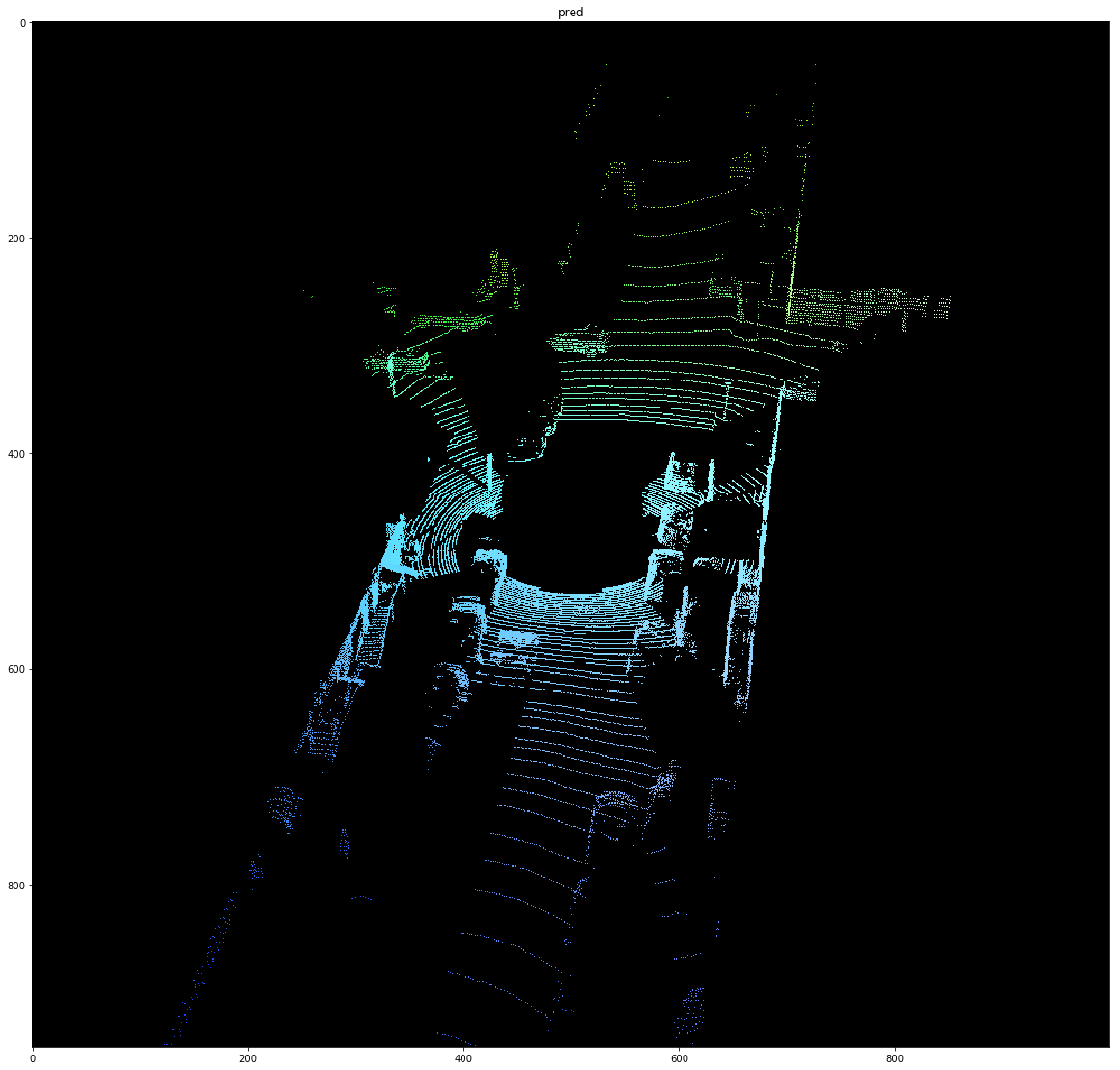} \\
		Ground Truth & Ours 3D-FCN+PCCN & Ground Truth & Ours 3D-FCN+PCCN \\
	\end{tabular}
	\vspace{-3mm}
	\caption{Lidar Flow Results on Driving Scene Dataset}
	\label{fig:flow-prediction}
\end{figure*}

\paragraph{Our Approaches:} We evaluate two versions of our approach.   Our first instance conducts continuous convolutions directly over the raw xyz-intensity lidar points (\textit{Ours PCCN}). 
Our second version (\textit{Ours 3D-FCN+PCCN}) performs continuous convolutions over the features extracted from  \textit{3D-FCN}. 
\textit{Ours PCCN} has 16 continuous conv layers with residual connections, batchnorm and ReLU non-linearities. We use the spatial support in $\bbR^3$ to define our kernel.  
We train the network with point-wise cross-entropy loss and Adam \cite{adam} optimizer. %
In contrast, \textit{Ours 3D-FCN+PCCN} model has  7 residual continuous convolutional layers on top of the trained \textit{3D-FCN} model and performs end-to-end fine-tuning using Adam optimizer. %

\paragraph{Results:} As shown in Tab.~\ref{tab-odtac}, by exploiting sophisticated feature via 3D convolutions, \textit{3D-FCN+PCCN} results in the best performance. %
\figref{fig-odtac} shows qualitative comparison between models. As shown in the figure, all models produce good results. Performance differences often result from ambiguous regions. 
In particular, we can see that the 3D-FCN model oversegements the scene: it mislabels a background pole as vehicle (red above egocar), nearby spurirous points as bicyclist (green above egocar), and a wall as pedestrian (purple near left edge). This is reflected in the confidence map (as bright regions). We observe a significant improvement in our \textit{3D-CNN + PCCN} model, with all of the above corrected with high confidence. 
For more results and videos please refer to the supplementary material. 

\paragraph{Model Sizes:} \shenlong{We also compare the model sizes of the competing algorithms in Tab.~\ref{tab-odtac}. 
\simon{In comparison to} the 3D-FCN approach, the end-to-end continuous convolution network's model size is eight times smaller
, while achieving comparable results. And the 3D-FCN+PCCN is just 0.01MB larger than 3D-FCN, but the performance is improved by a large margin in terms of mean IOU. }

\paragraph{Complexity and Runtime} We benchmark the proposed model's runtime over a GTX 1080 Ti GPU and Xeon E5-2687W CPU with 32 GB Memory. The forward pass of a 8-layer PCCN model (32 feature dim in each layer with 50 neighbours) takes 33ms.  The KD-Tree neighbour search takes 28 ms. The end-to-end computation takes 61ms. The number of operations of each layer is 1.32GFLOPs. 

\paragraph{Generalization:} To demonstrate the generalization ability of our approach, we evaluate our model, trained with only North American scenes, on the KITTI dataset \cite{kitti}, which was captured in Europe.
As shown in \figref{fig:kitti}, the model achieves good results, with well segmented dynamic objects, such as vehicles and pedestrians.

\subsection{Lidar Flow}
\paragraph{Dataset:} \shenlong{We also validate our proposed method over the task of lidar based motion estimation, refered to as lidar flow. In this task, the input is two consecutive frames of lidar sweep. The goal is to estimation the 3D motion field for each point in the first frame, to undo both ego-motion and the motion of dynamic objects. The ground-truth ego-motion is computed through a comprehensive filters that take GPS, IMU as well as ICP based lidar alignment against pre-scaned 3D geometry of the scene as input. And the ground-truth 6DOF dynamics object motion is estimated from the temporal coherent 3D object tracklet, labeled by in-house annotators. Combining both we are able to get the ground-truth motion field. ~\figref{fig:flow} shows the colormapped flow field and the overlay between two frames after undoing per-point motion. This task is crucial for many applications, such as multi-rigid transform alignment, object tracking, global pose estimation, \etc. The training and validation set contains 11,337 snippets while the test set contains 1,644 snippets.  %
We use 110k frame pairs for training and validation, and 16440 frame pairs for testing. End-point error, and outlier percentage at 10 cm and 20 cm are used as metric. %
}
\paragraph{Competing Algorithms:} We compare against the 3D-FCN baseline using the same architecture and volumetric representation as used in Sec.~\ref{sec:odtac}. We also adopt a similar 3D-FCN + PCCN architecture with 7 residual continuous convolution layers added as a polishing network. In this task, we remove the ReLU nonlinearity and supervise the PCCN layers with MSE loss at every layer. \shenlong{The training objective function is mean square error loss between the ground-truth flow vector and the prediction.}
\paragraph{Results:} Tab.~\ref{tab-lidarflow} reports the quantitative results. As shown in the table, our 3D-FCN+PCCN model outperforms the 3D-FCN by 0.351cm in end-point error \shenlong{and our method reduces approximately $20\%$ of the outliers. \figref{fig:flow-prediction} shows sample flow predictions compared with ground truth labels. As shown in the figure, our algorithm is able to capture both global motion of the ego-car including self rotation, and the motion of each dynamic objects in the scene. For more results please refer to our supplementary material.}

\begin{table}[]
\centering
\footnotesize
\begin{tabular}{c|cccc}
Method       			& EPE (cm)  			& Outlier$\%_{10}$ 	& Outlier$\%_{20}$	\\ \hline
3D-FCN       			& 8.161 				& 25.92\% 			& 7.12 \% \\ \hline
Ours 3D-FCN+PCCN 		& \textbf{7.810} 		& \textbf{19.84\%}	& \textbf{5.97\%} \\
\end{tabular}
\vspace{-3mm}
\caption{Lidar Flow Results on Driving Scenes Dataset}
\label{tab-lidarflow}
\end{table}

\section{Conclusions}
We have presented a new learnable convolution layer that operates over non-grid structured data. Our convolution kernel function is parameterized by multi-layer perceptrons and spans the full continuous domain. This allows us to design a new deep learning architecture that can be applied to arbitrary structured data, as long as the support relationships between elements are computable. We validate the performance on point cloud segmentation and motion estimation tasks, over very large-scale datasets with up to 200 bilion points. The proposed network achieves state-of-the-art performance on all the tasks and datasets.

{\small
\bibliographystyle{ieee}
\bibliography{top}
}
\clearpage

\appendix

\vspace*{0.2cm}
{ 
    \noindent 
    \Large 
    \textbf{Appendix}
} \\

\noindent
In this supplementary material, we first validate our method's generalization ability by testing the proposed models on KITTI dataset without re-training. We then provide more details about the lidar flow task. We also show additional results for all tasks, with analysis of failure modes. Lastly, we conduct a point cloud classification experiment using the deep parameteric continuous convolutional networks. In the supplementary video, we show our results on semantic segmentation and lidar flow results over a sequential data. We also show the generalization ability of the proposed method by training on our dataset and test it on KITTI as well as a truck sequence. 
\section{Generalization}
We show the generalization ability of our proposed model by training over one dataset and test it over another in our supplementary video. To be specific, we have used the following configurations:
\begin{itemize}
\item Train our proposed semantic labeling network on the driving scene data (several north America cities), and test it on KITTI (Europe). 
\item Train our proposed semantic labeling network on the driving scene data (non-highway road), and test it on a lidar sequence mounted on top of a high-way driving truck (highway road).
\item Train our proposed lidar flow network on the driving scene data (several north America cities), and test it on KITTI (Europe). 
\end{itemize}

\begin{figure*}
	\footnotesize
	\setlength\tabcolsep{0.5pt} %
	\renewcommand{\arraystretch}{0.8}
	\begin{tabular}{ccc}
  		\adjincludegraphics[width=.33\linewidth, trim={{.01\width} {.01\height} {.01\width} {.01\height}}, clip]{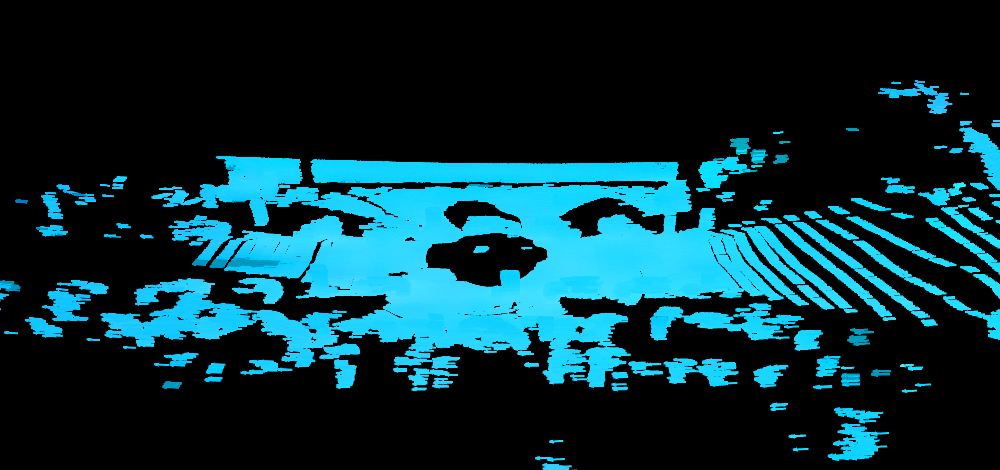} & 
  		\adjincludegraphics[width=.33\linewidth, trim={{.01\width} {.01\height} {.01\width} {.01\height}}, clip]{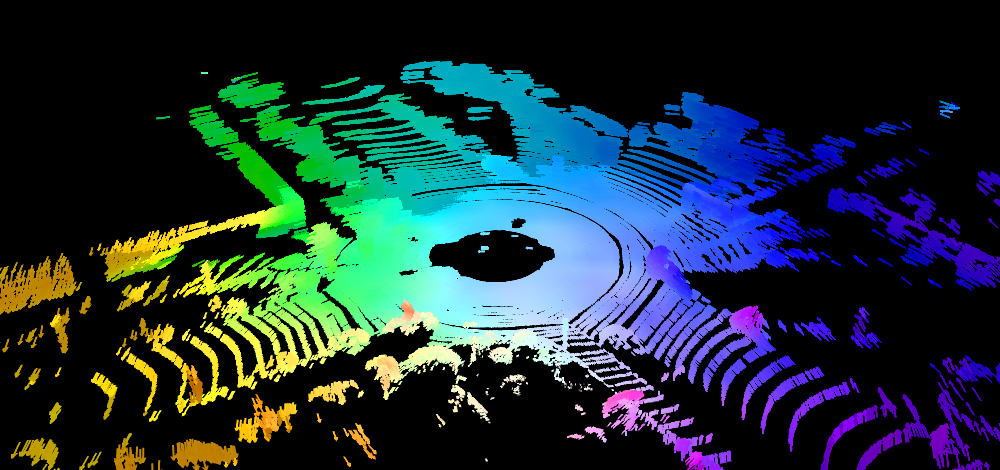} & 
  		\adjincludegraphics[width=.33\linewidth, trim={{.01\width} {.01\height} {.01\width} {.01\height}}, clip]{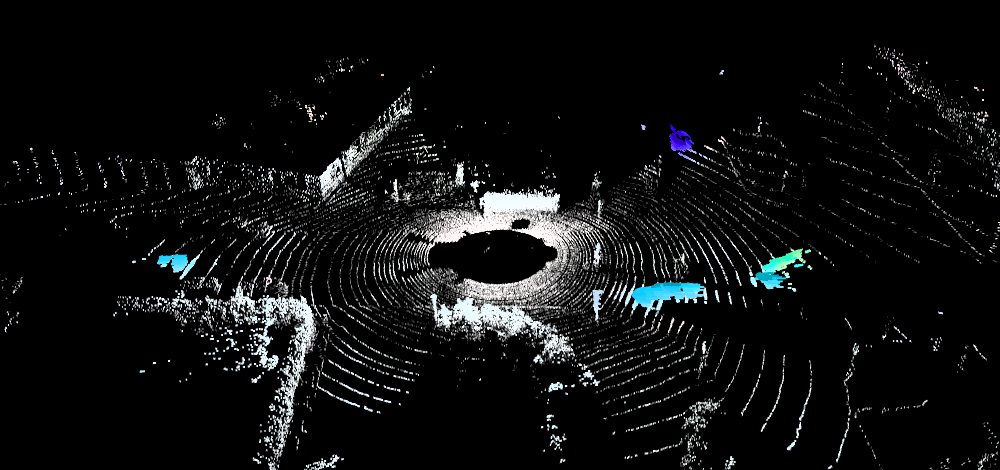} \\
   		\end{tabular}
	\vspace{-3mm}
	\caption{Lidar Flow Results on KITTI Dataset}
	\label{fig:flow-kitti}
\end{figure*}

\begin{figure*}
	\footnotesize
    \centering
	\setlength\tabcolsep{0.5pt} %
	\renewcommand{\arraystretch}{0.8}
	\begin{tabular}{ccc}
  		\adjincludegraphics[width=.33\linewidth, trim={{.01\width} {.01\height} {.01\width} {.01\height}}, clip]{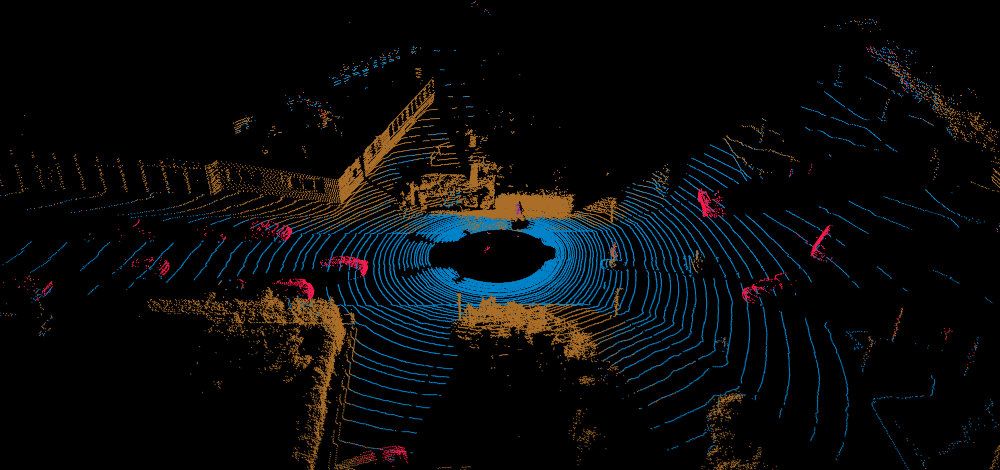} & 
  		\adjincludegraphics[width=.33\linewidth, trim={{.01\width} {.01\height} {.01\width} {.01\height}}, clip]{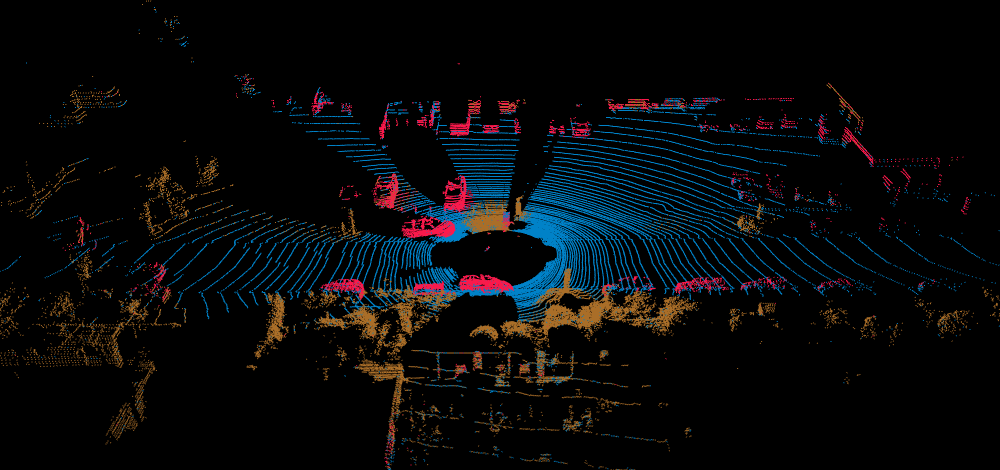} & 
  		\adjincludegraphics[width=.33\linewidth, trim={{.01\width} {.01\height} {.01\width} {.01\height}}, clip]{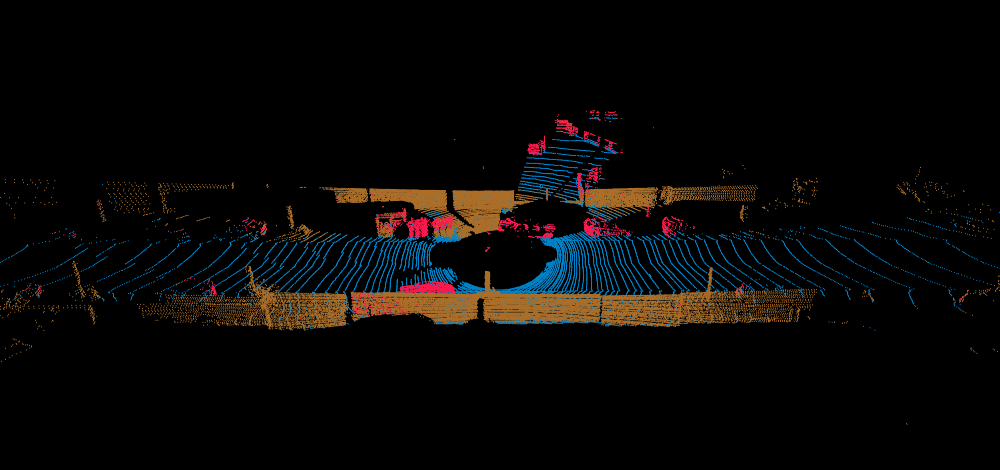} \\ 
  		\adjincludegraphics[width=.33\linewidth, trim={{.01\width} {.01\height} {.01\width} {.01\height}}, clip]{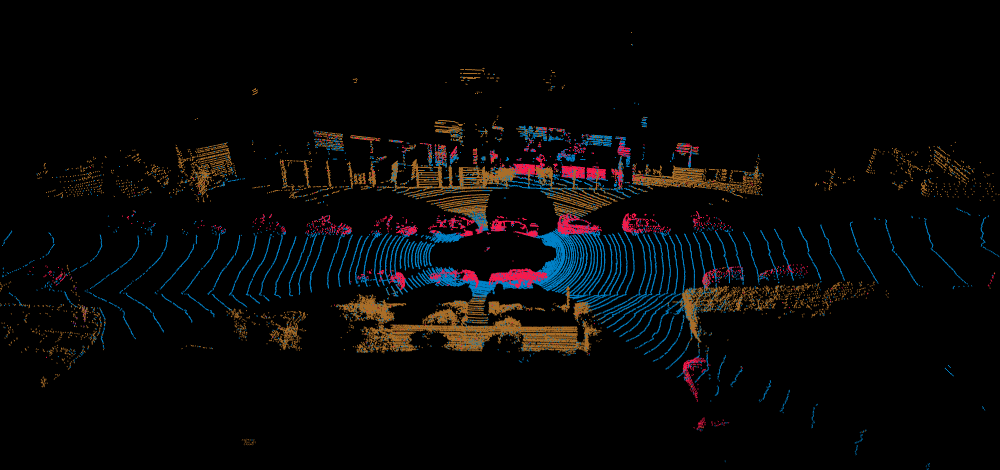} & 
  		\adjincludegraphics[width=.33\linewidth, trim={{.01\width} {.01\height} {.01\width} {.01\height}}, clip]{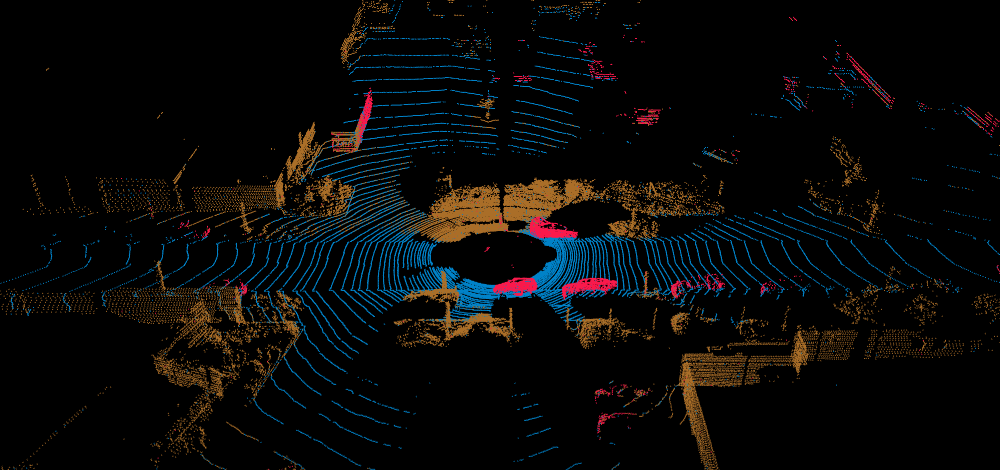} & 
  		\adjincludegraphics[width=.33\linewidth, trim={{.01\width} {.01\height} {.01\width} {.01\height}}, clip]{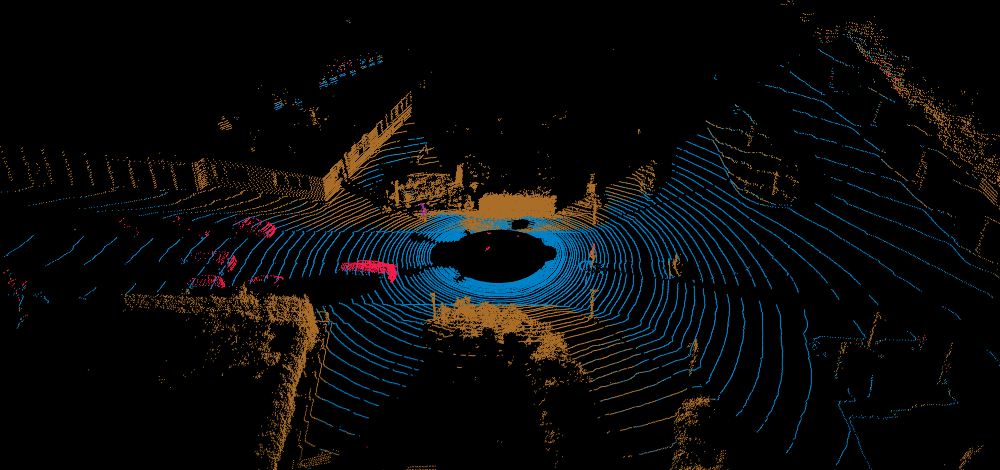} \\ 
   		\end{tabular}
	\vspace{-3mm}
	\caption{Semantic Segmentation on KITTI Dataset}
	\label{fig:seg-kitti}
\end{figure*}

Under all the settings, our algorithm is able to generalize well. \figref{fig:flow-kitti} shows our lidar flow model's performance on KITTI. As shown in \figref{fig:flow-kitti}, our Lidar flow model generalizes well to the unseen KITTI dataset. From left to right, the figures show the most common scenarios with moving, turning, and stationary ego-car. The model produces covincing flow predictions in all three cases. \figref{fig:seg-kitti} includes some additional segmentation results on KITTI. 

For more results over several sequences, please refer to our supplementary video. 

\section{Lidar Flow Data and Analysis}

\paragraph{Ground-truth Generation} In this paragraph, we describe how we generate the ground-truth lidar flow data in detail. For each consecutive frame, we first get the global vehicle pose transform from frame 0 to frame 1: $\mathbf{R}_{\mathrm{ego}}, \mathbf{t}_\mathrm{ego}$ with the help of additional sensors and prior information. This global vehicle pose transform represents how far away the vehicle moves and how the vehicle turns. 
This localization accuracy is at centi-meter scale. Therefore, the motion per each static point is: 
\[
\mathbf{f}_{\mathrm{static-gt}}^{(0)} = \mathbf{R}_\mathrm{ego}^T (\mathbf{x}^{(0)}_{static} - \mathbf{t}_\mathrm{ego}) - \bx^{(0)}_{static}
\]
where $\mathbf{f}_{\mathrm{gt}}^{(k)}$ is the ground-truth flow at the frame $k$ under the ego-car centered coordinate, $\mathbf{x}^{(k)}$ is the point's location at the frame $k$ under the ego-car centered coordinate.

\begin{figure*}
\centering
\includegraphics[width=.9\linewidth]{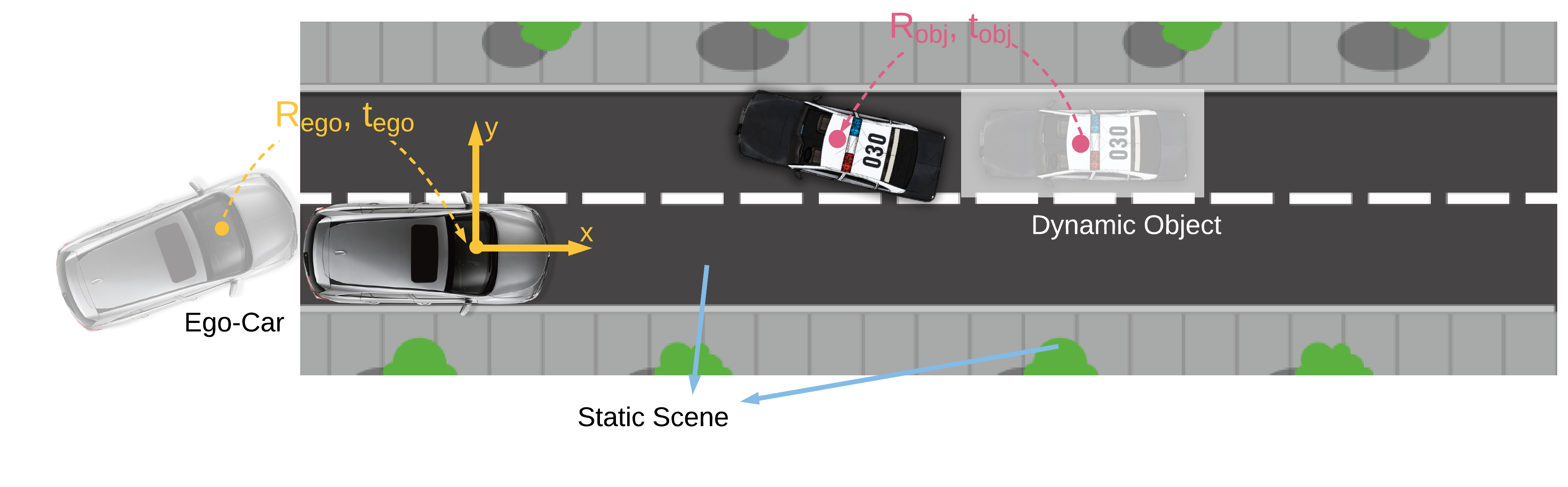} 
\caption{Flow data generation. The source of motion comes from two components: motion of the ego-car and motion of the dynamic objects. } %
\label{flow:data}
\end{figure*}

For dynamic objects in the scene, e.g. other vehicles and pedestrians, the motion between each lidar frame in the vehicle coordinate is not only due to the self-driving car's ego-motion. The movement of the dynamic objects themselves are also contributing to the motion. In this project, we assume rigid motion for all the objects. The labeling of the dynamics objects include two steps. Firstly, using the global pose that we get, we visualize the point cloud of 3D objects from two frames in the same reference coordinate and label the pose changes $\mathbf{R}_\mathrm{obj},\mathbf{t}_\mathrm{obj}$ between the objects at the different time. Secondly, both ego-motion and object motion are considered in order to generate the ground-truth flow vector:
\[
\mathbf{f}_{\mathrm{dynamic-gt}}^{(0)} = \mathbf{R}_\mathrm{obj}^T (\mathbf{R}_\mathrm{ego}^T (\mathbf{x}^{(0)}_{dynamic} - \mathbf{t}_\mathrm{ego}) - \mathbf{t}_\mathrm{obj}) - \bx^{(0)}_{dynamic}
\]
Please refer to Fig.~\ref{flow:data} for an illustration. 

\begin{figure*}
\centering
\includegraphics[width=.9\linewidth]{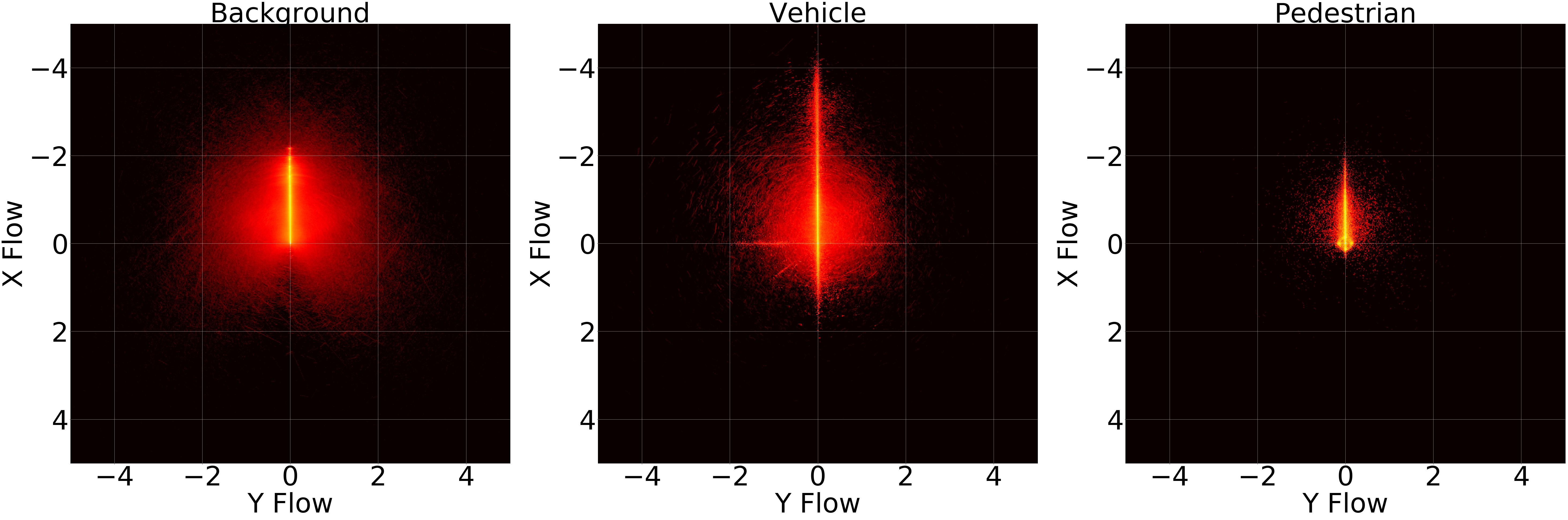} 
\caption{Ground-truth Motion Distribution of the Lidar Flow Dataset (unit in meters)} %
\label{flow:gtmotion}
\end{figure*}

\paragraph{Ground-truth Motion Analysis} We also conduct an analysis over the ground-truth motion distribution. In Fig.~\ref{flow:gtmotion} we show the 2D histogram of the GT 3D translation component along $x$ and $y$ axis respectively. We also show the motion distribution across different object types, e.g. static background, vehicle and pedestrian. As we can see, different semantic types have different motion patterns. And the heaviest density of distribution is on the y-axis, which suggests the forward motion is the major motion pattern of our ego-car. 

\begin{figure*}
	\footnotesize
	\setlength\tabcolsep{0.5pt} %
	\renewcommand{\arraystretch}{0.8}
	\begin{tabular}{ccc}
  		\adjincludegraphics[width=.33\linewidth, trim={{.01\width} {.01\height} {.01\width} {.01\height}}, clip]{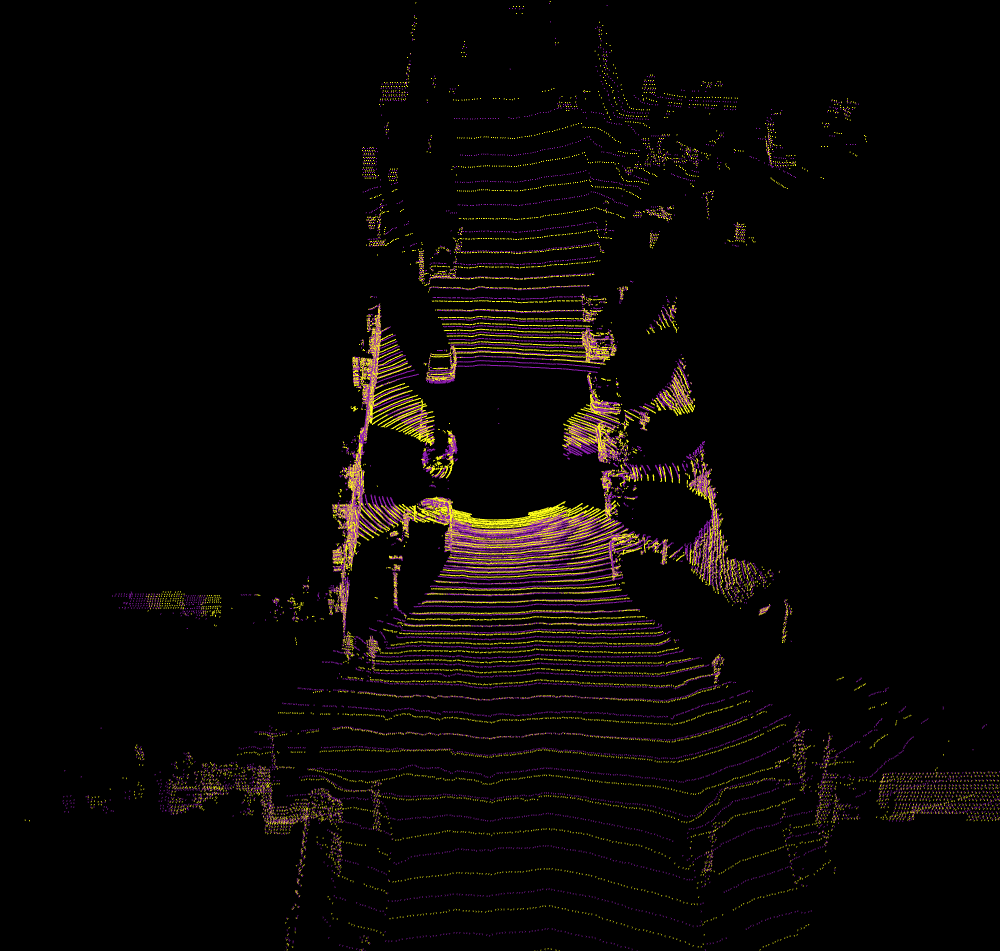} & 
      \adjincludegraphics[width=.33\linewidth, trim={{.01\width} {.01\height} {.01\width} {.01\height}}, clip]{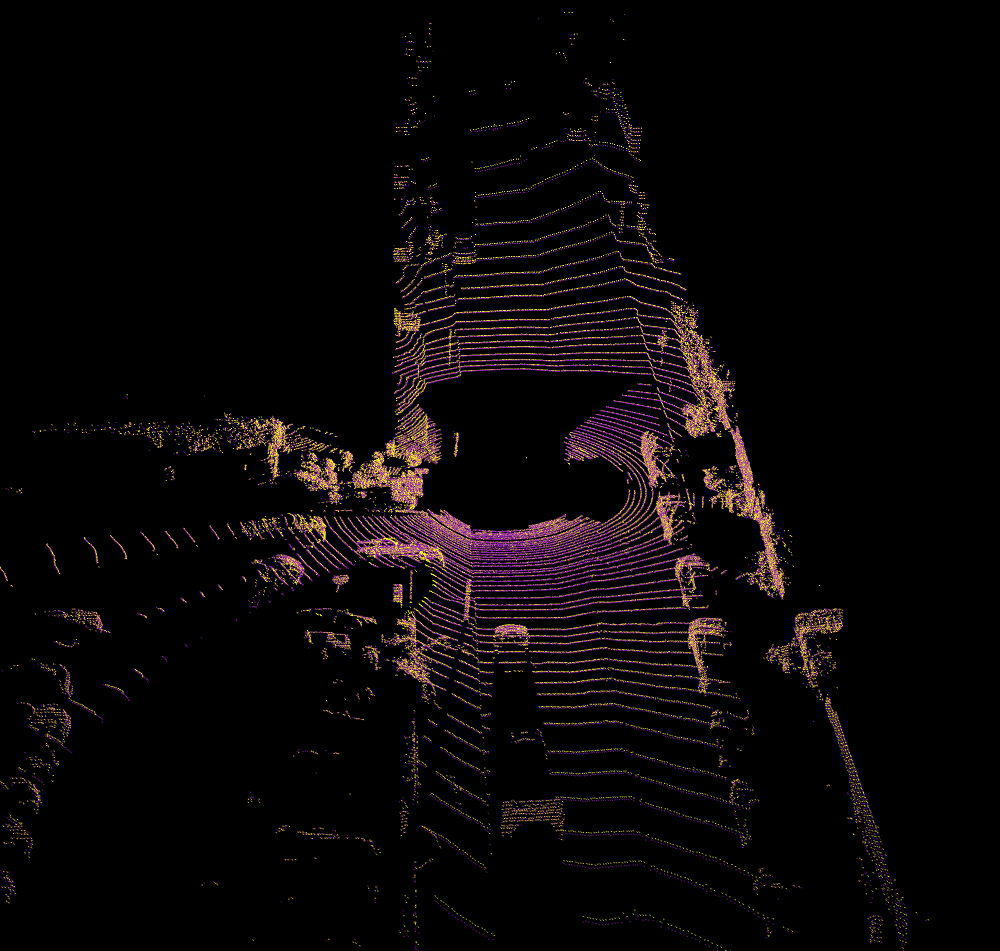} & 
      \adjincludegraphics[width=.33\linewidth, trim={{.01\width} {.01\height} {.01\width} {.01\height}}, clip]{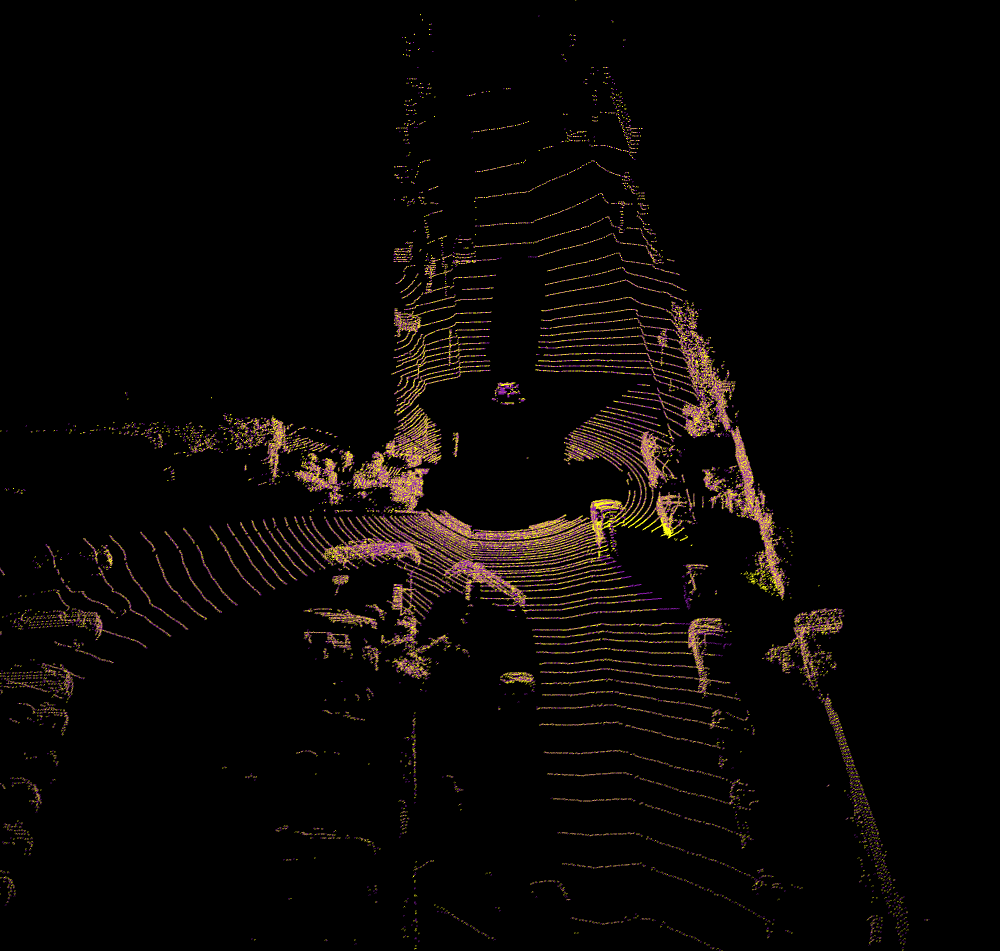} \\

      \adjincludegraphics[width=.33\linewidth, trim={{.3\width} {.3\height} {.3\width} {.3\height}}, clip]{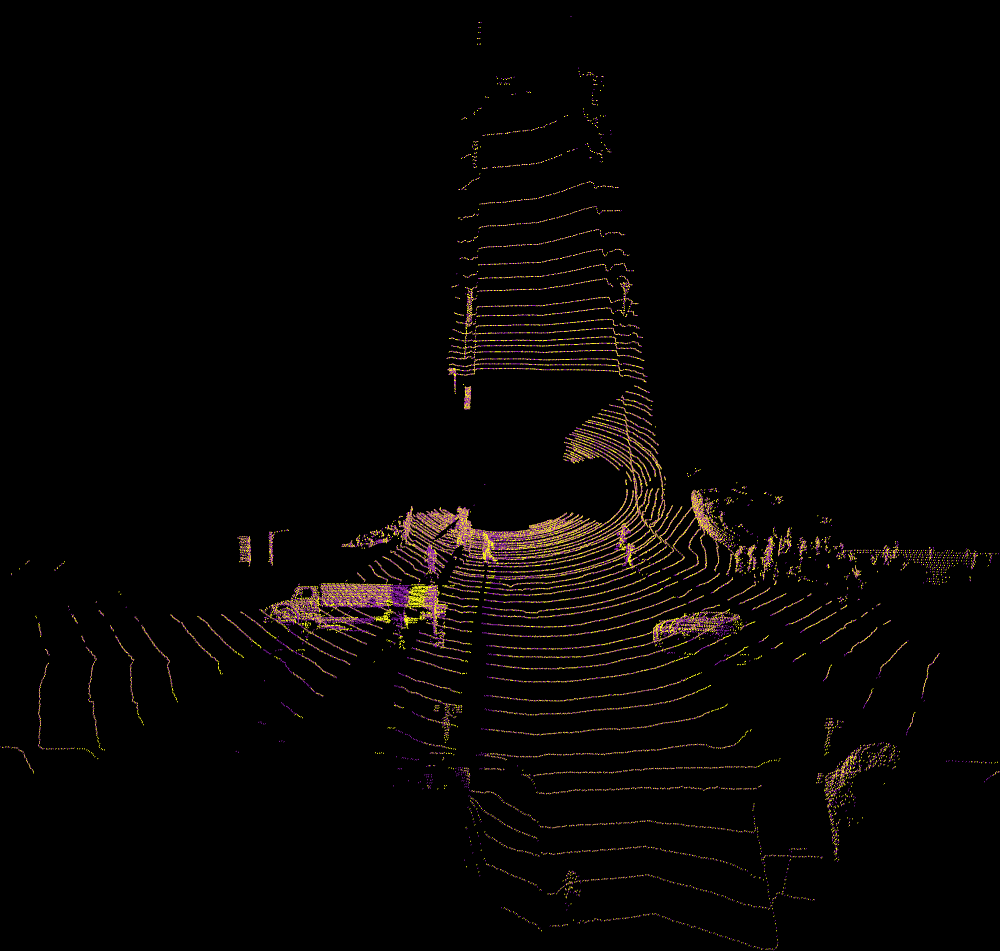} & 
  		\adjincludegraphics[width=.33\linewidth, trim={{.3\width} {.3\height} {.3\width} {.3\height}}, clip]{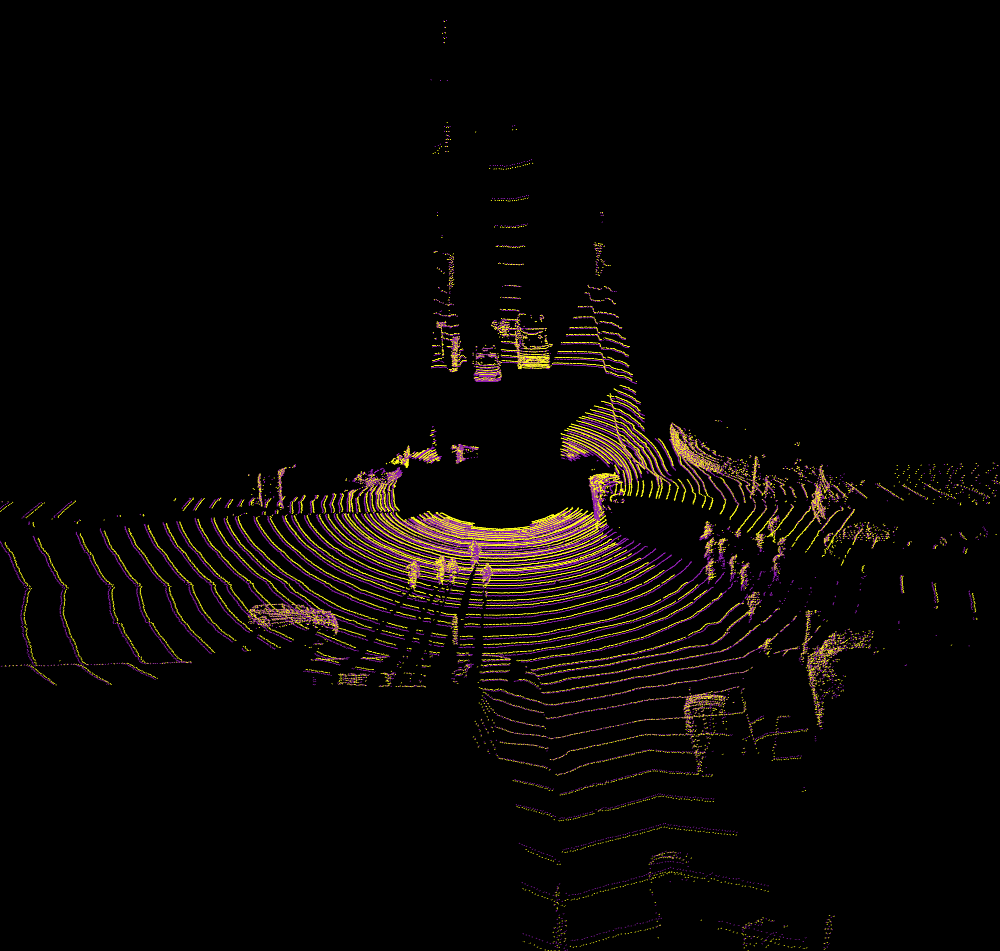} & 
  		\adjincludegraphics[width=.33\linewidth, trim={{.3\width} {.3\height} {.3\width} {.3\height}}, clip]{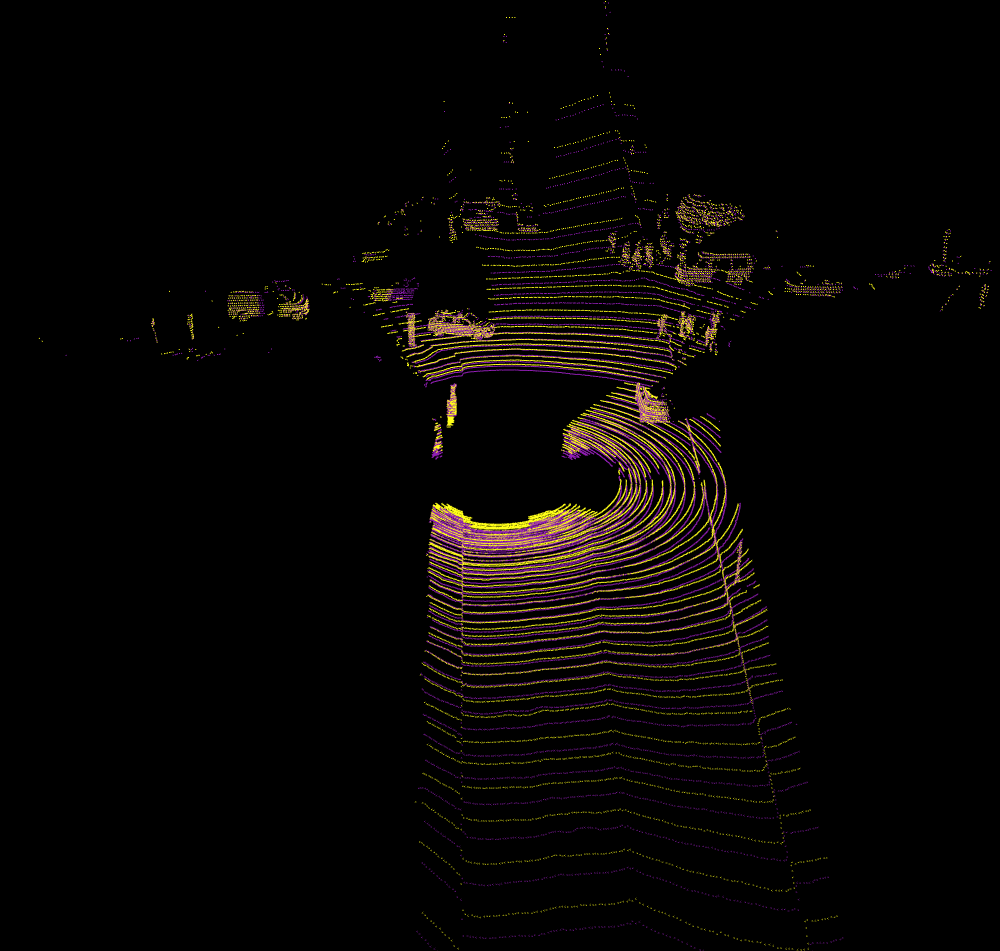} \\
   		\end{tabular}
	\vspace{-3mm}
	\caption{Flow Ground Truth Overlay of Entire Scene. Yellow: target frame, purple: warped source frame.}
	\label{fig:flow-gt-overlay}
\end{figure*}

\begin{figure*}
	\footnotesize
	\setlength\tabcolsep{0.5pt} %
	\renewcommand{\arraystretch}{0.8}
	\begin{tabular}{ccccc}
  		\adjincludegraphics[width=.20\linewidth, trim={{.01\width} {.01\height} {.01\width} {.01\height}}, clip]{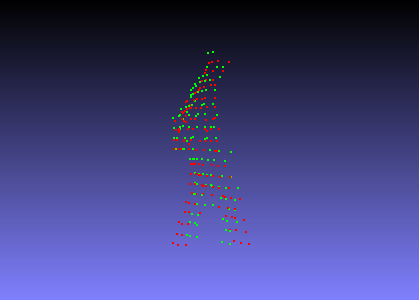} & 
  		\adjincludegraphics[width=.20\linewidth, trim={{.01\width} {.01\height} {.01\width} {.01\height}}, clip]{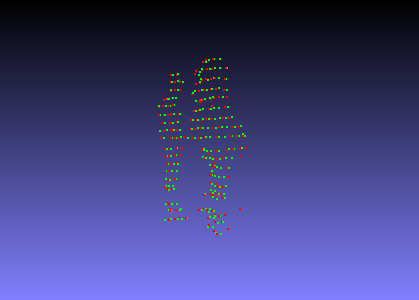} & 
  		\adjincludegraphics[width=.20\linewidth, trim={{.01\width} {.01\height} {.01\width} {.01\height}}, clip]{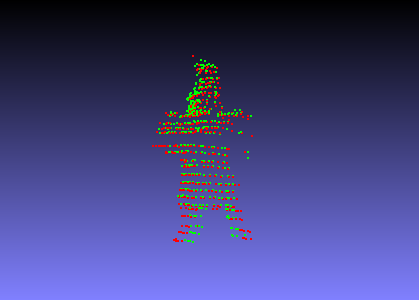} &
  		\adjincludegraphics[width=.20\linewidth, trim={{.01\width} {.01\height} {.01\width} {.01\height}}, clip]{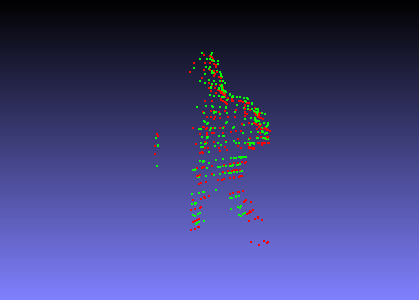} & 
  		\adjincludegraphics[width=.20\linewidth, trim={{.01\width} {.01\height} {.01\width} {.01\height}}, clip]{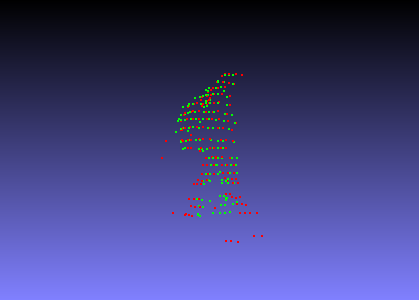} \\

  		\adjincludegraphics[width=.20\linewidth, trim={{.01\width} {.01\height} {.01\width} {.01\height}}, clip]{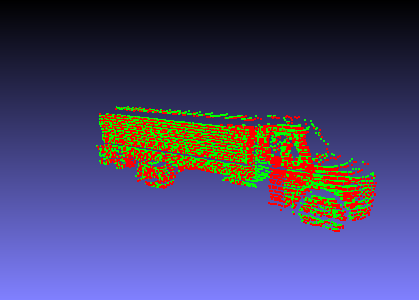} & 
  		\adjincludegraphics[width=.20\linewidth, trim={{.01\width} {.01\height} {.01\width} {.01\height}}, clip]{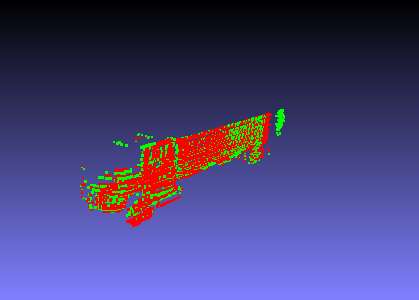} & 
  		\adjincludegraphics[width=.20\linewidth, trim={{.01\width} {.01\height} {.01\width} {.01\height}}, clip]{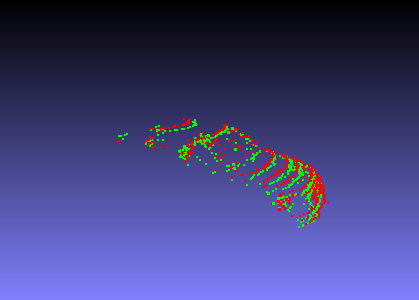} & 
  		\adjincludegraphics[width=.20\linewidth, trim={{.01\width} {.01\height} {.01\width} {.01\height}}, clip]{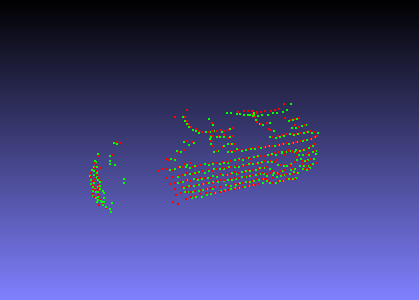} & 
  		\adjincludegraphics[width=.20\linewidth, trim={{.01\width} {.01\height} {.01\width} {.01\height}}, clip]{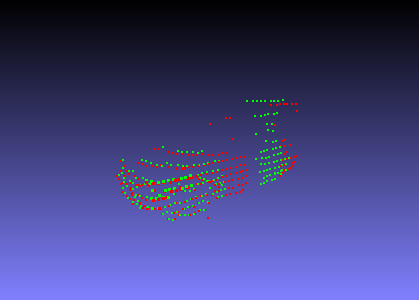} \\
   		\end{tabular}
	\vspace{-3mm}
	\caption{Flow Ground Truth Overlay of Individual Dynamic Objects. Green: target frame, red: warped source frame.}
	\label{fig:flow-gt-overlay-dynamic}
\end{figure*}

\paragraph{Ground-truth Validation and Visualization} 
We validate the quality of our ground-truth motion labels by overlaying target frame points ($\mathbf{x}^{(1)}$) and source frame points warped with ground-truth motion ($\mathbf{x}^{(0)} + \mathbf{f}_{\mathrm{gt}}^{(0)}$). \figref{fig:flow-gt-overlay} shows overlays of entire scenes and \figref{fig:flow-gt-overlay-dynamic} shows overlays of individual dynamic objects. For vehicles, points align perfectly across frames. For pedestrians, the correspondence is also near perfect: the only discrepancy is caused by non-rigid motion (e.g. changing posture).

\section{More Results}

In this section, we show additional qualitative results of the proposed algorithm over all the tasks.

\subsection{Semantic Segmentation for Indoor Scenes}

\begin{figure*}
	\footnotesize
	\setlength\tabcolsep{0.5pt} %
	\renewcommand{\arraystretch}{0.8}
	\begin{tabular}{ccc}
  		\adjincludegraphics[width=.33\linewidth, trim={{.01\width} {.01\height} {.01\width} {.01\height}}, clip]{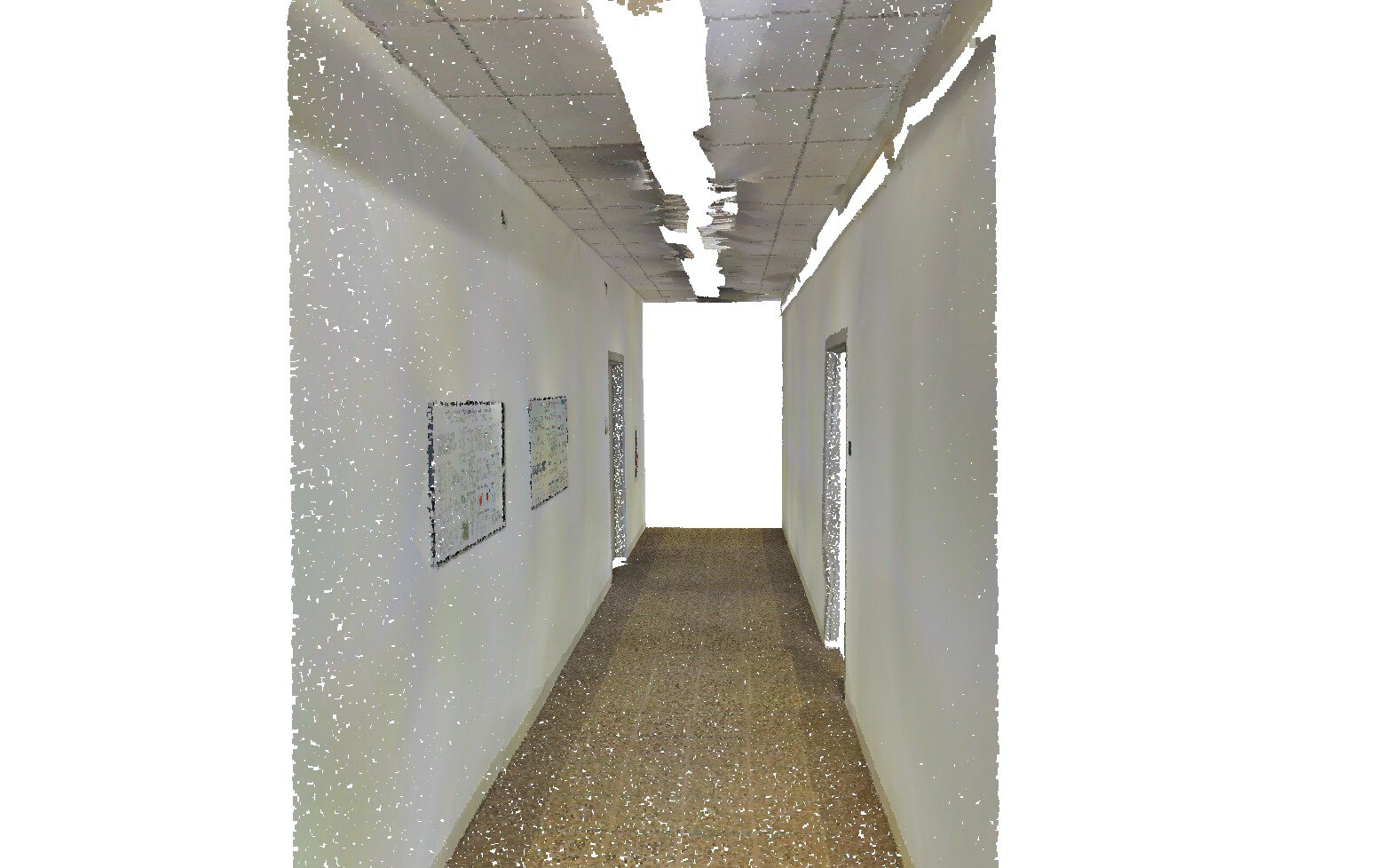} & 
  		\adjincludegraphics[width=.33\linewidth, trim={{.01\width} {.01\height} {.01\width} {.01\height}}, clip]{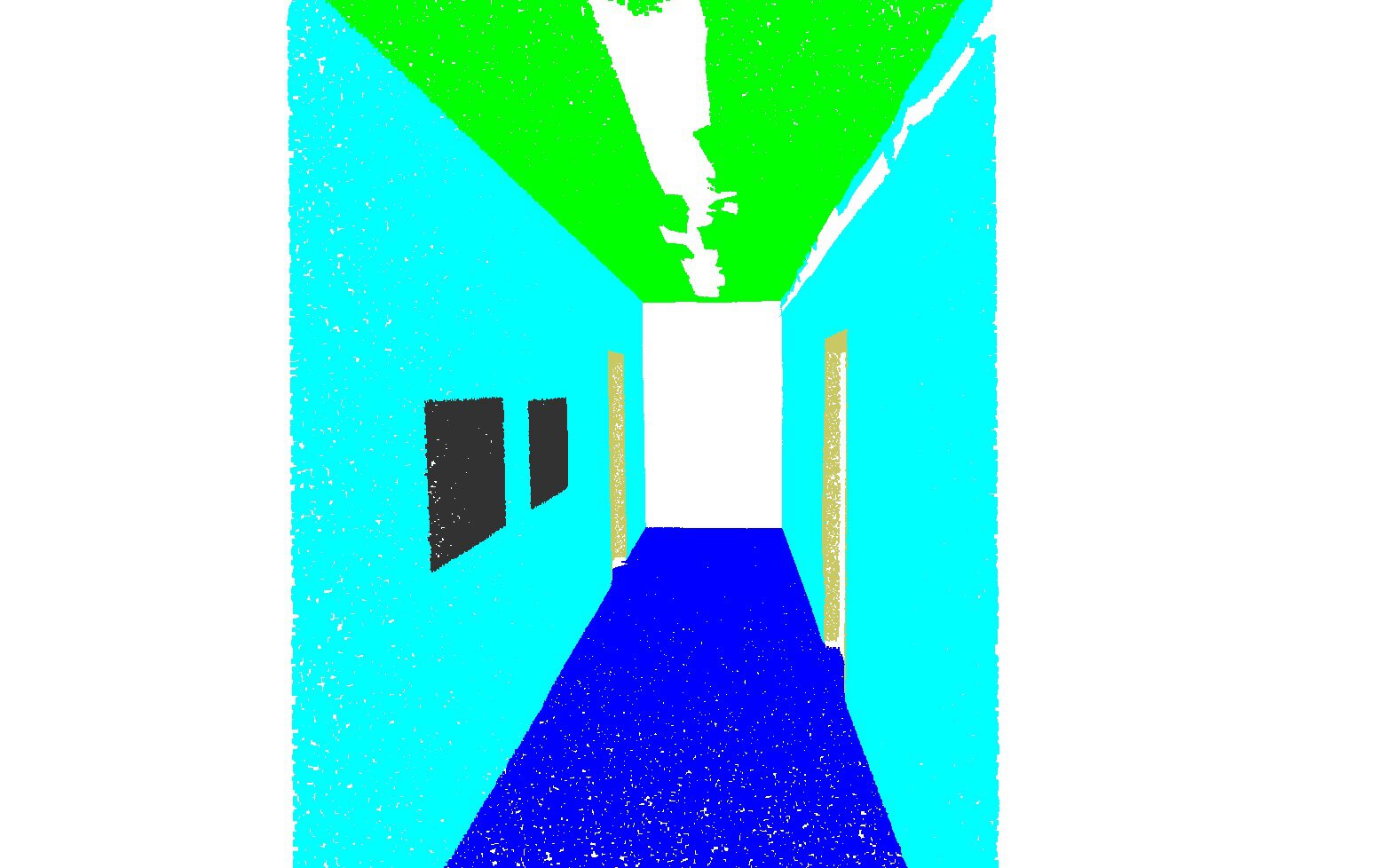} & 
  		\adjincludegraphics[width=.33\linewidth, trim={{.01\width} {.01\height} {.01\width} {.01\height}}, clip]{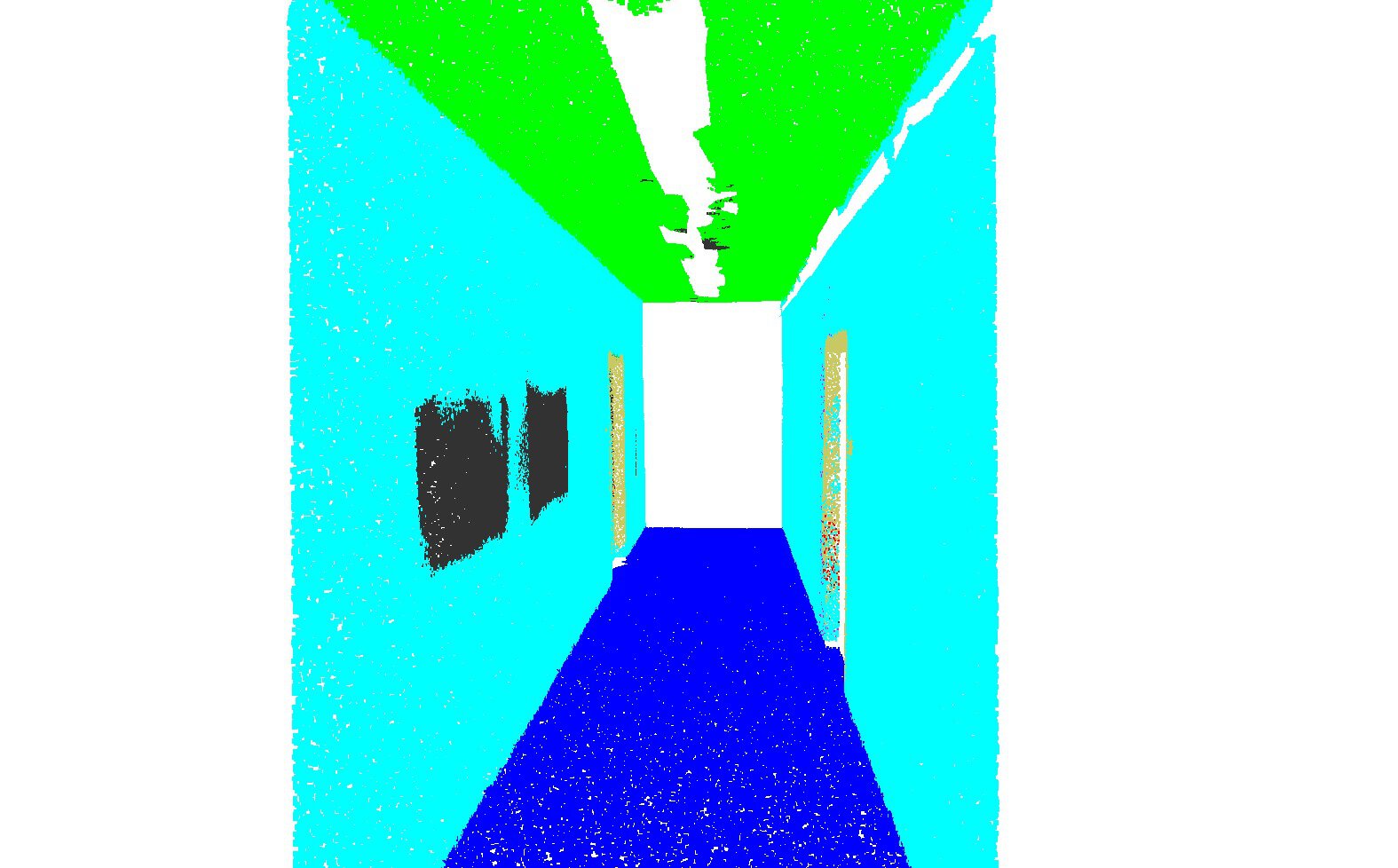} \\
  		\adjincludegraphics[width=.33\linewidth, trim={{.01\width} {.01\height} {.01\width} {.01\height}}, clip]{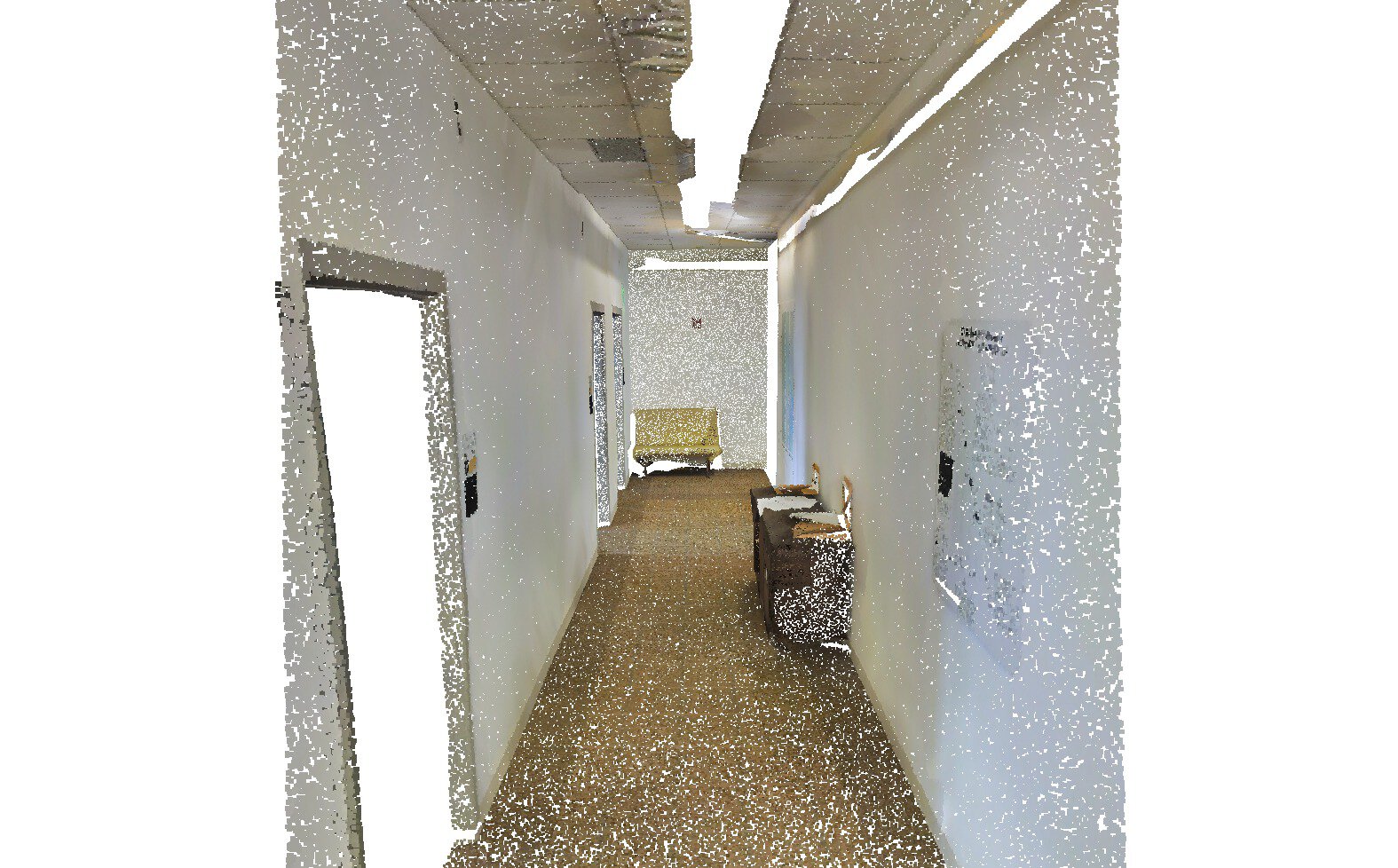} & 
  		\adjincludegraphics[width=.33\linewidth, trim={{.01\width} {.01\height} {.01\width} {.01\height}}, clip]{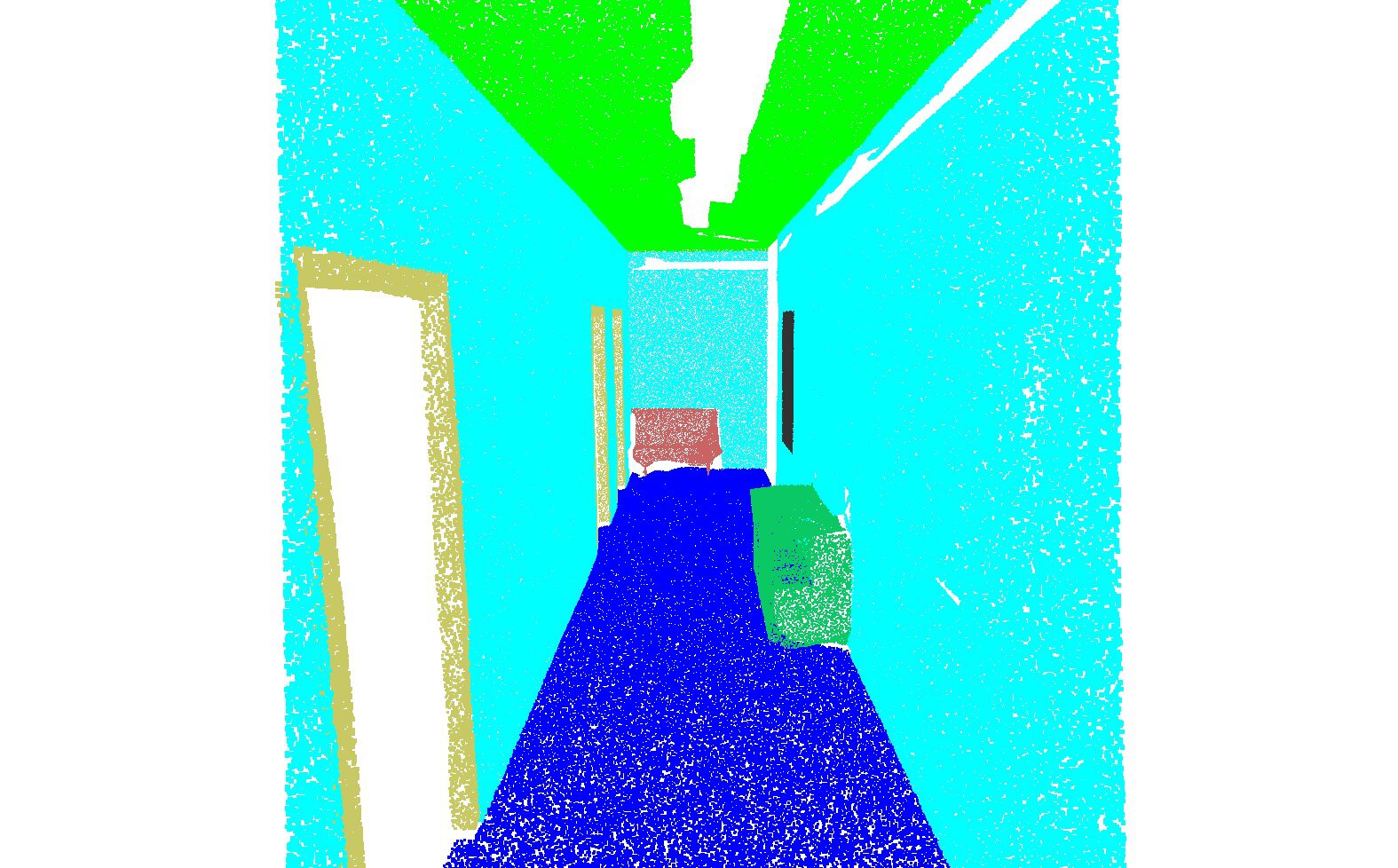} & 
  		\adjincludegraphics[width=.33\linewidth, trim={{.01\width} {.01\height} {.01\width} {.01\height}}, clip]{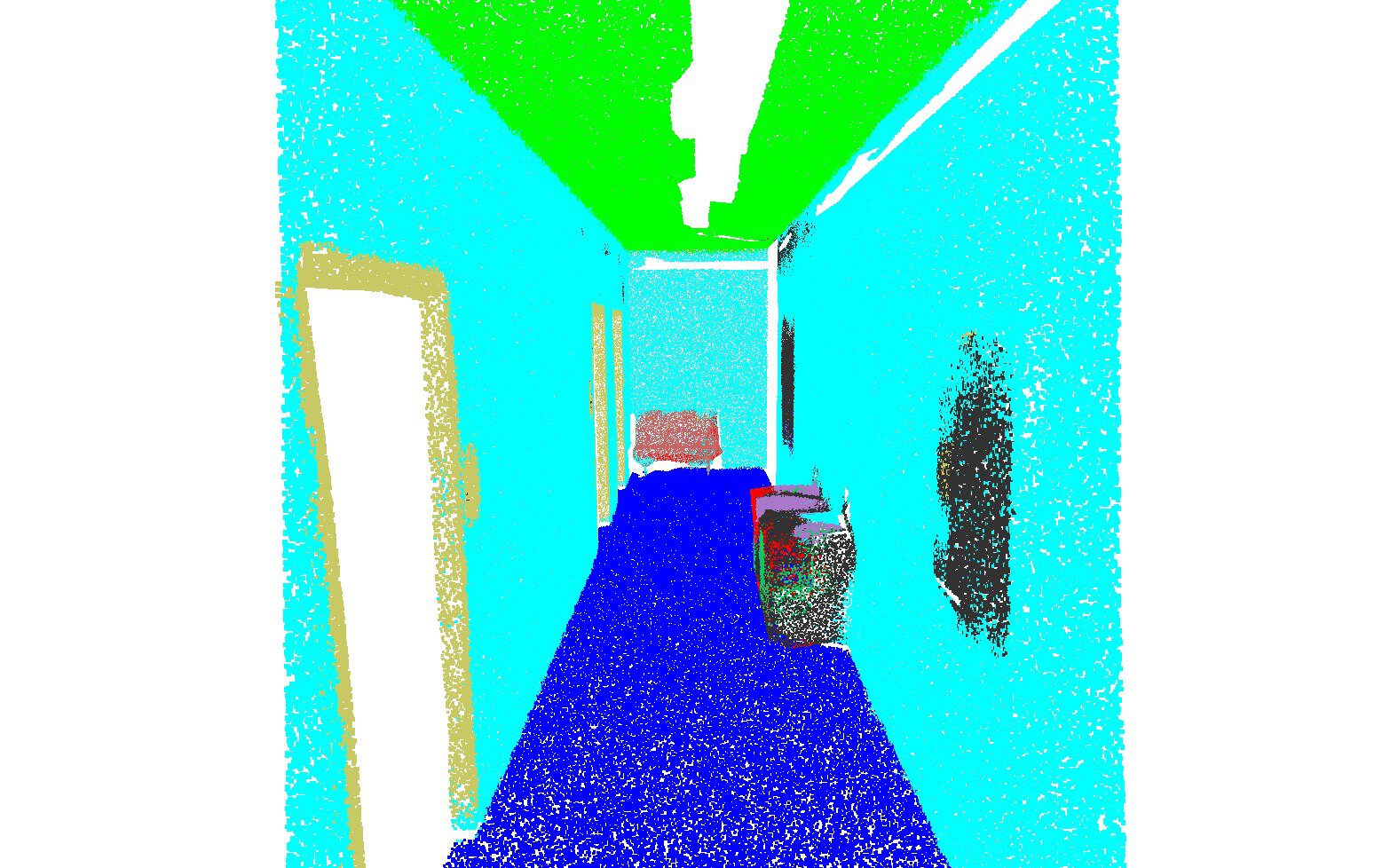} \\
  		\adjincludegraphics[width=.33\linewidth, trim={{.01\width} {.01\height} {.01\width} {.01\height}}, clip]{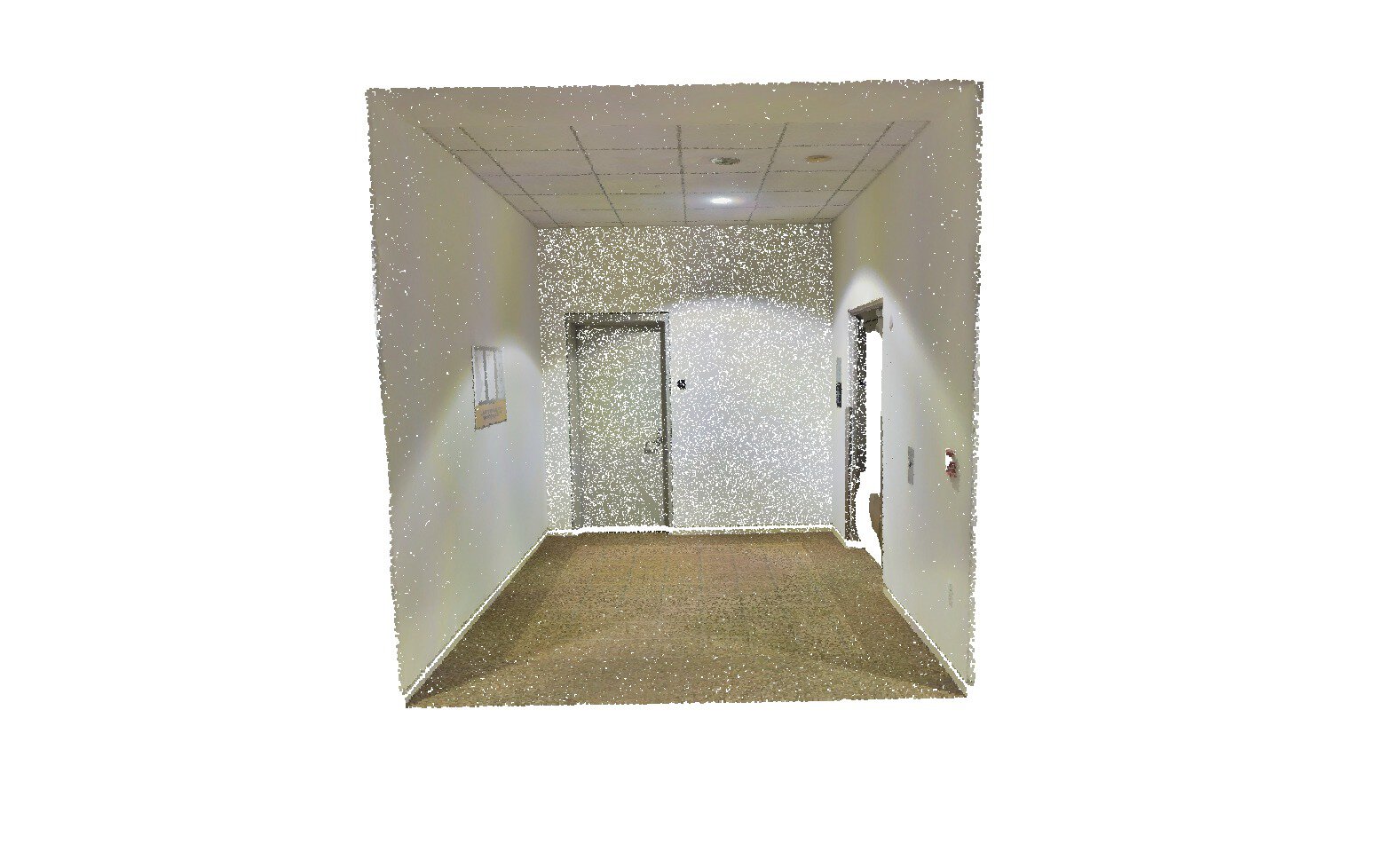} & 
  		\adjincludegraphics[width=.33\linewidth, trim={{.01\width} {.01\height} {.01\width} {.01\height}}, clip]{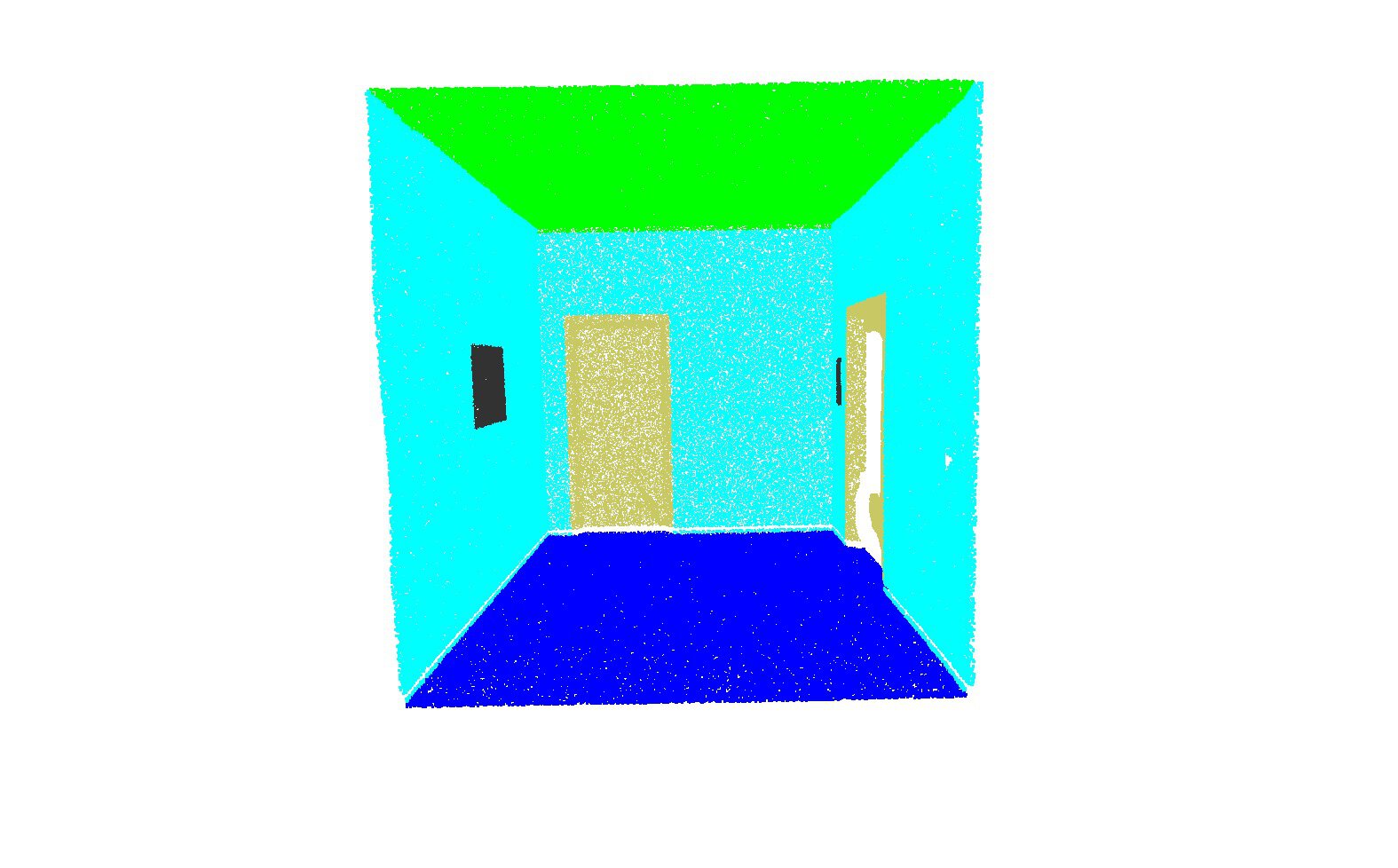} & 
  		\adjincludegraphics[width=.33\linewidth, trim={{.01\width} {.01\height} {.01\width} {.01\height}}, clip]{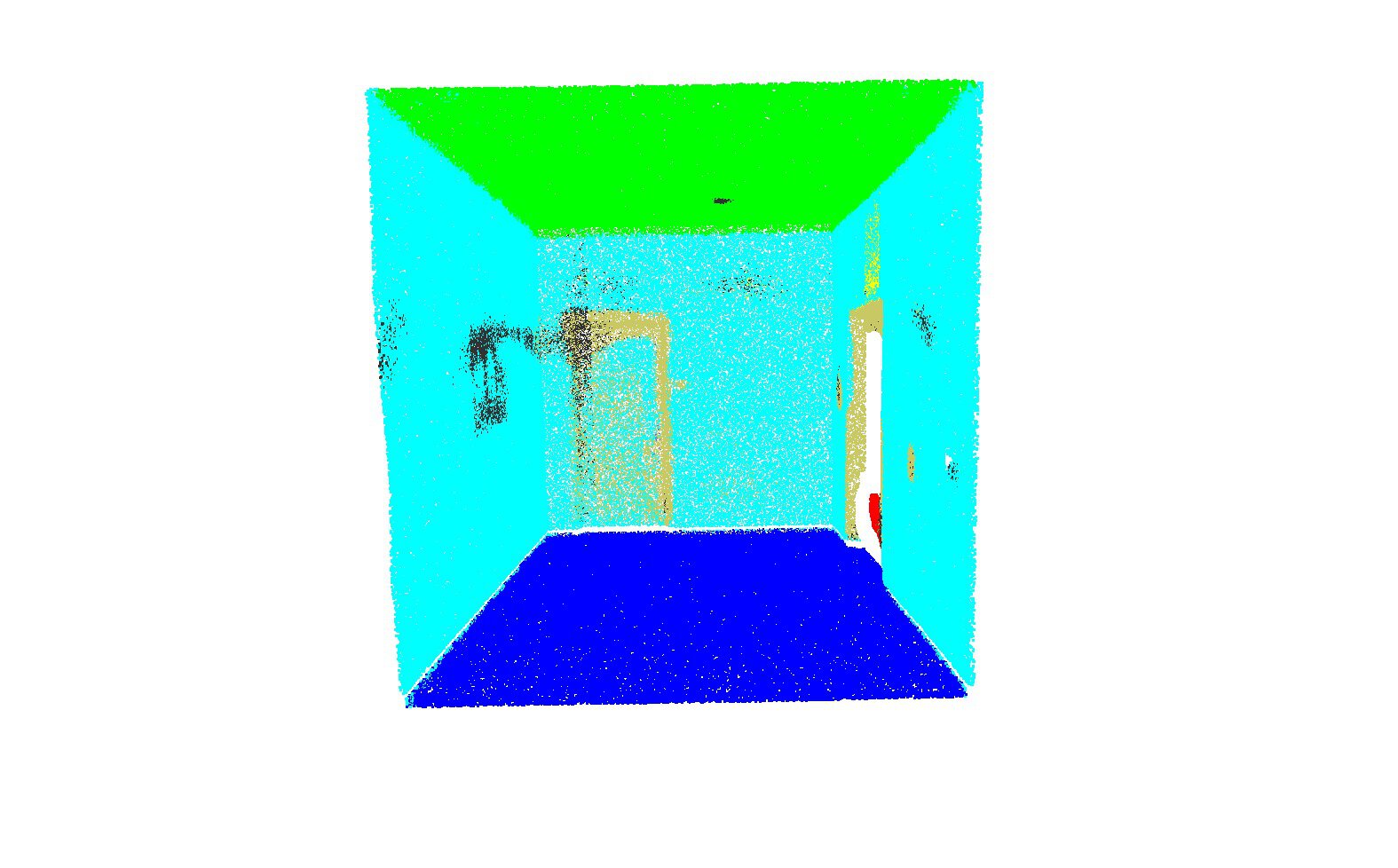} \\
  		\adjincludegraphics[width=.33\linewidth, trim={{.01\width} {.01\height} {.01\width} {.01\height}}, clip]{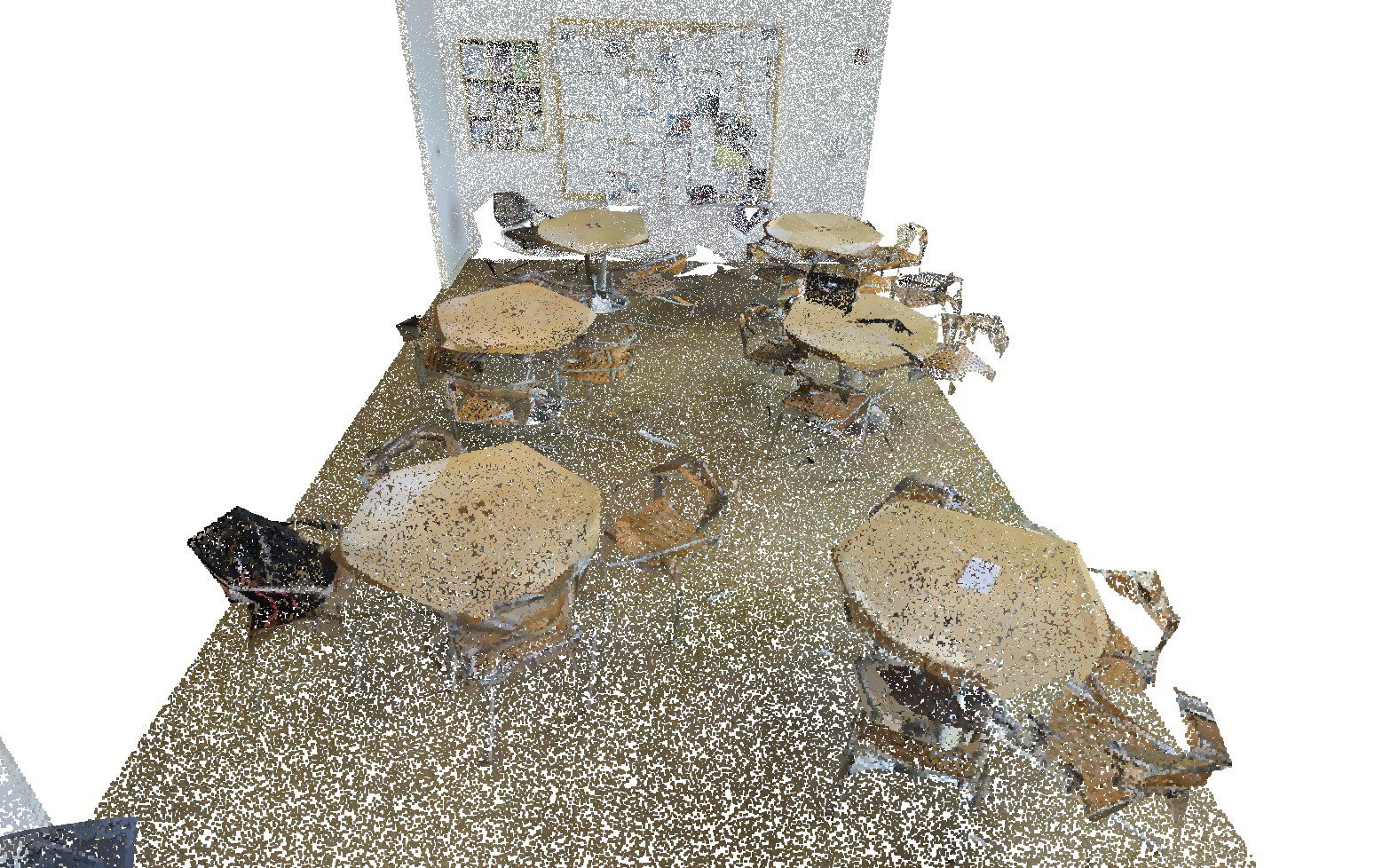} & 
  		\adjincludegraphics[width=.33\linewidth, trim={{.01\width} {.01\height} {.01\width} {.01\height}}, clip]{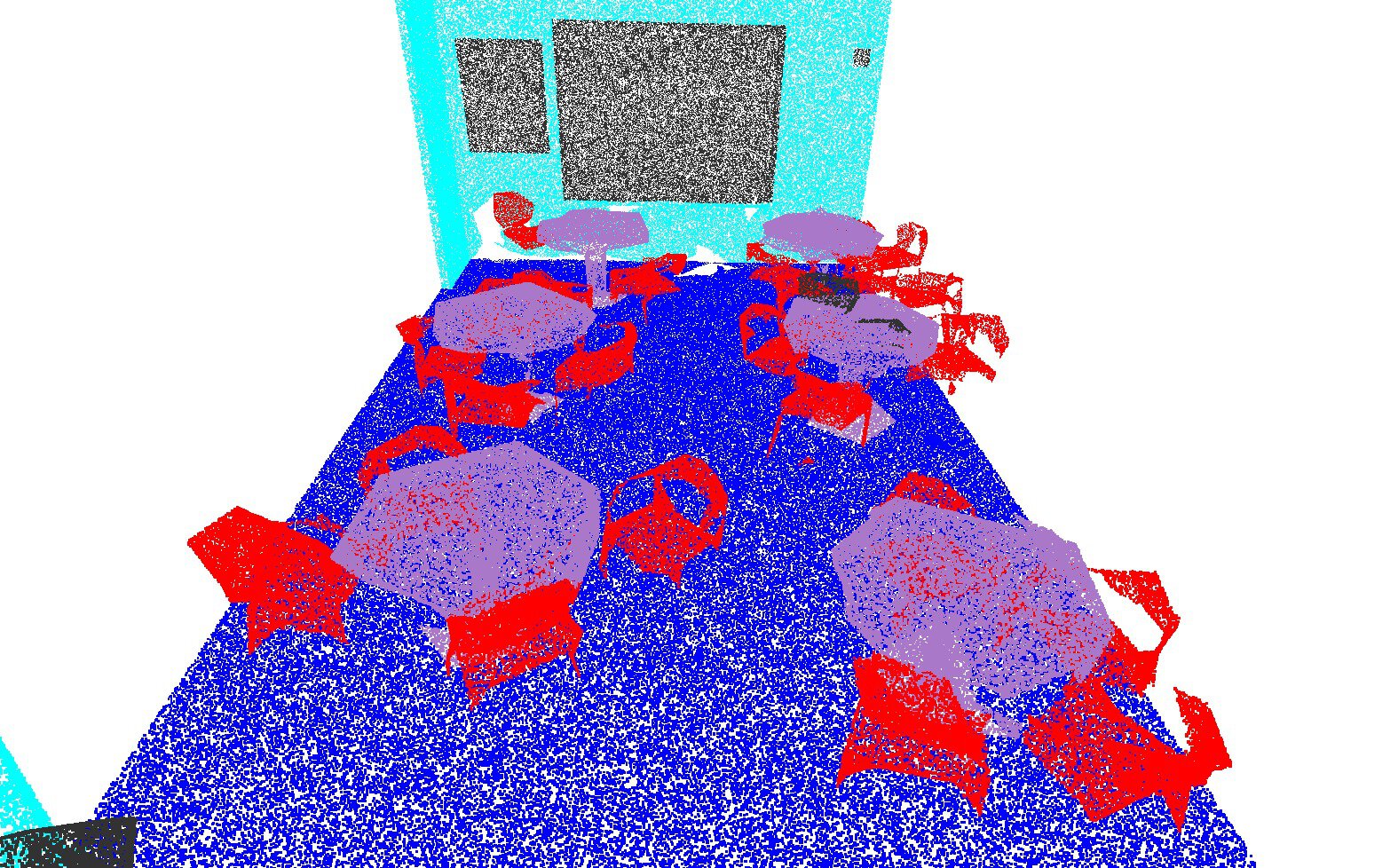} & 
  		\adjincludegraphics[width=.33\linewidth, trim={{.01\width} {.01\height} {.01\width} {.01\height}}, clip]{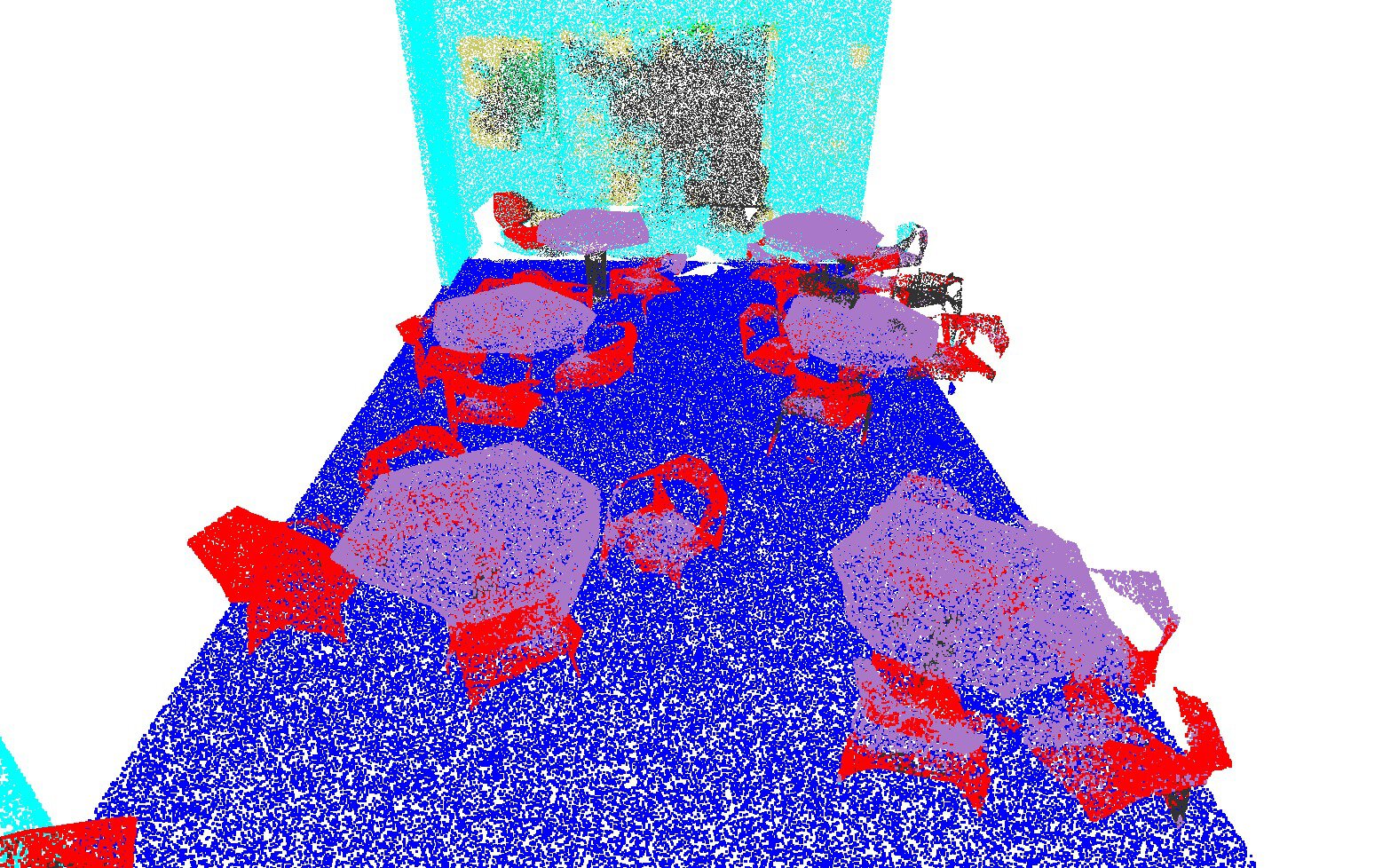} \\
  		\adjincludegraphics[width=.33\linewidth, trim={{.01\width} {.01\height} {.01\width} {.01\height}}, clip]{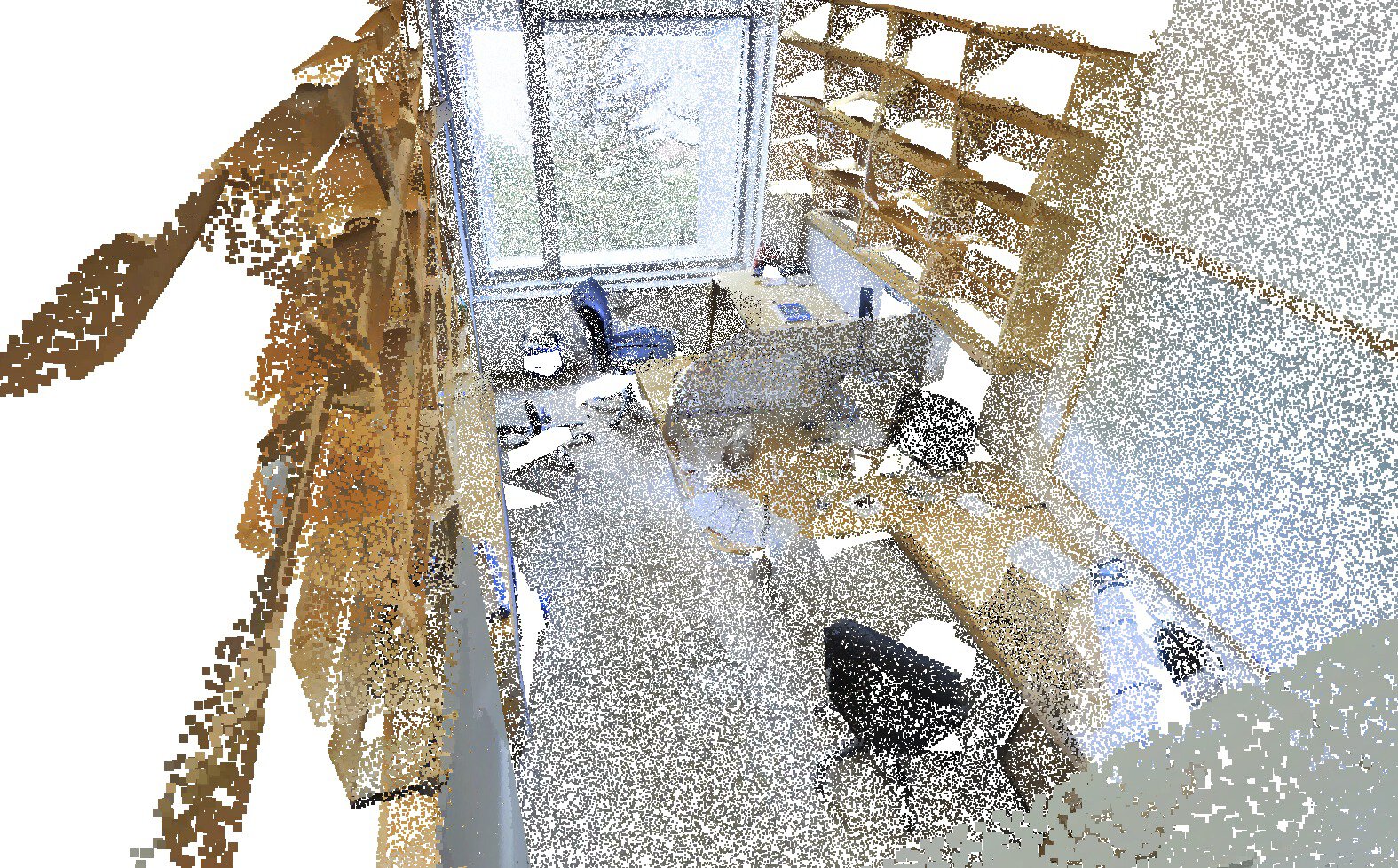} & 
  		\adjincludegraphics[width=.33\linewidth, trim={{.01\width} {.01\height} {.01\width} {.01\height}}, clip]{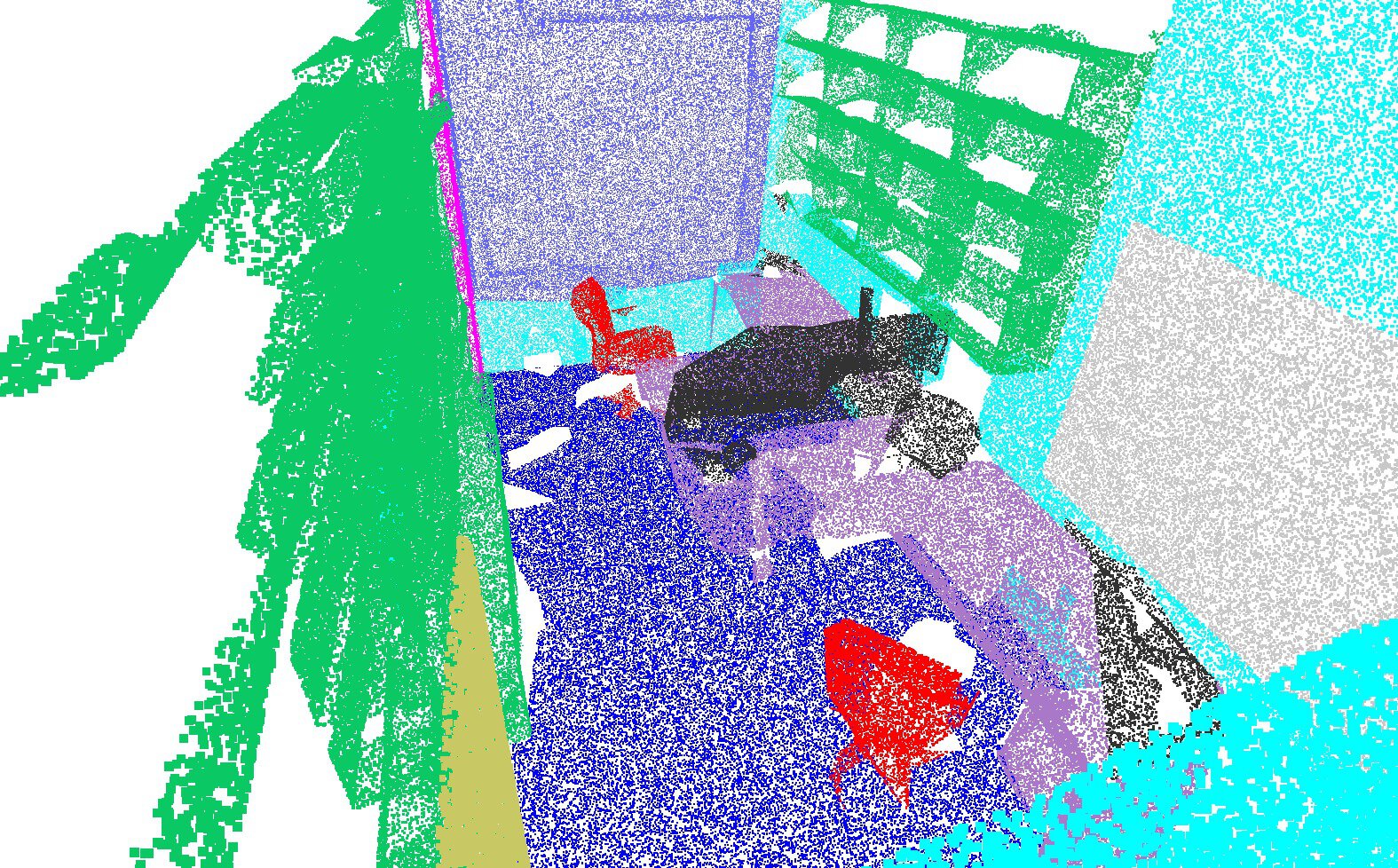} & 
  		\adjincludegraphics[width=.33\linewidth, trim={{.01\width} {.01\height} {.01\width} {.01\height}}, clip]{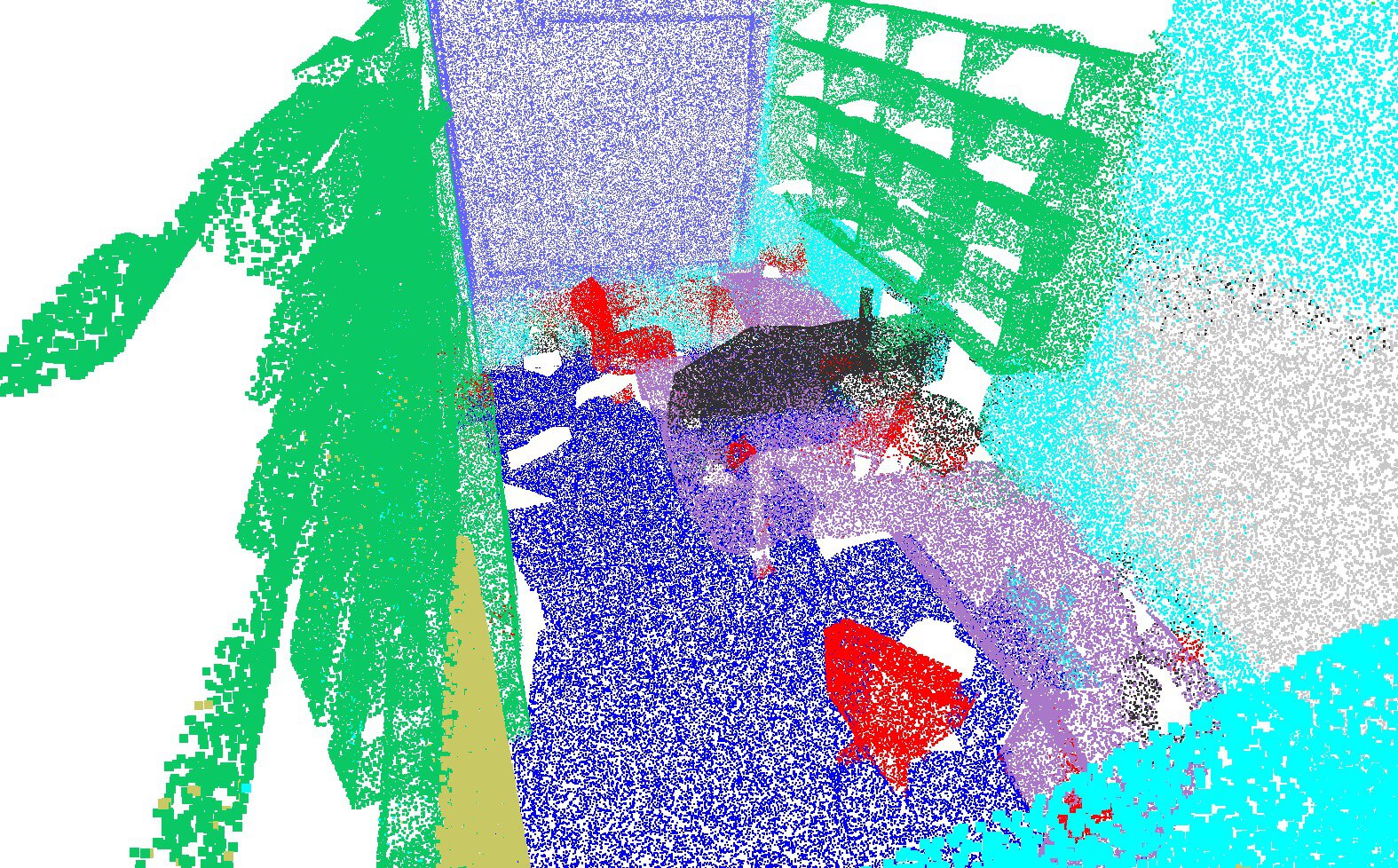} \\
  		\adjincludegraphics[width=.33\linewidth, trim={{.01\width} {.01\height} {.01\width} {.01\height}}, clip]{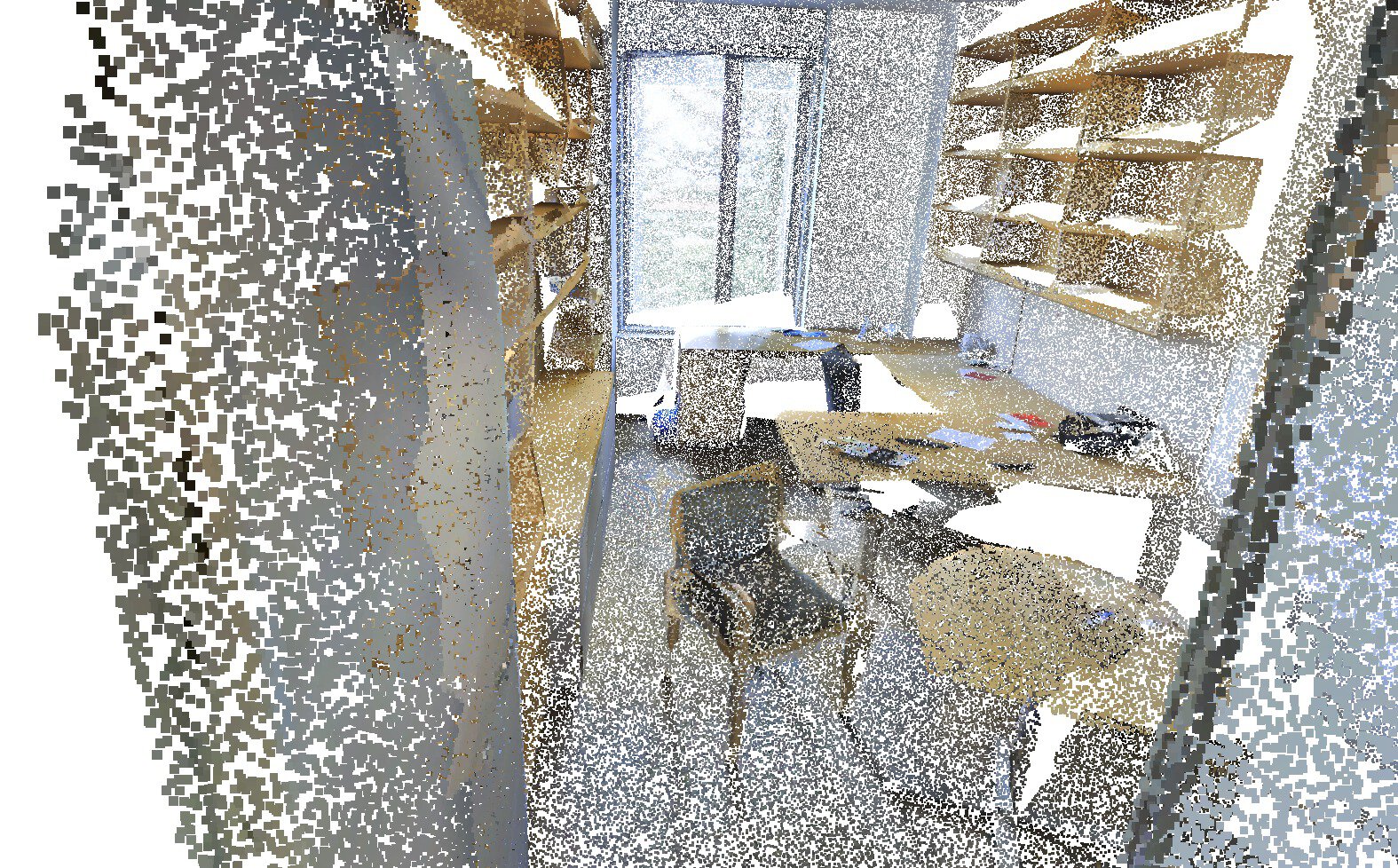} & 
  		\adjincludegraphics[width=.33\linewidth, trim={{.01\width} {.01\height} {.01\width} {.01\height}}, clip]{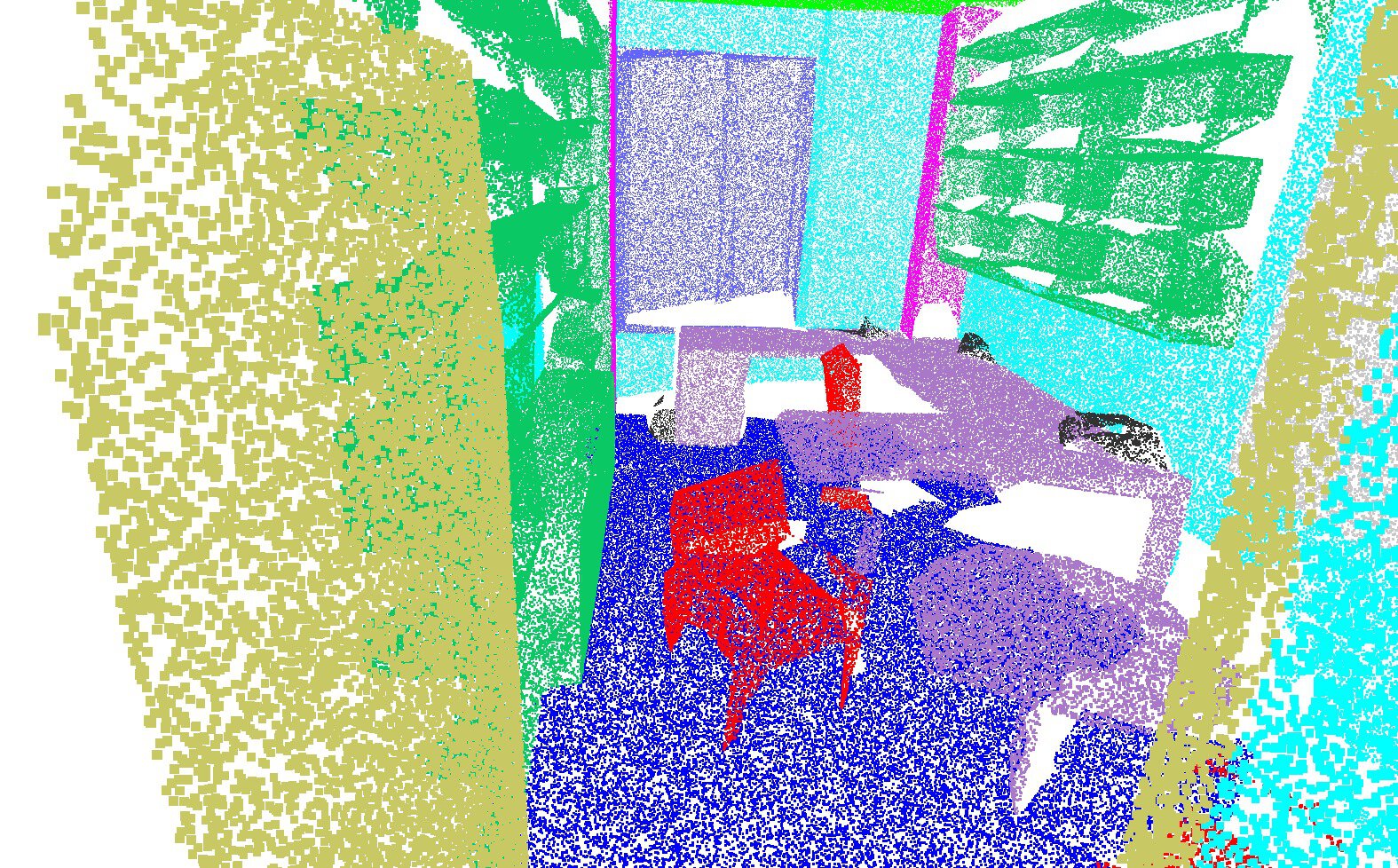} & 
  		\adjincludegraphics[width=.33\linewidth, trim={{.01\width} {.01\height} {.01\width} {.01\height}}, clip]{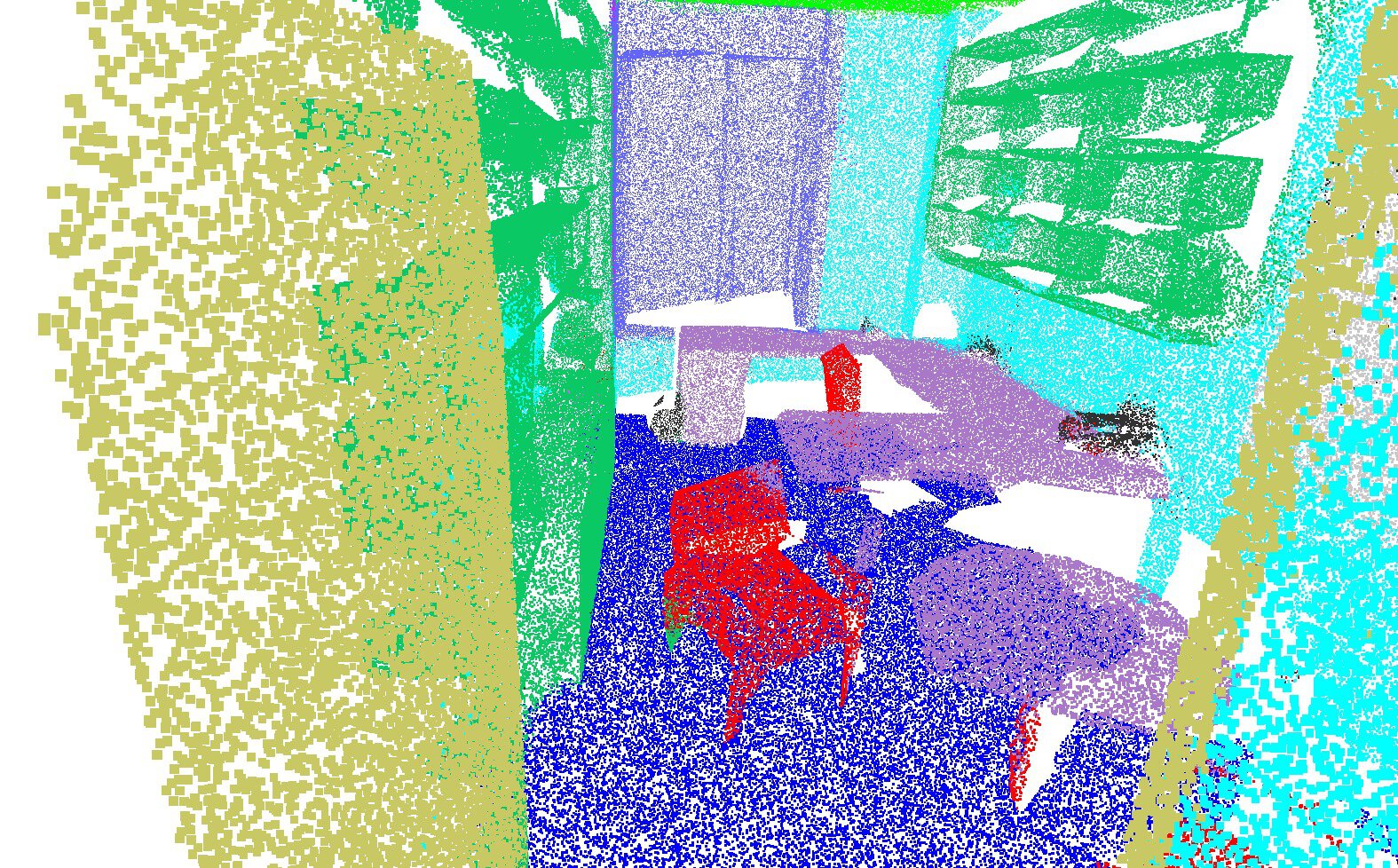} \\
  		Input & GT & Ours
   		\end{tabular}
	\vspace{-3mm}
	\caption{Semantic Segmentation Results on Stanford Indoor Dataset}
	\label{fig:indoor}
\end{figure*}

\begin{figure*}
	\footnotesize
	\setlength\tabcolsep{0.5pt} %
	\renewcommand{\arraystretch}{0.8}
	\begin{tabular}{ccc}
  		\adjincludegraphics[width=.33\linewidth, trim={{.01\width} {.01\height} {.01\width} {.01\height}}, clip]{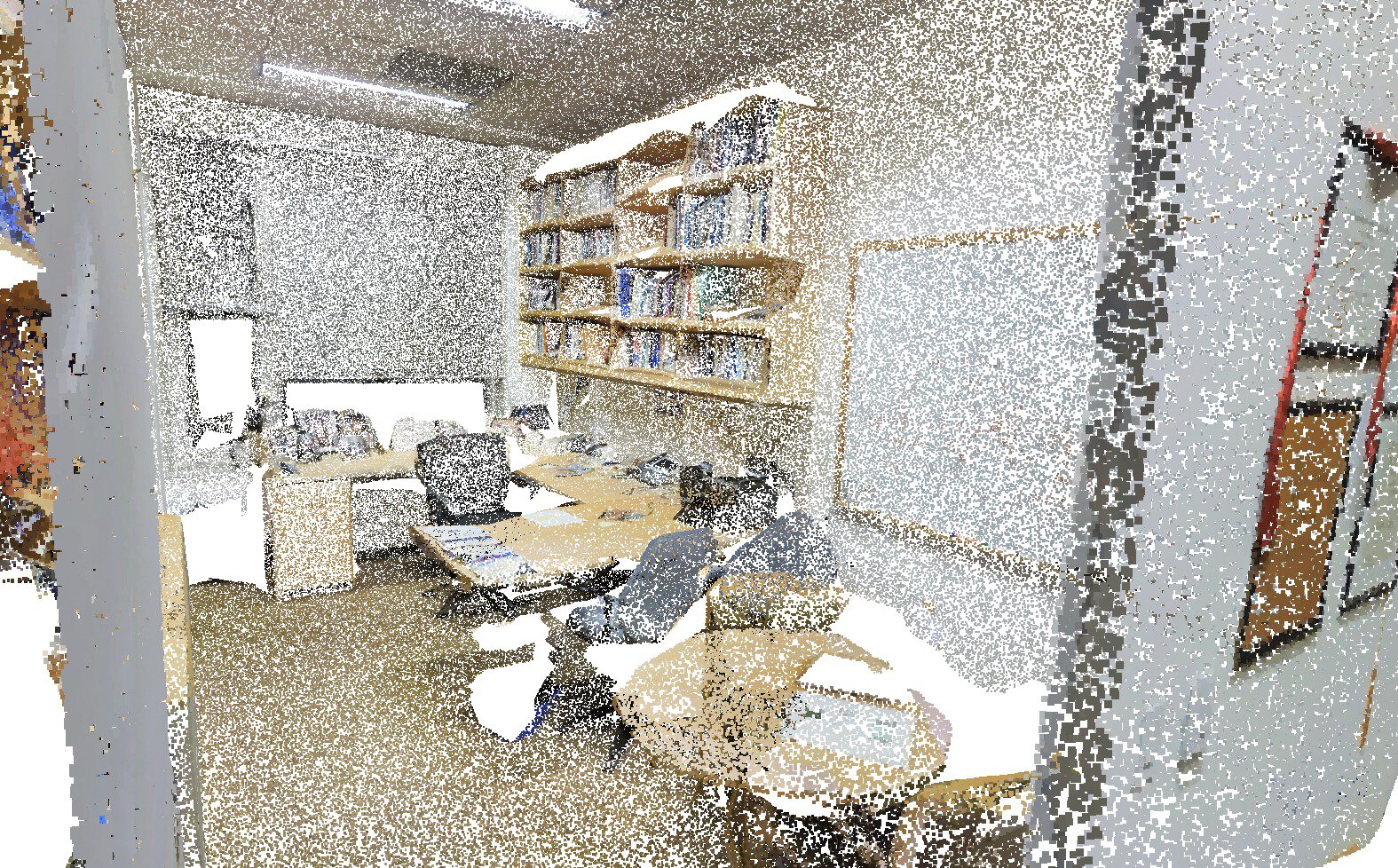} & 
  		\adjincludegraphics[width=.33\linewidth, trim={{.01\width} {.01\height} {.01\width} {.01\height}}, clip]{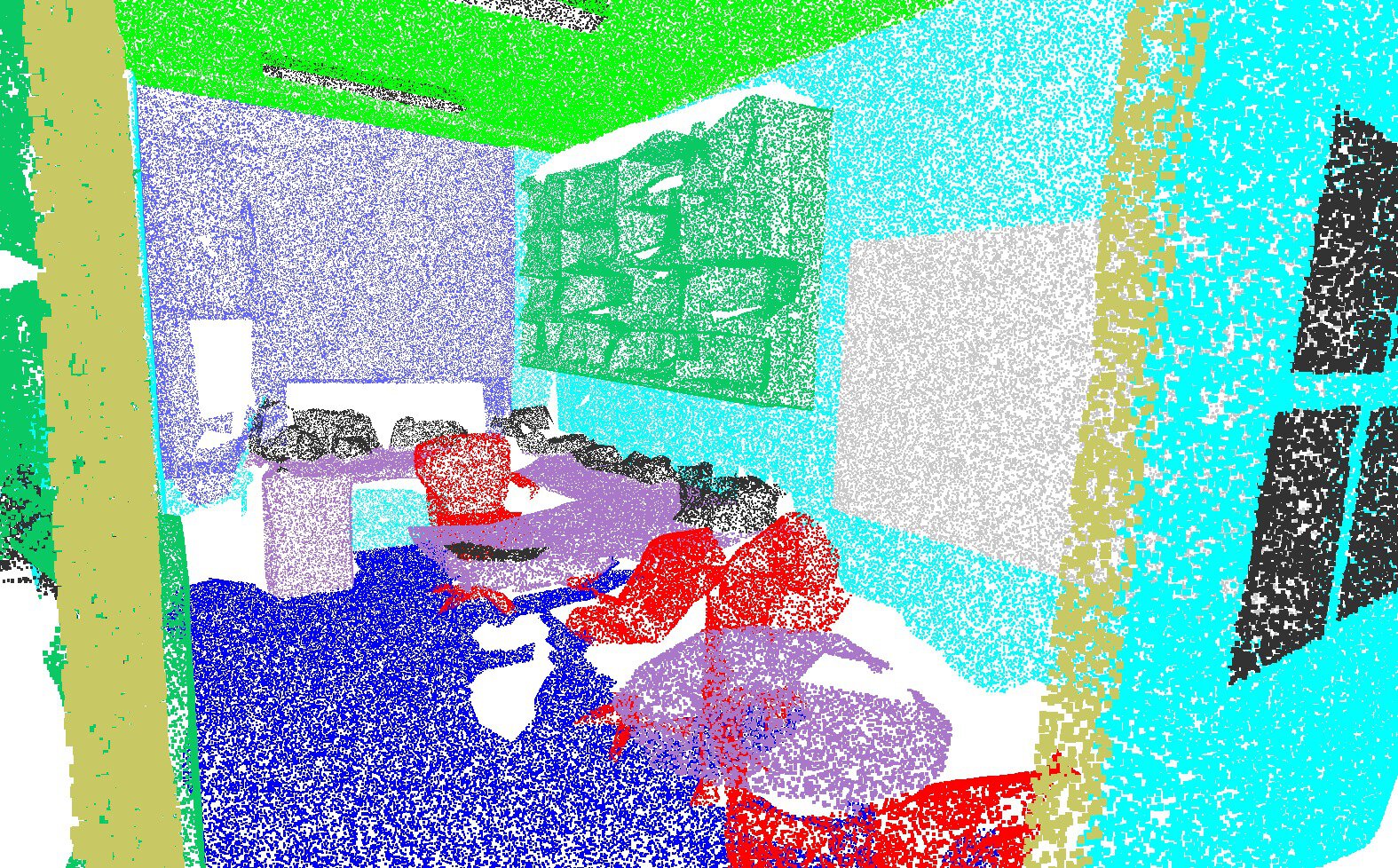} & 
  		\adjincludegraphics[width=.33\linewidth, trim={{.01\width} {.01\height} {.01\width} {.01\height}}, clip]{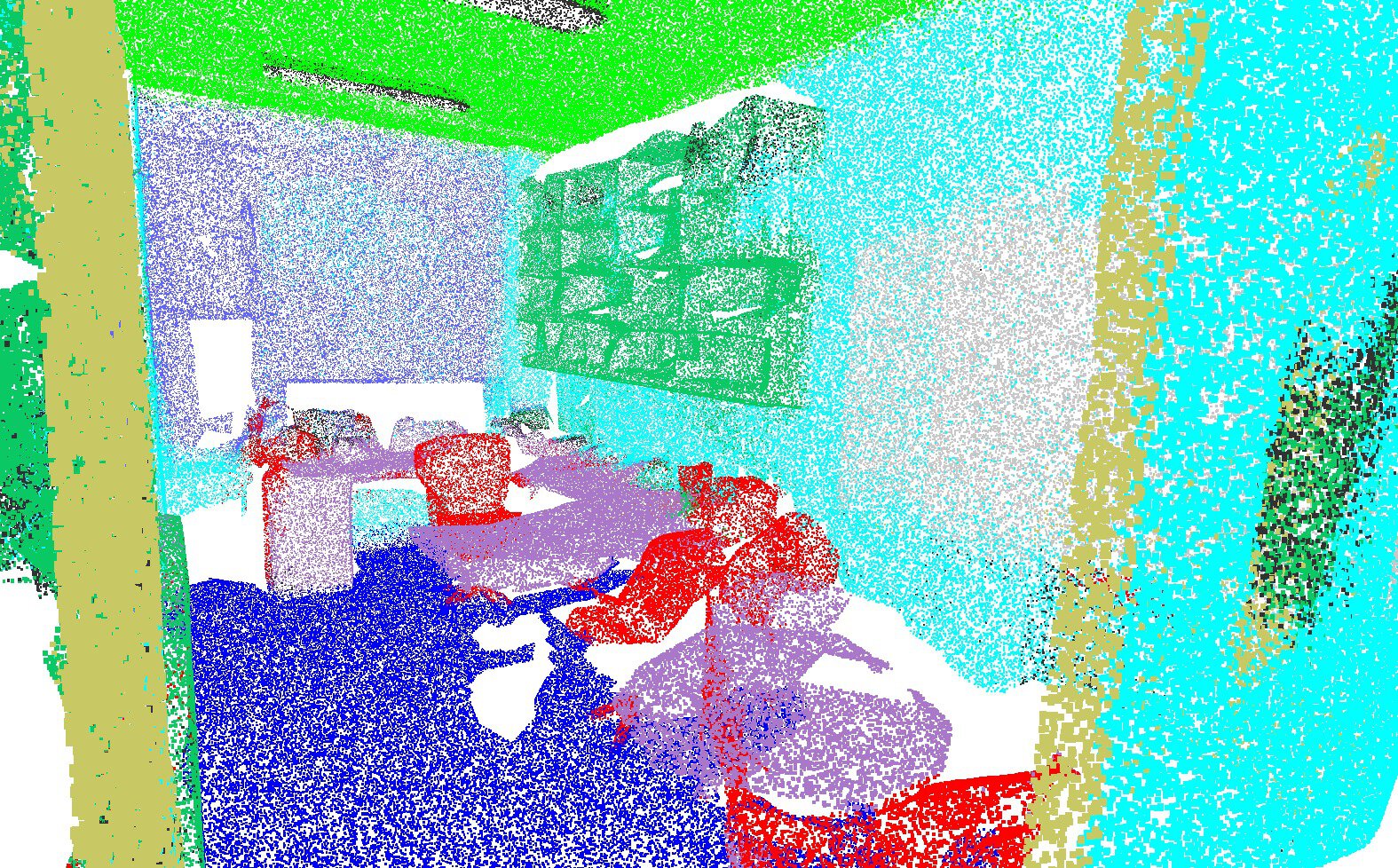} \\
  		\adjincludegraphics[width=.33\linewidth, trim={{.01\width} {.01\height} {.01\width} {.01\height}}, clip]{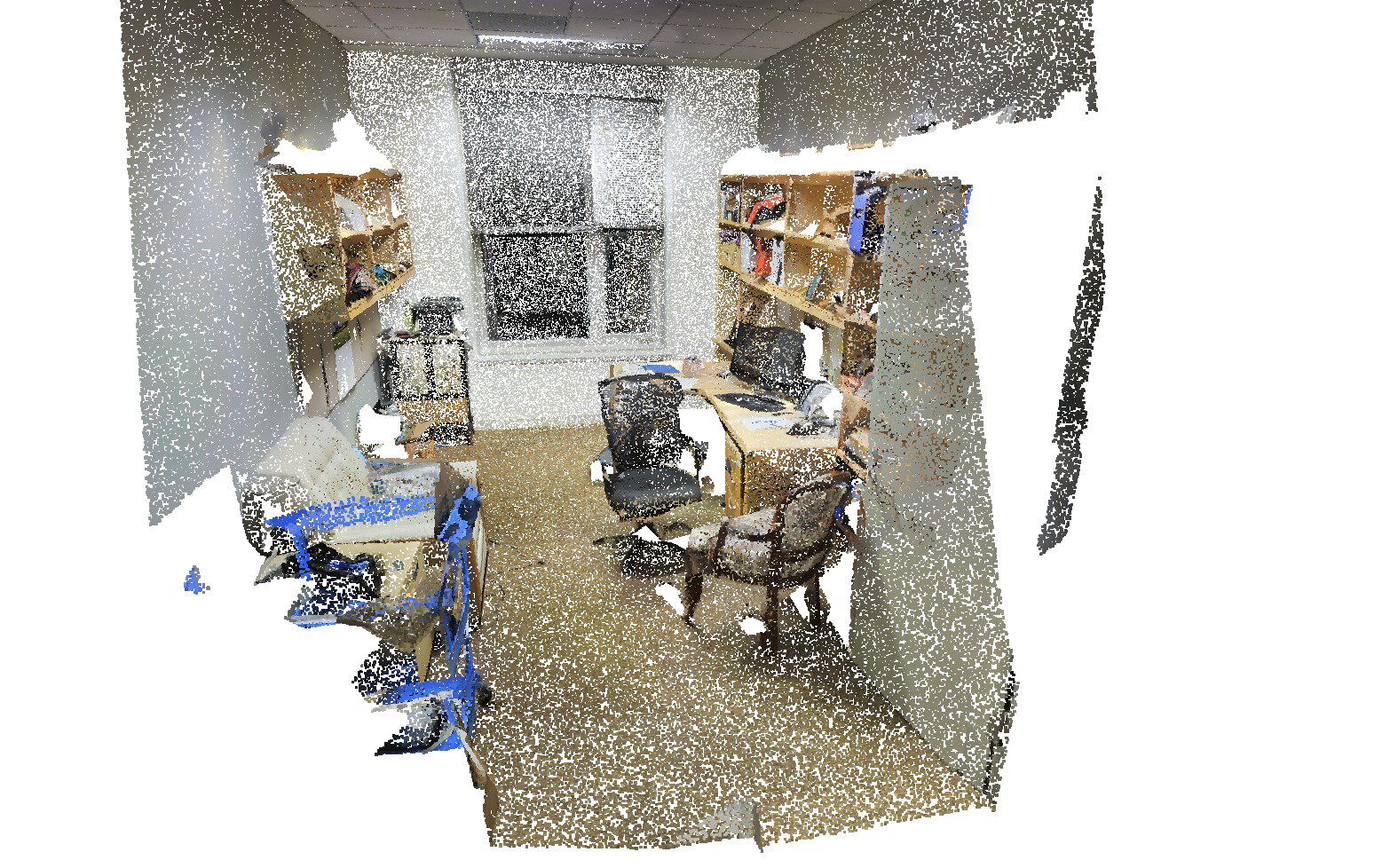} & 
  		\adjincludegraphics[width=.33\linewidth, trim={{.01\width} {.01\height} {.01\width} {.01\height}}, clip]{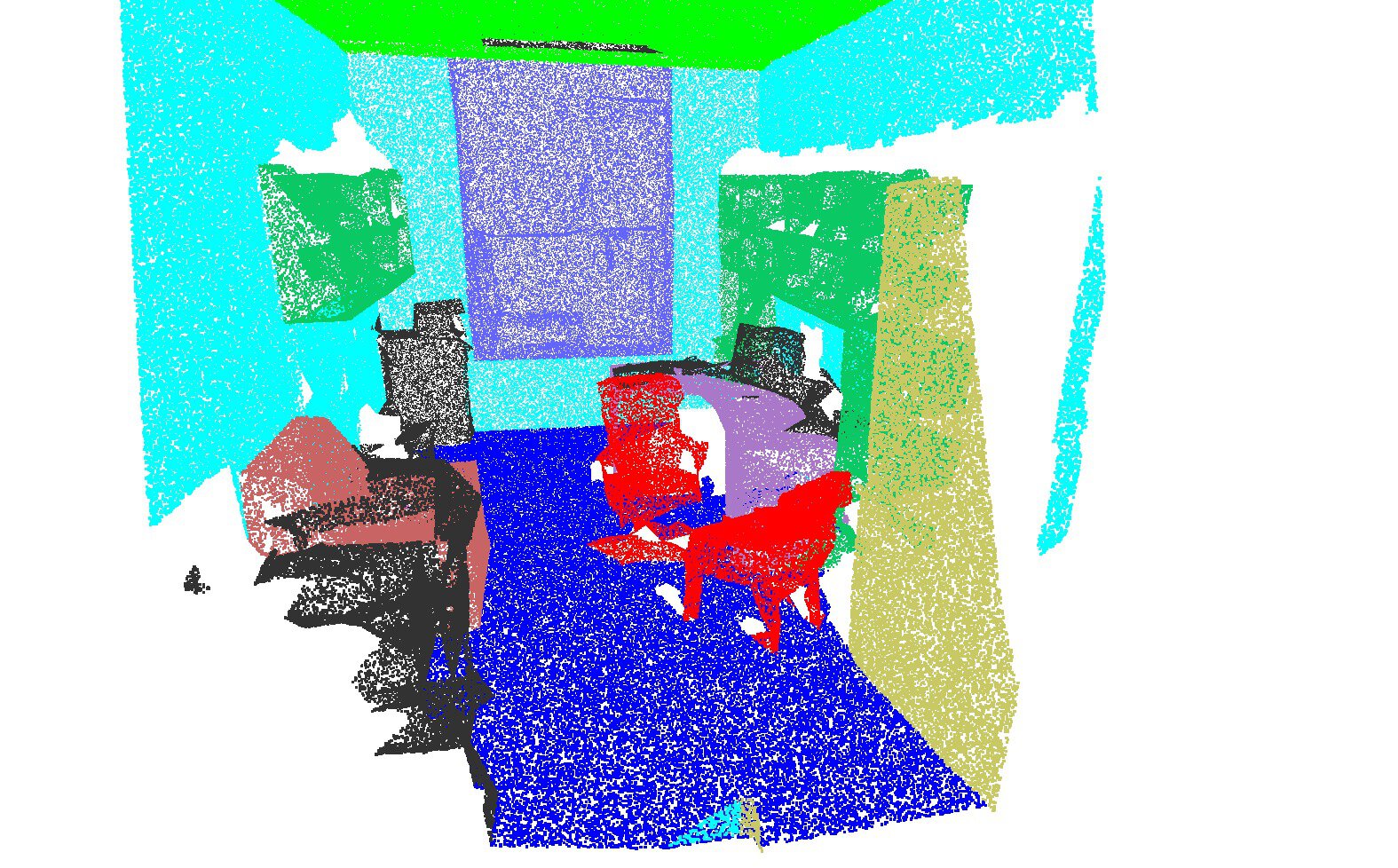} & 
  		\adjincludegraphics[width=.33\linewidth, trim={{.01\width} {.01\height} {.01\width} {.01\height}}, clip]{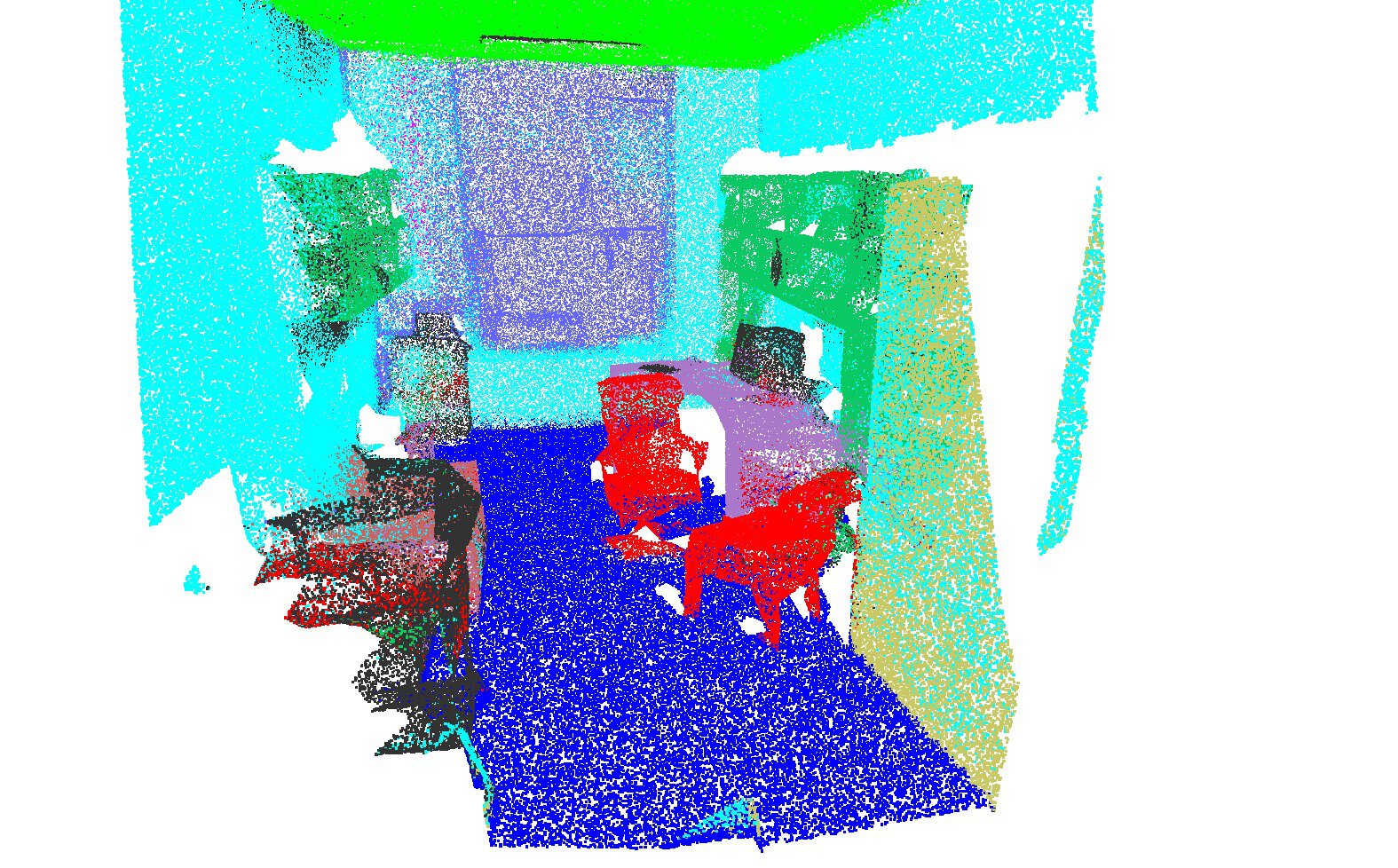} \\
  		\adjincludegraphics[width=.33\linewidth, trim={{.01\width} {.01\height} {.01\width} {.01\height}}, clip]{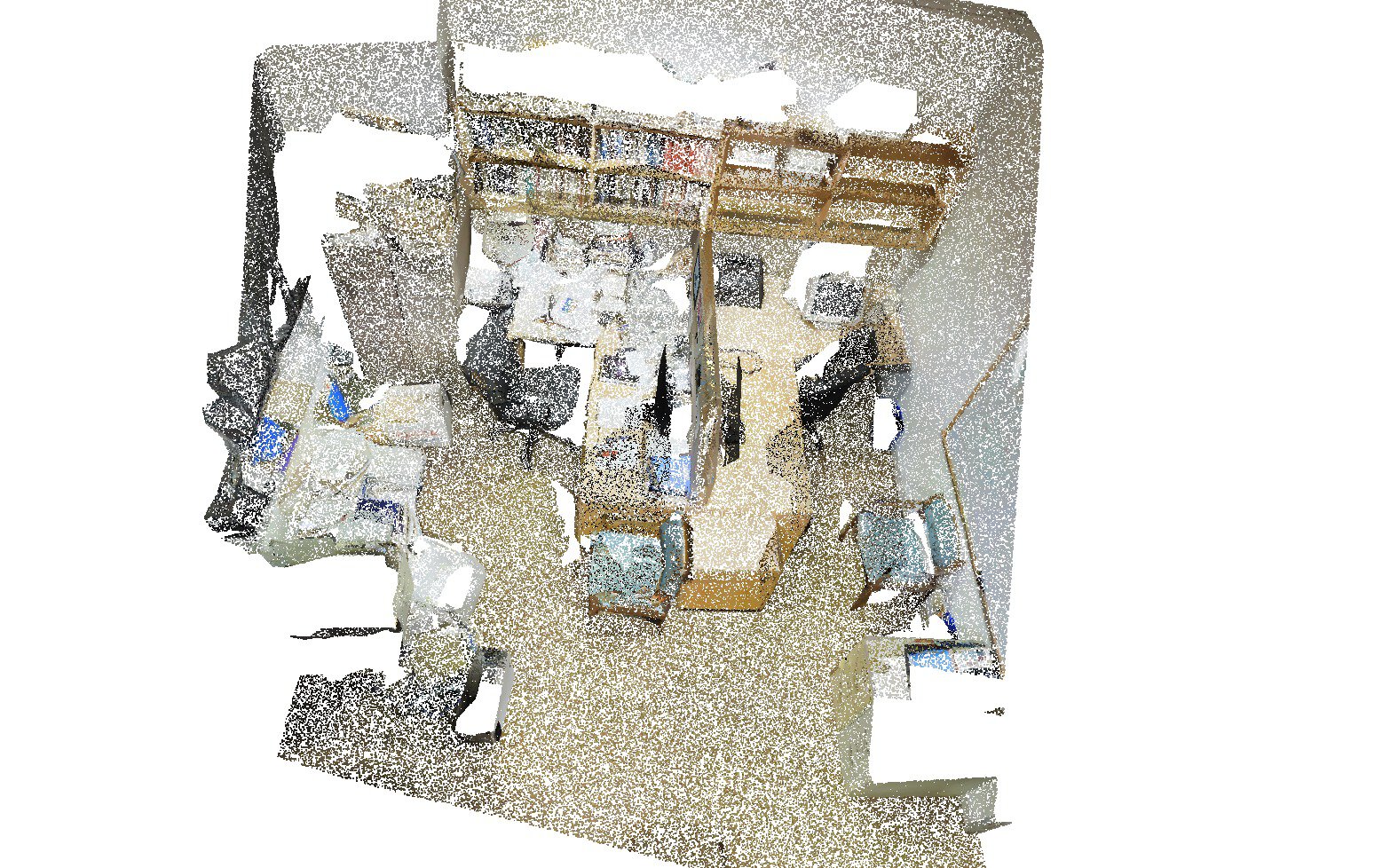} & 
  		\adjincludegraphics[width=.33\linewidth, trim={{.01\width} {.01\height} {.01\width} {.01\height}}, clip]{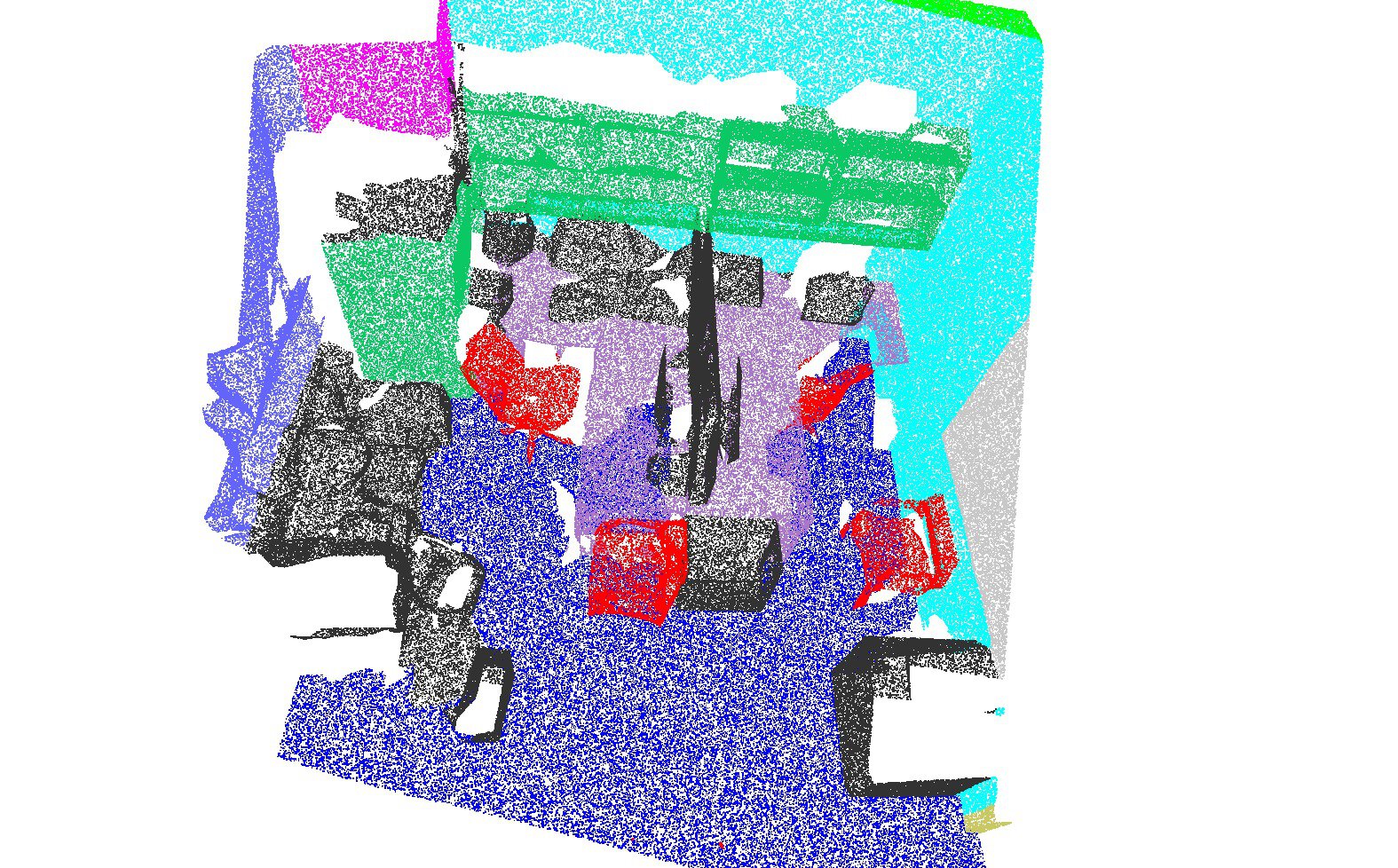} & 
  		\adjincludegraphics[width=.33\linewidth, trim={{.01\width} {.01\height} {.01\width} {.01\height}}, clip]{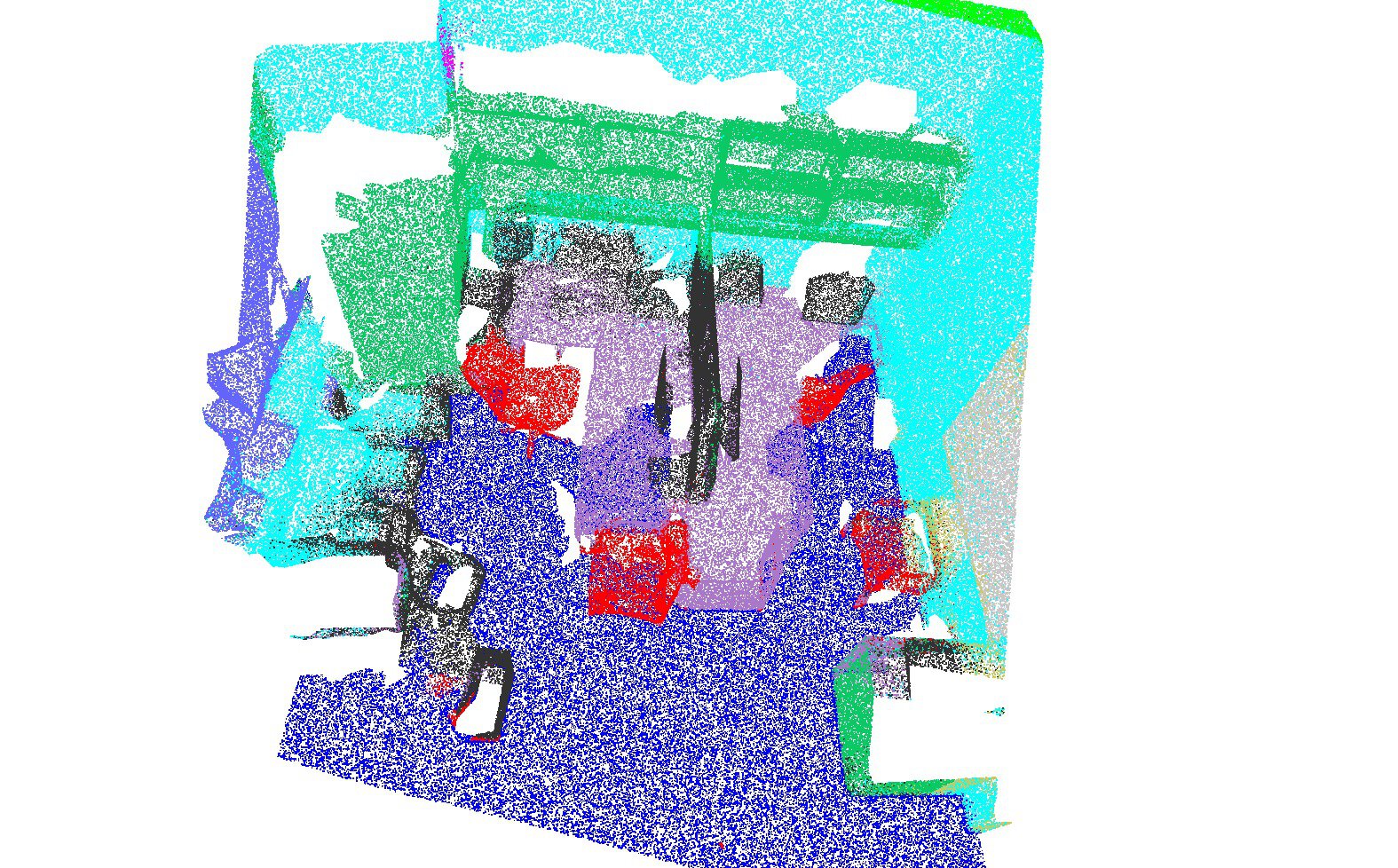} \\
  		\adjincludegraphics[width=.33\linewidth, trim={{.01\width} {.01\height} {.01\width} {.01\height}}, clip]{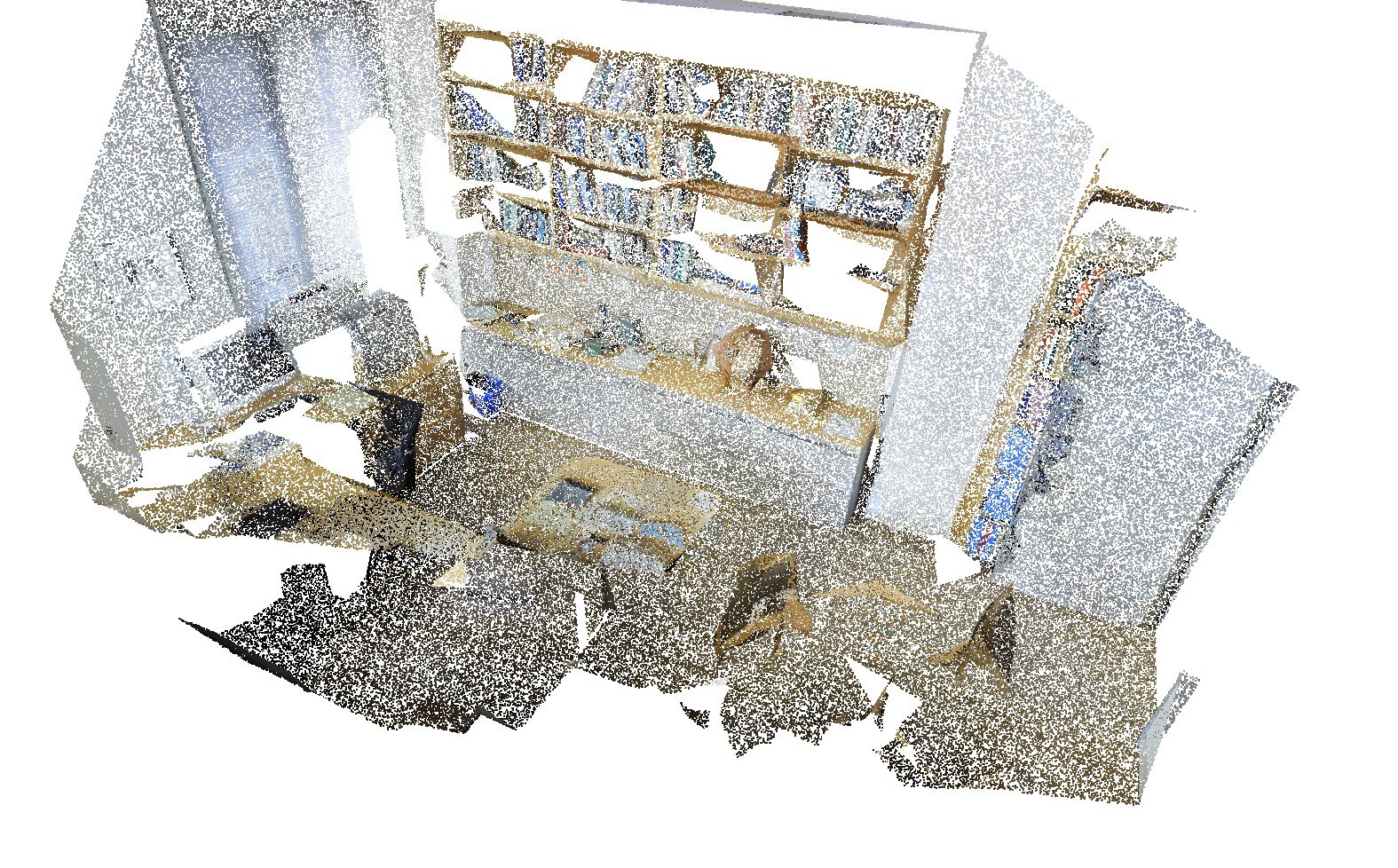} & 
  		\adjincludegraphics[width=.33\linewidth, trim={{.01\width} {.01\height} {.01\width} {.01\height}}, clip]{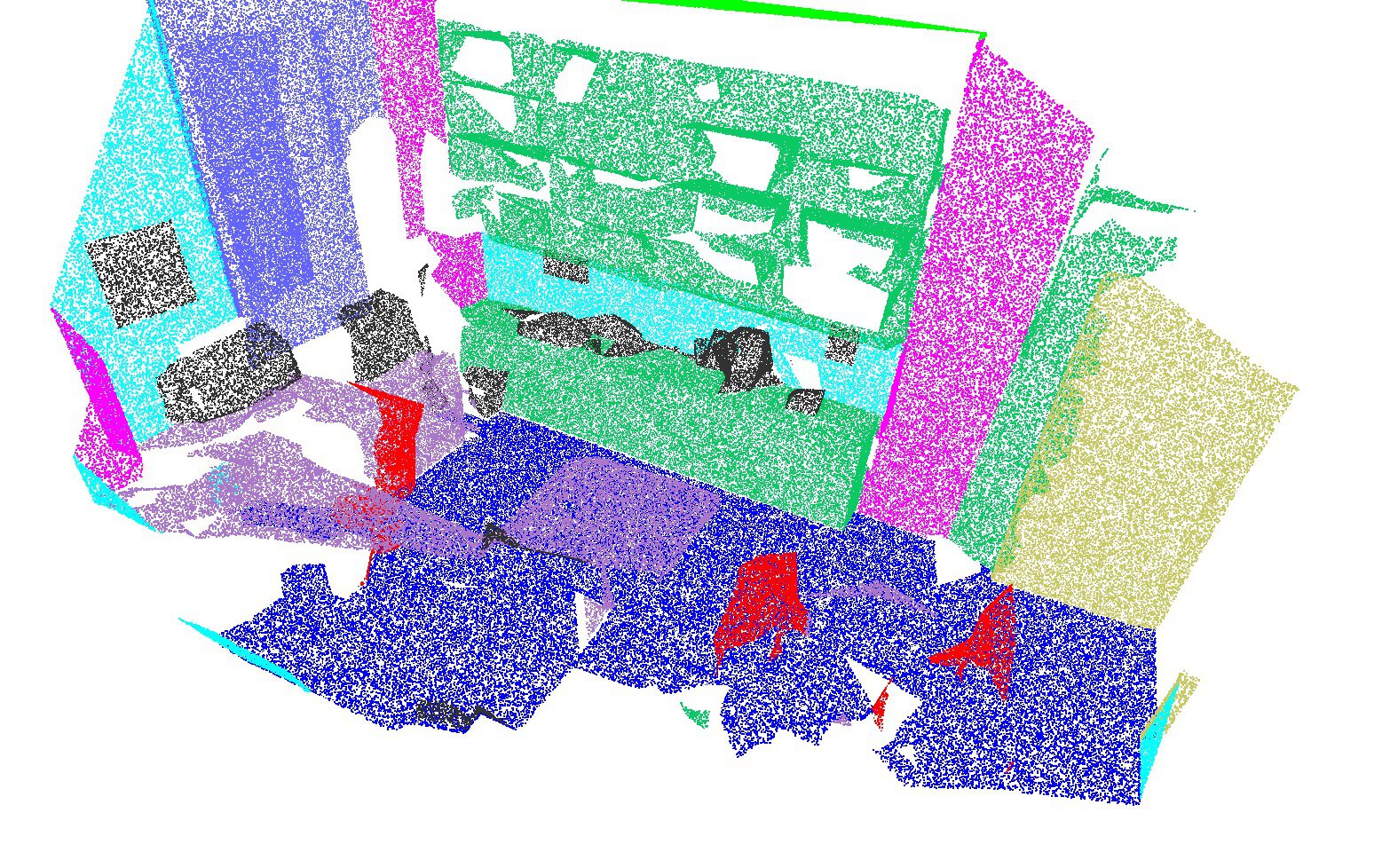} & 
  		\adjincludegraphics[width=.33\linewidth, trim={{.01\width} {.01\height} {.01\width} {.01\height}}, clip]{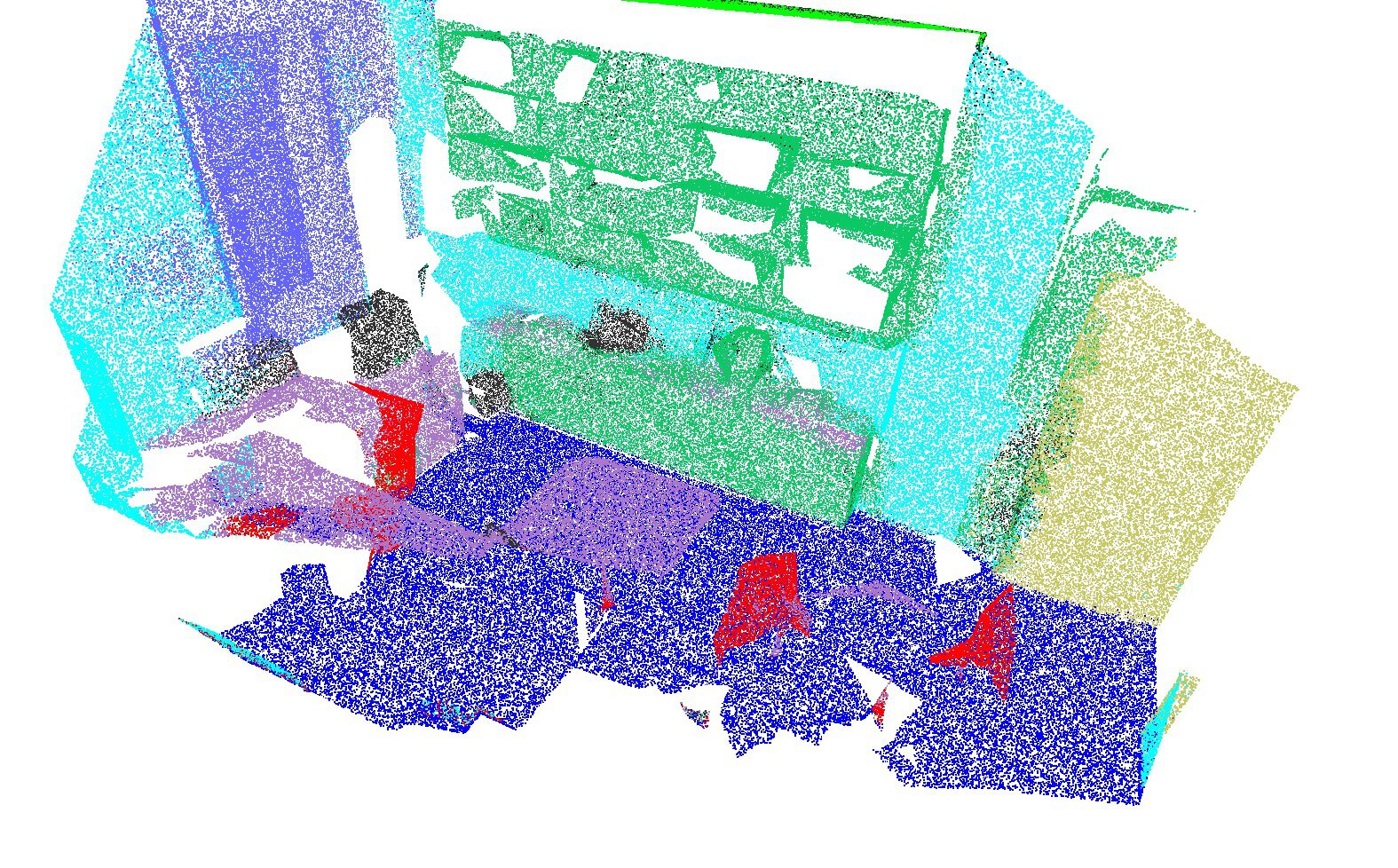} \\
  		\adjincludegraphics[width=.33\linewidth, trim={{.01\width} {.01\height} {.01\width} {.01\height}}, clip]{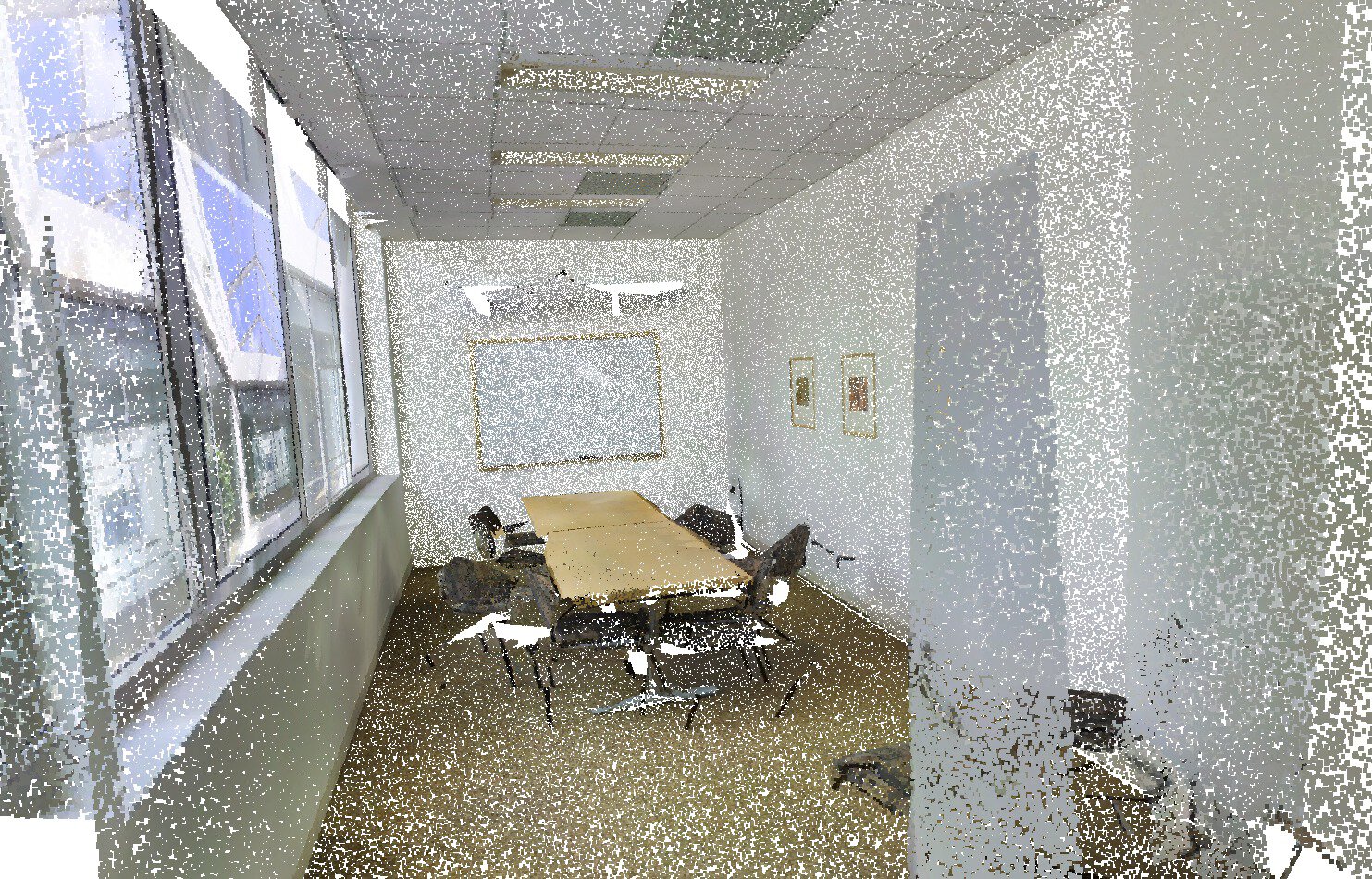} & 
  		\adjincludegraphics[width=.33\linewidth, trim={{.01\width} {.01\height} {.01\width} {.01\height}}, clip]{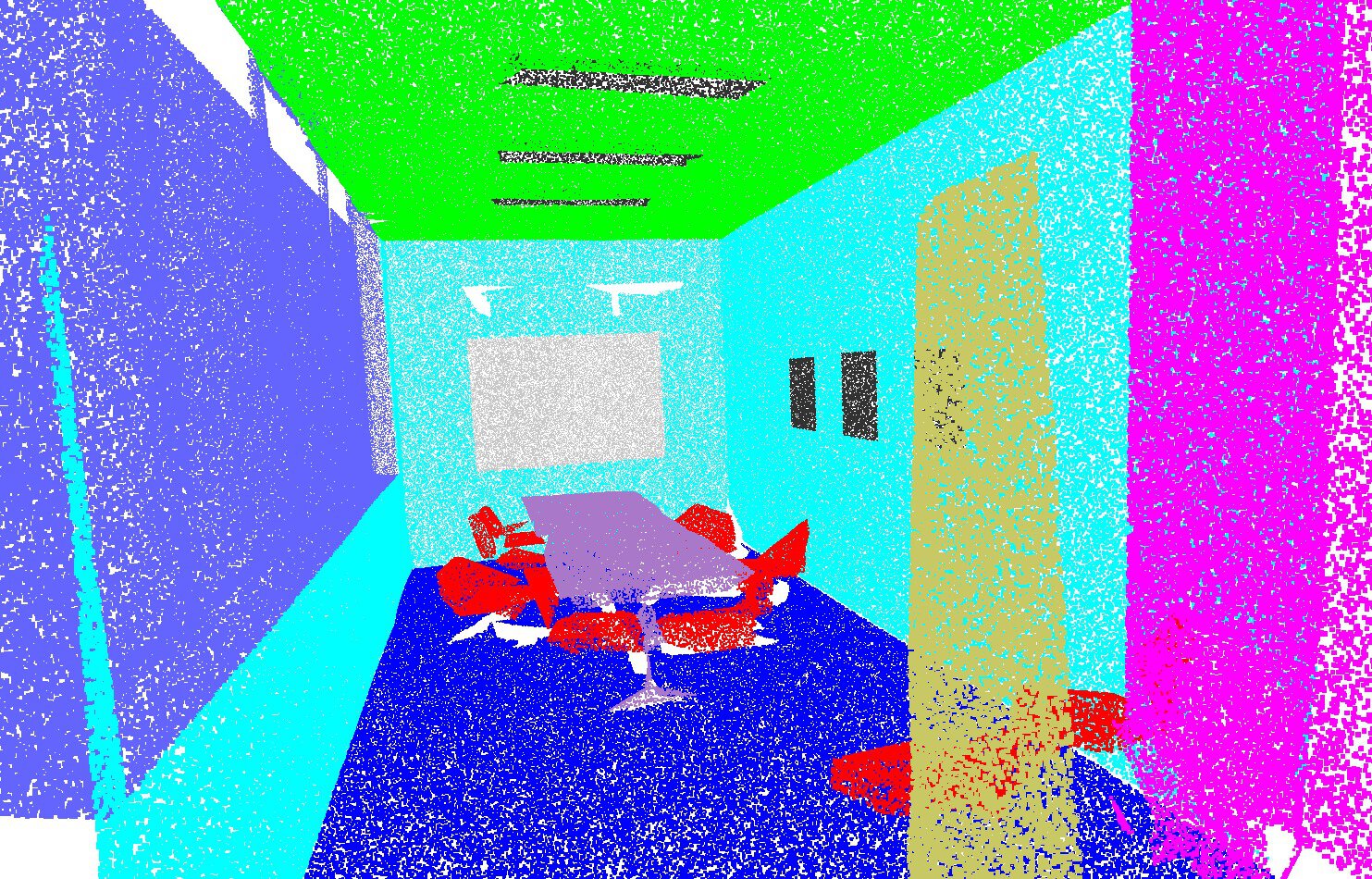} & 
  		\adjincludegraphics[width=.33\linewidth, trim={{.01\width} {.01\height} {.01\width} {.01\height}}, clip]{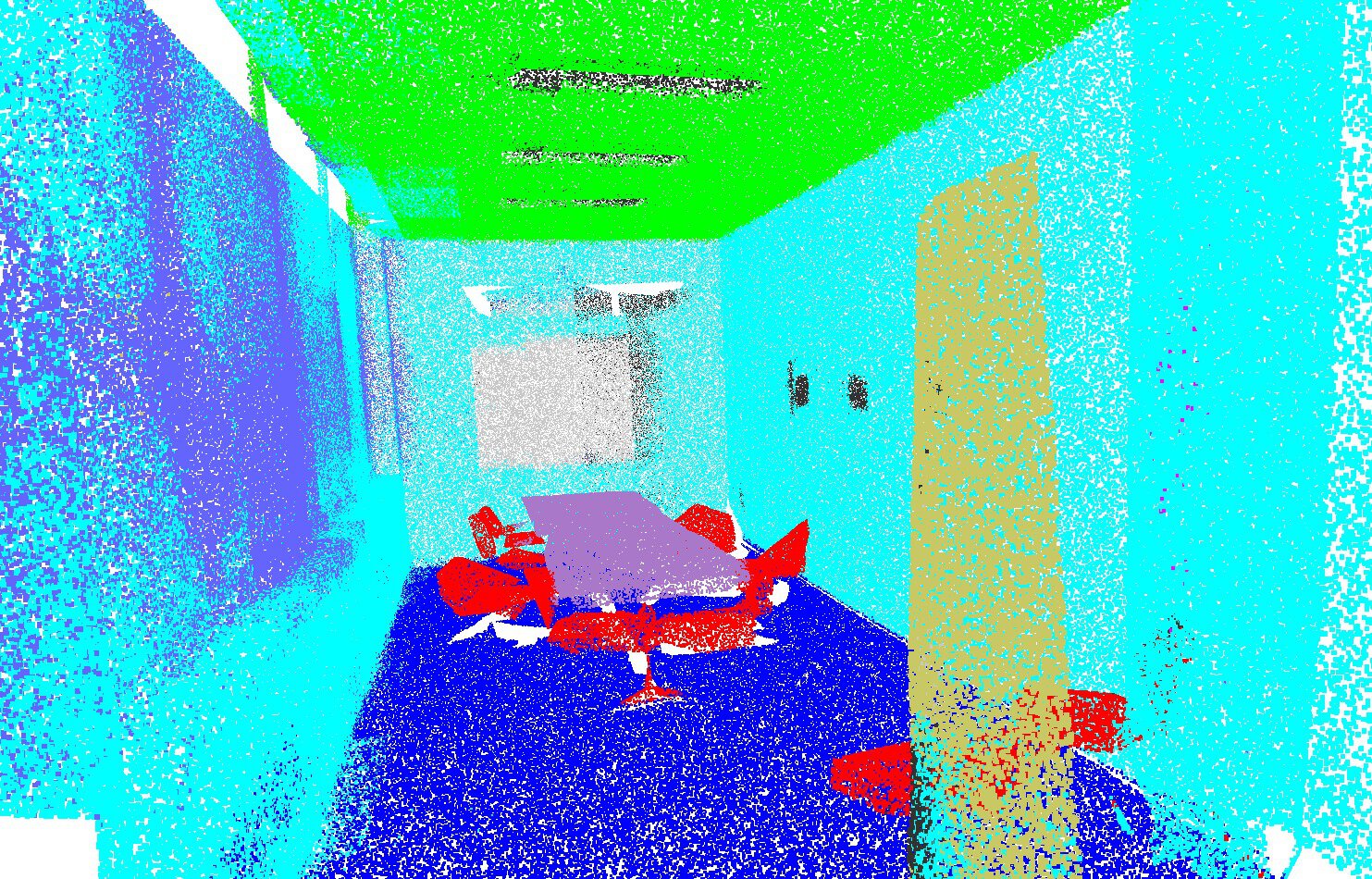} \\
  		\adjincludegraphics[width=.33\linewidth, trim={{.01\width} {.01\height} {.01\width} {.01\height}}, clip]{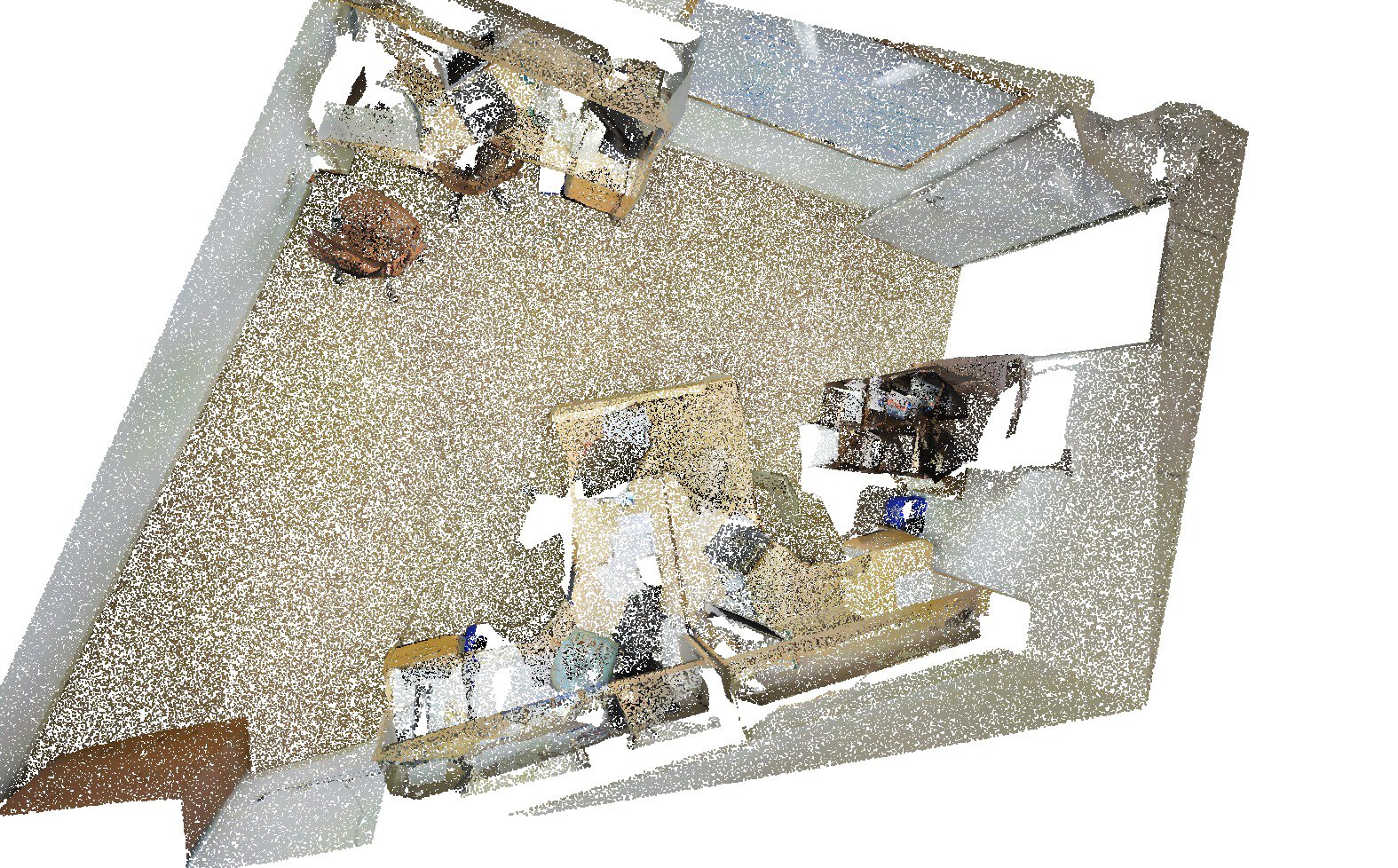} & 
  		\adjincludegraphics[width=.33\linewidth, trim={{.01\width} {.01\height} {.01\width} {.01\height}}, clip]{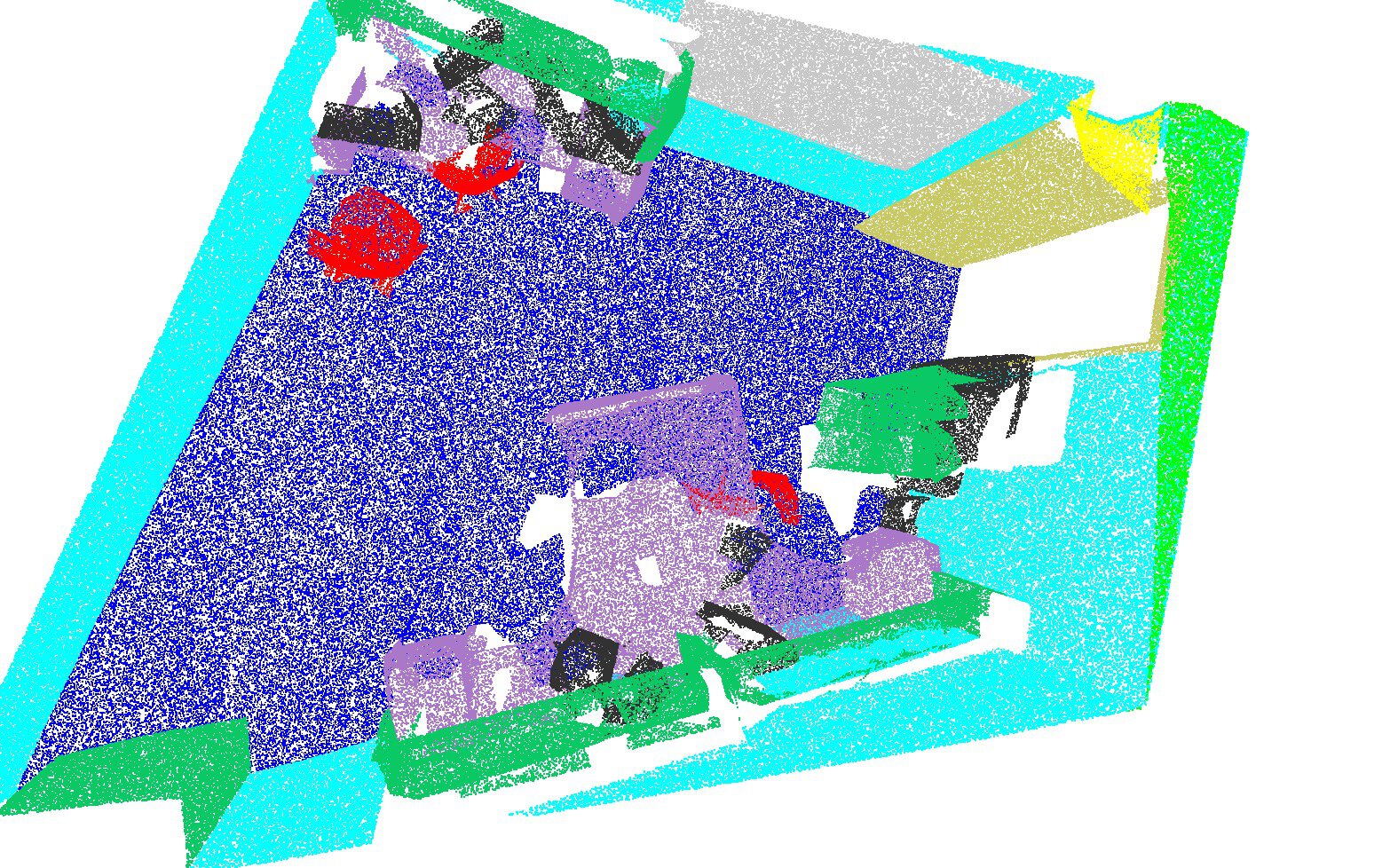} & 
  		\adjincludegraphics[width=.33\linewidth, trim={{.01\width} {.01\height} {.01\width} {.01\height}}, clip]{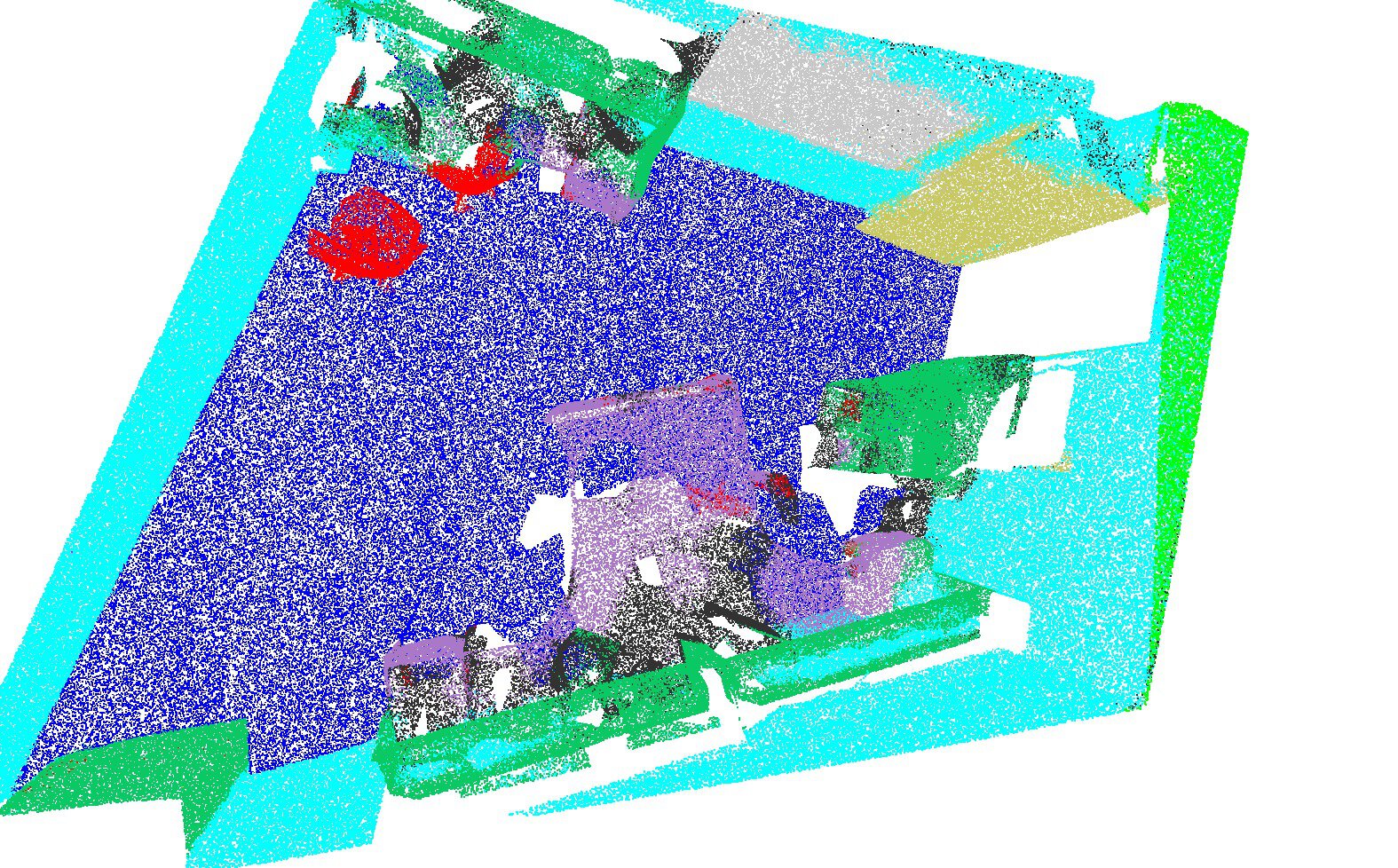} \\
  		Input & GT & Ours
   		\end{tabular}
	\vspace{-3mm}
	\caption{Semantic Segmentation Results on Stanford Indoor Dataset}
	\label{fig:indoor2}
\end{figure*}

Fig.~\ref{fig:indoor} and Fig.~\ref{fig:indoor2} show more qualitatitive results over the stanford dataset. As the figure shown, in most cases our model is able to predict the semantic labels correctly.  

\subsection{Semantic Segmentation for Driving Scenes}
\begin{figure*}
  \footnotesize
  \setlength\tabcolsep{0.5pt} %
  \renewcommand{\arraystretch}{0.8}
  \begin{tabular}{ccc}
      \adjincludegraphics[width=.33\linewidth, trim={{.15\width} {.25\height} {.15\width} {.25\height}}, clip]{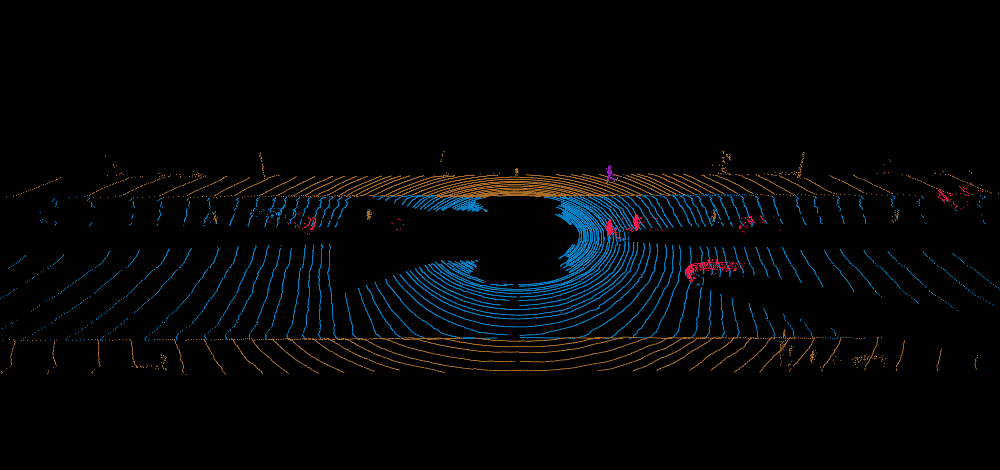} & 
      \adjincludegraphics[width=.33\linewidth, trim={{.15\width} {.25\height} {.15\width} {.25\height}}, clip]{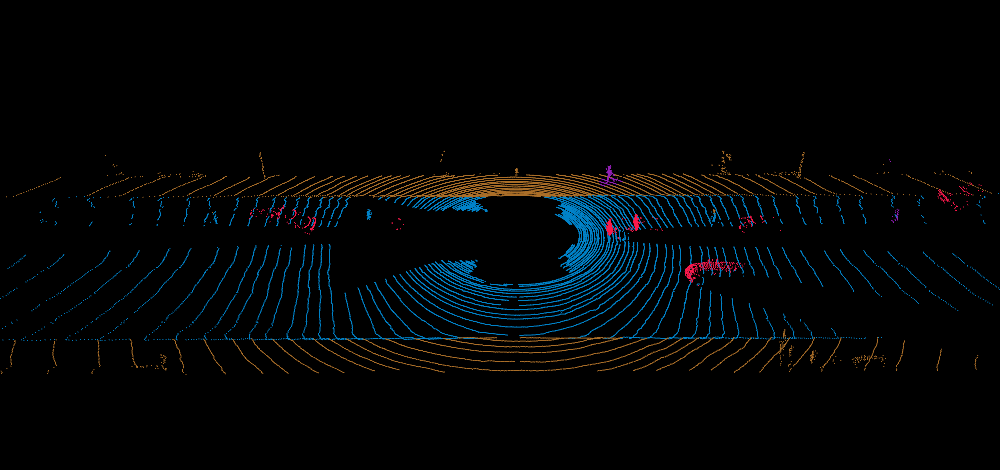} & 
      \adjincludegraphics[width=.33\linewidth, trim={{.15\width} {.25\height} {.15\width} {.25\height}}, clip]{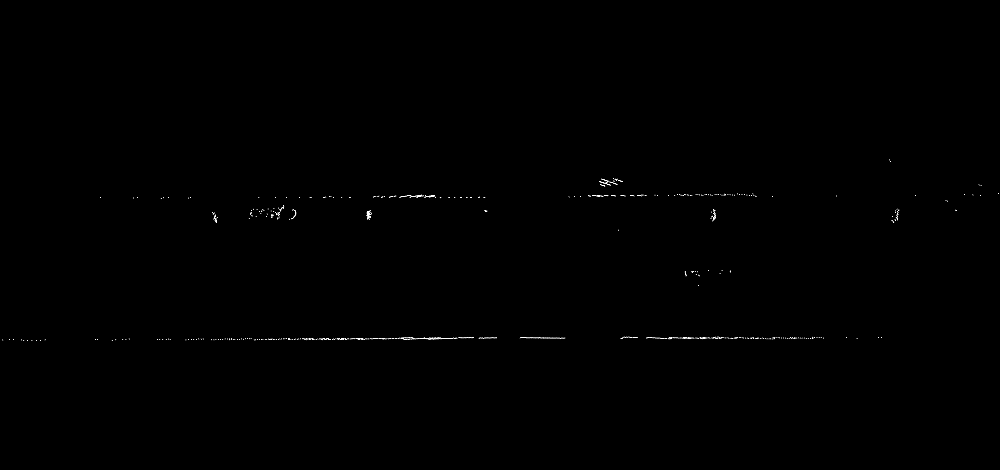} \\

      \adjincludegraphics[width=.33\linewidth, trim={{.15\width} {.25\height} {.15\width} {.25\height}}, clip]{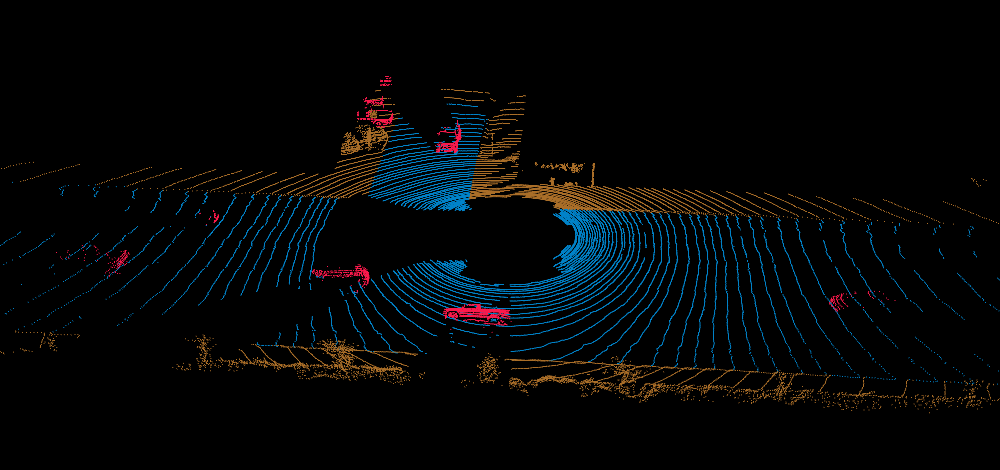} & 
      \adjincludegraphics[width=.33\linewidth, trim={{.15\width} {.25\height} {.15\width} {.25\height}}, clip]{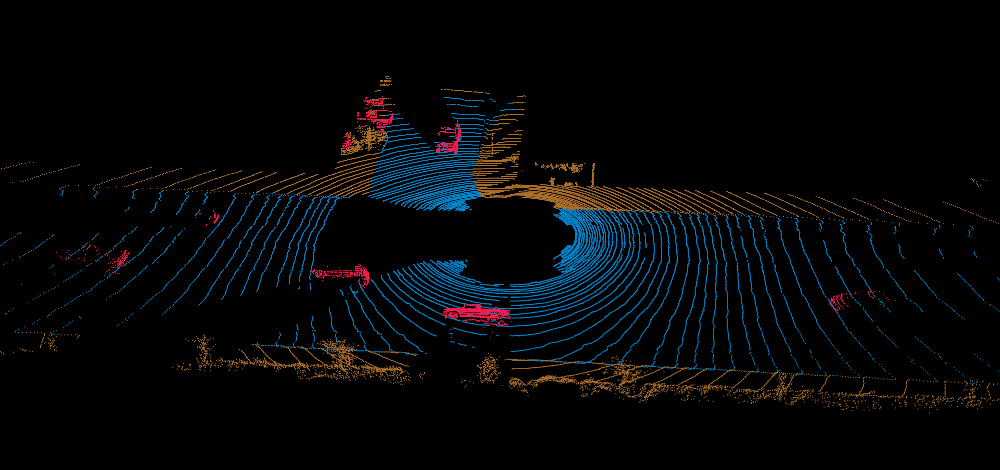} & 
      \adjincludegraphics[width=.33\linewidth, trim={{.15\width} {.25\height} {.15\width} {.25\height}}, clip]{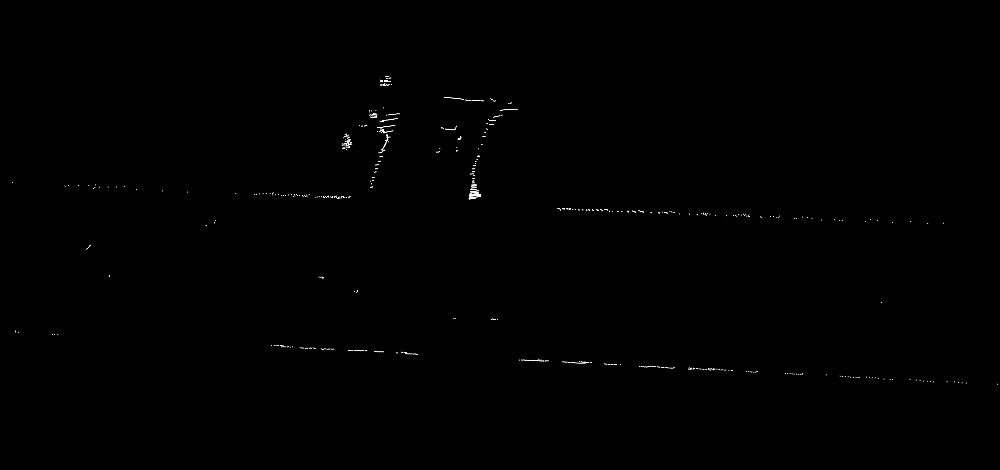} \\

      \adjincludegraphics[width=.33\linewidth, trim={{.15\width} {.25\height} {.15\width} {.25\height}}, clip]{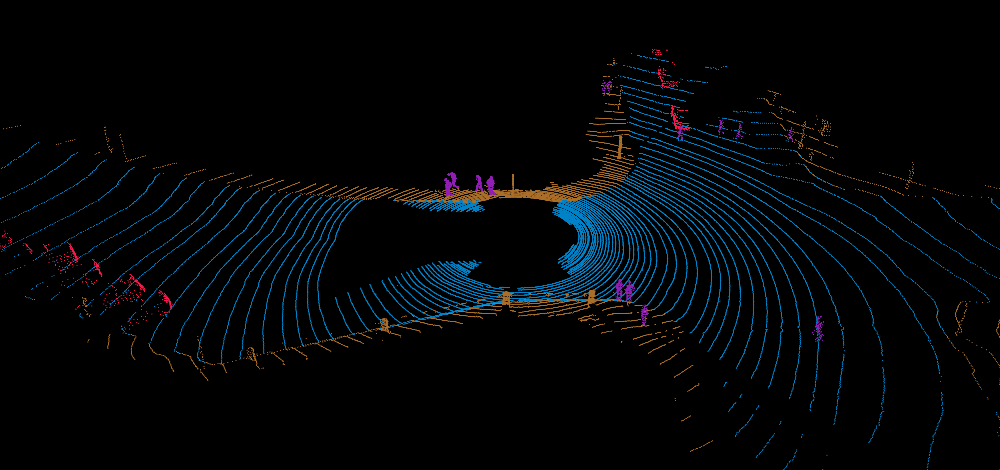} & 
      \adjincludegraphics[width=.33\linewidth, trim={{.15\width} {.25\height} {.15\width} {.25\height}}, clip]{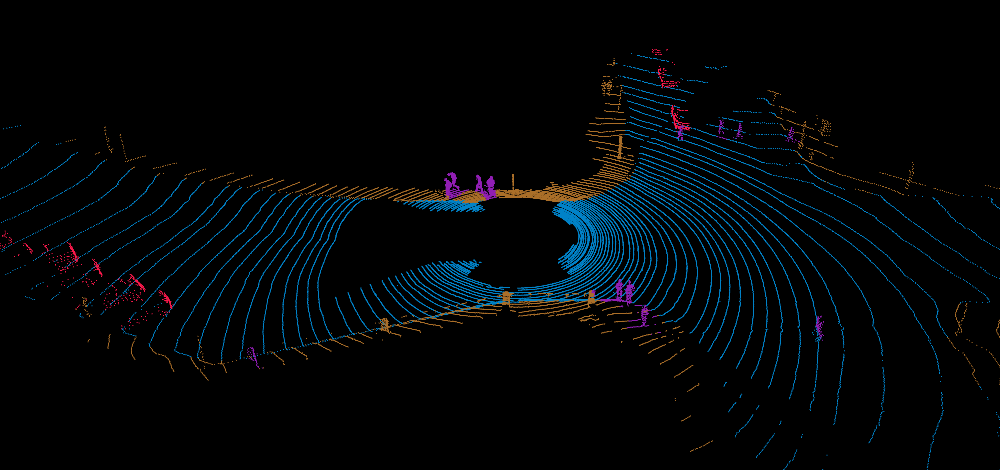} & 
      \adjincludegraphics[width=.33\linewidth, trim={{.15\width} {.25\height} {.15\width} {.25\height}}, clip]{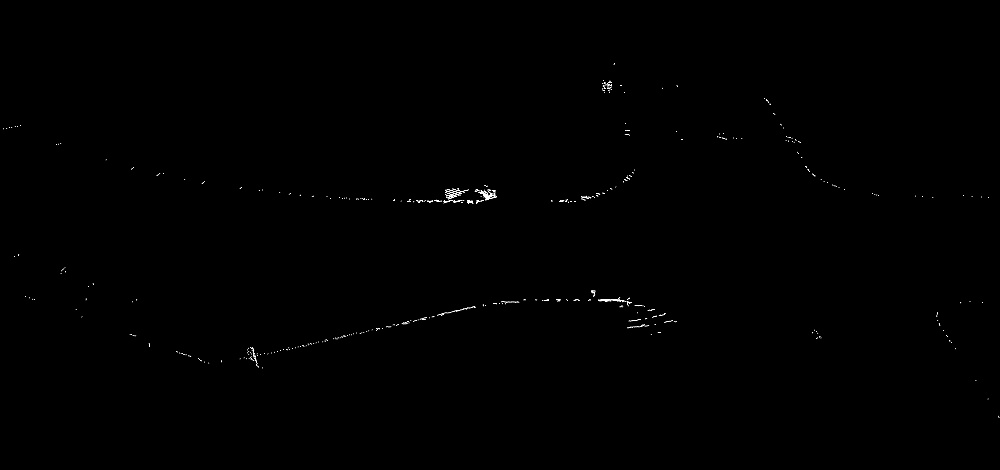} \\

      \adjincludegraphics[width=.33\linewidth, trim={{.15\width} {.25\height} {.15\width} {.25\height}}, clip]{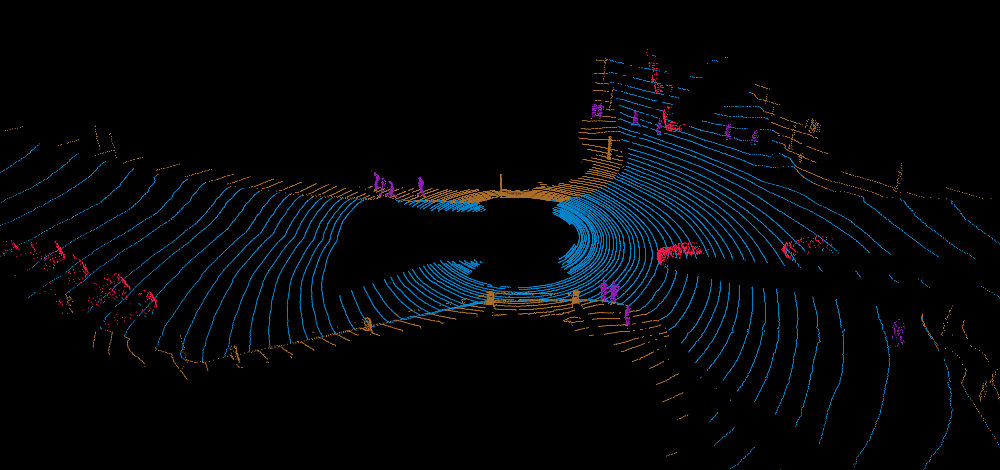} & 
      \adjincludegraphics[width=.33\linewidth, trim={{.15\width} {.25\height} {.15\width} {.25\height}}, clip]{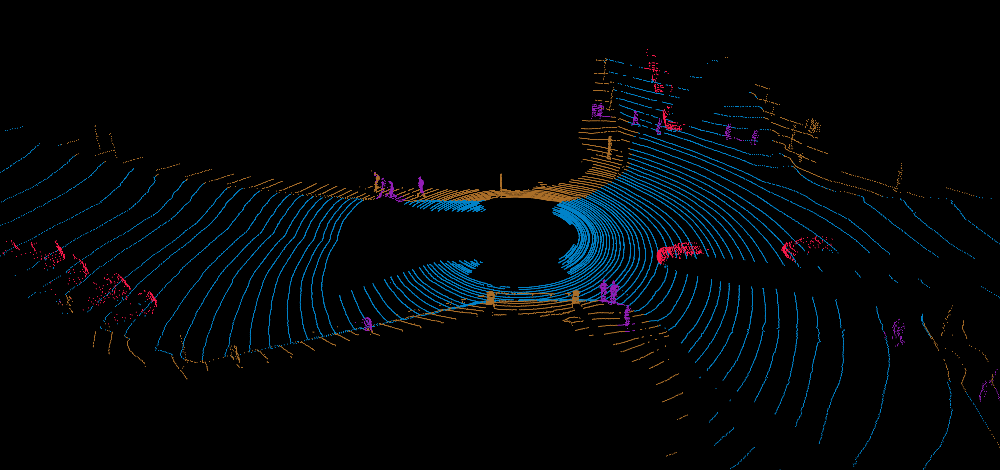} & 
      \adjincludegraphics[width=.33\linewidth, trim={{.15\width} {.25\height} {.15\width} {.25\height}}, clip]{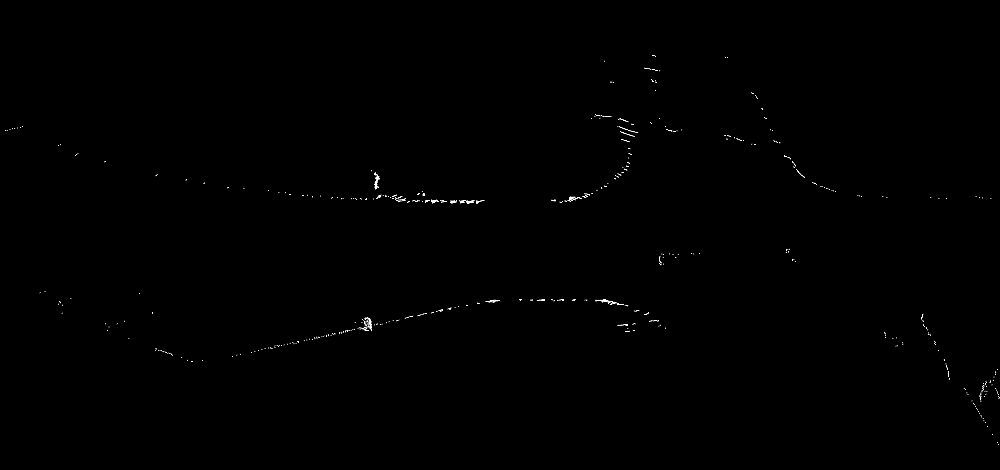} \\

      \adjincludegraphics[width=.33\linewidth, trim={{.15\width} {.25\height} {.15\width} {.25\height}}, clip]{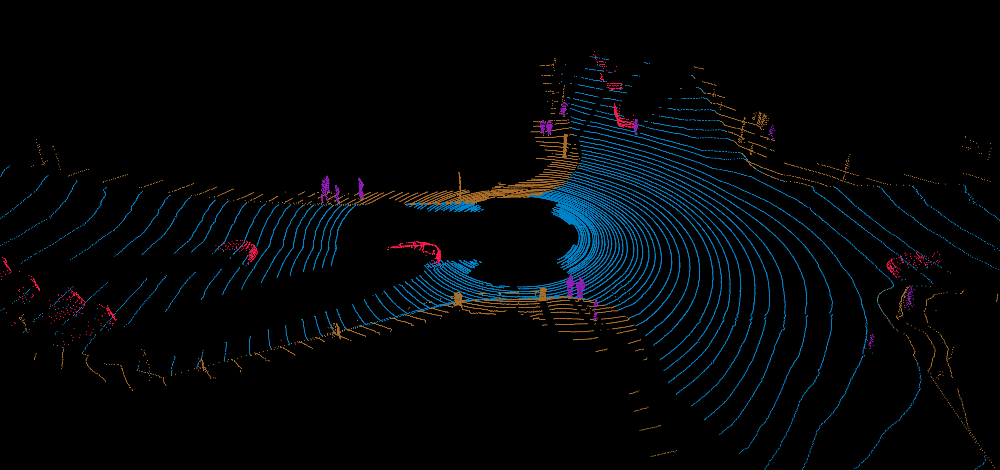} & 
      \adjincludegraphics[width=.33\linewidth, trim={{.15\width} {.25\height} {.15\width} {.25\height}}, clip]{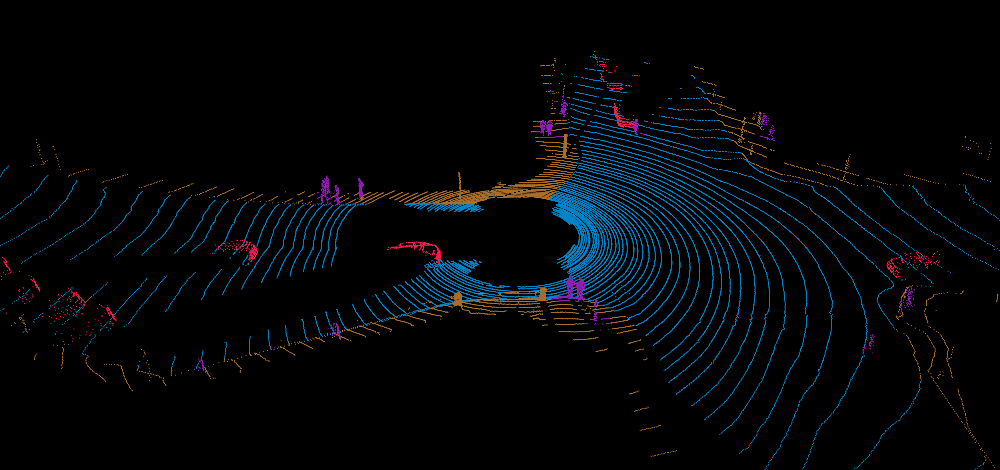} & 
      \adjincludegraphics[width=.33\linewidth, trim={{.15\width} {.25\height} {.15\width} {.25\height}}, clip]{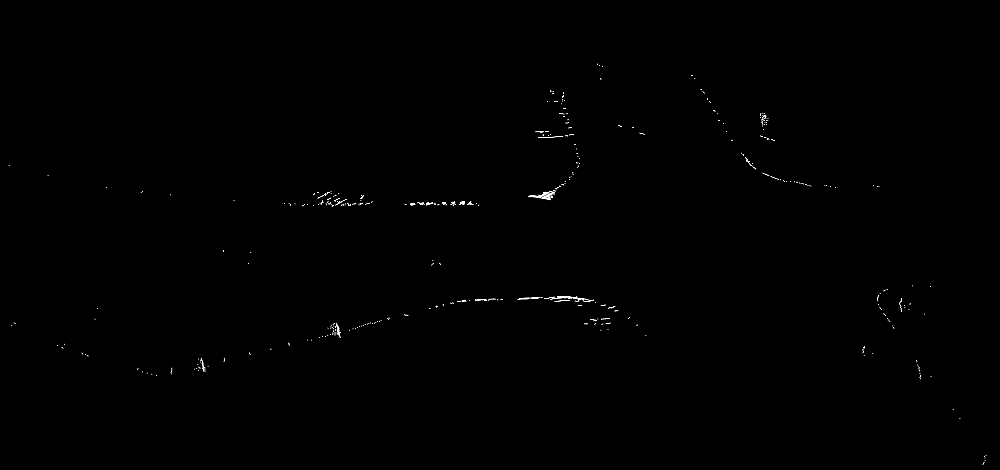} \\

      \adjincludegraphics[width=.33\linewidth, trim={{.15\width} {.25\height} {.15\width} {.25\height}}, clip]{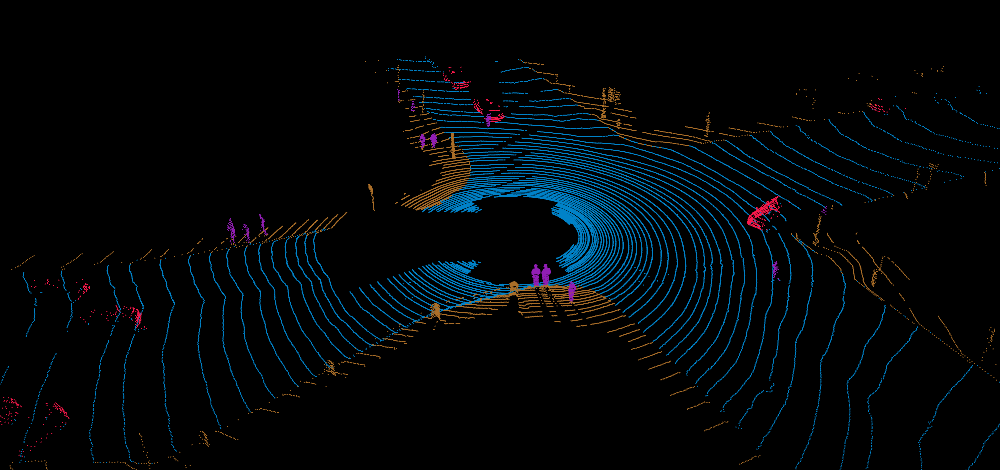} & 
      \adjincludegraphics[width=.33\linewidth, trim={{.15\width} {.25\height} {.15\width} {.25\height}}, clip]{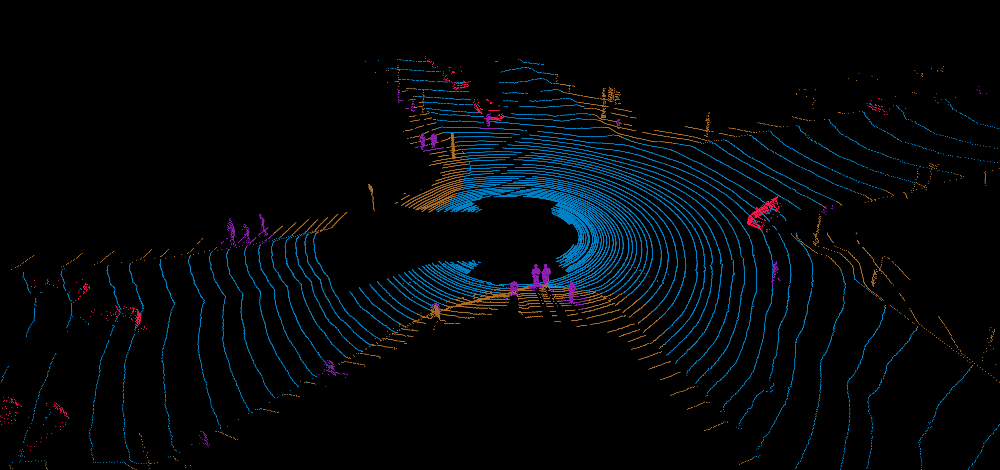} & 
      \adjincludegraphics[width=.33\linewidth, trim={{.15\width} {.25\height} {.15\width} {.25\height}}, clip]{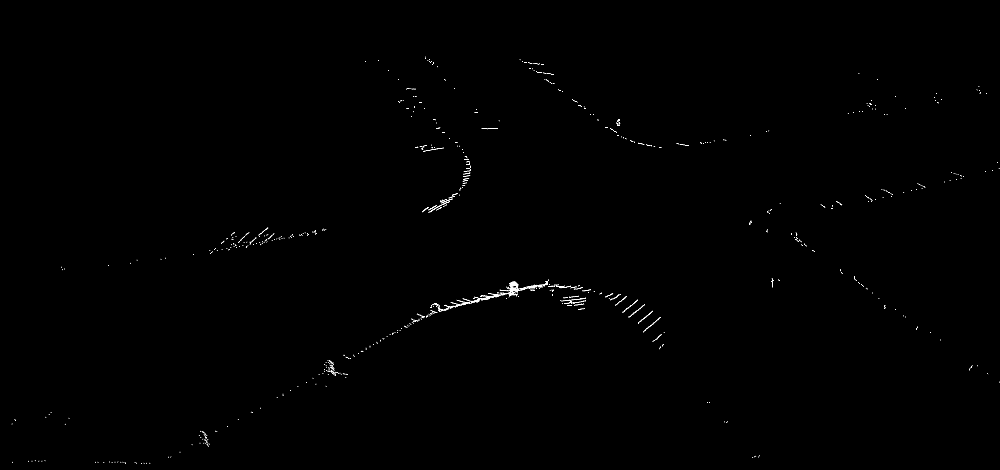} \\

      \adjincludegraphics[width=.33\linewidth, trim={{.15\width} {.25\height} {.15\width} {.25\height}}, clip]{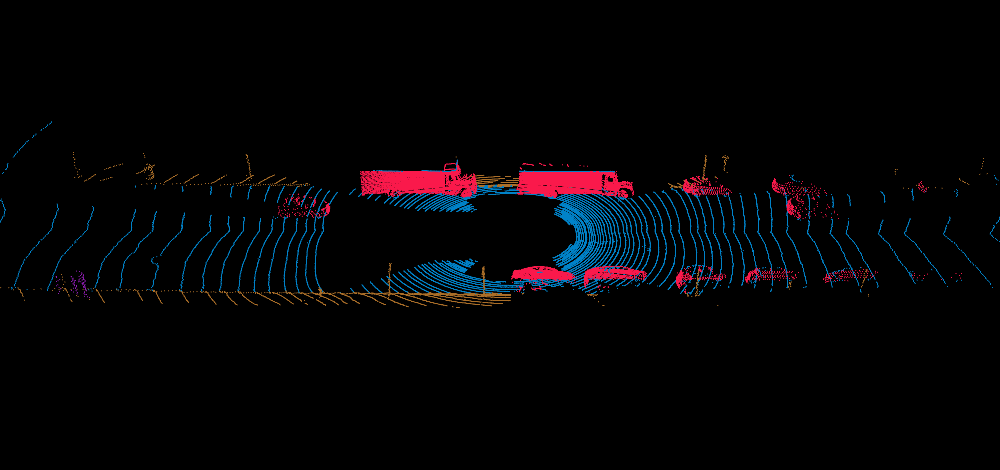} & 
      \adjincludegraphics[width=.33\linewidth, trim={{.15\width} {.25\height} {.15\width} {.25\height}}, clip]{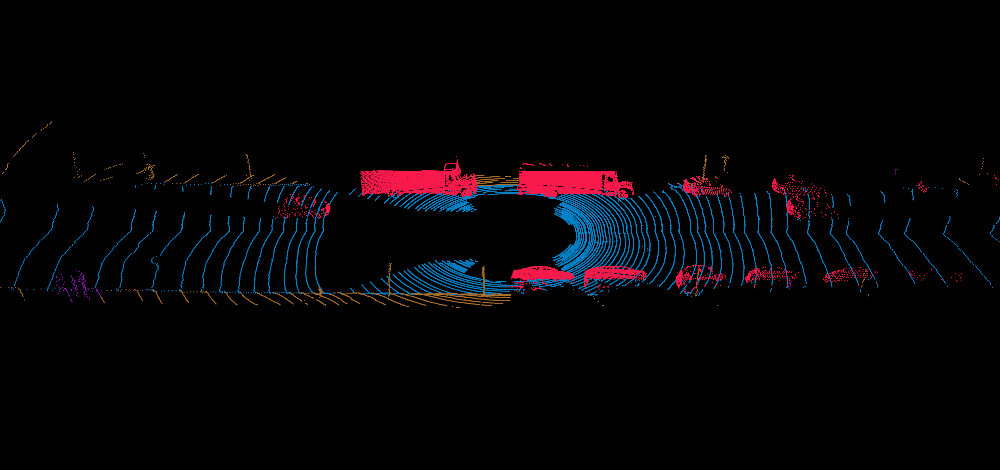} & 
      \adjincludegraphics[width=.33\linewidth, trim={{.15\width} {.25\height} {.15\width} {.25\height}}, clip]{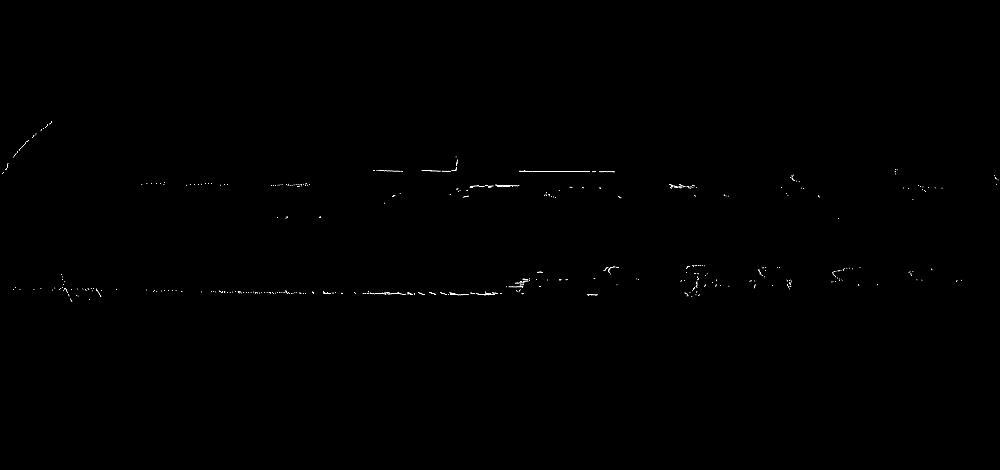} \\

      \adjincludegraphics[width=.33\linewidth, trim={{.15\width} {.25\height} {.15\width} {.25\height}}, clip]{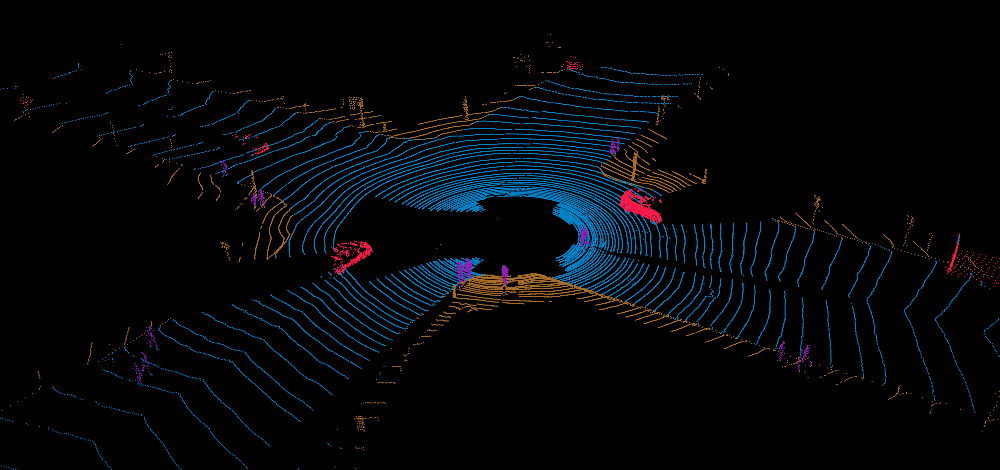} & 
      \adjincludegraphics[width=.33\linewidth, trim={{.15\width} {.25\height} {.15\width} {.25\height}}, clip]{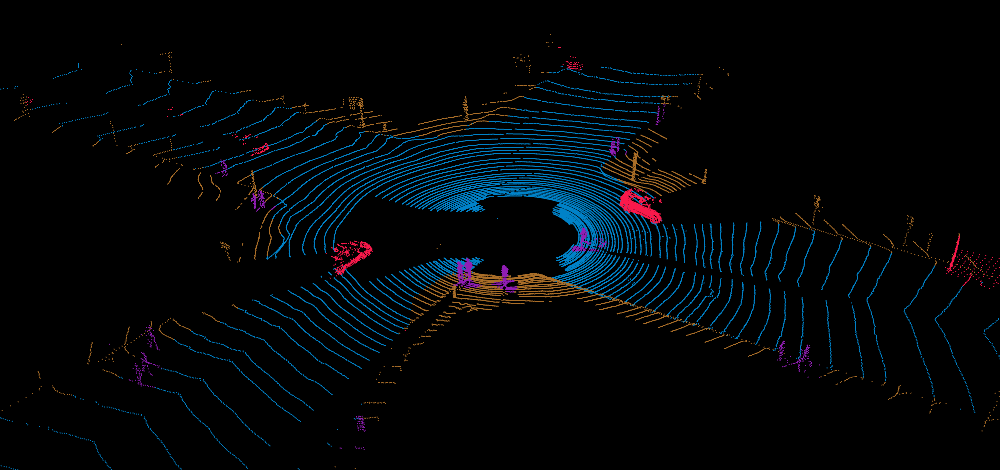} & 
      \adjincludegraphics[width=.33\linewidth, trim={{.15\width} {.25\height} {.15\width} {.25\height}}, clip]{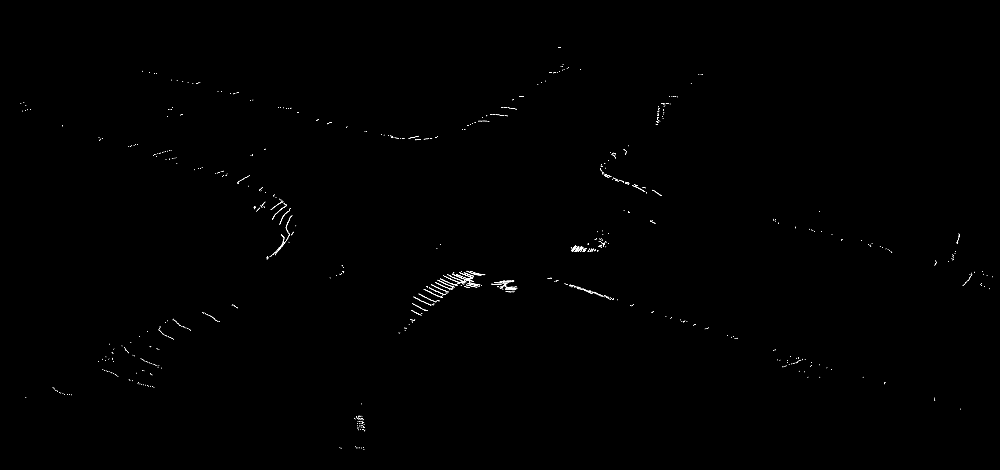} \\

      \adjincludegraphics[width=.33\linewidth, trim={{.15\width} {.25\height} {.15\width} {.25\height}}, clip]{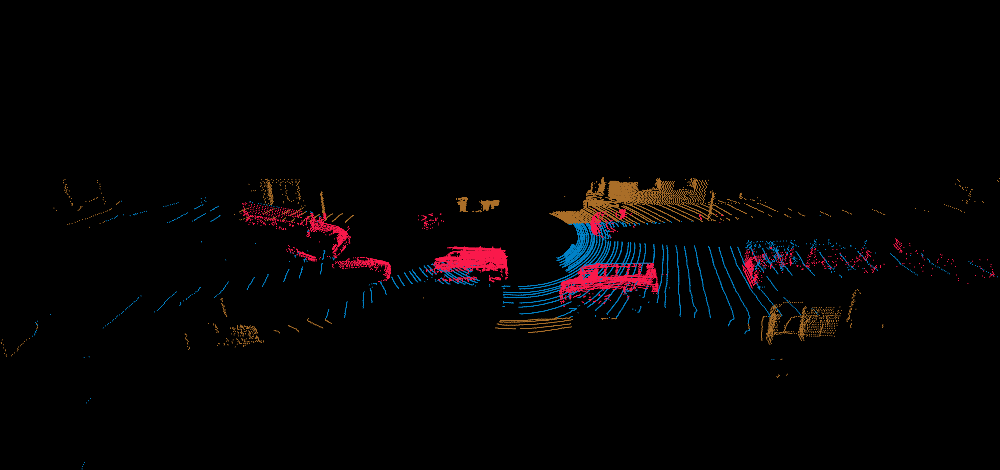} & 
      \adjincludegraphics[width=.33\linewidth, trim={{.15\width} {.25\height} {.15\width} {.25\height}}, clip]{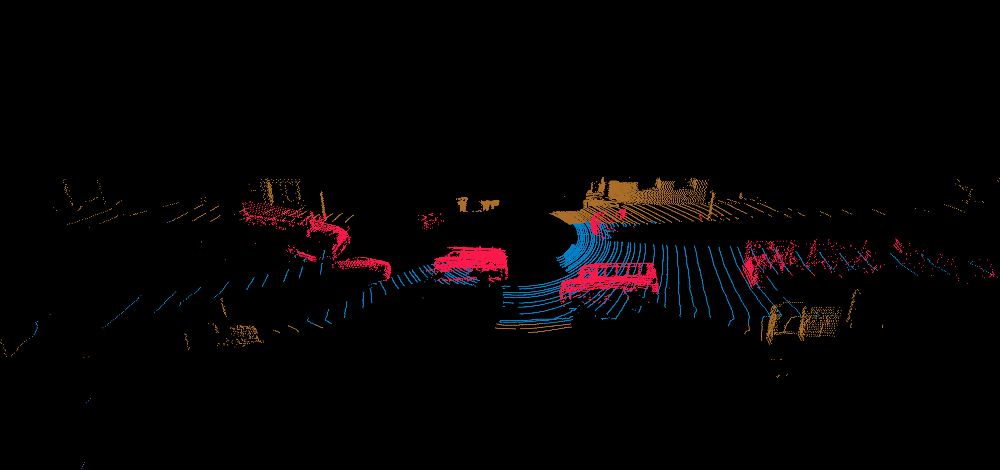} & 
      \adjincludegraphics[width=.33\linewidth, trim={{.15\width} {.25\height} {.15\width} {.25\height}}, clip]{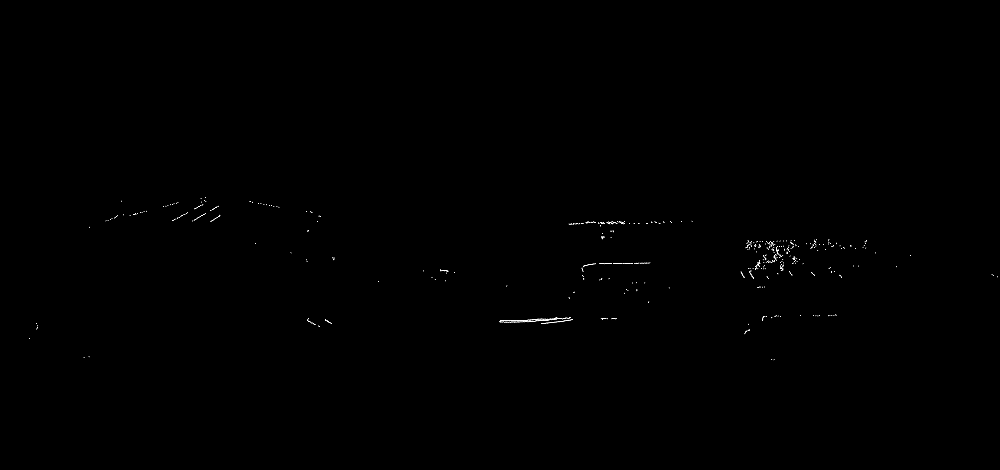} \\

      \adjincludegraphics[width=.33\linewidth, trim={{.15\width} {.25\height} {.15\width} {.25\height}}, clip]{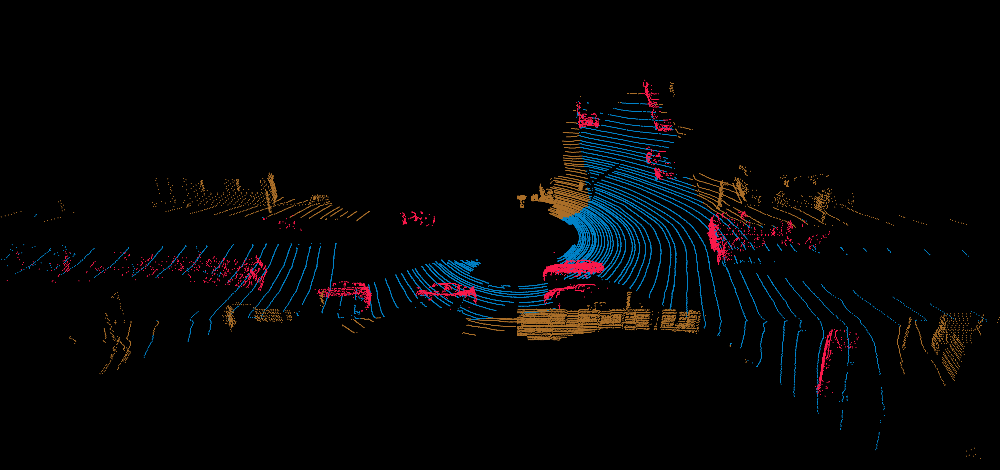} & 
      \adjincludegraphics[width=.33\linewidth, trim={{.15\width} {.25\height} {.15\width} {.25\height}}, clip]{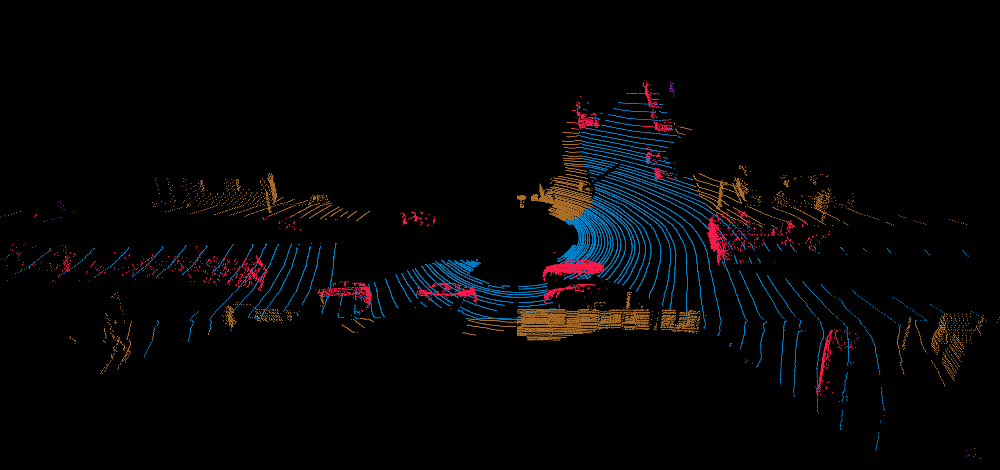} & 
      \adjincludegraphics[width=.33\linewidth, trim={{.15\width} {.25\height} {.15\width} {.25\height}}, clip]{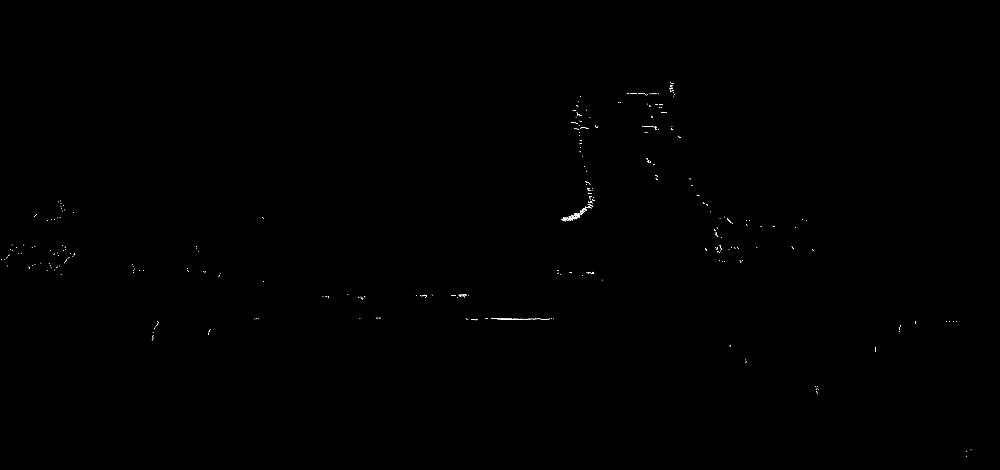} \\

      \adjincludegraphics[width=.33\linewidth, trim={{.15\width} {.25\height} {.15\width} {.25\height}}, clip]{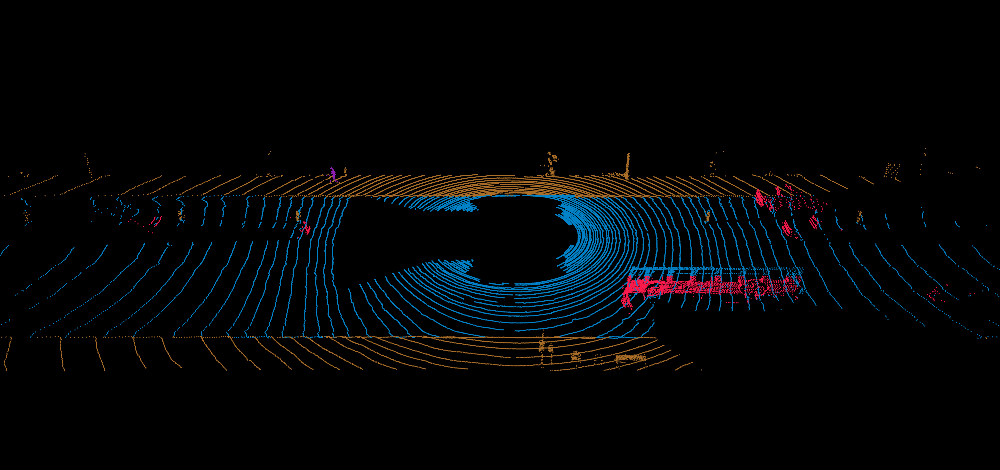} & 
      \adjincludegraphics[width=.33\linewidth, trim={{.15\width} {.25\height} {.15\width} {.25\height}}, clip]{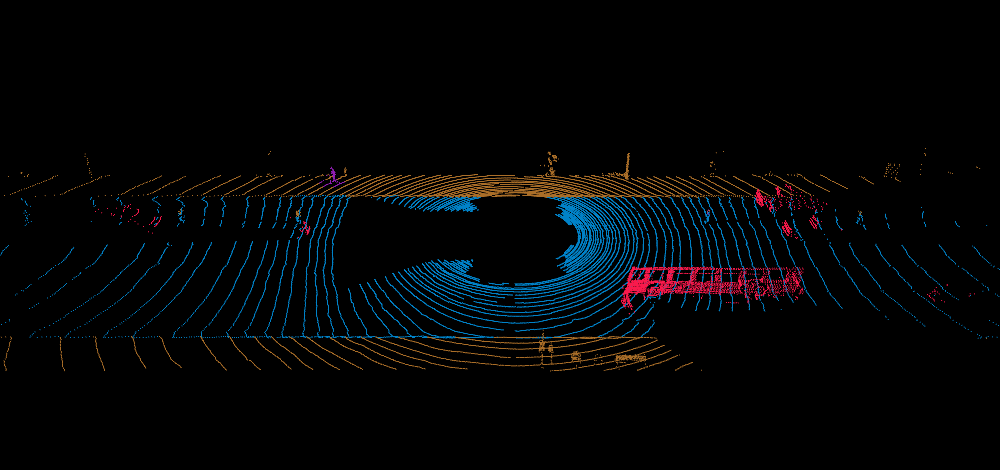} & 
      \adjincludegraphics[width=.33\linewidth, trim={{.15\width} {.25\height} {.15\width} {.25\height}}, clip]{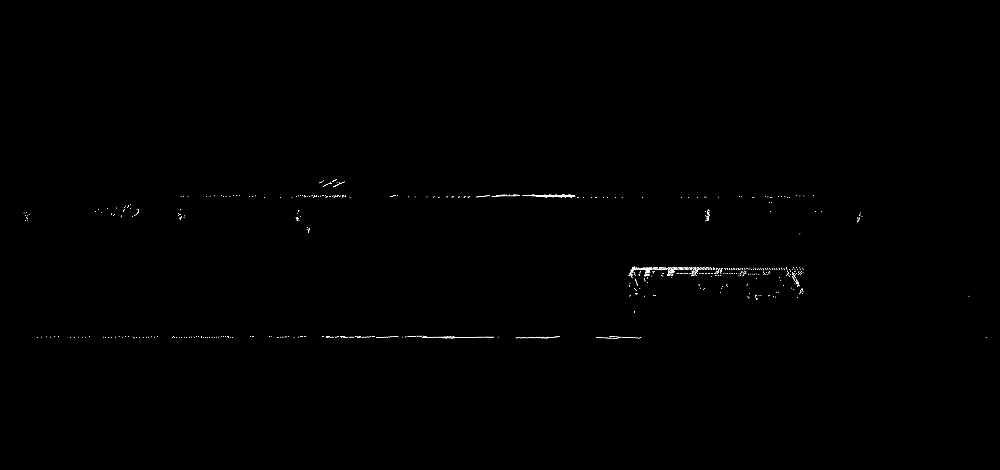} \\ 
      Ground Truth & Ours & Error Map \\
  \end{tabular}
  \vspace{-3mm}
  \caption{Semantic Segmentation Results on Driving Scenes Dataset}
  \label{fig:odtac}
\end{figure*}

\figref{fig:odtac} shows additional results for semantic labeling in driving scenes. As shown, the results capture very small dynamics, \eg pedestrians and bicyclists. This suggests our model's potential in object detection and tracking. The model is also able to distinguish between road and non-road through lidar intensity and subtle geometry structure such as road curbs. This validates our model's potential in map automation.

More specifically, we see that most error occur on road boundaries (bright curves in error map).

\subsection{Lidar Flow}
We show additional results on Lidar flow estimation in Fig.~\ref{fig:flow-prediction}. Unlike the visualization in the main submission, we visualize the colored vector in order to better depicts the magnitudes of the motion vector. As shown in the figure, our model is able to capture majority flow field. The majority of the error happens at the object boundary. This suggests that a better support domain that includes both space and intensity features could be potentially used to boost performance. 

\begin{figure*}
	\footnotesize
	\setlength\tabcolsep{0.5pt} %
	\renewcommand{\arraystretch}{0.8}
	\begin{tabular}{ccc}
  		\adjincludegraphics[width=.33\linewidth, trim={{.15\width} {.25\height} {.15\width} {.25\height}}, clip]{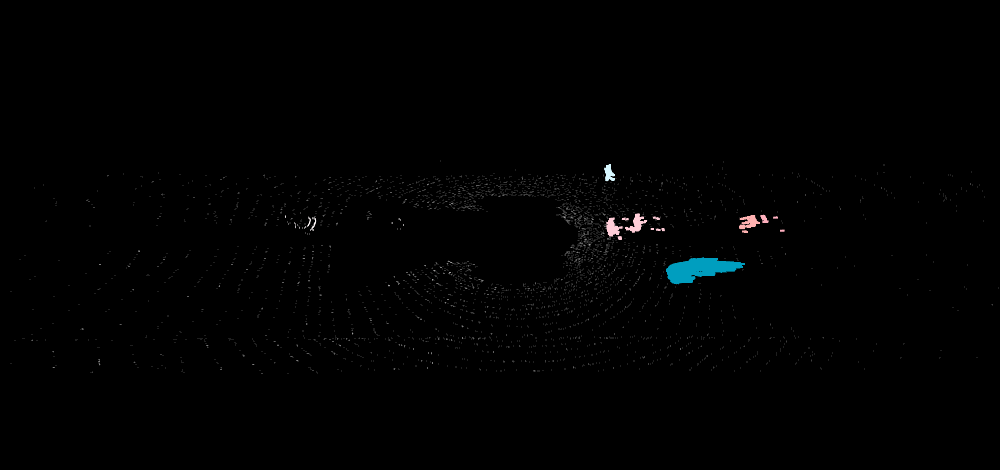} & 
   		\adjincludegraphics[width=.33\linewidth, trim={{.15\width} {.25\height} {.15\width} {.25\height}}, clip]{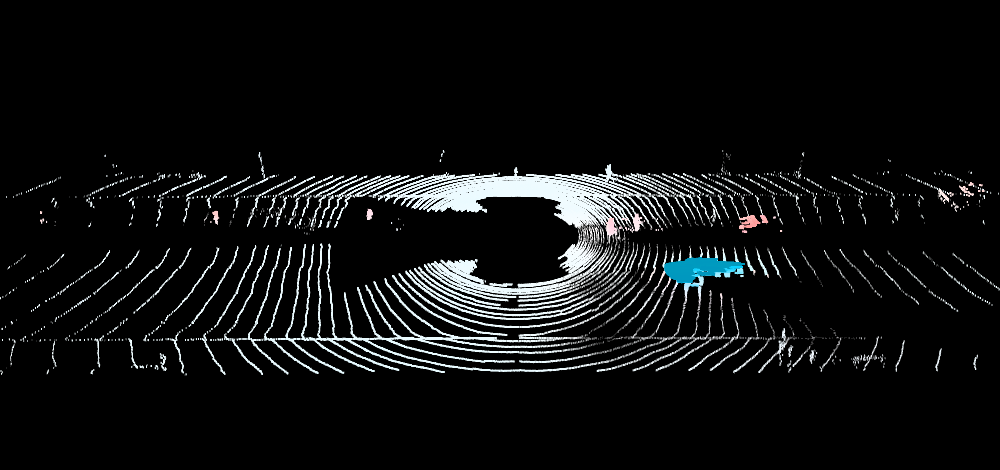} & 
   		\adjincludegraphics[width=.33\linewidth, trim={{.15\width} {.25\height} {.15\width} {.25\height}}, clip]{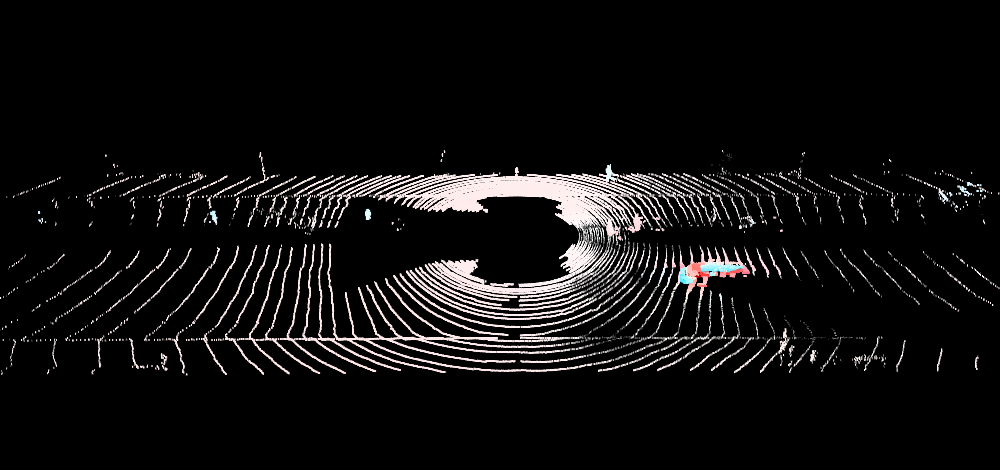} \\

      \adjincludegraphics[width=.33\linewidth, trim={{.15\width} {.25\height} {.15\width} {.25\height}}, clip]{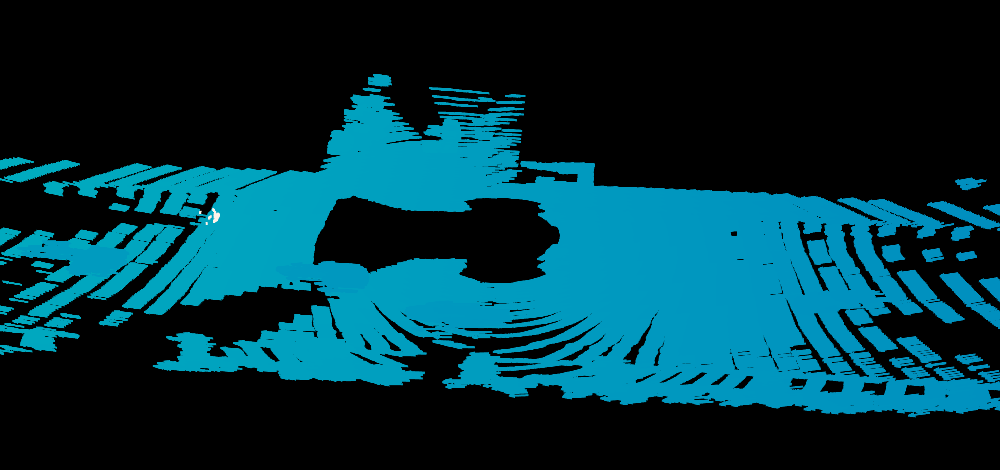} & 
      \adjincludegraphics[width=.33\linewidth, trim={{.15\width} {.25\height} {.15\width} {.25\height}}, clip]{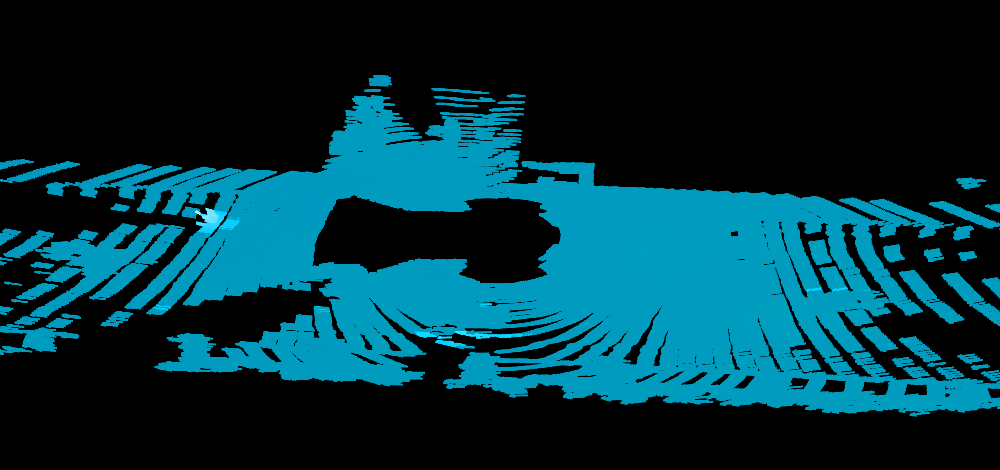} & 
      \adjincludegraphics[width=.33\linewidth, trim={{.15\width} {.25\height} {.15\width} {.25\height}}, clip]{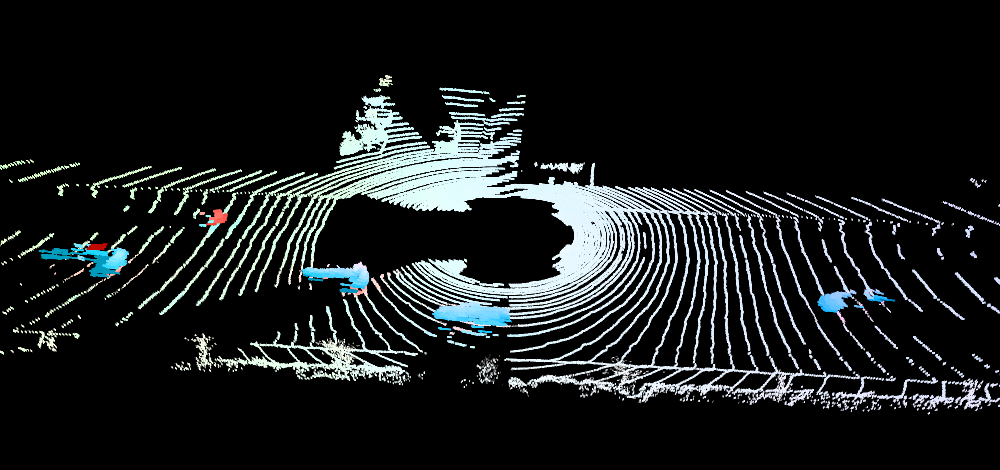} \\

   		\adjincludegraphics[width=.33\linewidth, trim={{.15\width} {.25\height} {.15\width} {.25\height}}, clip]{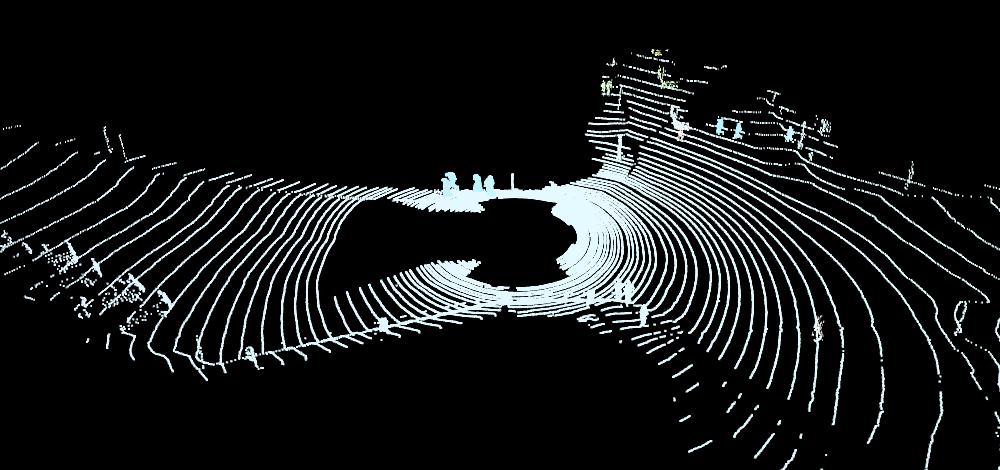} & 
   		\adjincludegraphics[width=.33\linewidth, trim={{.15\width} {.25\height} {.15\width} {.25\height}}, clip]{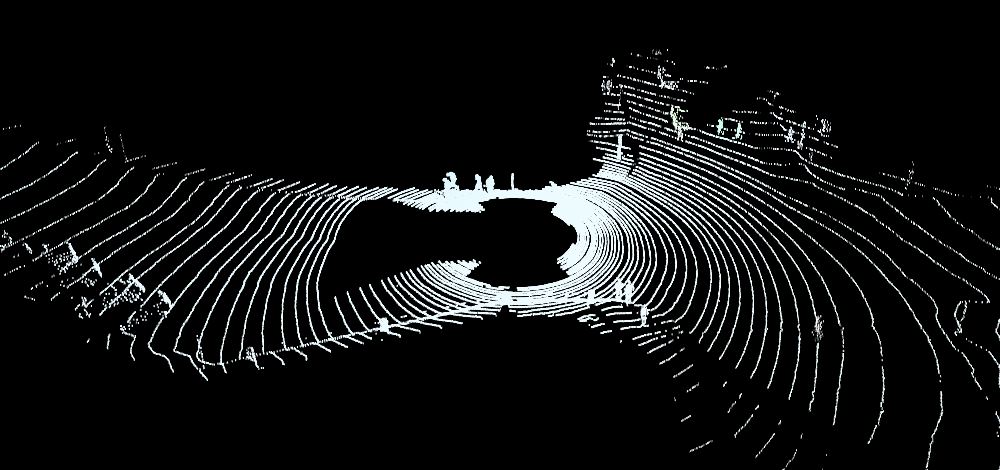} & 
   		\adjincludegraphics[width=.33\linewidth, trim={{.15\width} {.25\height} {.15\width} {.25\height}}, clip]{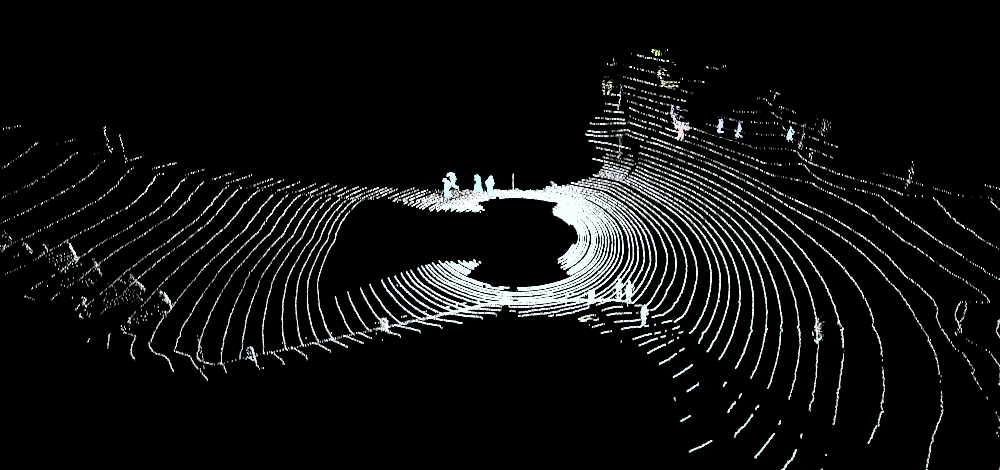} \\

      \adjincludegraphics[width=.33\linewidth, trim={{.15\width} {.25\height} {.15\width} {.25\height}}, clip]{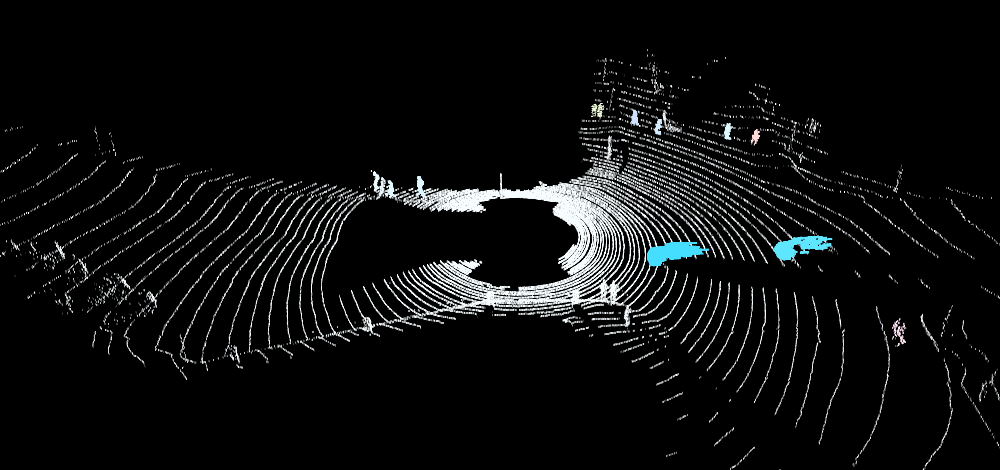} & 
      \adjincludegraphics[width=.33\linewidth, trim={{.15\width} {.25\height} {.15\width} {.25\height}}, clip]{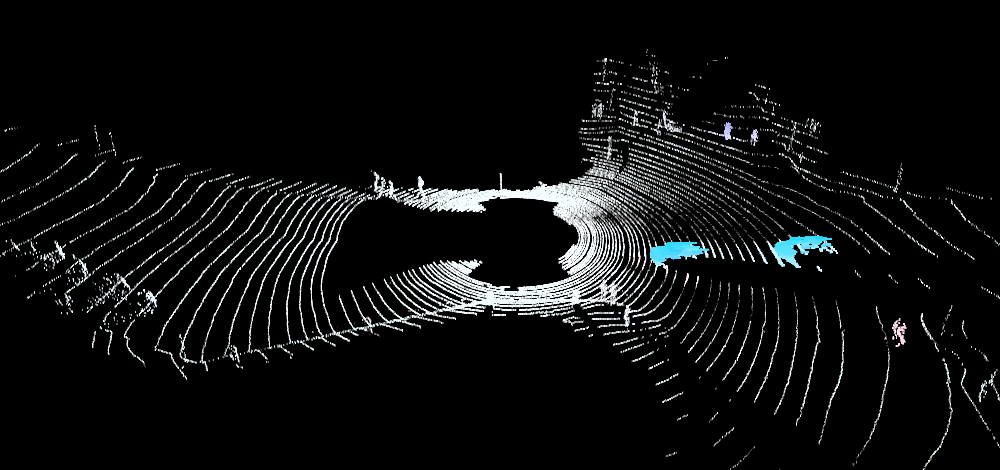} & 
      \adjincludegraphics[width=.33\linewidth, trim={{.15\width} {.25\height} {.15\width} {.25\height}}, clip]{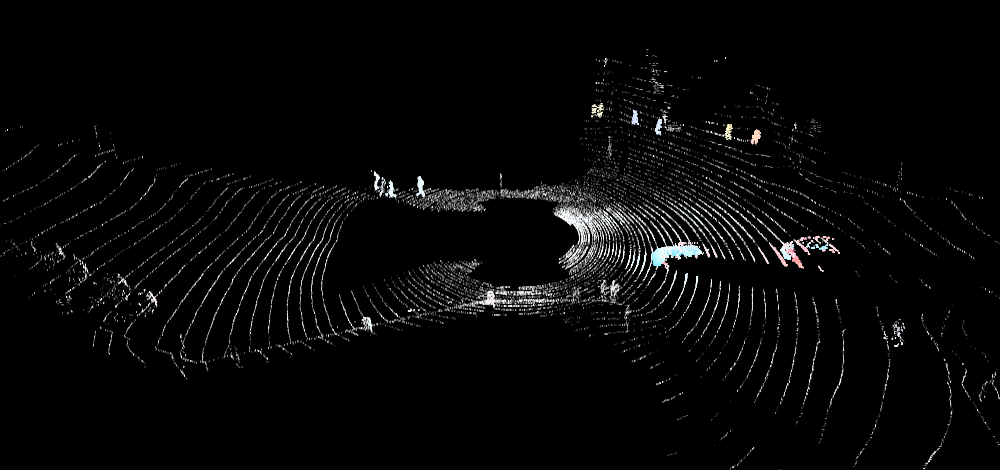} \\

   		\adjincludegraphics[width=.33\linewidth, trim={{.15\width} {.25\height} {.15\width} {.25\height}}, clip]{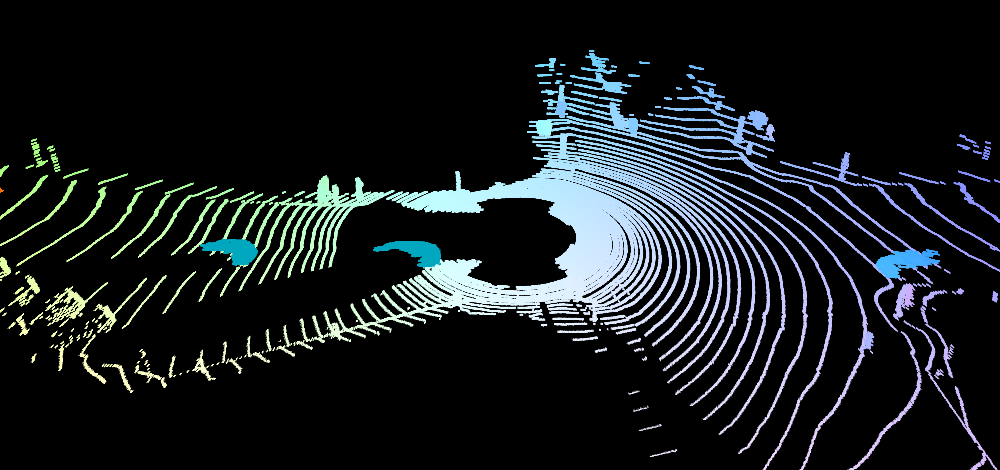} & 
   		\adjincludegraphics[width=.33\linewidth, trim={{.15\width} {.25\height} {.15\width} {.25\height}}, clip]{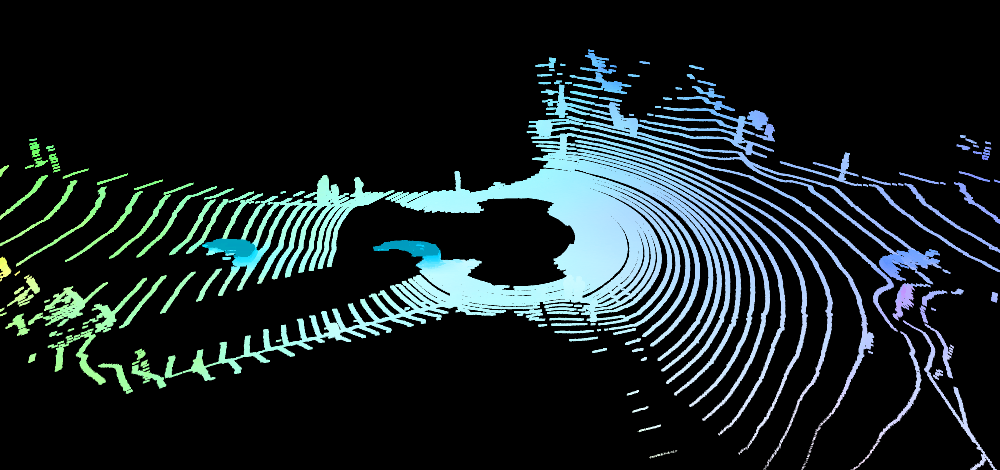} & 
   		\adjincludegraphics[width=.33\linewidth, trim={{.15\width} {.25\height} {.15\width} {.25\height}}, clip]{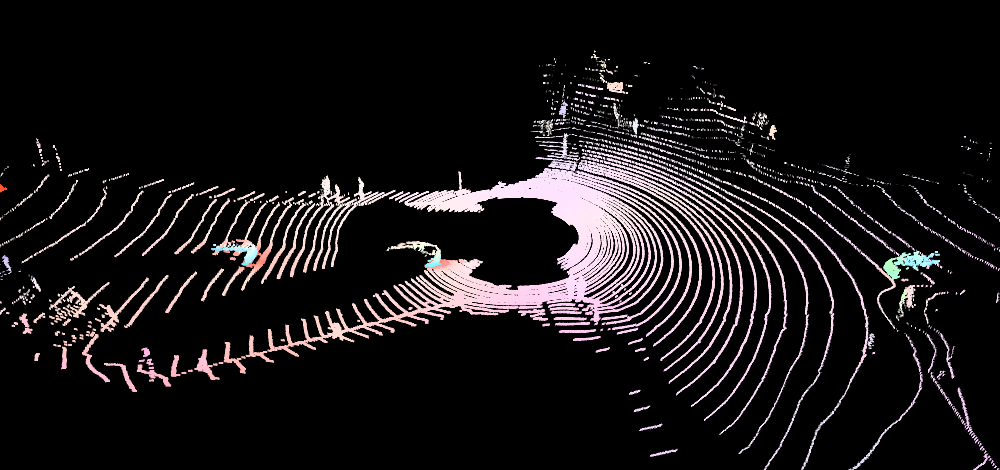} \\

      \adjincludegraphics[width=.33\linewidth, trim={{.15\width} {.25\height} {.15\width} {.25\height}}, clip]{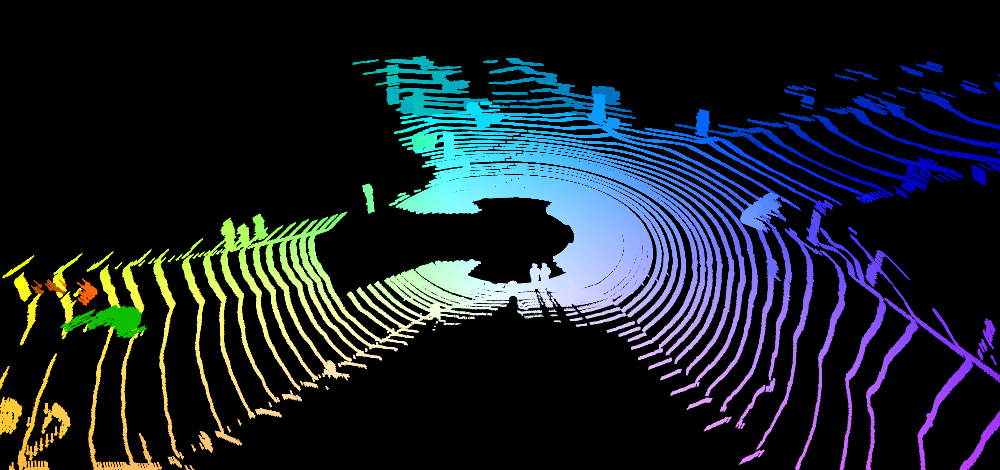} & 
      \adjincludegraphics[width=.33\linewidth, trim={{.15\width} {.25\height} {.15\width} {.25\height}}, clip]{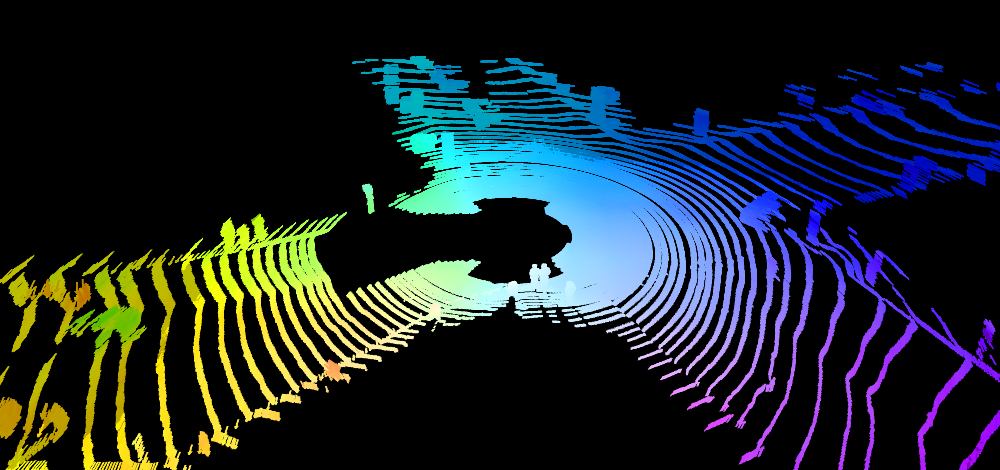} & 
      \adjincludegraphics[width=.33\linewidth, trim={{.15\width} {.25\height} {.15\width} {.25\height}}, clip]{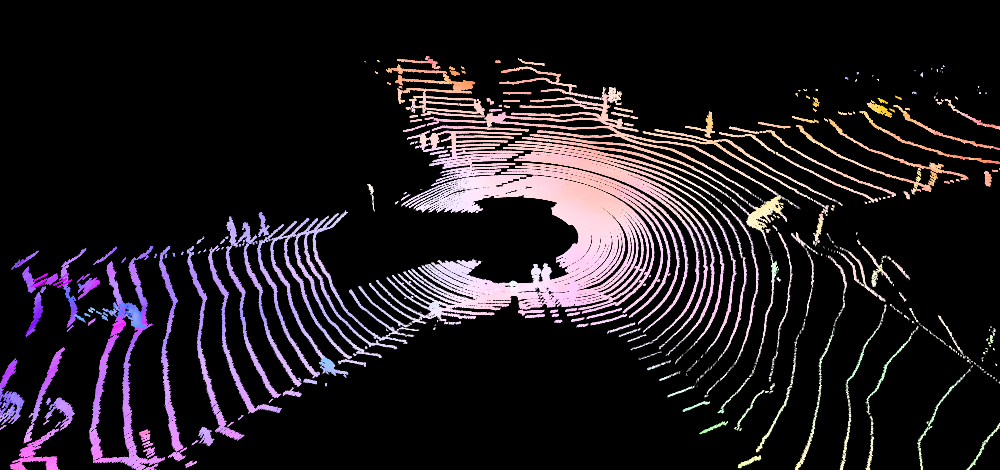} \\

   		\adjincludegraphics[width=.33\linewidth, trim={{.15\width} {.25\height} {.15\width} {.25\height}}, clip]{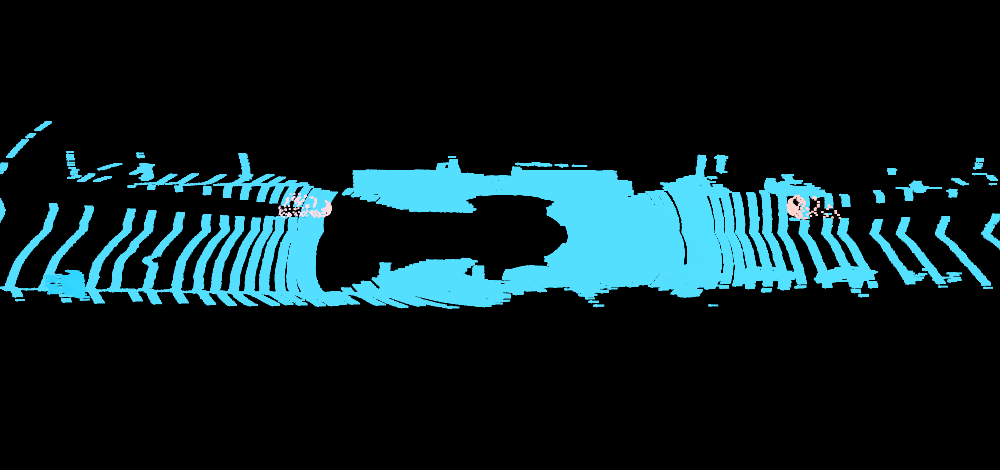} & 
   		\adjincludegraphics[width=.33\linewidth, trim={{.15\width} {.25\height} {.15\width} {.25\height}}, clip]{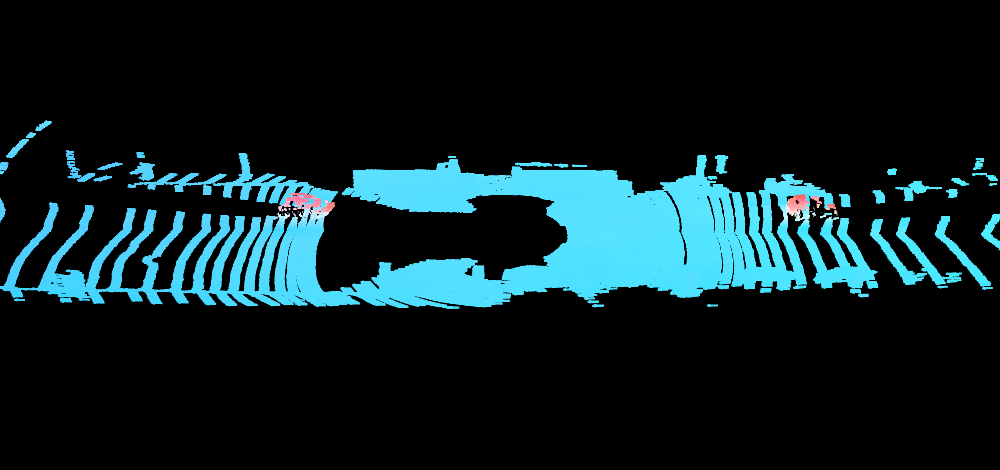} & 
   		\adjincludegraphics[width=.33\linewidth, trim={{.15\width} {.25\height} {.15\width} {.25\height}}, clip]{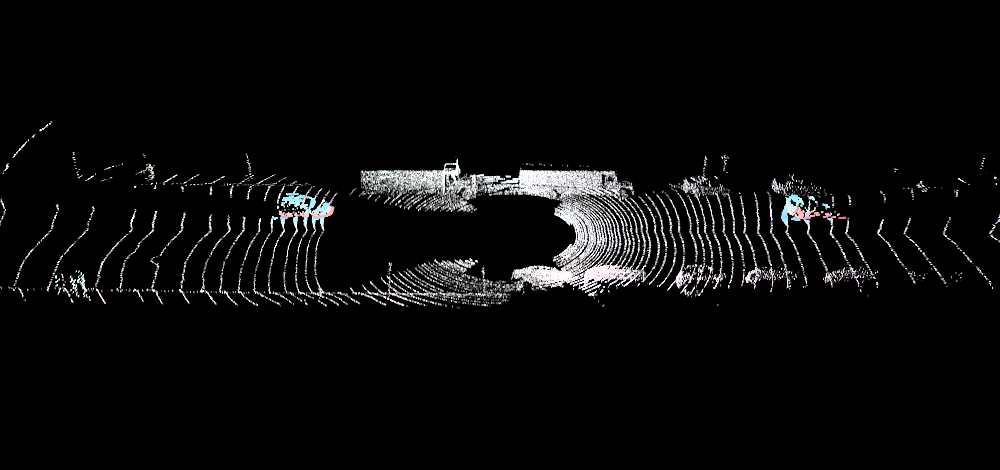} \\

   		\adjincludegraphics[width=.33\linewidth, trim={{.15\width} {.25\height} {.15\width} {.25\height}}, clip]{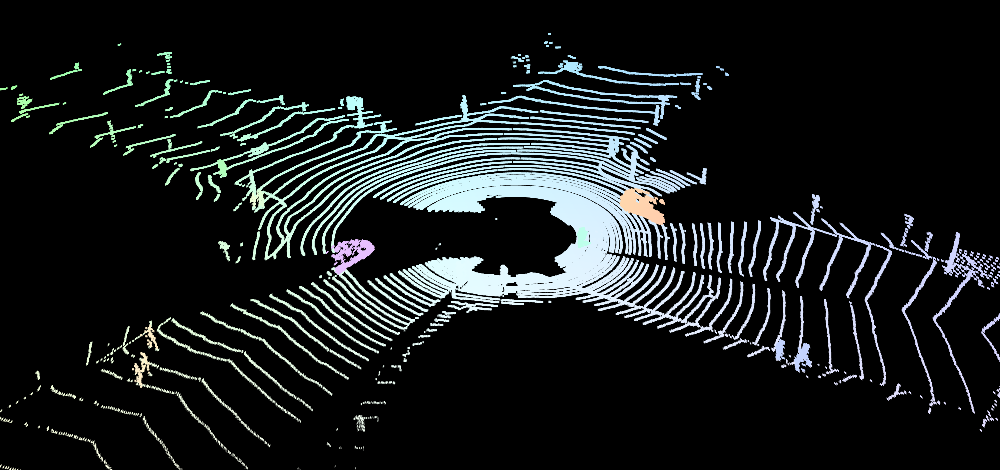} & 
   		\adjincludegraphics[width=.33\linewidth, trim={{.15\width} {.25\height} {.15\width} {.25\height}}, clip]{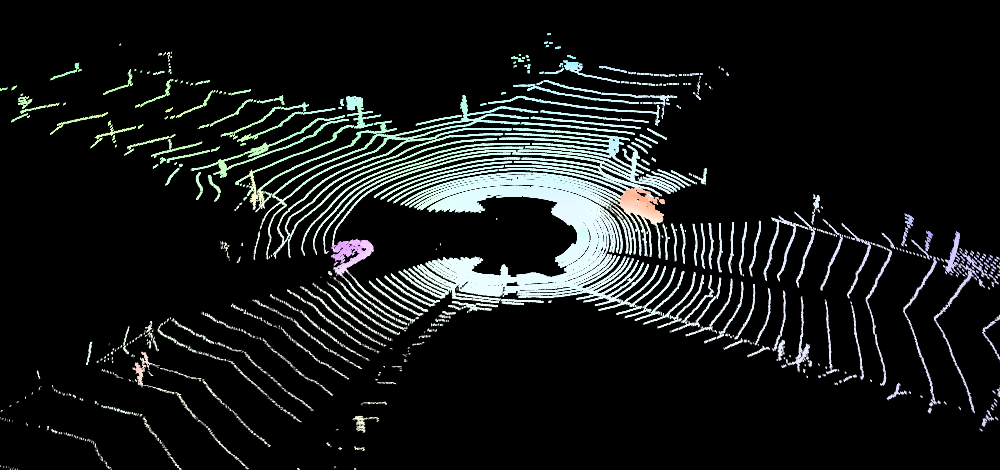} & 
   		\adjincludegraphics[width=.33\linewidth, trim={{.15\width} {.25\height} {.15\width} {.25\height}}, clip]{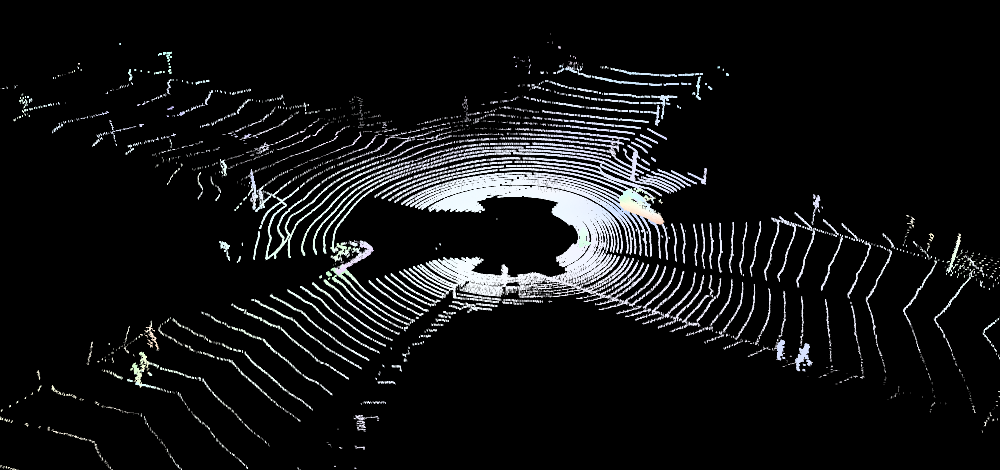} \\

      \adjincludegraphics[width=.33\linewidth, trim={{.15\width} {.25\height} {.15\width} {.25\height}}, clip]{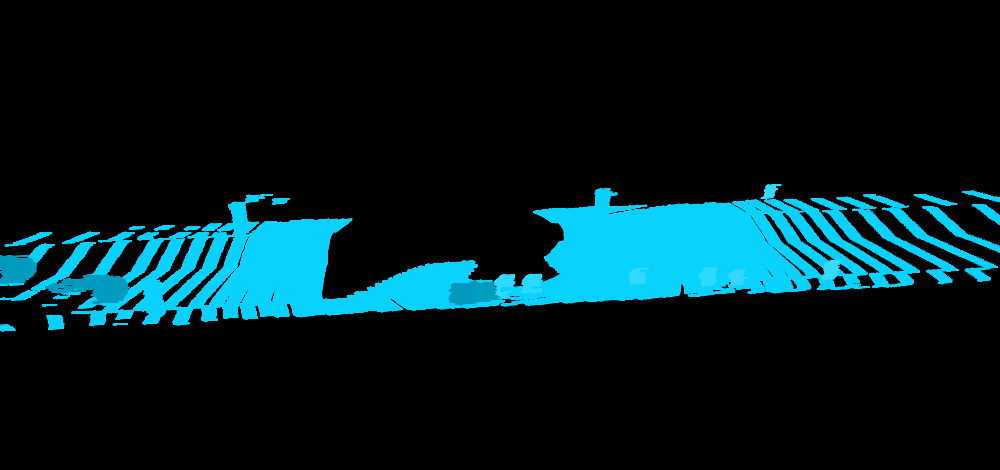} & 
      \adjincludegraphics[width=.33\linewidth, trim={{.15\width} {.25\height} {.15\width} {.25\height}}, clip]{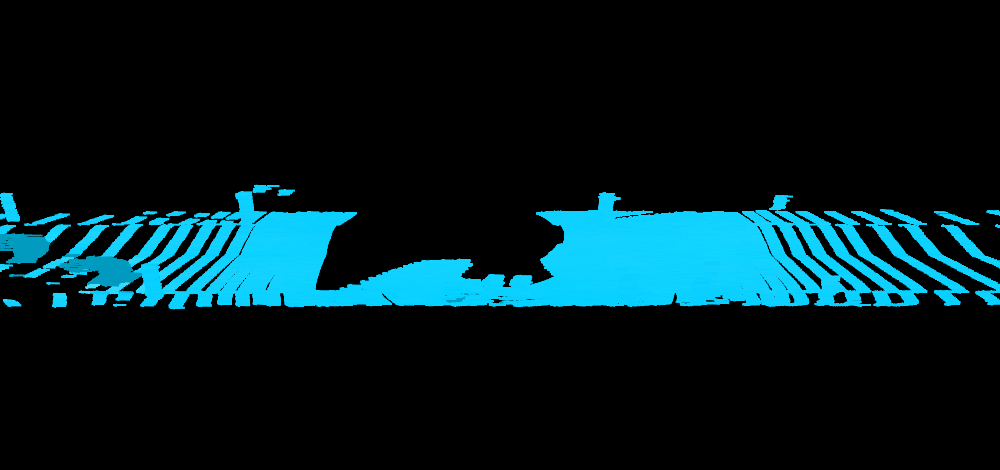} & 
      \adjincludegraphics[width=.33\linewidth, trim={{.15\width} {.25\height} {.15\width} {.25\height}}, clip]{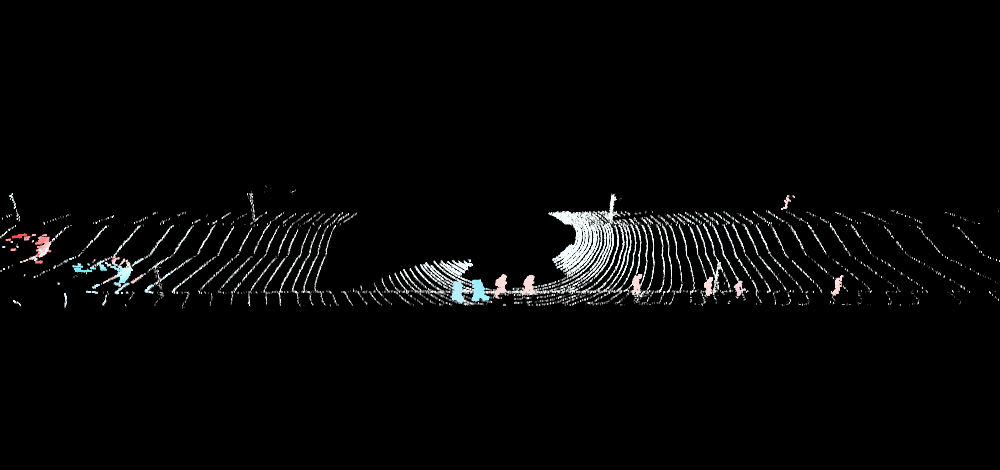} \\ 

      \adjincludegraphics[width=.33\linewidth, trim={{.15\width} {.25\height} {.15\width} {.25\height}}, clip]{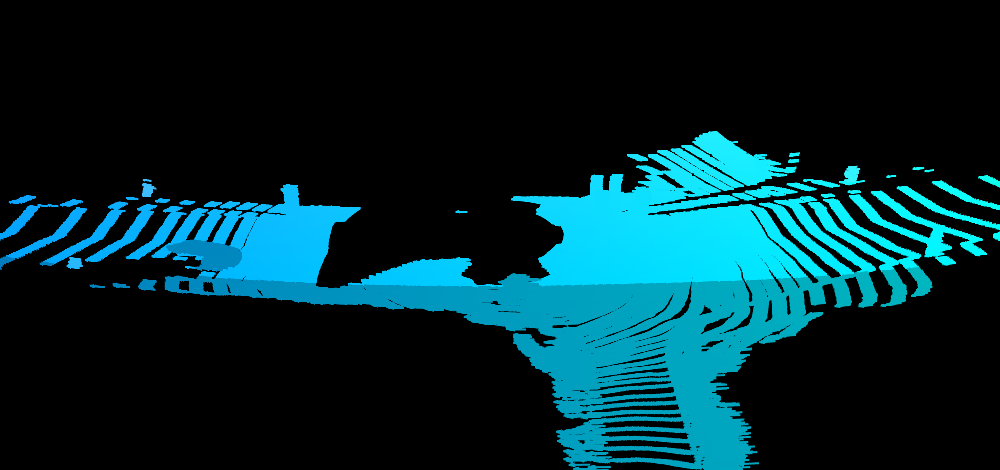} & 
      \adjincludegraphics[width=.33\linewidth, trim={{.15\width} {.25\height} {.15\width} {.25\height}}, clip]{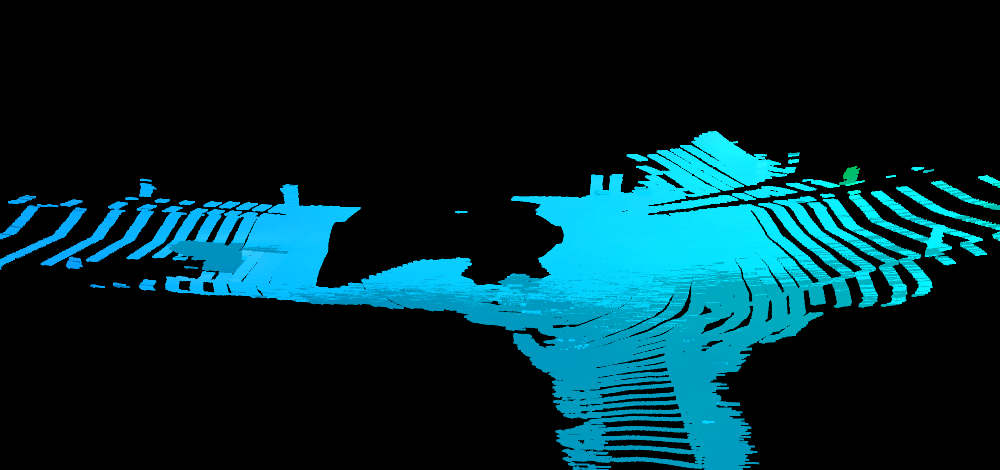} & 
      \adjincludegraphics[width=.33\linewidth, trim={{.15\width} {.25\height} {.15\width} {.25\height}}, clip]{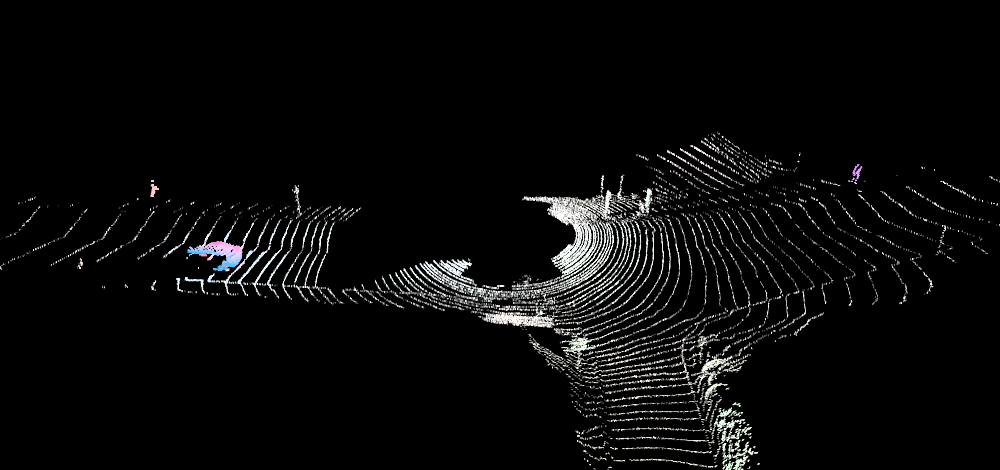} \\ 

      \adjincludegraphics[width=.33\linewidth, trim={{.15\width} {.25\height} {.15\width} {.25\height}}, clip]{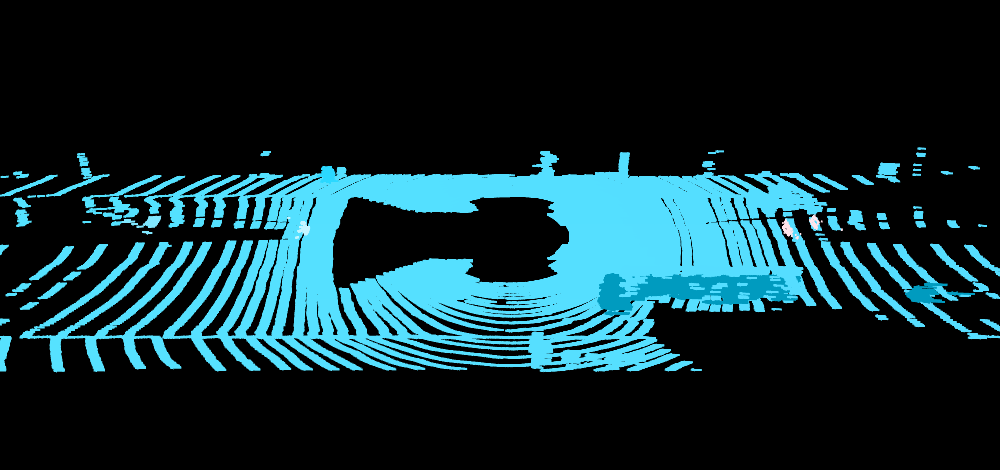} & 
      \adjincludegraphics[width=.33\linewidth, trim={{.15\width} {.25\height} {.15\width} {.25\height}}, clip]{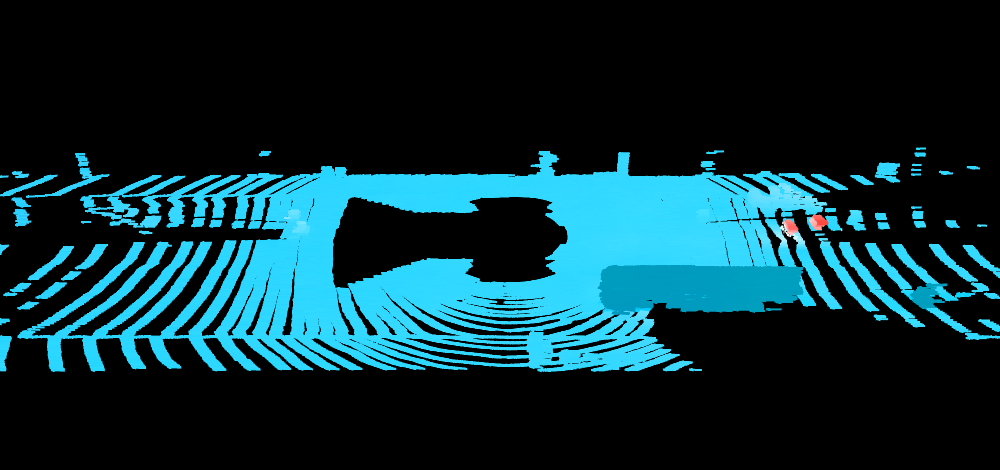} & 
      \adjincludegraphics[width=.33\linewidth, trim={{.15\width} {.25\height} {.15\width} {.25\height}}, clip]{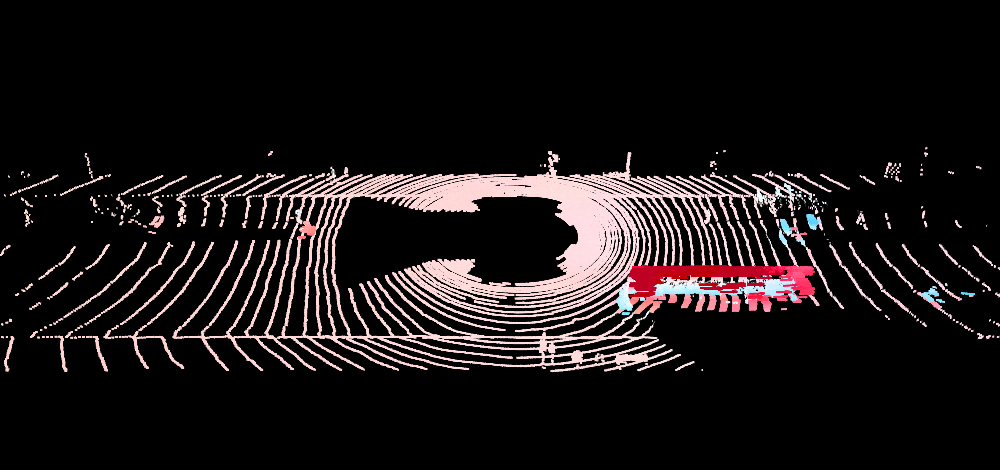} \\
   		Ground Truth & Ours & Error Map \\
	\end{tabular}
	\vspace{-3mm}
	\caption{Lidar Flow Results on Driving Scenes Dataset}
	\label{fig:flow-prediction}
\end{figure*}

\section{Activations}

\begin{figure*}
	\footnotesize
	\setlength\tabcolsep{0.5pt} %
	\renewcommand{\arraystretch}{0.8}
	\begin{tabular}{ccc}
  		\adjincludegraphics[width=.33\linewidth, trim={{.01\width} {.01\height} {.01\width} {.01\height}}, clip]{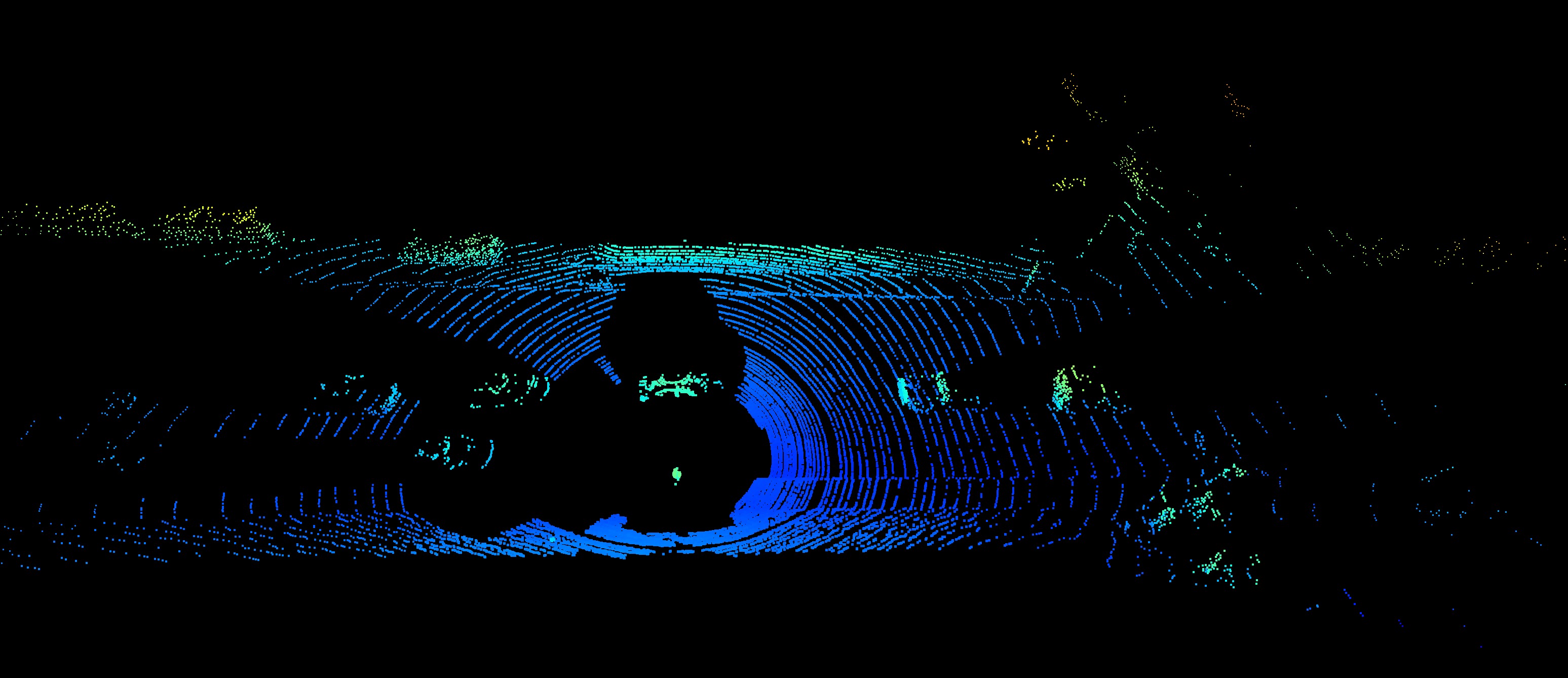} & 
  		\adjincludegraphics[width=.33\linewidth, trim={{.01\width} {.01\height} {.01\width} {.01\height}}, clip]{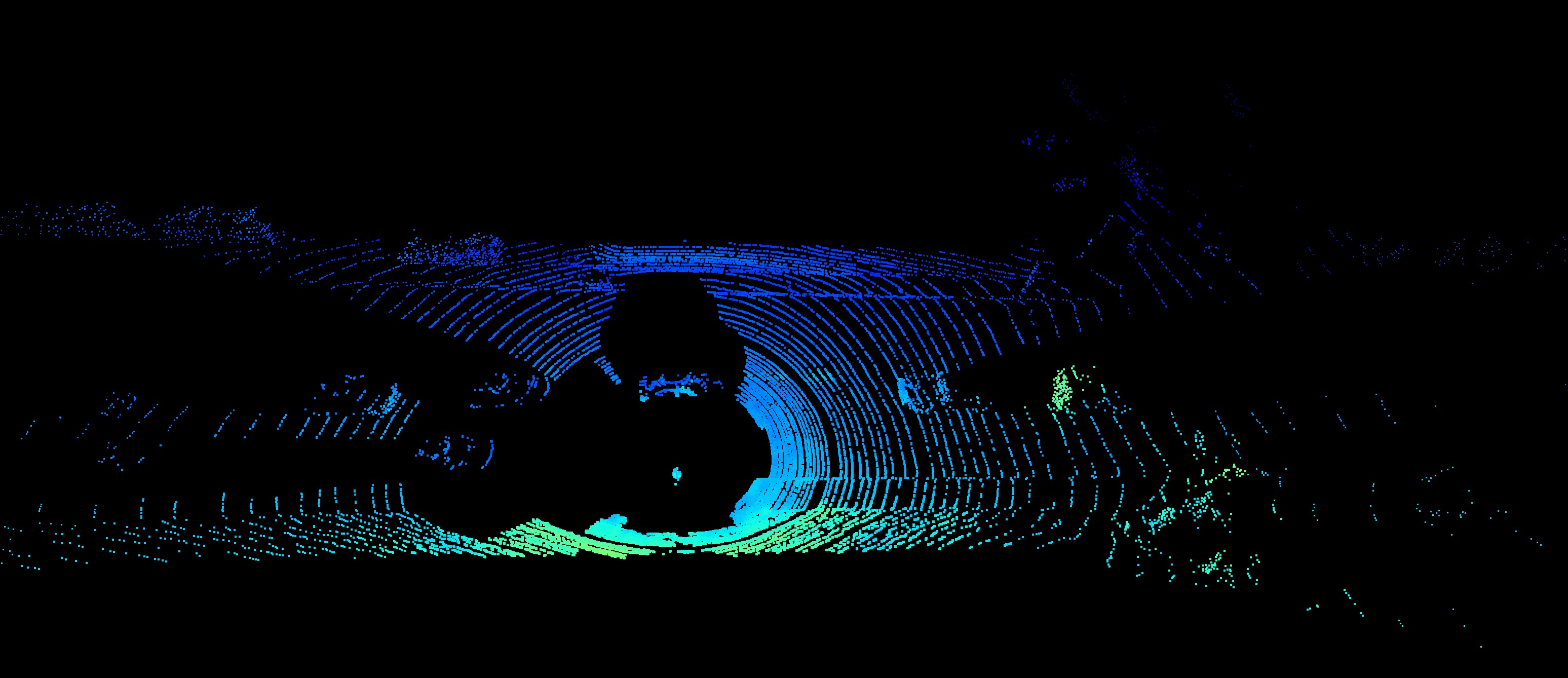} & 
  		\adjincludegraphics[width=.33\linewidth, trim={{.01\width} {.01\height} {.01\width} {.01\height}}, clip]{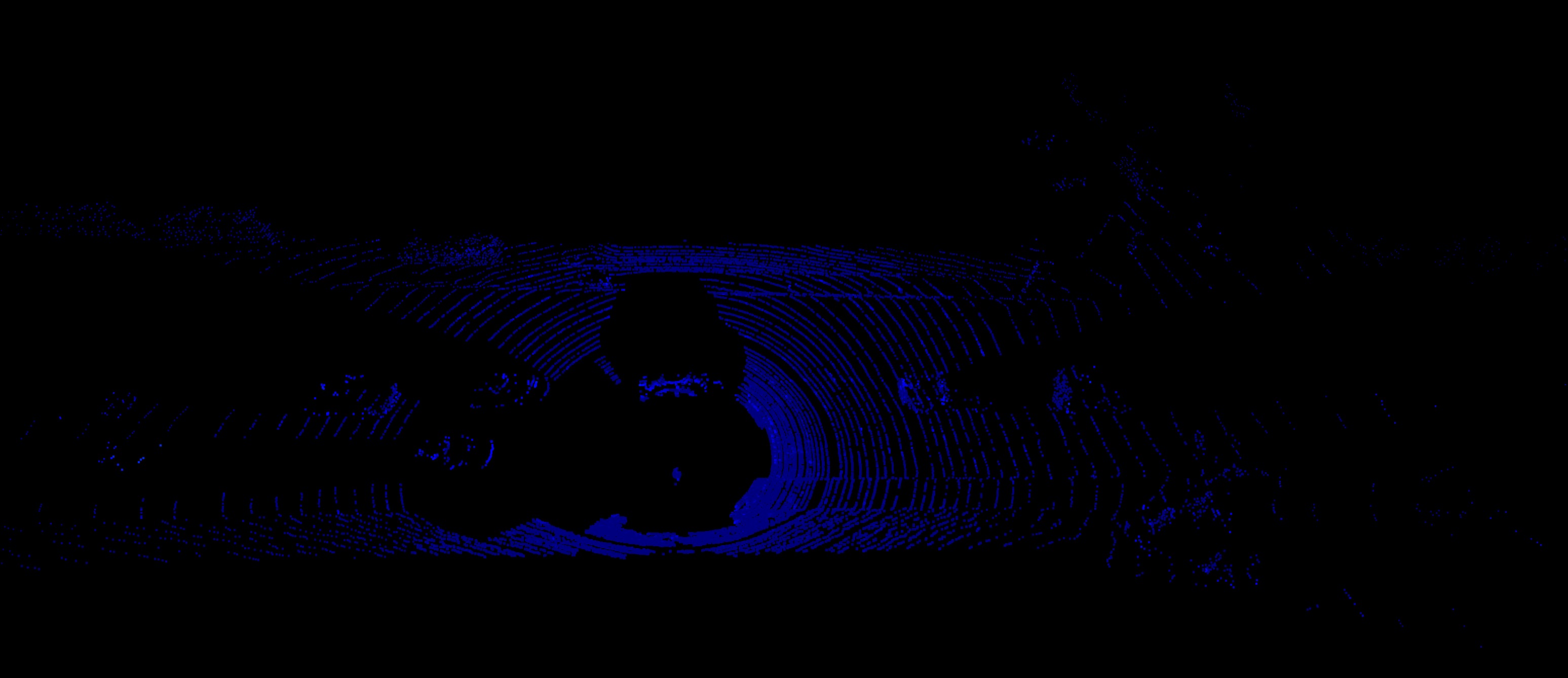} \\
  		\adjincludegraphics[width=.33\linewidth, trim={{.01\width} {.01\height} {.01\width} {.01\height}}, clip]{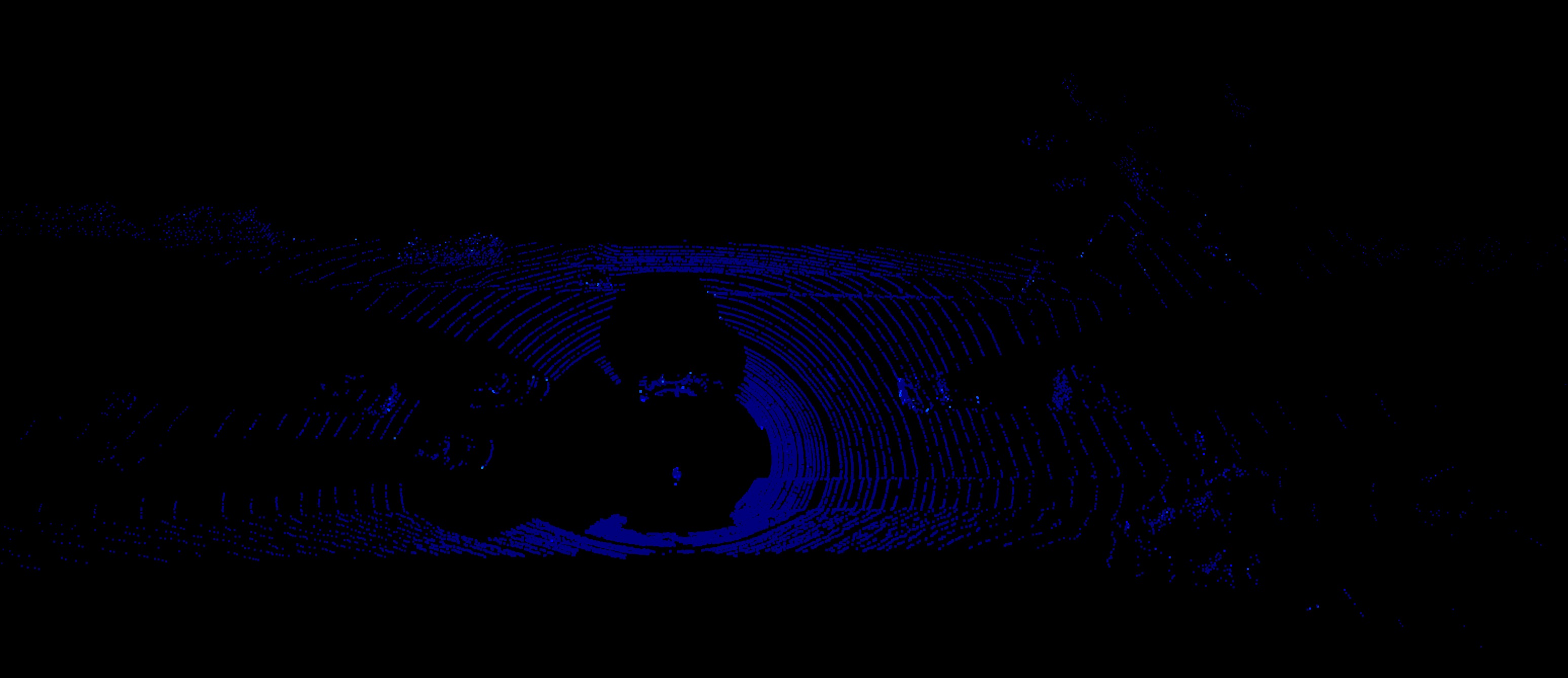} & 
  		\adjincludegraphics[width=.33\linewidth, trim={{.01\width} {.01\height} {.01\width} {.01\height}}, clip]{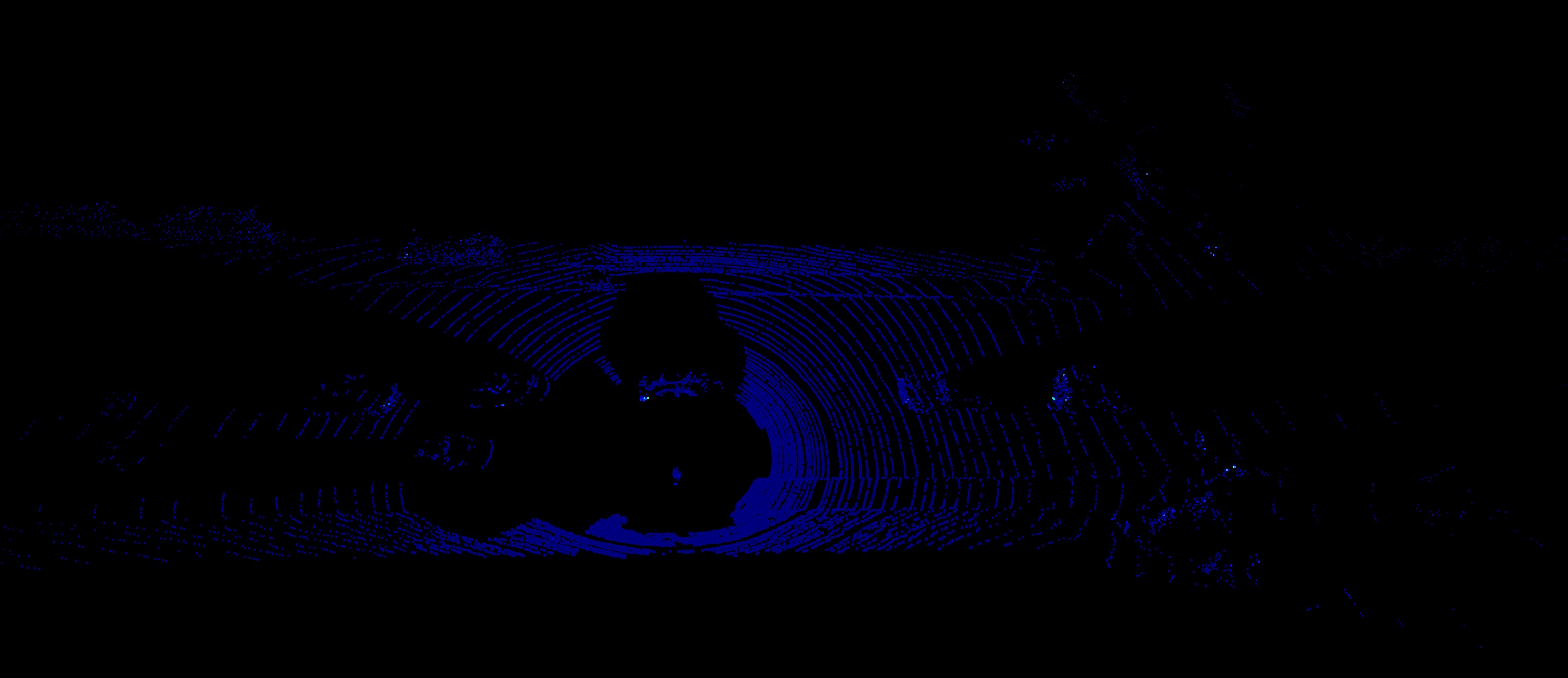} & 
  		\adjincludegraphics[width=.33\linewidth, trim={{.01\width} {.01\height} {.01\width} {.01\height}}, clip]{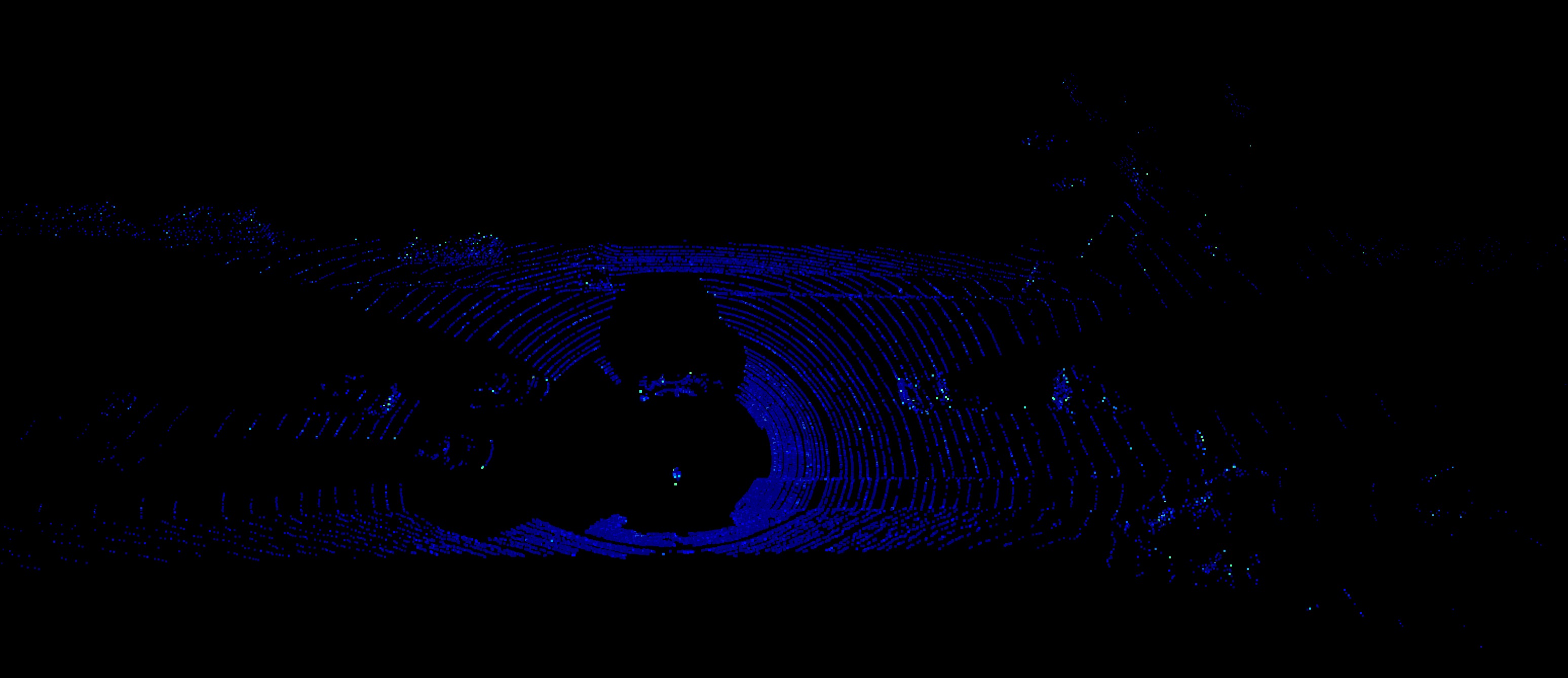} \\
   		\end{tabular}
	\vspace{-3mm}
	\caption{Activation Map of PCCN at Layer 1}
	\label{fig:layer1}
\end{figure*}

\begin{figure*}
	\footnotesize
	\setlength\tabcolsep{0.5pt} %
	\renewcommand{\arraystretch}{0.8}
	\begin{tabular}{ccc}
  		\adjincludegraphics[width=.33\linewidth, trim={{.01\width} {.01\height} {.01\width} {.01\height}}, clip]{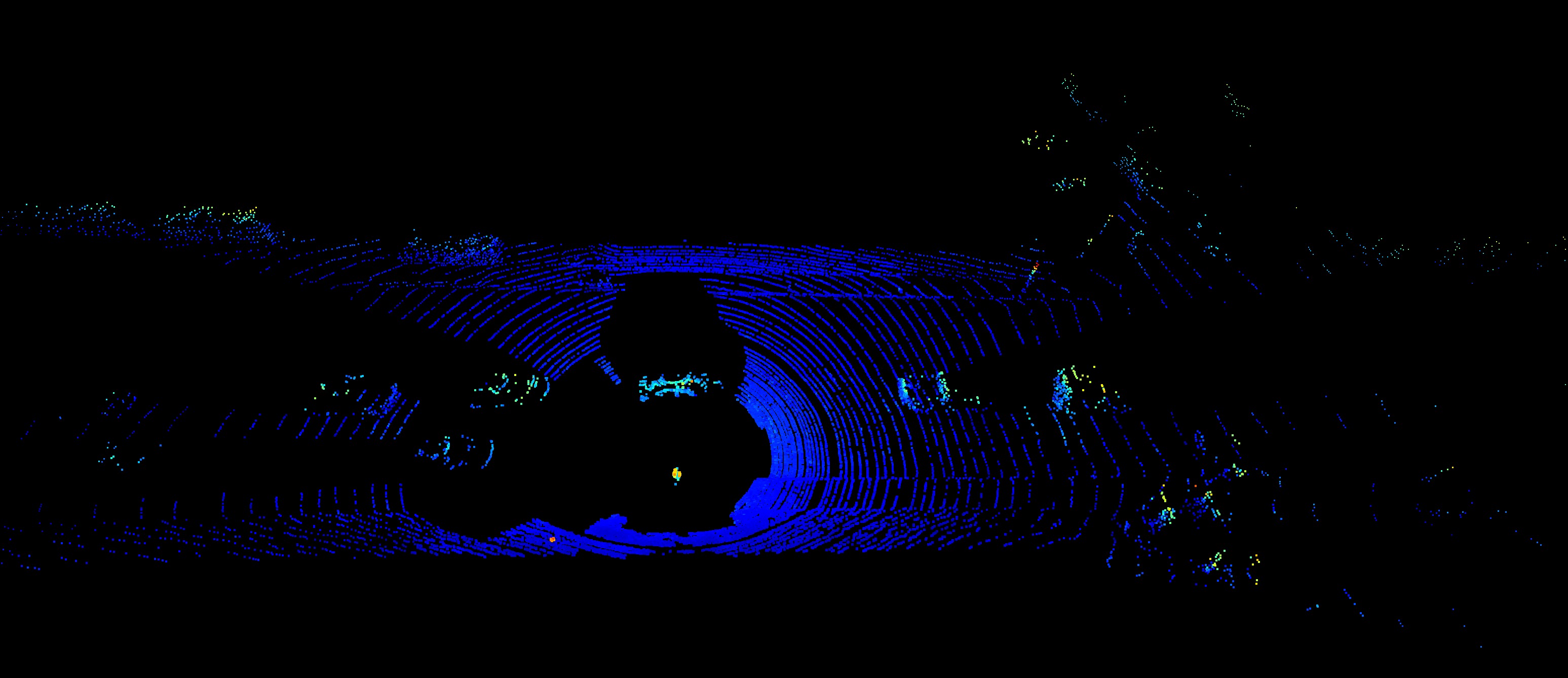} & 
  		\adjincludegraphics[width=.33\linewidth, trim={{.01\width} {.01\height} {.01\width} {.01\height}}, clip]{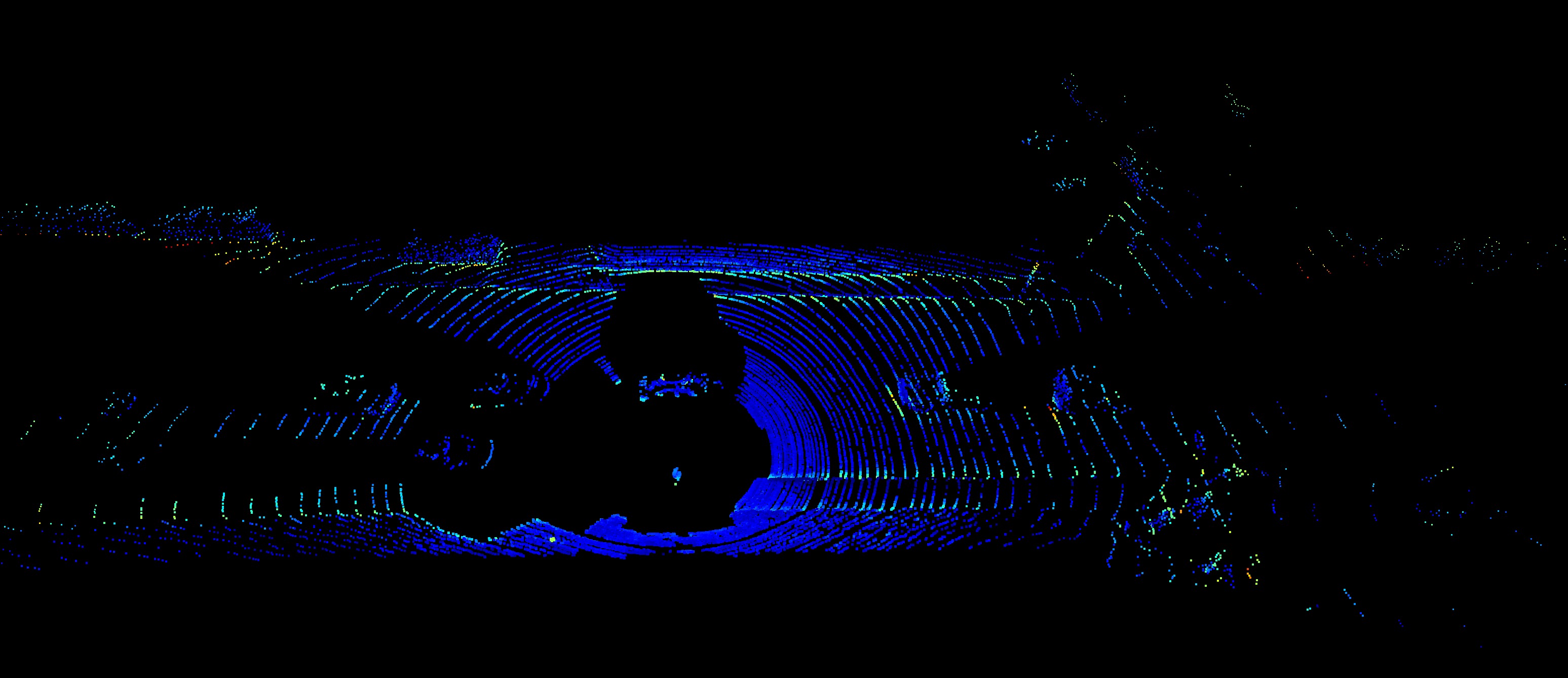} & 
  		\adjincludegraphics[width=.33\linewidth, trim={{.01\width} {.01\height} {.01\width} {.01\height}}, clip]{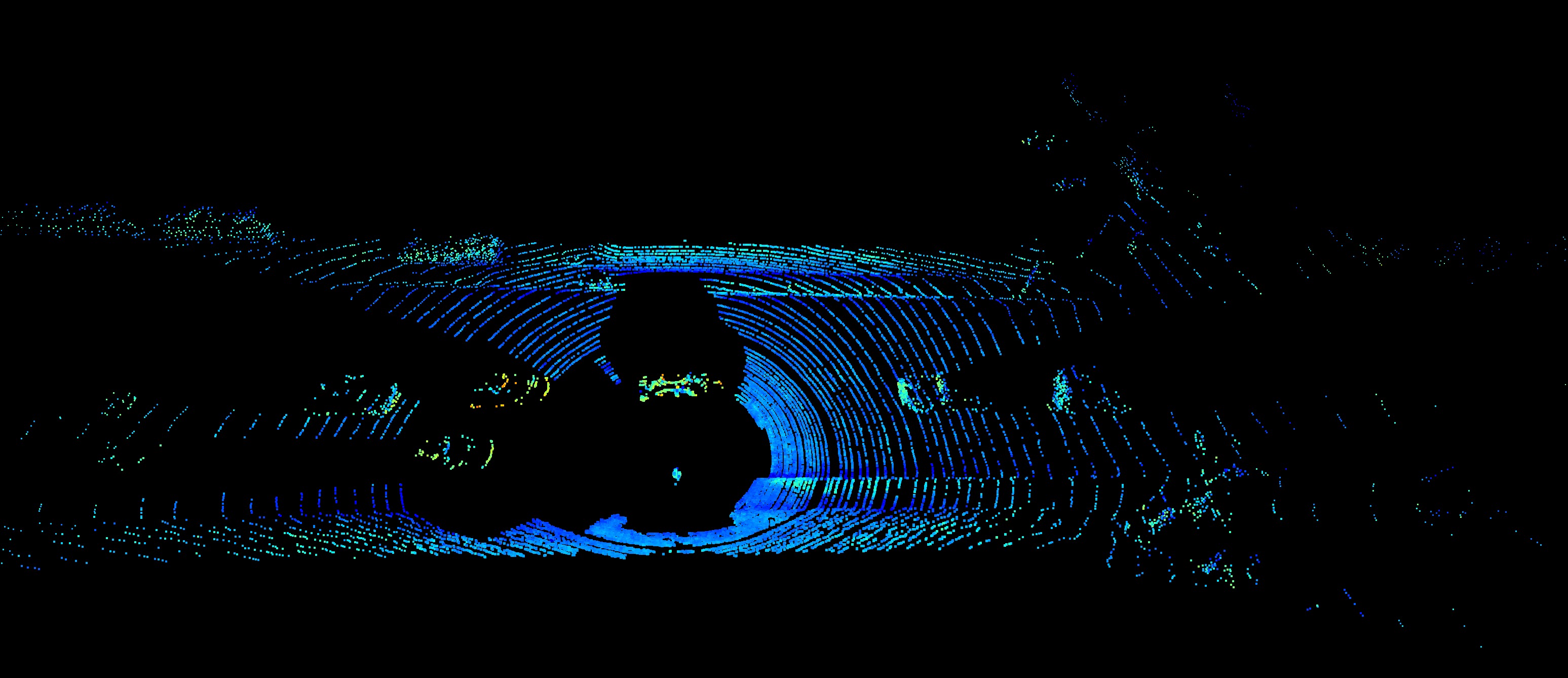} \\
  		\adjincludegraphics[width=.33\linewidth, trim={{.01\width} {.01\height} {.01\width} {.01\height}}, clip]{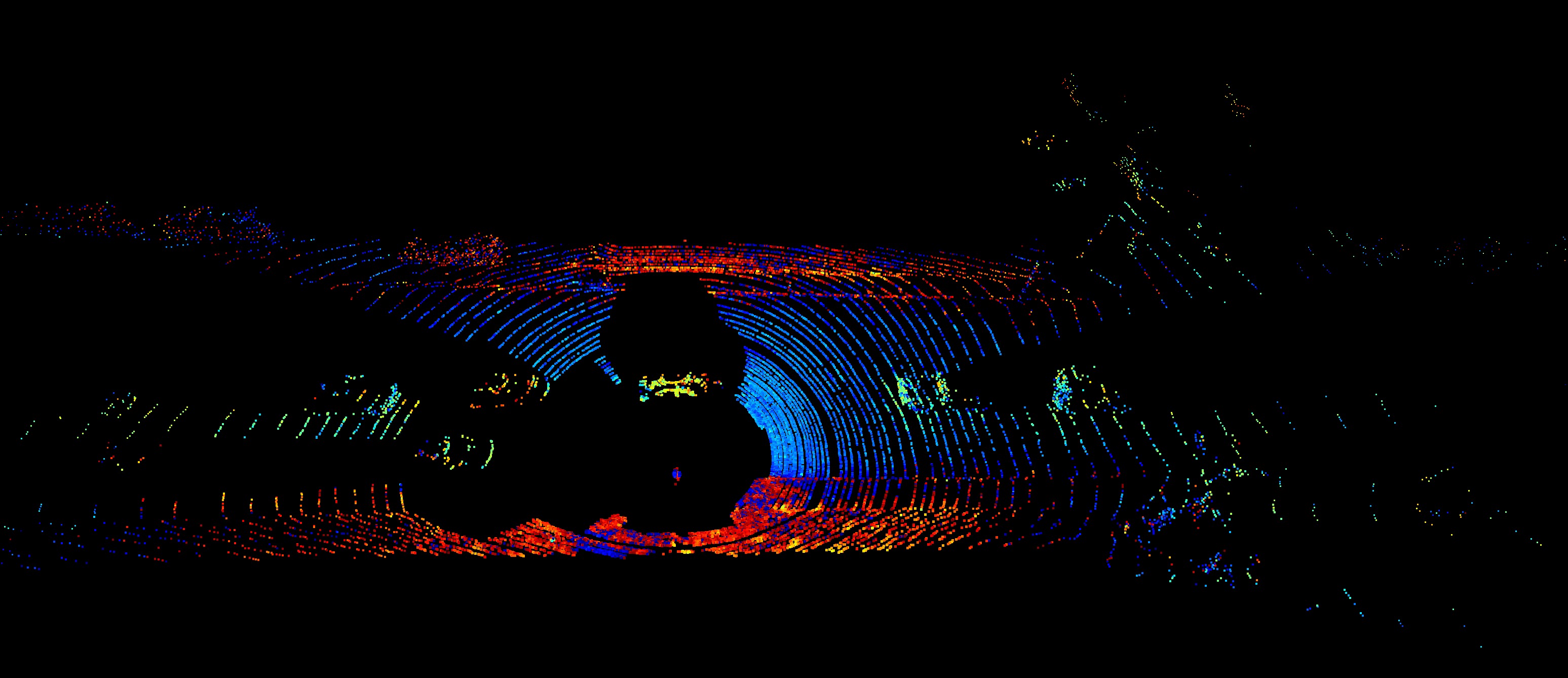} & 
  		\adjincludegraphics[width=.33\linewidth, trim={{.01\width} {.01\height} {.01\width} {.01\height}}, clip]{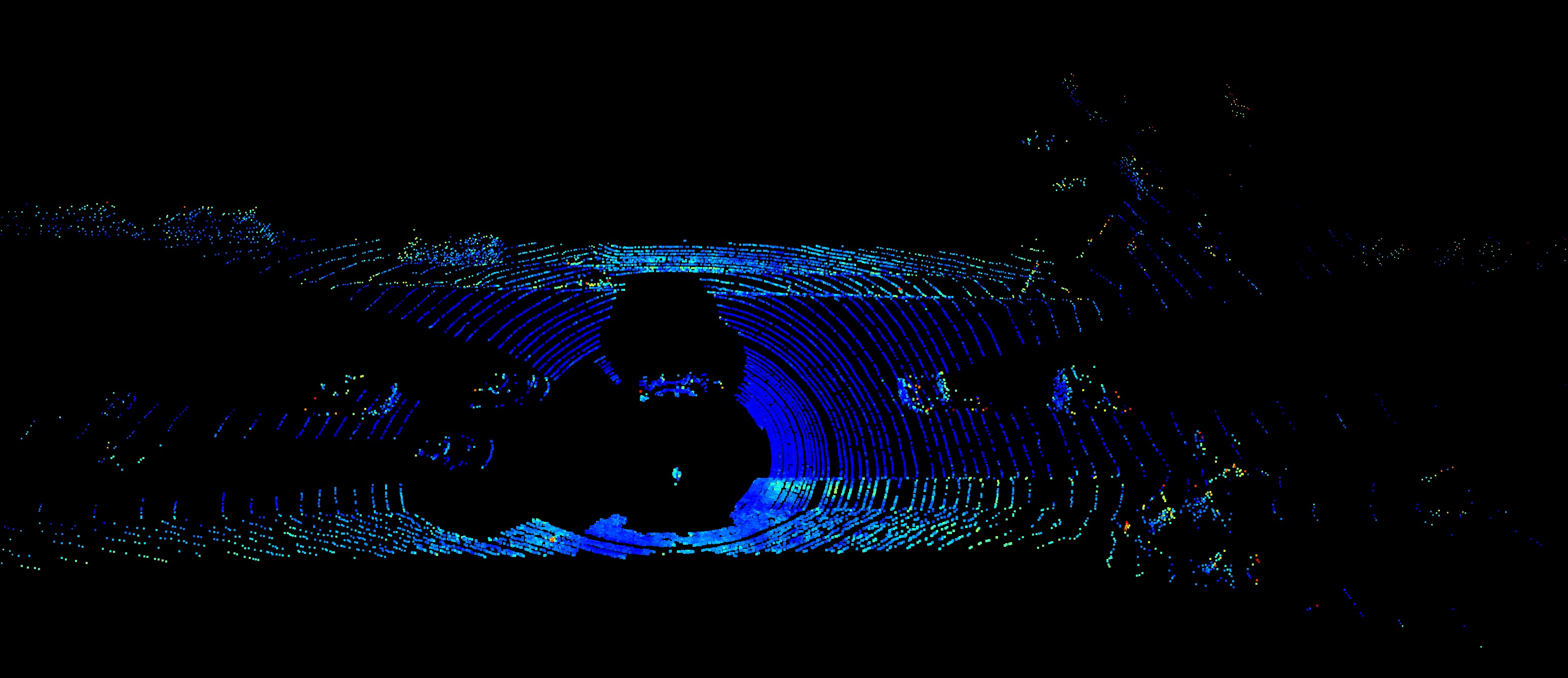} & 
  		\adjincludegraphics[width=.33\linewidth, trim={{.01\width} {.01\height} {.01\width} {.01\height}}, clip]{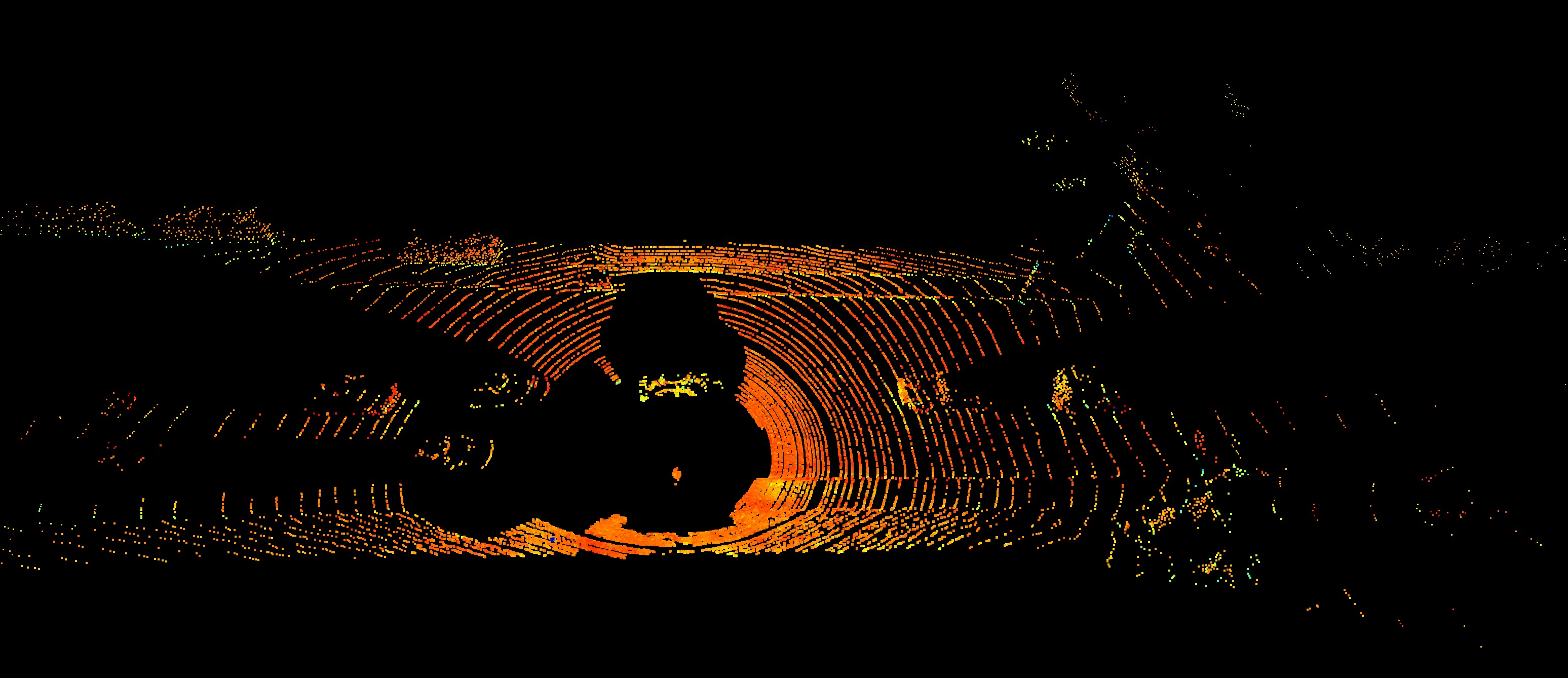} \\
   		\end{tabular}
	\vspace{-3mm}
	\caption{Activation Map of PCCN at Layer 8}
	\label{fig:layer8}
\end{figure*}

We visualize the activation maps of the trained PCCN network over a single lidar frame from the driving scene dataset for segmentation. Fig.~\ref{fig:layer1} depicts the activation map at layer 1 of PCCN. As we can see, at the early conv layer the method mainly captures low-level geometry details, \eg the z coordinate, the intensity peak, \etc. Fig.~\ref{fig:layer8} shows the activation map at layer 8 of PCCN. The conv layers begin to capture information with more semantic meaning, e.g. the road curb and the dynamic objects.  

\section{Point Cloud Classification}

To verify the applicability of the proposed parameteric continuous convolution over global prediction task, we conduct a simple point cloud classification task on the ModelNet40 benchmark. This dataset contains CAD models from 40 categories. The state-of-the-art and most representative algorithms conducted on ModelNet40 are compared \cite{mvcnn}. We randomly sampled 2048 points for each training and testing sample over the 3D meshes and feed the point cloud into our neural network. The architecture contains 6 continuous convolution layers with 32-dimensional hidden features, followed by two layers with 128-dimensions and 512 dimensions respectively. The output of the last continuous convolution layer is fed into a max pooling layer to generate the global 512-dimensional feature, followed by two fc layers to output the final logits. \tabref{tab-modelnet40} reports the classification performance.  As we can see in the table, the performance is comparable with PointNet and slightly below PointNet++. Here we use a naive global max pooling to aggregate global information for our method. We expect to achieve better results with more comprehensive and hierachical pooling strategies. 

\begin{table}[]
\centering
\caption{ModelNet40 Point Cloud Classification}
\label{tab-modelnet40}
    \begin{tabular}{c|cc}
        Method           & Input            & Accuracy \\ \hline
        MVCNN \cite{mvcnn}           & Multi-view Image & 90.1\%      \\
        3DShapeNet \cite{modelnet40}       & Volume           & 84.7\%      \\
        VoxNet \cite{voxnet}           & Volume           & 85.9\%      \\
        Subvolume \cite{subvolume}        & Volume           & 89.2\%      \\
        ECC \cite{ecc}              & Point            & 87.4\%      \\
        PointNet vanilla \cite{pointnet} & Point            & 87.2\%      \\
        PointNet \cite{pointnet}         & Point            & 89.2\%      \\
        PointNet++ \cite{pointnet2}       & Point            & \textbf{91.9\%}      \\
        Ours             & Point            & 88.9\%     
    \end{tabular}
\end{table}

\end{document}